\documentclass{article}
\usepackage{arxiv}
\usepackage[utf8]{inputenc} 
\usepackage[T1]{fontenc}    
\usepackage{hyperref}       
\usepackage{url}            
\usepackage{booktabs}       
\usepackage{amsfonts}       
\usepackage{nicefrac}       
\usepackage{microtype}      
\usepackage{lipsum}		    
\usepackage{graphicx}
\usepackage[numbers, sort&compress]{natbib}

\usepackage{doi}
\usepackage{enumerate}
\usepackage[nolist,nohyperlinks]{acronym}
\usepackage{cite}
\usepackage{algorithm}
\usepackage{algorithmicx}
\usepackage{algpseudocode}
\usepackage{amsmath}
\usepackage{amssymb}
\usepackage{multicol}
\usepackage{multirow}
\usepackage{makecell}
\usepackage{moresize}
\usepackage{bm}
\usepackage[table]{xcolor}
\usepackage{placeins}
\usepackage{subfig}

\usepackage[graphicx]{realboxes}

\def\vs{\emph{vs.~}}

\newcommand{\nl}{\vspace{0.5mm}{\noindent}}
\newcommand{\trtitle}{Continual Facial Expression Recognition: A Benchmark}

\renewcommand{\shorttitle}{Continual Facial Expression Recognition}

\title{\trtitle}

\author{
Nikhil~Churamani$^{*}$, Tolga~Dimlioglu$^{\dagger}$, German~I.~Parisi$^{\ddagger}$ 
and Hatice Gunes$^{*}$\\
$^{*}${Department of Computer Science and Technology, University of Cambridge, UK}\\ 
$^{\dagger}${Department of Electrical and Computer Engineering, New York University, USA} \\
$^{\ddagger}${Department of Informatics, University of Hamburg, Germany} 
\\\tt{Email: nikhil.churamani@cl.cam.ac.uk}
}
\hypersetup{
pdftitle={\trtitle},
pdfsubject={Continual Facial Expression Recognition},
pdfauthor={Nikhil~Churamani, Tolga~Dimlioglu, German I.Parisi, and Hatice~Gunes},
pdfkeywords={},
}

\begin{document}

\maketitle
\begin{abstract}

Understanding human affective behaviour, especially in the dynamics of real-world settings, requires \acf{FER} models to continuously adapt to individual differences in user expression, contextual attributions, and the environment. Current (deep) \ac{ML}-based \ac{FER} approaches pre-trained in isolation on benchmark datasets fail to capture the nuances of real-world interactions where data is available only incrementally, acquired by the agent or robot during interactions. New learning comes at the cost of previous knowledge, resulting in \textit{catastrophic forgetting}. Lifelong or \acf{CL}, on the other hand, enables adaptability in agents by being sensitive to changing data distributions, integrating new information without interfering with previously learnt knowledge. Positing \ac{CL} as an effective learning paradigm for \ac{FER}, this work presents the \ac{ConFER} benchmark that evaluates popular \ac{CL} techniques on \ac{FER} tasks. It presents a comparative analysis of several \ac{CL}-based approaches on popular \ac{FER} datasets such as \acs{CK+}, \acs{RAF-DB}, 
and AffectNet and present strategies for a successful implementation of \ac{ConFER} for \acf{AC} research. \ac{CL} techniques, under different learning settings, are shown to achieve \ac{SOTA} performance across several datasets, thus motivating a discussion on the benefits of applying \ac{CL} principles towards human behaviour understanding, particularly from facial expressions, as well the challenges entailed.

\textbf{Keywords:} Continual Learning, Lifelong Learning, Affective Computing, Facial Expression Recognition.

\end{abstract}

\section{Introduction}
Rapid improvements in \acf{AI} and \acf{ML} have lead to intelligent and adaptive systems increasingly becoming integrated in and contributing towards critical decisions of daily human life~\citep{buolamwini2018gender, Howard2017Addressing, McKinsey2021State}. It is thus important for these systems to be able to understand user idiosyncrasies and behaviour-patterns to improve their reliability. This is particularly true when these systems are used to observe and analyse human behaviour under conversational or assistive settings~\citep{McColl2015} to improve user experience. Understanding the psychological and cognitive underpinnings of human affective behaviour forms the central focus of \acf{AC} research~\citep{Picard:1997:AC:265013}. Aimed at developing solutions that can analyse human affective behaviour, \ac{AC} research evaluates different inward (biological signals like galvanic skin-responses or heart rate) or outward signals (physical gestures like facial expressions, body gestures and speech intonations)~\citep{Gunes2011Emotion} to perceive the expressed affective state of users. 

Amongst other signals such as speech~\citep{Schuller2018SER} and body signalling~\citep{Noroozi2021Survey}, analysing human facial expressions is a straightforward and perhaps, the most commonly used method to appraise human affective behaviour~\citep{Sariyanidi2015Automatic}. 
As a result, \acf{FER} has increasingly become a popular area of investigation, providing opportunities for developing agents that can interact with humans in an engaging manner. From traditionally applying image processing-based techniques such as shape-based representations, spectral analysis, and other feature-based transformations (see~\citep{Sariyanidi2015Automatic, zeng2009survey} for a detailed analysis), \ac{FER} approaches have now `evolved' into using more data-driven methods~\citep{Kollias2018IJCNN,Li2020Deep}. This has been particularly accelerated by the advances in Deep Learning research~\citep{LeCun2015, Schmidhuber15} that allow models to learn important information directly from the data, making the models much more robust to changes in environmental and background settings~\citep{Dhall2012,Kollias2018Deep, KOSSAIFI201723, Li2020Deep}. 

However, conventional deep learning methods, despite achieving \ac{SOTA} performance on benchmark evaluations, face \textit{three} major challenges. Firstly, the training pipeline of such models is constructed around using large benchmark datasets to train the models offline, in isolation, for later application in real-world situations, with little to no \textit{adaptation}~\citep{Churamani2020CLAC, Churamani2020CL4AR, Schmidhuber15}. The learning depends on training the model on well-balanced mini-batches of \acf{i.i.d} data samples from stationary data distributions~\citep{SeffBSL17}. The real-world applications, however, require learning models to \textit{continually} encounter new information and learn and adapt to such dynamics. Furthermore, as data is only made available incrementally, and in most cases, sequentially~\citep{Parisi2020OCL}, these models are not able to adapt their learning to rapidly changing data distributions~\citep{Parisi2018b}. Incrementally exposed to information pertaining to one task or class at a time, learning is adapted to perfect only this task, overwriting any previous knowledge~\citep{kemker2018measuring}. This also leads to the second challenge, that is, \textit{interference}, where the newly acquired information interferes with the previously learnt knowledge, which may not be desirable~\citep{LopezPaz2017GEM}. Even if the new task is compatible with previously learnt information, incrementally receiving data samples only pertaining to this new task gradually overwrites previous knowledge, deteriorating the performance of the model and causing \textit{catastrophic forgetting}~\citep{McClelland1995,MCCLOSKEY1989109}. Finally, as the third challenge, these models may experience \textit{capacity saturation} where, as they acquire more data, and adapt to this new information, their overall capacity to represent and preserve knowledge saturates~\citep{Sodhani2020Toward}. This can result from the complexity of the model not being enough to retain information or the learnt feature representations not able to adequately distinguish between learnt tasks~\citep{Parisi2018a}. Since the hyper-parameters of the models are decided a priori, and in most cases, cannot be changed later, this results in the model being unable to accommodate the increasing number and changing complexity of the tasks to learn.


Addressing the above-mentioned challenges faced by \ac{ML} models, particularly when it comes to learning online in real-world situations, \textit{Lifelong} or \textit{\acf{CL}} research~\citep{HASSABIS2017245,Parisi2018b, THRUN199525, Hadsell2020CL} focuses on developing capabilities that allow models to \textit{continually} learn and adapt, acquiring new information while retaining previously learnt knowledge. Such online adaptation capabilities are essential for systems that interact with uncertain and changing environments, such as long-term, repeated interactions with humans, as they are expected to consistently learn new information and concepts while, at the same time, remembering past interactions with the users. Starting with a very limited understanding, out-of-the-box, \ac{CL}-based learning models can be sensitive to individual user characteristics such as culture, gender, personality~\citep{Churamani2020CL4AR,Churamani2022DICL4Bias} or expressivity~\citep{pmlrbarros19a,Churamani2020CLIFER} and adapt over continued interactions. Much like humans, \textit{continually learning} about the users and their behaviours, and \textit{dynamically adapting} to changing interaction conditions become the desiderata from such affective learning systems~\citep{Churamani2020CL4AR}. Recent advances in \ac{CL}~\citep{LESORT2020CL4R, Parisi2018b} research focus on this very challenge of embedding lifelong learning and adaptability in \ac{ML} systems by balancing novel learning on incrementally acquired streams of data, with the preservation of past knowledge.

In this work, we discuss the specific challenges faced by \acf{FER} models, particularly when it comes to real-world applications. We discuss the current state-of-the-art and highlight how current approaches for \ac{FER} may be insufficient in realising real-time adaptation especially with respect to their use in systems that interact with users. As a solution to this problem, we propose, formalise and evaluate the use of \ac{CL} as a learning paradigm for training \ac{FER} models that can learn with streams of data, online, while preserving previously seen information. To summarise, the main contributions of this work are as follows:

\begin{enumerate}
    \itemsep-0.2em
    \item We present a novel application of \ac{CL} for \ac{FER}, under Task- (\acs{Task-IL}) and \acf{Class-IL} settings, discussing the challenges addressed by \ac{CL} for realising real-world application of \ac{FER} and present a discussion on how \ac{CL} can be leveraged for developing systems that can incrementally learn to recognise human facial expressions.
 
    \item We then present \ac{ConFER} as the first comprehensive benchmark evaluating \ac{CL} as a learning paradigm for \ac{FER} evaluating $11$ \ac{CL} strategies, across \acs{Task-IL} and \acs{Class-IL} settings, on $3$ popular \ac{FER} benchmarks\footnote{Additional experiments with \acs{iCV-MEFED}~\citep{Guo2018icv}, \acs{FER}- 2013~\citep{Goodfellow2013Challenges} and BAUM-1~\citep{ZhalehpourBaum2017} datasets can be found in Appendix~\ref{app:additional}.}: \acs{CK+}~\citep{lucey2010extended}, \acs{RAF-DB}~\citep{li2017reliable}, 
    and AffectNet~\citep{2018Affectnet}, covering a variety of background and data recording settings. 
    
\end{enumerate}

The rest of the paper is structured as follows: Section~\ref{sec:fer} discusses the challenges faced by state-of-the-art deep learning approaches for \ac{FER} in real-world settings and presents an overview of popular learning strategies used to remedy these challenges. Section~\ref{sec:cl} provides an overview on \acf{CL}, the existing state-of-the-art and challenges with applied \ac{CL}. Section~\ref{sec:confer} then formalises \acf{ConFER}, underlining on how \ac{CL} can be leveraged for \ac{FER} under Task-(\acs{Task-IL}) and \ac{Class-IL} settings and briefly describes the different strategies compared in this paper to evaluate their application for \ac{FER} under the different learning settings. It provides the details for the experiment set-up and other implementation details while Section~\ref{sec:conferresults} presents the results from the different experiments. Section~\ref{sec:discussion} provides an overall discussion on the experimental evaluation conducted and provides insights for the successful application of \ac{CL} for \ac{FER}. Finally, Section~\ref{sec:conclusion} summarises the learning from this work and provides a discussion on the limitations and future directions for \ac{ConFER}.

\section{Facial Expression Recognition: Challenges for Online Learning}
\label{sec:fer}

\acf{FER} is one of the most popular and straightforward tools used for evaluating human affective behaviour. Analysing the contraction and relaxation of facial muscles can provide insights into the affective state expressed by the user. With deep learning approaches becoming more and more popular, \ac{FER} methods have achieved near-perfect performance on benchmark evaluations~\citep{Li2020Deep}. 
Human affective interactions, however, are diverse, complex and unique with respect to each interaction context and individual. As a result, the high performance of \ac{FER} models on benchmark evaluations is not translated to real-world applications of such systems.
As a result, \ac{AC} research, over the last decade or so, has shifted its focus towards developing techniques that are able to recognise expressions \textit{in-the-wild}~\citep{Dhall2014, KOSSAIFI201723, Li2020Deep, Zafeiriou2017AFF}, robust to movements of the observed person, noisy environments and occlusions~\citep{Zen2016Learning}. These benchmarks represent varying background and recording conditions, recording an individual's expressions in a spontaneous and naturalistic manner which offers a much harder problem for \ac{FER} algorithms to solve as they need to generalise their learning across different contextual settings as well as varying lighting and background conditions, robust to the high variation in data. Yet, despite pushing \ac{FER} models towards more naturalistic and real-world applications, the training and evaluation paradigm for \ac{FER} models, albeit \textit{in-the-wild}, still does not support real-time adaptation. The models now rely on even larger amounts of data, representing high variation in recording conditions, still to be trained off-line and later applied to real-world settings with little-to-no adaptation capabilities~\citep{Churamani2020CLAC}. Furthermore, as recording and data collection are not controlled, such datasets may suffer from imbalanced data distributions, not just with respect to different class labels but also with respect to several demographic attributes such as gender, race and age resulting in biases affecting model performance for under-represented demographic groups~\citep{buolamwini2018gender, Cheong2021Hitchhikers, Churamani2022DICL4Bias, Li2020Deeper}.  

Real-world \ac{FER} applications are presented with two main challenges. Firstly, models need to learn novel information such as learning new expression classes or \textit{affective concepts}~\citep{scherer2000psychological} expressed by individuals as well as the variation within those expression classes. Secondly, to personalise this learning towards each individual by being sensitive to individual attributes of gender, race, personality or even expressivity~\citep{pmlrbarros19a, Chu2017Selective, Churamani2020CLIFER}. Such adaptive systems may enable an improved understanding of human affective behaviour, enhancing the overall user experience of these systems~\citep{Rudoviceaao6760, Churamani2020CL4AR}. 

To address some of the above-mentioned challenges vis-\`a-vis continual adaptation, several different learning strategies have been proposed that may allow models to `progressively' learn and adapt \textit{on-the-fly}. Here, we discuss some of these strategies highlighting the challenges addressed by each of these as well as discussing their limitations.
\vspace{-1mm}

\subsection{Transfer Learning}
\acf{TL} focuses on the challenge of re-using existing knowledge from one domain or task towards solving another task, without having to completely retrain the model~\citep{Bengio2013Representation, Silver2011IUT}. Most transfer learning-based \ac{FER} approaches focus on re-using high-level feature extractors trained for a different task such as image recognition~\citep{Chatfield2014Return, Alex2012, simonyan2014very} or face recognition~\citep{Cao2018VGGFace2,Parkhi15} and applying these for \ac{FER} by finetuning the top layers to learn expression classes~\citep{Aly2019FER, Barros2019Exploring, KAYA201766, Rescigno2020Personalized, Xu2016Video}. 
Despite reusing and \textit{transferring} previously learnt knowledge to novel tasks, transfer learning models do not particularly guard against \textit{catastrophic forgetting}. As the model is \textit{fine-tuned} towards a new task, it may forget previous tasks and would need to be trained again. This is witnessed particularly in the case of Zero-shot~\citep{Lu2019Zero, Wang2019SZL}, One-shot~\citep{Cruz2014One,Maddula2021Emotion, Vinyals2016MNO} or Few-shot~\citep{ciubotaru2019revisiting} learning approaches that reduce dependency on the availability of labelled data to learn new tasks but do not prevent catastrophic forgetting. 

One solution to this problem is to create a copy of the model for each new task retaining the feature-extraction layers and fine-tune this copy alone towards solving the new task. This would create an \textit{ensemble} of different models, each contributing towards solving one particular task~\citep{Coop2013Ensemble, REN2017398,SBMW18}. Yet, despite restricting the size of the models, there is redundancy as the same model maybe copied, albeit fine-tuned individually, a number of times. A solution to this problem is offered by \textit{progressive networks}~\citep{Rusu2016ProgressiveNN} that allocate additional neural resources to handle a new task, preserving old knowledge. For each task, the model adds another branch to the output model which is trained on the new task~\citep{Barros2019Exploring}. These branches share common low-level features, allowing for reuse of domain knowledge while high-level and task-dependent features are learnt and preserved in the different branches of the model. Despite successfully re-using and adapting models to solve different tasks, their ability to guard against catastrophic forgetting is sub-optimal as they are optimised only for successful \textit{forward transfer} of information~\citep{Pan2010TL}. Even with ensemble methods and progressive networks offering positive examples, choosing the right strategy towards re-using pre-trained models and the constraints on the scalability of these models still remains largely unsolved, limiting their ability to adapt in real-time~\citep{Parisi2018b}. 

\vspace{-1mm}
\subsection{Multi-task Learning}
Another popular strategy employed to tackle the problem of learning several tasks, jointly at once~\citep{Han2018DMTL, Kendall2018MTL} or in succession~\citep{CHEN2021MTL, Honari2018MTL}, is \acf{MTL}. \ac{MTL} aims to improve the performance and learning efficiency of models by introducing multiple objectives, one or more for each task, and jointly optimising model using shared feature representations between the tasks while maintaining individual task-specific output layers~\citep{Kendall2018MTL}. Similar to \acs{TL}, \ac{MTL} aims to exploit the relationships between the tasks to learn robust feature representations that positively contribute towards these tasks~\citep{Han2018DMTL}, yet it focuses on jointly optimising the model on multiple objectives rather than sequentially transferring knowledge from previously learnt tasks to novel ones. This makes it difficult to train the models, often resulting in sub-optimal generalisation capabilities~\citep{Kokkinos_2017_CVPR}. 

Deep \ac{MTL} models focusing on facial analysis combine several related tasks such as facial landmark detection~\citep{CHEN2021MTL, Devries2014MTL, Hu2018DeepMTL}, facial attribute detection~\citep{Han2018DMTL, Zhang2016Learning} or learning demographic attributes~\citep{Han2015Demographic, Han2018DMTL}. Even though this facilitates learning robust feature representations, all tasks still need to be known apriori with annotations for each of the tasks to be solved. Furthermore, despite the focus on enabling an efficient \textit{transfer} of knowledge between the tasks, several tasks are learnt at the same time instead of incrementally with no protection against \textit{forgetting} of past knowledge.

While some \ac{MTL} approaches have focused on sequentially multi-task learning~\citep{CHEN2021MTL,Honari2018MTL}, applying the learning from one task to enhance the models ability to learn the other task using a \textit{residual} learning module, they assume ground-truth knowledge for all the tasks are available for each individual data sample a priori. This does not allow models to adapt with incrementally acquired information, making real-world adaptation arduous. Furthermore, predictions from one task are used to guide learning for the subsequent ones~\citep{Honari2018MTL}, assuming a direct hierarchy or inter-relationship between the learnt tasks which may not always exist in real-world applications.

\vspace{-1mm}
\subsection{Curriculum Learning}
Curriculum Learning proposes to enhance model learning by incrementally increasing the complexity of the tasks learnt. Simple multi-stage learning strategies can improve the generalisation capabilities of the model and lead to faster convergence on complex tasks~\citep{Bengio2009CL}. These approaches take inspiration from how curriculum-based learning is applied for humans to incrementally build complex reasoning and problem-solving capabilities~\citep{ELMAN199371}. However, the performance of these models depends heavily on the choice of the curricula used. To incrementally build task-solving capabilities, it is important to define which tasks the model learns first and what order is followed. \citet{GravesBMMK17} suggest following a difficulty-based ranking mechanism, determined by a stochastic policy, to order the tasks.

Inspired by these methods, several curriculum-learning based approaches have been proposed for \ac{FER}. \citet{Gui2017Curriculum} define facial expressions with high intensities as \textit{simpler} to classify than expressions with low intensities and use this assumption to structure a curriculum for their model by splitting the dataset into three tasks (high, medium and low intensity) and train their model iteratively learning \textit{easy} to \textit{hard} tasks. \citet{Liu2019Improved} employ a density distance clustering algorithm to determine cluster centres for each of the categories and complexity (simple, hard or complex) of the samples, for each category, is determined using the distance from these cluster centres such that the further away the sample, higher the complexity. In practice, this may be similar to \acs{TL}~\citep{Parisi2018b, Weiss2016} where learning from one task, in this case the \textit{simpler} tasks, informs learning on subsequent \textit{harder} tasks, following a particular curriculum. Yet, much like \acs{TL}, there are no active mechanisms to prevent against \textit{catastrophic forgetting}. Furthermore, despite curriculum learning based methods allowing for models to incrementally learn novel tasks, these still require complete knowledge of all the tasks in order to rank them, a priori, on a liner \textit{difficulty axis}~\citep{GravesBMMK17}. This becomes problematic during online learning settings as the tasks may not be known in advance and, depending upon the application context, it may not always be possible to \textit{rank} the tasks following the same complexity criteria. 

\section{Continual Learning: An Overview}
\label{sec:cl}

Conventional (deep) \ac{ML} approaches work on a pivotal assumption that all data samples for the tasks to be learnt by the models are available apriori, that is, there is a clearly separated \textit{training phase} for the models with training data samples drawn from relatively stationary data-distributions. The real-world, however, is not stationary~\citep{Hadsell2020CL} and changes continuously. As a result, learning systems continuously, and more importantly, sequentially, encounter novel information and tasks, requiring them to adapt, not only integrating this new learning \textit{on-the-fly}, but also ensuring that past knowledge is not forgotten~\citep{Parisi2018b}. Such sequential learning in real-world settings violates the quintessential \ac{ML} assumption of training samples being \acf{i.i.d} to train the model. As a result, any adaptation or integration of new information is achieved at the cost of previous knowledge being forgotten or overwritten, leading to \textit{catastrophic forgetting}~\citep{McClelland1995,MCCLOSKEY1989109}.

\textit{Lifelong} or \textit{\acf{CL}} research~\citep{HASSABIS2017245,Parisi2018b,THRUN199525, Hadsell2020CL} aims to address this very problem of continuous adaptability in agents, enabling them to continually learn and adapt throughout their lifetime, balancing learning of novel information with the retention of previously acquired knowledge. Distinct from \acs{TL} or \ac{MTL} where the emphasis is on reusing previously acquired knowledge to assist in learning new tasks or curriculum learning where all the tasks to be learnt are known apriori and a curriculum can be carefully constructed, \ac{CL} focuses on learning with continuous and sequential streams of data acquired from non-stationary or changing environments.

\subsection{Background}

Contrary to computational learning systems, humans are seen to learn throughout their lifetime, acquiring and integrating new information without affecting previous knowledge~\citep{Power2017Neural}. Different areas of the brain are seen to be responsible for balancing two complementary tasks; maintain high plasticity levels in order to learn novel information while, at the same time, balance this new learning with previous knowledge to stabilise neural adaptation~\citep{KUMARAN2016512, McClelland1995}. This conundrum around learning to regulate neural plasticity levels to balance learning is termed as the \textit{stability-plasticity} dilemma~\citep{Mermillod2013}. The \acf{CLS} theory~\citep{McClelland1995} posits that the \textit{hippocampal} and \textit{neocortical} areas of the brain form a complementary learning system encoding information at different speeds and temporal resolutions~\citep{OReilly2014CLS}. The hippocampus learns statistical regularities and specifics, forming an \textit{episodic memory} of novel and arbitrary information. This information is stored and replayed for the \textit{slow} learning of structured \textit{semantic} knowledge in the neocortex. 
Furthermore, the \ac{PFC} also contributes towards the consolidation of specific memories into long-term understanding and is responsible for selective memory recall~\citep{Kitamura73}. 

Investigating neuro-physiological mechanisms that regulate synaptic plasticity~\citep{Power2017Neural, Wixted2004} in the human brain has lead to the development of several \ac{CL} methods (see \citet{Parisi2018b}) which demonstrate similar, albeit simplified, learning mechanisms for artificial agents. As we improve our understanding of the human brain, and the different underlying learning mechanisms, their translation to artificial agents will help us enhance their performance as well~\citep{Hadsell2020CL, HASSABIS2017245}.

\vspace{-1mm}
\subsubsection{\acs{CL} Problem Formulation}
Incrementally learning novel tasks involves agents to encounter observations from a dynamic and infinite sequence of relatively stationary data distributions. At any given time, the agent observes a data distribution $D_i$ and samples a training set $Tr_i$ from this distribution to learn a target function ($h^{*}(x, t)$) that solves a novel task $t$. \citet{LESORT2020CL4R} formulate a \ac{CL} algorithm $A_i^{CL}$ to learn a general (target) model $h^*$ as follows:

\begin{equation}
    \forall D_i \in \mathcal{D},\mspace{8mu} A_i^{CL}:\mspace{2mu} 
    \langle h_{i-1}, Tr_i, M_{i-1}, t_i\rangle \xrightarrow{}\langle h_i,M_i\rangle,
    \label{eq:CL_1}
\end{equation}

{\noindent}where $h_i$ is the current model hypothesis, $M_i$ is the memory storing training samples up to time-step $i$; $t_i$ is the current task label; $Tr_i$ is the training set of examples $e_j^i =\langle x_j^i,y_j^i\rangle$ with $j\in[1\dots,m]$ drawn from the current stationary data distribution $D_i$. For each $Tr_i$, $A_i^{CL}$ adapts its model hypothesis acquiring this new information ($h_{i-1}\xrightarrow{}h_i$), as well as updates its memory to represent past learning ($M_{i-1}\xrightarrow{}M_i$). 

As the agent interacts with its environment, it may encounter a sequence of experiences ($\mathcal{D}=\{D_1,\dots, D_N\}$) requiring the model to learn to solve novel tasks while ensuring that previous knowledge is not overwritten with this new learning. 

\vspace{-1mm}
\subsubsection{Learning Scenarios and Types}
\label{sec:cl-scenarios}

\begin{figure}[t!]
    \centering
    \includegraphics[width=0.8\textwidth]{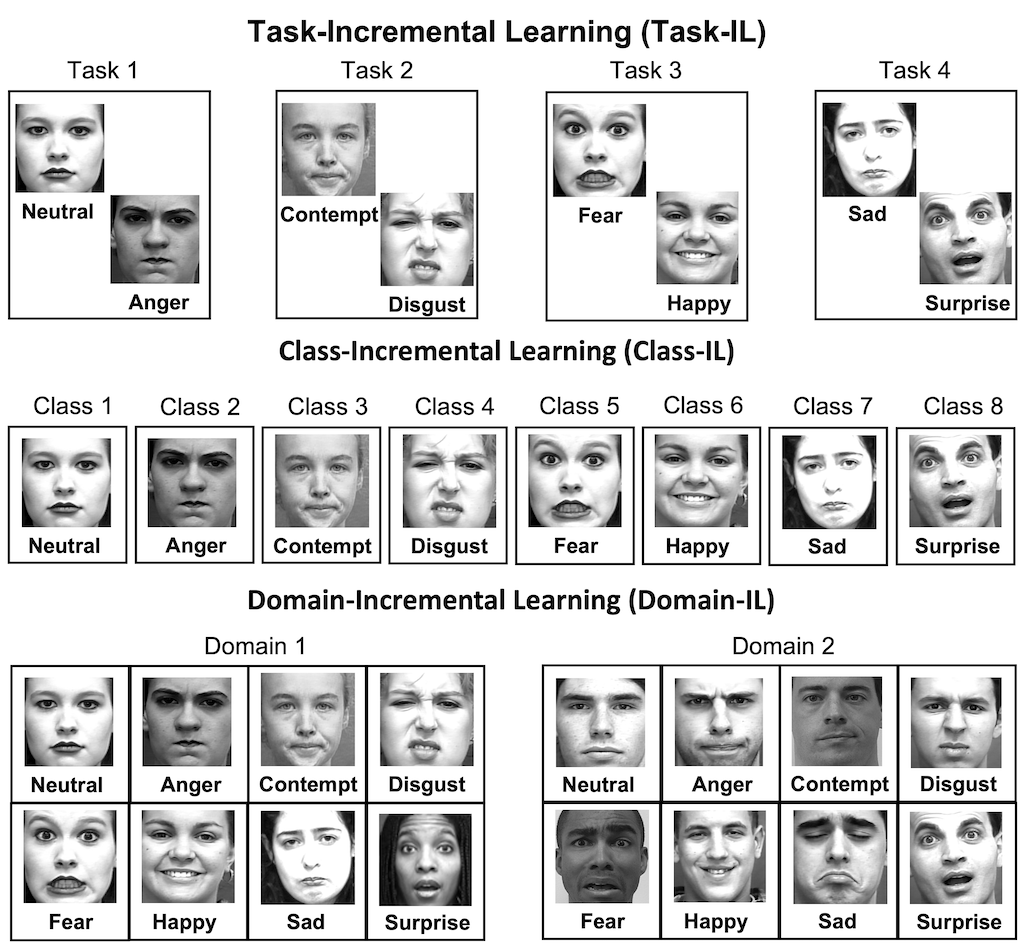}
    \caption[Task boundaries for \acs{Task-IL} \acs{Class-IL}, and \acs{Domain-IL} Continual Learning.]{Task boundaries for \acs{Task-IL} (top), \acs{Class-IL} (middle) and \acs{Domain-IL} (bottom) scenarios for \acf{FER}. Adapted from~\citep{van2019three} with images from the \acs{CK+}~\citep{lucey2010extended} dataset.}
    \label{fig:conferscenarios}
\end{figure}

\ac{CL}-based approaches focus on models incrementally acquiring information about new tasks, while trying to retain performance on previously seen tasks. As the model continues to learn, it may be exposed to data pertaining to only one class or multiple classes at a time during training. Based on what information the model might have available and what it needs to learn, that is, whether the model has the information about which task or class it is witnessing or if it needs to infer this information during learning, \citet{van2019three} formalised three different \textit{learning scenarios} for \ac{CL} approaches. These scenarios are illustrated in Figure~\ref{fig:conferscenarios} in the context of \ac{FER} where the model needs to learn different expression categories.

\begin{enumerate}
    \item [\bf a)] \textbf{\acf{Task-IL}:} where the model is incrementally presented with individual tasks consisting of learning one or more sub-tasks or classes while being explicitly provided information about the task identity.
    \item [\bf b)] \textbf{\acf{Domain-IL}:} where the task to be learnt does not change but the input data distribution changes and the model needs to manage such shifts in data distributions.
    \item [\bf c)] \textbf{\acf{Class-IL}:} where the model needs to incrementally learn new classes without being given any information on the task.

\end{enumerate}

Furthermore, based on the nature and availability of data to the model, \citet{MALTONI2019CL} provide another nomenclature for splitting the learning to three \textit{learning types}:

\begin{enumerate}
    \item [\bf a)] \textbf{\acf{NI}:} where the model receives samples from all the tasks in the first instance and subsequently receives novel instances of these already learnt tasks.
    
    \item [\bf b)] \textbf{\acf{NC}:} where the model incrementally receives samples only from a novel task or \textit{concept} and needs to integrate this novel learning.
    
    \item [\bf c)] \textbf{\acf{NIC}:} where the model receives samples from novel tasks to be learnt as well as from already learnt tasks with each sequential batch of input. 
\end{enumerate}
 
 Both \ac{Task-IL} and \ac{Class-IL} scenarios are compatible with the \ac{NC} learning type as the model needs to incrementally learn novel classes and tasks, while the \ac{Domain-IL} scenario is compatible with the \ac{NI} learning type as it requires the model to adapt its learning to novel instances of already seen tasks drawn from shifting data distributions~\citep{COSSU2021CL}.

\vspace{-1mm}
\subsection{The State-of-the-art}
\ac{CL}-based methods employ (deep) \ac{ML} models with learning capabilities that can integrate novel information while preserving past knowledge. This may be achieved by \textit{regulating} model updates to control model plasticity, storing and \textit{replaying} already seen information to simulate \ac{i.i.d} learning settings or dynamically \textit{expanding} neural resources for the model to accommodate novel information without interfering with past knowledge. 


\vspace{-1mm}
\subsubsection{Regularisation-based Methods}
Regularisation is a popular technique for neural models~\citep{Schmidhuber15} that adds upper-bounds on weight-updates while training the network to guard against over-fitting. Additional loss terms are applied, penalising large changes in the weights during training. For \ac{CL}, regularisation can reduce \textit{destructive interference} from the newly learnt tasks. Different approaches achieve this by either freezing parts of the model that correspond to previously learnt information~\citep{Fernando2017PathNet} and updating only newly added parameters~\citep{Razavian2014CVPRW}, penalising weight updates that deteriorate performance at previously learnt tasks~\citep{Li2018LWF} or prioritising weight-updates for different parameters based on their relevance to different tasks~\citep{kirkpatrick2017overcoming,zenke2017continual}.  

Thus, by constraining weight-updates of the model (or parts of it), previous knowledge can be preserved, avoiding catastrophic changes to the learnt parameters. Despite the competitive performance of regularisation approaches, they become computationally expensive as the number of tasks grow, limiting their ability to learn new tasks incrementally in an online manner~\citep{kemker2018measuring}. 




\vspace{-1mm}
\subsubsection{Replay-based Methods}
Under the \ac{CL} paradigm, whenever a new task is encountered, conventional \ac{ML} models suffer from \textit{catastrophic forgetting} as new learning may overwrite previously acquired knowledge. This problem does not occur in \textit{offline} settings where all the data is available to the model at all times. Thus, to mitigate catastrophic forgetting, the most straightforward approach can be to physically store the encountered data samples in a memory buffer and regularly replay this data,  interleaving it with newly encountered samples. Such a \textit{rehearsal} mechanism~\citep{Robins1993Rehearsal} allows the model to replicate offline i.i.d settings as the model is trained on mixed batches of data consisting of samples from all tasks learnt so far~\citep{Hsu18_EvalCL}. Although this works for a small number of tasks~\citep{Gepperth2016,rebuffi2017icarl}, it does not scale well as the number of tasks increase. Furthermore, in the case of high-dimensional data samples (for example, images and videos) corresponding to a large number of tasks, it becomes computationally and memory-wise intractable to physically store and replay training samples. Some of the more recent methods aim to address these issues by proposing improved rehearsal mechanisms that either impose memory budgets on the storage samples from previously seen task~\citep{LopezPaz2017GEM} or reduce the memory load by \textit{rehearsing} low-dimensional feature representations extracted from the input data in the form of activations of a given \textit{latent replay layer}~\citep{Pellegrini2020Latent} instead of replaying high-dimensional input data patterns. 

As an alternative to maintaining memory buffers, a generative or probabilistic model can be used to learn data statistics to draw \textit{pseudo-samples}~\citep{Robins1995} from the memory, reducing the cost of these models significantly~\citep{Churamani2020CLIFER, Nguyen2018variational, Rolnick2018Experienced, SeffBSL17,NIPS2017_6892,vandeVen2018CL, Wu2018MRG}. As actual samples of the data are not needed to be stored, such a \textit{pseudo-rehearsal} of information provides a viable and scalable solution to alleviate catastrophic forgetting. Yet, as the number of tasks increase, it becomes progressively harder to train the generative or probabilistic model that can be used to generate and replay high-dimensional data samples from all the seen tasks. Furthermore, even though \textit{pseudo-rehearsal} significantly reduces the memory consumption, it has the added computational overload of training and maintaining a generative model. Recent methods~\citep{Stoychev2023LGRFG, vandeVen2020Brain} have tried to address this problem by combining latent replay~\citep{Pellegrini2020Latent} and deep generative replay~\citep{NIPS2017_6892} by using a generative model that learns to reconstruct and replay low-dimensional latent representations instead of high-dimensional input data to further reduce the memory and computational footprint of them models.

\vspace{-1mm}
\subsubsection{Dynamic Architectures}
The focus of regularisation and replay-based approaches is on regulating learning in the existing model architecture to accommodate new information, while, at the same time alleviating catastrophic forgetting. However, as the complexity of the data and the tasks increase, these strategies are not able to scale up as a result of \textit{capacity saturation} in these models~\citep{kirkpatrick2017overcoming}, leading to newly learnt information interfering with previous knowledge. This may result from reduced plasticity in model parameters as more and more weights are `\textit{frozen}' to reduce negative interference on previously learnt tasks, or the samples stored for each task in the memory buffer being not representative enough to preserve previous knowledge, or the generative model facing capacity saturation as well, not being able to reconstruct and replay data samples from previously seen tasks, efficiently. 

To alleviate this problem, additional neural resources may need to be allocated to these models, either by expanding the trainable parameters~\citep{Draelos2017NDL,yoon2018lifelong}, or allowing the model architecture itself to grow and expand~\citep{PARISI2017Lifelong,Part2017SOINN,Xiao2014EIL,zhou12b} to account for the increased complexity of the task. Starting with a relatively simple architecture, the model may need to be extended by allocating additional neurons~\citep{PARISI2017Lifelong,Part2017SOINN, zhou12b} or even complete network layers and branches for the model~\citep{Fernando2017PathNet,Rusu2016ProgressiveNN} as and when the model is not able to scale to a new task. This growth can be regulated using different strategies, such as performance on tasks~\citep{zhou12b}, neural activations in response to data samples~\citep{PARISI2017Lifelong} or evaluating the contribution of existing parameters towards solving new tasks~\citep{schwarz2018progress}. Despite the additional overhead of adding new neural resources and the inherent computational expense, these models are shown to work well towards mitigating catastrophic forgetting, enabling continual learning of information~\citep{Parisi2018b}.

\vspace{-1mm}
\subsubsection{Neuro-Inspired and Complementary Learning-based Approaches}
An enhanced understanding of the different learning processes in the brain has inspired a variety of current research in \ac{CL}~\citep{Kudithipudi2022} where different methods attempt to emulate the complementary learning of novel experiences along with the long-term retention of knowledge, as posited by the \ac{CLS} theory~\citep{McClelland1995,MCCLOSKEY1989109}. To alleviate forgetting, learning is implemented over multiple memory models to represent the different areas of the human brain, each of which is tuned to learn at different speeds and temporal resolutions. A hippocampal \textit{episodic memory} is implemented~\citep{PARISI2017Lifelong, LopezPaz2017GEM, chen2020bypassing, dAutume2019MbPA, Gepperth2016, Churamani2020CLIFER} to represent active learning of novel information encoding the current sensory experiences into \textit{pattern-separated}, non-overlapping representations. The neocortical \textit{semantic memory}, on the other hand, focuses on long-term retention of information~\citep{PARISI2017Lifelong} by slowly replaying the experiences encoded in the episodic memory to learn \textit{pattern-complete}, overlapping representations that can generalise across multiple samples for a particular task~\citep{ROLLS20164}. Furthermore, a \acf{PFC}-based model~\citep{Kamra2017DeepGD,Kemker2018FearNetBM} may also be applied that may be used for long-term retention of previously seen knowledge. A replay of experiences can be realised using generative or probabilistic models~\citep{Churamani2020CLIFER,Kamra2017DeepGD,Kemker2018FearNetBM, NIPS2017_6892} that enable a transfer of experiences from the \textit{episodic} to the \textit{semantic} memory or the \acs{PFC} via \textit{pseudo-rehearsal}, facilitating long-term preservation of knowledge.


As our understanding of neuro-cognitive mechanisms of learning enhances, the ability of such \ac{CLS}-based multi-memory systems to continually learn and adapt to information will be augmented, resulting in improved performances for artificial agents in uncertain and unpredictable environments~\citep{Hadsell2020CL, Parisi2018b}.

\section{The \acf{ConFER} Benchmark}
\label{sec:confer}

Addressing the challenges faced by conventional (deep) \ac{ML}-based \ac{FER} models, particularly when it comes to online learning (see Section~\ref{sec:fer}), \ac{CL}-based methods can be leveraged to allow models to \textit{continually} learn and adapt with new information while retaining previously learnt knowledge. Such adaptation is essential for systems that operate under uncertain and changing contexts of long-term and sustained interactions with humans. 
This paper 
evaluates, the use of \ac{CL} as a learning paradigm for training \ac{FER} models that can learn with streams of data, online, while preserving previously seen information. Benchmark results are presented comparing the \ac{SOTA} in \ac{CL} on $3$ different \ac{FER} datasets\footnote{Additional experiments with \acs{iCV-MEFED}~\citep{Guo2018icv}, \acs{FER}- 2013~\citep{Goodfellow2013Challenges} and BAUM-1~\citep{ZhalehpourBaum2017} datasets can be found in Appendix~\ref{app:additional}.} to evaluate their ability to continually learn $8$ facial expression classes. These include the $6$ `universal emotions' proposed by \citet{Ekman1971Constants}, namely, \textit{Surprise, Fearful, Disgust, Happiness, Sadness, Anger} along with two additional classes, namely, \textit{Contempt} and \textit{Neutral}, under different learning settings. This allows the first comprehensive evaluation of \ac{CL} for \ac{FER}, across different learning settings, laying the foundations for novel inquiry into \acf{ConFER}.

\subsection{Learning Settings}
\label{sec:learningsettings}
Interacting with humans, agents may learn about facial expressions, one or more categories at a time or may learn with different individuals or groups under varying contextual settings. The \ac{ConFER} benchmark explores Task- (\acs{Task-IL}) and \acf{Class-IL} settings where popular \ac{CL}-based methods are evaluated on their ability to incrementally learn facial expression categories, one or more classes at a time. \acf{Domain-IL} settings~\citep{van2019three}, on the other hand, may help evaluate how agents may translate their learning of different facial expression categories from one domain-group to the other, which was explored extensively in our previous works~\citep{Kara2021Towards,Churamani2022DICL4Bias}. 


\subsubsection{Task-Incremental Learning for \acs{ConFER}}
Under \ac{Task-IL} settings, learning is split into a sequence of \textit{tasks}, where at any given time, the model receives data for only one task. Each \textit{task} may consist of learning one or more distinct \textit{classes} with the data for this task $\bm{\langle\mathcal{X}^{(t)},\mathcal{Y}^{(t)}\rangle}$ being randomly drawn from the training data distribution $\bm{\mathcal{D}^{(t)}}$. The models, at all times, are aware of the task they need to learn, that is, task-labels are always available. Algorithm~\ref{algo:task-il} outlines the training procedure for a learning model $\bm{M}$, on $\bm{T}$ tasks under \ac{Task-IL} settings. The $8$ expression classes are split into $4$ tasks consisting of $2$ classes each. After each task, average performance metrics (see Section~\ref{sec:confermetrics}) are computed using the test-set for \textit{all} the tasks seen so far. For example, after task $\bm{t_n}$, performance metrics are computed by averaging the results obtained using the test-data from tasks~$\bm{t_1 - t_{n}}$.

\begin{algorithm}[h!]
    \caption{\acf{Task-IL} for \acs{ConFER}}
    \label{algo:task-il}
    {\small
        \begin{algorithmic}[1]
            \State Partition data into $T$ \textit{tasks}, each consisting one or more classes. 
            \State Initialise model parameters ($\theta_{M}$).
            \For{$t=1$ to $T$}
                \State Train $M$ with ($\mathcal{X}^{(t)},\mathcal{Y}^{(t)}$) $\in \mathcal{D}^{(t)}$. {\footnotesize\texttt{// Data from only task $t$.}}
                \For{$i=1$ to $t$}
                    \State Compute performance metrics using test-set for task $i$.
                \EndFor
                \State Aggregate (average) performance metrics for all seen tasks. {\footnotesize\texttt{// Tasks $i$ to $t$.}}
            \EndFor
        \end{algorithmic}
    }
\end{algorithm}

\subsubsection{Class-Incremental Learning for \acs{ConFER}}
\label{sec:conferclassil}
\acf{Class-IL} settings can be considered to be a special case of \ac{Task-IL} where each task consists of data from exactly one class~($\bm{\langle\mathcal{X}^{(c)},\mathcal{Y}^{(c)}\rangle\in \mathcal{D}^{(c)}}$). \ac{Class-IL} is a complex scenario as the model needs to incrementally learn new classes without receiving any information about the class or task~\citep{van2019three}. Algorithm~\ref{algo:class-il} outlines the training procedure for a learning model $\bm{M}$, on $\bm{C}$ classes under \ac{Class-IL} settings.

\begin{algorithm}[h!]
    \caption{\acf{Class-IL} for \acs{ConFER}}
    \label{algo:class-il}
    {\small
        \begin{algorithmic}[1]
            \State Partition data into $C$ \textit{classes}, each consisting of data only from only class $c\in C$.
            \State Initialise model parameters ($\theta_M$).
            \For{$c=1$ to $C$}
                \State Train $M$ with ($\mathcal{X}^{(c)},\mathcal{Y}^{(c)}$) $\in \mathcal{D}^{(c)}$. {\footnotesize\texttt{// Data from only class $c$.}}
                \For{$i=1$ to $c$}
                    \State Calculate performance metrics using test-set for class $i$.
                \EndFor
                \State Aggregate (average) performance metrics for all seen classes. {\footnotesize\texttt{// Classes $i$ to $c$.}}
            \EndFor
        \end{algorithmic}
    }
\end{algorithm}
\noindent For \acs{Class-IL} evaluations, the models incrementally learn the $8$ expression classes one class at a time. After learning each new class, the average performance metrics are computed for all the classes seen so far.

\subsection{Compared Approaches}
\label{sec:comparedCL}
To understand the applicability of \ac{CL} towards \ac{FER}, several \ac{CL} approaches are compared on popular \ac{FER} benchmarks to evaluate how these methods fair when applied under \acs{Task-IL} and \acs{Class-IL} settings. These are as follows:

\subsubsection{Baselines}
Baseline evaluations enable a comparison of \ac{CL}-based methods with \textit{conventional} \ac{ML} strategies. Similar to \citep{van2019three, Delange2022CLSurvey} the following baselines are used:

\paragraph{\acf{LB}:} For the first baseline, the model is trained sequentially on all tasks, \textit{fine-tuning} with the data for only that task. This represents the \ac{LB} for the evaluation.

\paragraph{\acf{UB}:} For the second baseline, the model is trained using data from all the tasks seen so far. At any give time, the model has complete access to all previously seen data samples. This represents the \textit{joint-training} \ac{UB}.

    

\subsubsection{Regularisation-based Approaches}
\label{sec:conferregularisationmethods}
Regularisation-based methods impose constraints on weight-updates during training in a manner that new learning does not interfere with past knowledge. The following regularisation-based approaches are compared:

\paragraph{\acf{EWC}:} The \ac{EWC} approach, as proposed by~\citet{kirkpatrick2017overcoming} regularises model updates by imposing a quadratic penalty on parameter updates between old and new tasks. Changes in parameters that contribute the most towards old tasks are discouraged by penalising updates in these parameters when learning a new task. For each parameter $\bm{\theta}$, an importance value is computed using the task's training data $\bm{\mathcal{D}^{(t)}}$, modelled as the posterior distribution $p(\theta | \mathcal{D}^{(t)})$. 
Laplace approximation is used to approximate this as a \textit{Gaussian Distribution} with its mean given by parameters $\bm{\theta_{t}^*}$ (parameters of task $\bm{t}$) and the importance of the parameters determined by the diagonal of the Fischer Information Matrix. As a result, the loss function for \ac{EWC} becomes:
\begin{equation}
    L(\theta) = L_{t+1} (\theta) + \frac{1}{2} \lambda \sum_i F_i (\theta_i - \theta_{t,i}^*)^2,
\end{equation}
where $\bm{L_{t+1}}$ is the loss for task $\bm{t_{t+1}}$,  $\bm{\lambda}$ is the regularisation coefficient that assigns a relevance value to the old tasks with respect to the new one, $\bm{i}$ denotes the index of the parameter $\bm{\theta}$ and $\bm{F_i}$ is the $\bm{i^{th}}$ diagonal element of the Fischer Matrix. 

\paragraph{\ac{EWC}-Online:} One of the main disadvantages for the \ac{EWC} method is that as the number of tasks increases, the number of quadratic terms in the regularisation term grows. \citet{schwarz2018progress} propose a modification to \ac{EWC} where, instead of several quadratic terms, a single quadratic penalty is applied, determined by a \textit{running sum} of the Fischer Information Matrices of previous tasks. Thus, the updated regularisation term of the \ac{EWC}-Online approach is given as:
\begin{equation}
    L_{reg}^T = \sum_i \tilde{F}_{i}^{(T-1)} (\theta_i - \theta_{i}^{(T-1)})^2,
\end{equation}
where $\bm{\theta_{i}^{(T-1)}}$ is the $\bm{i^{th}}$ parameter after learning task $\bm{T-1}$ and $\bm{\tilde{F}_{i}^{(T-1)}}$ is the running sum of the diagonal elements of Fischer Matrices of all previous tasks, calculated as:
\begin{equation}
    \tilde{F}_{i}^{(T)} = \gamma  \tilde{F}_{i}^{(T-1)} + F_i^{T},
\end{equation}
where $\bm{\gamma}$ controls the contribution of previously learnt tasks.

\paragraph{\acf{SI}:} Much like \ac{EWC}, \ac{SI}~\citep{zenke2017continual} also regularises the model by penalising changes in relevant network parameters (or synapses) such that new tasks are learnt without forgetting the old. Alleviating \ac{CF}, for each synapse, its importance for solving a learnt task is computed and changes in most important synapses, that is, synapses that contribute the most to previously learnt tasks, are discouraged. The cost function $\bm{L_n^*}$ is used with a surrogate loss term which approximates the summed loss functions of all the previous tasks $\bm{L_o^*}$:
\begin{equation}
    L_n^* = L_n + c\sum_i \Omega_k^n (\theta_k^* - \theta_k)^2,
\end{equation}
where $\bm{\theta_k}$ represents the parameters for the new task,  $\bm{\theta_k^*}$ represents the parameters at the end of the previous task, $\bm{\Omega_k^n}$ is the parameter regulation strength and $\bm{c}$ is the weighting factor balancing new \vs old learning.

\paragraph{\acf{LwF}:} The \ac{LwF} approach proposed by \citet{Li2018LWF} maintains three sets of network parameters; shared parameters ($\bm{\theta_s}$) that are applied across all tasks, task-specific parameters ($\bm{\theta_o}$) for previously learnt (old) tasks, and task-specific parameters ($\bm{\theta_n}$) for the current task to be learnt. \ac{LwF} aims to optimise $\bm{\theta_s}$ and $\bm{\theta_n}$ on the new task, constraining the predictions on the new task's samples using $\bm{\theta_s}$ and $\bm{\theta_o}$ in order to learn parameters that work well on old and new tasks, using samples only from the new task. This way, the updates to $\bm{\theta_s}$ and $\bm{\theta_o}$ are also constrained in order to avoid forgetting. As a result, the \acs{LwF} computes model parameters as follows:
\begin{equation}
    \theta_{s}^{*}, \theta_{o}^{*}, \theta_{n}^{*} \gets \operatorname*{argmin}_{\hat{\theta}_s, \hat{\theta}_o, \hat{\theta}_n} \Bigl( \lambda_o \mathcal{L}_{old} (Y_o, \hat{Y}_{o}) + \mathcal{L}_{new} (Y_{n}, \hat{Y}_{n}) + \mathcal{R} (\hat{\theta}_s, \hat{\theta}_o, \hat{\theta}_n) \Bigr)
\end{equation}
where $\hat{Y}_o \equiv F(X_n, \hat{\theta}_s, \hat{\theta}_o)$ is the \textit{old task output}, $\hat{Y}_n \equiv F(X_n, \hat{\theta}_s, \hat{\theta}_n)$ is the \textit{new task output}, $\bm{\mathcal{L}_{new}}$ is a logistic loss for the new task, $\bm{\mathcal{L}_{old}}$ is a cross-entropy-based \textit{distillation} loss for the old tasks, $\mathcal{R}=0.0005$ is a regularisation term for weight-decay for \ac{SGD}-based optimisation and $\lambda_o=1$ balances the priority of preserving old task performance over the new task.

\subsubsection{Replay-based Approaches}
\label{sec:conferreplaymethods}

Replay-based methods aim to simulate \ac{i.i.d} data settings by either physically storing samples from previously seen tasks in a memory buffer (\textit{rehearsal}), or train a generative model to learn the inherent data statistics and draw pseudo-samples (\textit{pseudo-rehearsal}) for previously seen tasks. Previous task data is replayed to the model, interleaving it with novel data, to mitigate forgetting. The following replay-based approaches are compared:


\paragraph{\acf{NR}:} Different from regularisation-based approaches, the \ac{NR} approach~\citep{Hsu18_EvalCL} aims to approximate \textit{i.i.d} settings for learning by maintaining a replay buffer of size $\bm{B_{size}}$ to randomly store previously seen data samples. Samples from the replay buffer are replayed to the model constructing mini-batches of data using an equal number of samples from the new as well as previously seen data. This interleaving of old and new data ensures that past knowledge is not forgotten.

\paragraph{Averaged Gradient Episodic Memory (A-\acs{GEM}):} The \ac{GEM} approach by~\citet{LopezPaz2017GEM} aims to minimise forgetting in the model by employing an \textit{episodic memory} $\bm{\mathcal{M}}$ that stores a subset of observed samples for all known $\bm{T}$ tasks. For the model $\bm{f_\theta}$ parameterised by $\theta\in\mathbb{R}^p$, the loss for the memories from the $k^{th}$ task is given as:
\begin{equation}
    l(f_\theta,\mathcal{M}_k) = \frac{1}{|\mathcal{M}_k|} \sum_{(x_i,k,y_i) \in \mathcal{M}_k} l(f_\theta(x_i,k),y_i)
    \label{eq:gem}
\end{equation}
For any given task $\bm{t}$, the \ac{GEM} tries to minimise the loss for that task while treating the losses on the episodic memories of the previously seen $\bm{k}$ tasks (equation~\ref{eq:gem}) as \textit{inequality constraints}, avoiding increase in the losses but allowing for their decrease. This is given as:
\begin{equation}
    \operatorname*{minimise}_\theta \hspace{2mm} l(f_\theta,\mathcal{D}_t) \hspace{4mm} s.t \hspace{4mm} l(f_\theta, \mathcal{M}_k) \leq l(f_{\theta}^{t-1}, \mathcal{M}_k) \hspace{5mm}  \forall k<t
\end{equation}
where $\bm{f_{\theta}^{t-1}}$ is the model trained up to task $\bm{t-1}$. This allows a positive backward transfer, unlike other approaches like the \ac{EWC}. However, this requires considerably more memory than regularisation approaches as an episodic memory $\bm{\mathcal{M}_k}$ is required for each task $\bm{k}$ that enables the models to successfully mitigate forgetting. \citet{chaudhry2019efficient} propose an improvement over \ac{GEM} by introducing the Averaged-\acs{GEM} that ensures that ``at every training step, the \textit{average} episodic memory loss over the previous tasks does not increase''. This is different from \ac{GEM} that evaluates the loss for each seen task individually. 

The training objective of A-\acs{GEM} now becomes:
 \begin{equation}
    \operatorname*{minimise}_\theta \hspace{2mm} l(f_\theta,\mathcal{D}_t) \hspace{4mm} s.t \hspace{4mm} l(f_\theta, \mathcal{M}) \leq l(f_{\theta}^{t-1}, \mathcal{M}) \hspace{4mm}  where \hspace{4mm} \mathcal{M}=\cup_{k<t} \mathcal{M}_k
\end{equation}
    
\paragraph{\acf{LR}:} Storing and replaying samples from previously seen tasks have shown to greatly improve the ability of models to avoid forgetting, albeit incurring a high memory cost. To balance this, \citet{Pellegrini2020Latent} proposed the \acf{LR} strategy which, instead of maintaining copies of input data patterns, stores extracted feature representations in the form of activations of a given \textit{latent replay layer} in a memory buffer of size $\bm{B_{size}}$. Freezing the lower layers (below the latent replay layer) and rehearsing the compact latent activations offers a significant reduction in the memory and computation cost of the model without significantly compromising on model performance.

\paragraph{\acf{DGR}:} The \ac{DGR} approach~\citep{NIPS2017_6892} presents a \textit{pseudo-rehearsal} strategy where instead of explicitly storing samples from previously seen tasks, a generative model is used to generate synthetic samples representative of previously learnt tasks. The \acs{DGR} approach is centred around a \textit{scholar} ($\bm{H=\langle G,S\rangle}$) that consists of two parts; a \textit{generator} ($\bm{G}$) that generates pseudo-samples for previously seen tasks and a \textit{solver} ($\bm{S})$ that sequentially learns to solve the task at hand. A new scholar model is maintained for each task, using the learning of the previous scholar (on the previous task) and the new data for the next task. The learning for the scholar is given as:
\begin{equation}
    L (\theta_i) = r\mathbb{E}_{(x,y)\sim\mathcal{D}_i}[L(S(\bm{x};\theta_i),\bm{y})] + (1-r)\mathbb{E}_{\bm{x'}\sim G_{i-1}}[L(S(\bm{x'};\theta_i),S(\bm{x'};\theta_{i-1}))]
\end{equation}
where $\bm{\theta_i}$ are model parameters for the $\bm{i}$-th scholar and $\bm{r}$ is the ratio of mixing real and synthetic data ($\bm{\bm{x'}}$) generated by the generator ($\bm{G_{i-1}}$). For each task $\bm{t_i}$, synthetic data is sampled for previously seen tasks using $\bm{G_{i-1}}$ and interleaved with the new data $\bm{D_i}$ to train both the generator $\bm{G_i}$ and the solver $\bm{S_i}$.

\paragraph{\acf{DGR+D}:} The \acs{DGR+D} approach~\citep{vandeVen2018CL} adapts the \ac{DGR} learning strategy by forming the replay targets as `\textit{soft-targets}' (softmax output vector) rather than using `\textit{hard-targets}' (labelled replay samples) as in the case of \ac{DGR}. This provides a more fine-grained update on the model (old to new) and the new solver is trained to match the output of old solver using a cross-entropy loss. Furthermore, a temperature $\bm{T}$ is added to the softmax output of the solver to further smoothen the probability distribution. The soft-targets are given as:
\begin{equation}
    \tilde{y_c} = p_{\hat{\theta^{(K-1)}}}^{T} (Y = c|\bm{x})
    \label{eq:distill}
\end{equation}
where $\bm{\theta^{(K-1)}}$ is the vector with parameter values at the end task $\bm{K-1}$ and $\bm{p_{\theta}^{T}}$ is the conditional probability for the model (with parameters $\bm{\theta}$) with a temperature $\bm{T}$ added to its softmax output. The distillation loss is thus given as:
\begin{equation}
    L_{distil} (\bm{x},\bm{\tilde{y}},\bm{\theta}) = -T^{2} \sum_{c=1}^{N} \tilde{y_c} \log p_{\theta}^{T}(Y=c|\bm{x})
\end{equation}
where $\bm{N}$ is the number of classes.

\paragraph{\acf{LGR}:} The \acf{LGR} approach \citep{Stoychev2021LGR,Stoychev2023LGRFG}, is proposed as a resource-efficient pseudo-rehearsal strategy that combines the benefits of \ac{DGR}~\citep{NIPS2017_6892} with using low-dimensional latent representations~\citep{Pellegrini2020Latent}, reducing both memory and computational expenses. \ac{LGR} is based on the assumption that once feature extraction layers (the \textit{root}) of a model are trained sufficiently and relatively \textit{frozen}, the extracted feature representations can be used effectively to rehearse past knowledge~\citep{Pellegrini2020Latent}. The \textit{scholar}-based architecture of the \ac{DGR}~\citep{NIPS2017_6892} is adapted to use \ac{LGR} instead by splitting the \textit{scholar} into three components: (i) a \textit{generator} ($\bm{G}$) to reconstruct and replay \textit{latent representations} instead of high-dimensional input patterns, (ii) a pre-trained and \textit{frozen} \textit{root} ($\bm{R}$) to extract task-independent latent representations, and (iii) the \textit{top} ($\bm{T}$) for the \textit{solver} to learn task-discriminative information. This is different from \ac{DGR} where the scholar consists of a single, combined solver while its generator is used to reconstruct high-dimensional pseudo-samples representing the input data. The learning objective for the \ac{LGR}-based \textit{scholar} is modified as follows:
\begin{equation}
    L(\theta_i)=r\mathbb{E}_{(R_x,y_i)\sim D_i}[L(T(R_x;\theta_i),y_i) + (1 - r) \mathbb{E}_{R^{\prime}_{x}\sim G_{old}}[L(T(R^\prime_x;\theta_i),y^\prime)]
\end{equation}
where $\bm{\theta_i}$ are model parameters for the $\bm{i}$-th scholar and $\bm{r}$ is the ratio of mixing real ($\bm{R_x}$) and synthetic ($\bm{R^\prime_x}$) latent representations from the generator ($\bm{G_{old}}$). For each task $\bm{t_i}$, $\bm{R^{\prime}_x}$ are passed through the \textit{top} of the solver to obtain labels $\bm{T(R^\prime_x)}$ for them. Once the generator is updated, the \textit{top} is updated to learn the new task. The current task data $\bm{\langle R(x), y\rangle}$ is combined with the labelled \textit{latent} pseudo-samples $\bm{\langle R^\prime_x, T(R^\prime_x)\rangle}$ generated for previously seen tasks. 

\paragraph{\acf{LGR+D}:} The \ac{DGR+D}~\citep{vandeVen2018CL} approach is also adapted to use \ac{LGR} using `soft' instead of `hard' targets~\citep{Stoychev2021LGR,Stoychev2023LGRFG}. A distillation loss similar to \ac{DGR+D} is used to provide more fine-grained updates to the model.

\begin{figure}
    \centering
    \includegraphics[width=0.5\textwidth]{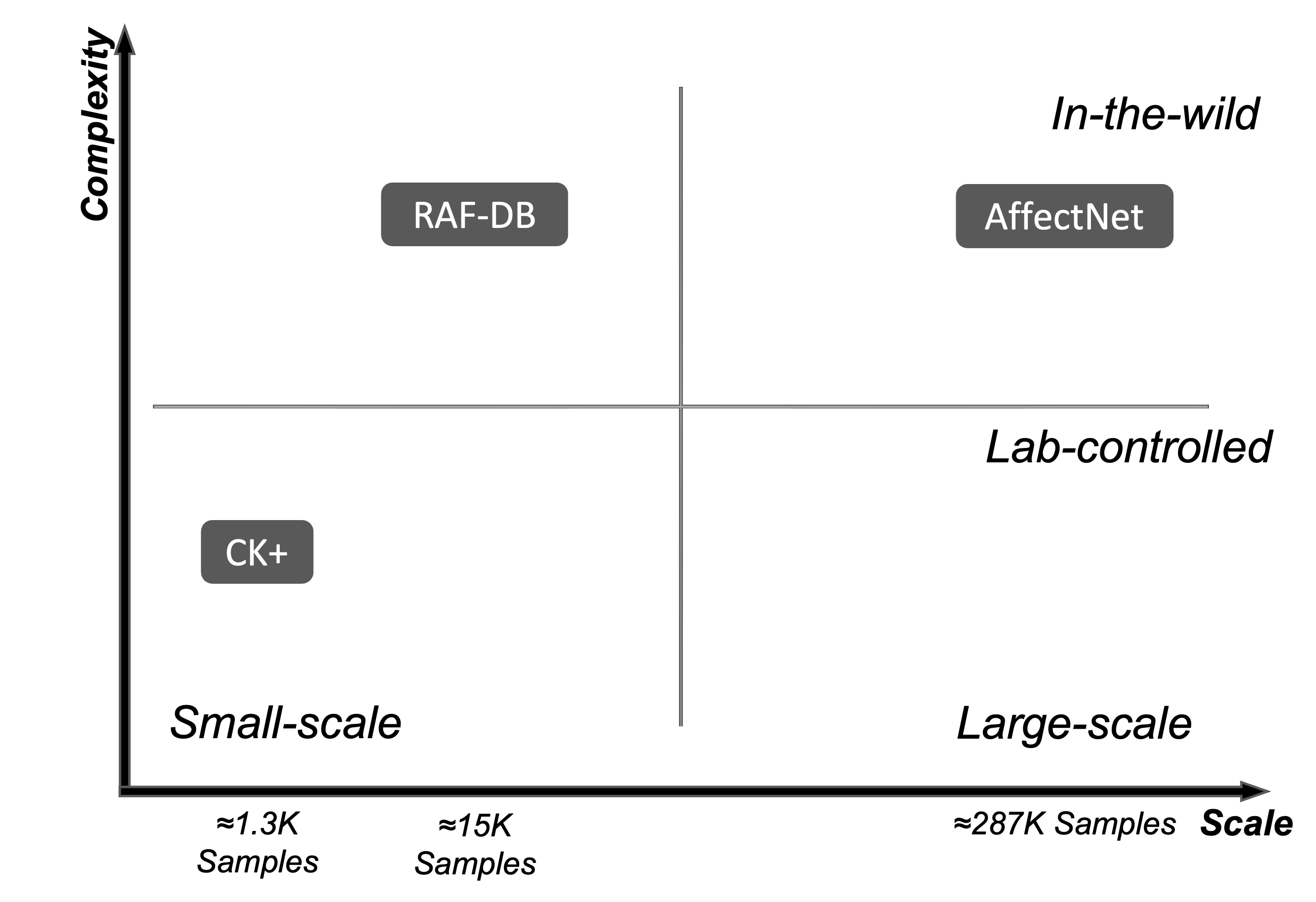}
    \caption{Benchmark Datasets for Facial Expression Recognition.}
    \label{fig:benchmarks}
\end{figure}

\subsection{Datasets}
The different \ac{CL}-based approaches described above are compared on $3$ popular \ac{FER} benchmarks: the \acs{CK+}~\citep{lucey2010extended}, \acs{RAF-DB}~\citep{li2017reliable,li2019reliable}, and AffectNet~\citep{2018Affectnet} datasets. These datasets represent a variety in terms of \textit{scale}, \textit{complexity} and data acquisition settings and are chosen following a two-fold selection criteria looking at: (a)~the complexity of data settings, that is, whether the images are recorded in lab-controlled settings (\acs{CK+}) or represent \textit{in-the-wild} settings (\acs{RAF-DB} and AffectNet), and (b) the \textit{scale} of the dataset, that is, relatively small-scale (\acs{CK+} and \acs{RAF-DB} with $n_{samples} \leq 20K$), or large-scale (AffectNet with $n_{samples} > 80K$). Additional experiments with $3$ medium-scale (with $20K > n_{samples} \leq 80K$) datasets, namely, \acs{iCV-MEFED}~\citep{Guo2018icv}, \acs{FER}-2013~\citep{Goodfellow2013Challenges}, and the BAUM-1 dataset~\citep{ZhalehpourBaum2017} can be found in Appendix~\ref{app:additional}.

\subsubsection{\acf{CK+} Dataset:} 
The \ac{CK+} dataset~\citep{lucey2010extended} is one of the most popularly used \ac{FER} dataset consisting of video recordings of $123$ subjects, annotated for $7$ expression classes namely, \textit{Anger, Surprise, Fearful, Disgust, Happiness, Sadness} and \textit{Contempt}. In total, there are $327$ video sequences showing a shift from \textit{neutral} to peak expression intensity. Similar to other frame-based approaches (\citep{Li2020Deep}), the last $3$ frames for each sequence are taken to represent the corresponding expression class while the first frame is used to constitute \textit{neutral} class samples. This results in a total of $\approx1.3K$ frames, annotated for $8$ expression classes to learn. A cross-subject split is performed, assigning $86$ subjects to the train set and $37$ to the test dataset, resulting in a $70:30$ data-split while maintaining the data distributions. 

\subsubsection{\acf{RAF-DB}:}
The \ac{RAF-DB} dataset~\citep{li2017reliable,li2019reliable} consists of $\approx15K$ facial images ``downloaded from the internet'' and manually labelled by multiple annotators for $6$ expression classes namely, \textit{Anger, Surprise, Fearful, Disgust, Happiness} and \textit{Sadness} along with \textit{Neutral} faces. As these images represent \textit{in-the-wild} settings, there is great variability within the samples of each class with respect to gender, race, head pose, facial attributes like facial hair and skin colour, and variations resulting from lighting conditions and artefacts such as glasses. The training and test-split provided by the authors of the dataset is used for the experiments presented in this work. 

\subsubsection{AffectNet Dataset:}
The full AffectNet dataset~\citep{2018Affectnet} consists of more than a million facial images downloaded from the internet. Half of the images ($\approx420$K) are manually labelled by multiple annotators for arousal and valence labels as well as eight facial expression categories namely \textit{Anger, Surprise, Fearful, Disgust, Happy, Sad, Contempt} and \textit{Neutral}. The rest of the images are automatically annotated by training a ResNet-based neural model trained on the manually labelled images. AffectNet dataset samples represent \textit{in-the-wild} settings with great diversity with respect to gender, race, head pose, facial attributes like facial hair and skin colour, and variations resulting from lighting conditions. As these images are captured randomly from the web, for several samples the face is either occluded or not present in the data samples. Thus, from the manually labelled samples, used for the experiments conducted, $\approx287k$ provide expression labels. 

\subsection{Pre-processing and Data Augmentation}
\label{sec:aug}
For the training and evaluation of the models on all the datasets, face-centred RGB images are used, resized to ($100\times 100 \times 3$). The image size is inspired from other approaches in literature~\citep{Li2020Deep, nott44740} and limitations in terms of the available GPU resources. Imbalances with respect to the distribution of the data samples for each class is a major problem in \ac{FER} datasets~\citep{Li2020Deep}. This is also seen in the datasets evaluated here. For instance, for \acs{RAF-DB}, $\approx31\%$ of the samples correspond to \textit{Happiness} while only $\approx2\%$ samples are available for \textit{Fearful}. 
For AffectNet as well, \textit{Happiness} samples constitute $\approx32\%$ of the usable samples. Thus, for a fair comparison in both balanced and imbalanced settings, data-augmentation is employed for all methods where additional samples for each class in the training set are generated by randomly ($p=0.5$) flipping samples horizontally and randomly ($p=0.5$) rotating them by an angle of up to $30^\circ$ in either direction. For AffectNet, however, the authors of the dataset already provide a \textit{downsampled} evaluation strategy~\citep{2018Affectnet} in an attempt to minimise class-imbalances by keeping an upper-limit for samples from each class to $15000$ samples. For all the datasets, all methods are compared both without and with data-augmentation.

\begin{table}[t!]
\centering
\setlength\tabcolsep{2.0pt}
\caption{Learning order followed for the \acs{Class-IL} and \acs{Task-IL} Experiments.}
\label{tab:orders}
{
\footnotesize

\begin{tabular}{l|llllllll}\toprule

\multicolumn{9}{c}{\textbf{Order~1}}\\\midrule

\textbf{\acs{Class-IL}} &  \textbf{C1:}~\textit{Neutral} & \textbf{C2:}~\textit{Anger} & \textbf{C3:}~\textit{Contempt}$^*$ & \textbf{C4:}~\textit{Disgust}  & \textbf{C5:}~\textit{Fearful} & \textbf{C6:}~\textit{Happiness} & \textbf{C7:}~\textit{Sadness} & \textbf{C8:}~\textit{Surprised} \\\midrule

\textbf{\acs{Task-IL}} & \multicolumn{2}{l}{\textbf{T1:}~\{\textit{Neutral, Anger}\}} & \multicolumn{2}{l}{\textbf{T2:}~\{\textit{Contempt$^*$, Disgust}\}} & \multicolumn{2}{l}{\textbf{T3:}~\{\textit{Fearful, Happiness}\}} & \multicolumn{2}{l}{\textbf{T4:}~\{\textit{Sadness, Surprised}\}} \\ \midrule

\multicolumn{9}{c}{\textbf{Order~2}}\\\midrule

\textbf{\acs{Class-IL}} &  \textbf{C1:}~\textit{Neutral} & \textbf{C2:}~\textit{Happiness} & \textbf{C3:}~\textit{Surprised} & \textbf{C4:}~\textit{Anger} & \textbf{C5:}~\textit{Fearful} & \textbf{C6:}~\textit{Sadness} & \textbf{C7:}~\textit{Disgust} & \textbf{C8:}~\textit{Contempt}$^*$ \\ \midrule

\textbf{\acs{Task-IL}} & \multicolumn{2}{l}{\textbf{T1:}~\{\textit{Neutral, Happiness}\}} & \multicolumn{2}{l}{\textbf{T2:}~\{\textit{Surprised, Anger}\}} & \multicolumn{2}{l}{\textbf{T3:}~\{\textit{Fearful, Sadness}\}} & \multicolumn{2}{l}{\textbf{T4:}~\{\textit{Disgust, Contempt$^*$}\}} \\ \midrule

\multicolumn{9}{c}{\textbf{Order~3}}\\\midrule

\textbf{\acs{Class-IL}} &  \textbf{C1:}~\textit{Neutral} & \textbf{C2:}~\textit{Contempt}$^*$ & \textbf{C3:}~\textit{Sadness}  & \textbf{C4:}~\textit{Anger} & \textbf{C5:}~\textit{Fearful} & \textbf{C6:}~\textit{Disgust} & \textbf{C7:}~\textit{Happiness} & \textbf{C8:}~\textit{Surprised} \\ \midrule
\textbf{\acs{Task-IL}} & \multicolumn{2}{l}{\textbf{T1:}~\{\textit{Neutral, Contempt$^*$}\}} & \multicolumn{2}{l}{\textbf{T2:}~\{\textit{Sadness, Anger}\}} & \multicolumn{2}{l}{\textbf{T3:}~\{\textit{Fearful, Disgust}\}} & \multicolumn{2}{l}{\textbf{T4:}~\{\textit{Happiness, Surprised}\}} \\ \bottomrule

\end{tabular}
}
\end{table}

\subsection{Experiment Settings}
\label{sec:expsetup}
Several studies investigating different \ac{CL} scenarios have shown that the order in which the different tasks or classes are presented to the models may impact learning dynamics~\citep{Delange2022CLSurvey,Churamani2020CLIFER}. This can be particularly important for \ac{FER} where the learnt feature representations by the models might overlap substantially (similar facial features albeit with different expressions). In this work, $3$ different class and task orderings are explored to understand their impact on the compared \ac{CL} methods. 
For the first ordering, the class labels provided by the authors of the \acs{CK+} dataset~\citep{lucey2010extended} (O1) are used for all the datasets. Additionally, $2$ more class and task orderings are explored that, starting from \textit{neutral}, either randomly learn \textit{positive} expressions first (O2) or randomly learn \textit{negative} expressions first (O3). The three orderings used in the experiments are given in Table~\ref{tab:orders}.



Contempt labels are unavailable for \ac{RAF-DB} and hence \textit{skipped}$^*$ for evaluations on this dataset resulting in the models evaluated for $7$ classes for \ac{Class-IL} and $4$ tasks for \ac{Task-IL} settings with Task $4$ consisting of only $1$ class.

\subsection{Implementation Details}
Each of the \ac{CL} approaches described in Section~\ref{sec:comparedCL} is implemented using a relatively simple \ac{CNN} model following the \textit{Lenet} architecture proposed by~\citet{lecun1998gradient}. This makes it faster to train the models incrementally and reduces the dependency on specialised training regimes, other than those focusing on \ac{CL}. The model consists of $2$ convolutional (\verb|conv|) layers with $20$ and $50$ filters, respectively, of size $(5\times5)$. Each \verb|conv| layer is followed by a \verb|BatchNormalisation| layer and uses a \verb|ReLU| activation. \verb|Max-pooling| is implemented after each \verb|conv| layer. The final \verb|Max-pooling| layer output is connected to a \acf{FC} layer consisting of $500$ nodes followed by a \verb|BatchNormalisation| layer and \verb|ReLU| activation. The \verb|FC| layer is followed by a \verb|softmax| output layer for classification. All models are trained using the \verb|Adam| optimiser ($\beta_1 = 0.9$, $\beta_2 = 0.999$). 
Table~\ref{tab:hyperConFER} presents the different hyper-parameters used for the experiments. These values are set based on separate hyper-parameter grid-searches for each model and selecting the best-performing values. All experiments are \textit{repeated} $3$ times and the results are \textit{averaged} across the repetitions to account for the random seeds. All models are implemented using the PyTorch Python Library\footnote{\url{https://pytorch.org}} adapting the code repository made available by~\citet{vandeVen2018CL}. To standardise the results, all experiments for an individual dataset are conducted on one of two systems equipped with: (a)~Nvidia Quadro RTX $8000$ GPU, $8$-core Intel Xeon Gold CPU @$2.30$GHz and $64$ GB RAM or (b)~Nvidia GeForce GTX $1080$Ti GPU, $12$-core Intel i7-8700 CPU @$3.20$GHz and $32$ GB RAM.

\begin{table}[t]
    \centering
    \caption[Model Hyper-parameters for ConFER Experiments.]{Model Hyper-parameters for \acs{Task-IL} and \acs{Class-IL} Experiments.}    
    \label{tab:hyperConFER}
    {
    \footnotesize
    \begin{tabular}{r|l|c|c|c}
        \toprule
        \textbf{Method} &  \textbf{Hyper-Parameter}	& \textbf{Learning-Rate}	    &	\textbf{Batch-Size} & \textbf{Iterations} \\ \midrule 
        
        \acs{EWC}       & $\lambda=5000$            & \multirow{11}[6]{*}{$2.5 e^{-4}$}  &  \multirow{11}[6]{*}{$128$ ($32$ for \acs{CK+})} & \multirow{11}[6]{*}{$500$}   \\ 
        
        \acs{EWC}-Online        & $\lambda=5000, \gamma=1$          &  {}       &  {} &  {} \\
        \acs{SI}                & $c=1$                             &  {}       &  {} &  {} \\
        \acs{LwF}               & $\lambda_{0}=1$                   &  {}       &  {} &  {} \\ 
        \acs{NR}                & \makecell[l]{$B_{size}=1500$ {\tiny(\acs{Task-IL})}\\$B_{size}=1000$ {\tiny(\acs{Class-IL})}}  &  {}  &    {} &  {} \\
        A-\acs{GEM}             & $M_{size}=2000$                   &  {}       &  {} &  {} \\
        \acs{LR}                & $B_{size}=1000$                   &  {}       &  {} &  {} \\
        \acs{DGR}               & $G_{FC} = 1600, G_{OUT}=30000$    &  {}       &  {} &  {} \\
        \acs{LGR}               & $G_{FC} = 200, G_{OUT}=200$       &  {}       &  {} &  {} \\
        \acs{DGR+D}             & $G_{FC} = 1600, G_{OUT}=30000$    &  {}       &  {} &  {} \\
        \acs{LGR+D}             & $G_{FC} = 200, G_{OUT}=200$       &  {}       &  {} &  {} \\ \bottomrule
    \end{tabular}
    }
\end{table}

\subsection{Evaluation Metrics}
\label{sec:confermetrics}
As the models successfully learn novel expression classes, it is important to ensure that past knowledge is not overwritten. Thus, the different \ac{CL} methods compared are evaluated both in terms of \acf{Acc} achieved as well as \acf{CF}, measuring their ability to maintain performance on previously learnt information while learning novel facial expression categories. 

\subsubsection{\acf{Acc}:}
The \acf{Acc} represents the fraction of correctly classified samples by the model on all the tasks seen so far. If $c_i$ be the number of correctly classified samples and $m_i$ be the total number of test-set samples for task $i$ then \acs{Acc} at the end of task $i$ is calculated as:
\begin{equation}
    Acc =\frac{1}{n}\sum_{i=1}^{n}{\frac{c_i}{m_i}} 
\end{equation}
\noindent where $n \in [1, N] $ is the number of tasks seen so far by the model and $N$ is the total number of tasks. At each step, average test-set accuracy scores are reported across all classes for all seen tasks for \ac{Task-IL} and for all seen classes in the case of \ac{Class-IL} evaluations. 

\subsubsection{\acf{CF}:}
\acf{CF}~\citep{kirkpatrick2017overcoming} can be defined as the tendency of a model to forget previously learnt information when presented, sequentially, with a new task. In \ac{CL} settings, where models are constantly presented with new tasks and classes to learn, it is important that the efficacy of these models is evaluated using both their accuracy scores as well as their ability to mitigate forgetting. Thus, \ac{CF} scores (computed as the negative of the \acf{BWT} metric proposed by~\citep{Diaz-Rodriguez2018}) are also reported, measuring the impact of learning new tasks on past knowledge. If $A$ is a $n\times n$ matrix where $n$ is the number of classes or tasks to be learnt, each row of the matrix $A$ stores the accuracy scores ($a_{i,j}$) for each class or task learnt up to that point. This results in $A$ being a lower-triangular matrix. If $i$ is the index of the current class or task and $j$ denotes the immediately previous task, then \ac{CF} is calculated as follows:
{
\[
    CF = \frac{\sum_{j=1}^{i-1} a_{j,j} - a_{i,j}}{i-1} 
    \textrm{,  where  } 
    A= 
    \begin{bmatrix}
        a_{11} & 0  & \dots  & 0 \\
        a_{21} & a_{22}  & \dots  & 0 \\
        \vdots & \vdots  & \ddots & \vdots \\
        a_{n,1} & a_{n,2}  & \dots  & a_{n,n}
    \end{bmatrix}
\]
}
The \ac{CF} score $\in [-1,1]$ measures the amount of forgetting that the model suffers from, with positive values denoting forgetting and negative values suggesting that newly learnt tasks improve the model's performance on previous task(s). 

As a guide to interpret the results, a model achieving low \textit{Acc} but high \textit{\acs{CF}} scores indicates that the model forgets previously seen tasks and also fails to learn the novel task. Low \textit{Acc} and low \textit{\ac{CF}} scores suggest that even though the model is able to mitigate forgetting, it struggles while learning new tasks. High \textit{Acc} and high \textit{\ac{CF}} scores, on the other hand, suggest that the past knowledge is \textit{overwritten} in the model as it learns the novel task. And finally, the ideal or sought after case for continually learning models is when it achieves a high \textit{Acc} with low \textit{\ac{CF}} scores. This suggests that the model, while successfully learning a novel task or class, is able to preserve its performance on previously seen tasks, mitigating forgetting in the model.

\section{Results}
\label{sec:conferresults}
This section presents both \acs{Task-IL} and \ac{Class-IL} results across the $3$ datasets, 
both without and with data-augmentation.
\subsection{\acs{CK+} Results}
The \acs{CK+} dataset represents relatively \textit{simplistic} learning settings consisting of images recorded in lab-controlled settings. Experiments under \ac{Task-IL} and \ac{Class-IL} settings are discussed below.

\begin{table}[t!]
\centering
\setlength\tabcolsep{3.0pt}
\caption[\acs{Task-IL} \acs{Acc} for \acs{CK+}  W/ and W/O Data-Augmentation.]{\acs{Task-IL} \acf{Acc} for \acs{CK+}  W/ and W/O Data-Augmentation. \textbf{Bold} values denote best (highest) while [\textit{bracketed}] denote second-best values for each column. 
}
\label{tab:task-il-ck-acc}

{
\scriptsize
\begin{tabular}{l|cccc|cccc}\toprule
\multicolumn{1}{c|}{\multirow{2}{*}{\textbf{Method}}}            & \multicolumn{4}{c|}{\textbf{\acs{Acc} W/O Data-Augmentation}} 
& \multicolumn{4}{c}{\textbf{\acs{Acc} W/ Data-Augmentation}} \\ \cmidrule{2-9}

\multicolumn{1}{c|}{ }                          & 
\multicolumn{1}{c|}{\textbf{Task 1}}            & \multicolumn{1}{c|}{\textbf{Task 2}}          &
\multicolumn{1}{c|}{\textbf{Task 3}}            & \multicolumn{1}{c|}{\textbf{Task 4}}           &
\multicolumn{1}{c|}{\textbf{Task 1}}            & \multicolumn{1}{c|}{\textbf{Task 2}}          &
\multicolumn{1}{c|}{\textbf{Task 3}}            & \multicolumn{1}{c}{\textbf{Task 4}}          \\ \midrule

\multicolumn{9}{c}{\textbf{Baseline Approaches}} \\ \midrule

LB      & $0.92\pm0.04$ & $0.83\pm0.04$ & $0.89\pm0.03$ & $0.81\pm0.03$ 
        & $0.94\pm0.01$ & $0.82\pm0.01$ & $0.88\pm0.00$ & $0.84\pm0.05$ \\
UB      & \cellcolor{gray!25}$\bm{0.94\pm0.01}$ & \cellcolor{gray!25}$\bm{0.96\pm0.01}$ & \cellcolor{gray!25}$\bm{0.97\pm0.02}$ &         
        \cellcolor{gray!25}$\bm{0.97\pm0.02}$ 
        & \cellcolor{gray!25}$\bm{0.96\pm0.01}$ & \cellcolor{gray!25}$\bm{0.98\pm0.00}$ & \cellcolor{gray!25}$\bm{0.99\pm0.00}$ & \cellcolor{gray!25}$\bm{0.99\pm0.00}$ \\ \midrule

\multicolumn{9}{c}{\textbf{Regularisation-Based Approaches}} \\ \midrule

EWC       
                                        & [$0.93\pm0.04$] & $0.83\pm0.03$ & $0.91\pm0.05$ & $0.84\pm0.04$ 
                                        & [$0.95\pm0.01$] & $0.89\pm0.03$ & $0.94\pm0.01$ & $0.90\pm0.01$ \\
EWC-Online   & [$0.93\pm0.04$] & $0.82\pm0.04$ & $0.89\pm0.07$ & $0.84\pm0.04$ 
                                        & [$0.95\pm0.01$] & $0.90\pm0.03$ & $0.92\pm0.01$ & $0.90\pm0.01$ \\
SI            & [$0.93\pm0.05$] & $0.82\pm0.04$ & $0.89\pm0.03$ & $0.84\pm0.01$ 
                                        & [$0.95\pm0.01$] & $0.82\pm0.00$ & $0.87\pm0.01$ & $0.82\pm0.05$ \\
LwF           & [$0.93\pm0.04$] & [$0.94\pm0.02$] & [$0.95\pm0.02$] & [$0.95\pm0.02$] 
&[$0.95\pm0.01$] & [$0.97\pm0.01$] & $0.97\pm0.00$ & $0.97\pm0.00$  \\ \midrule
\multicolumn{9}{c}{\textbf{Replay-Based Approaches}} \\ \midrule

 NR   & $0.92\pm0.04$ & \cellcolor{gray!25}$\bm{0.96\pm0.02}$ & \cellcolor{gray!25}$\bm{0.97\pm0.01}$ & 
                        \cellcolor{gray!25}$\bm{0.97\pm0.02}$ 
                         & [$0.95\pm0.01$] & \cellcolor{gray!25}$\bm{0.98\pm0.01}$ & [$0.98\pm0.00$] & [$0.98\pm0.00$] \\
A-GEM    & $0.92\pm0.04$ & \cellcolor{gray!25}$\bm{0.96\pm0.02}$ & [$0.95\pm0.02$] & $0.94\pm0.02$ 
                         & [$0.95\pm0.01$] & [$0.97\pm0.01$] & [$0.98\pm0.00$] & $0.97\pm0.00$ \\
LR    & $0.78\pm0.02$ & $0.88\pm0.03$ & $0.88\pm0.01$ & $0.90\pm0.02$ 
                         & $0.89\pm0.02$ & $0.95\pm0.01$ & $0.95\pm0.01$ & $0.96\pm0.00$ \\
DGR    & \cellcolor{gray!25}$\bm{0.94\pm0.03}$ & $0.85\pm0.03$ & $0.84\pm0.03$ & $0.81\pm0.00$ 
                         & [$0.95\pm0.01$] & $0.81\pm0.00$ & $0.85\pm0.02$ & $0.82\pm0.02$ \\
LGR    & $0.81\pm0.03$ & $0.89\pm0.00$ & $0.89\pm0.02$ & $0.88\pm0.01$ 
                         & $0.88\pm0.02$ & $0.93\pm0.01$ & $0.94\pm0.01$ & $0.95\pm0.01$ \\
DGR+D    & \cellcolor{gray!25}$\bm{0.94\pm0.03}$ & $0.87\pm0.01$ & $0.88\pm0.02$ & $0.86\pm0.01$ 
                         & [$0.95\pm0.01$] & $0.85\pm0.01$ & $0.88\pm0.01$ & $0.89\pm0.02$ \\
LGR+D    & $0.80\pm0.02$ & $0.88\pm0.02$ & $0.90\pm0.02$ & $0.92\pm0.00$ 
                         & $0.89\pm0.01$ & $0.94\pm0.01$ & $0.95\pm0.00$ & $0.96\pm0.00$ \\

\bottomrule

\end{tabular}


}
\end{table}

\begin{table}[t!]
\centering
\caption[\acs{Task-IL} \acs{CF} scores for \acs{CK+}  W/ and W/O Data-Augmentation.]{\acs{Task-IL} \acs{CF} scores for \ac{CK+}  W/ and W/O Data-Augmentation. \textbf{Bold} values denote best (lowest) while [\textit{bracketed}] denote second-best values for each column. }
\label{tab:task-il-ck-cf}

{
\scriptsize
\setlength\tabcolsep{3.5pt}

\begin{tabular}{l|rrr|rrr}\toprule
\multicolumn{1}{c|}{\multirow{2}{*}{\textbf{Method}}}          & \multicolumn{3}{c|}{\textbf{\acs{CF} W/O Data-Augmentation}} 
& \multicolumn{3}{c}{\textbf{\acs{CF}  W/ Data-Augmentation}} \\ \cmidrule{2-7}

\multicolumn{1}{c|}{ }                          & 
\multicolumn{1}{c|}{\textbf{Task 2}}          &
\multicolumn{1}{c|}{\textbf{Task 3}}            & \multicolumn{1}{c|}{\textbf{Task 4}}           &
\multicolumn{1}{c|}{\textbf{Task 2}}          &
\multicolumn{1}{c|}{\textbf{Task 3}}            & \multicolumn{1}{c}{\textbf{Task 4}}          \\ \midrule

\multicolumn{7}{c}{\textbf{Baseline Approaches}} \\ \midrule

                      LB  
                      &  $0.22\pm0.02$ &  $0.32\pm0.02$ &  $0.27\pm0.04$ 
                                        &  $0.32\pm0.04$ &  $0.29\pm0.10$ &  $0.31\pm0.00$ \\
                      UB  
                      &  [$0.00\pm0.04$] &  [$0.00\pm0.02$] &  [$0.00\pm0.04$] 
                                        & \cellcolor{gray!25}$\bm{-0.05\pm0.02}$ & \cellcolor{gray!25}$\bm{-0.03\pm0.01}$ & \cellcolor{gray!25}$\bm{-0.03\pm0.01}$ \\\midrule

\multicolumn{7}{c}{\textbf{Regularisation-Based Approaches}} \\ \midrule

EWC    
                                        &  $0.15\pm0.08$ &  $0.27\pm0.04$ &  $0.27\pm0.02$ 
                                        &  $0.13\pm0.04$ &  $0.17\pm0.03$ &  $0.17\pm0.06$ \\
EWC-Online   
                                        &  $0.22\pm0.14$ &  $0.26\pm0.03$ &  $0.27\pm0.03$ 
                                        &  $0.19\pm0.05$ &  $0.17\pm0.03$ &  $0.16\pm0.07$ \\
         SI   
                                        &  $0.22\pm0.03$ &  $0.26\pm0.04$ &  $0.29\pm0.06$ 
                                        &  $0.34\pm0.04$ &  $0.33\pm0.11$ &  $0.32\pm0.01$ \\
                 LwF  
                                        &  $0.06\pm0.01$ &  $0.04\pm0.01$ &  $0.04\pm0.01$ 
                                        &  $0.05\pm0.02$ &  $0.04\pm0.00$ &  $0.01\pm0.00$ \\ \midrule

\multicolumn{7}{c}{\textbf{Replay-Based Approaches}} \\ \midrule

               NR   
                                        & \cellcolor{gray!25}$\bm{-0.01\pm0.03}$ &  \cellcolor{gray!25}$\bm{0.00\pm0.01}$ & \cellcolor{gray!25}$\bm{-0.01\pm0.01}$ 
                                        &  [$0.00\pm0.01$] &  [$0.00\pm0.01$] & [$-0.02\pm0.00$] \\
   A-GEM    
                                        &  $0.06\pm0.02$ &  $0.06\pm0.01$ &  [$0.00\pm0.01$] 
                                        &  $0.01\pm0.01$ &  $0.03\pm0.01$ &  $0.01\pm0.01$ \\
       LR    
                                        &  $0.06\pm0.02$ &  $0.03\pm0.03$ & \cellcolor{gray!25}$\bm{-0.01\pm0.03}$ 
                                        &  $0.02\pm0.02$ &  [$0.00\pm0.01$] & [$-0.02\pm0.01$] \\
             DGR    
                                        &  $0.38\pm0.16$ &  $0.33\pm0.05$ &  $0.24\pm0.09$ 
                                        &  $0.40\pm0.08$ &  $0.32\pm0.05$ &  $0.34\pm0.01$ \\
           LGR    
                                        &  $0.07\pm0.06$ &  $0.10\pm0.01$ &  $0.01\pm0.03$ 
                                        &  $0.04\pm0.03$ &  $0.02\pm0.01$ &  $0.02\pm0.01$ \\
          DGR+D    
                                        &  $0.24\pm0.09$ &  $0.22\pm0.05$ &  $0.19\pm0.02$ 
                                        &  $0.29\pm0.02$ &  $0.20\pm0.04$ &  $0.25\pm0.01$ \\
         LGR+D
                                        &  $0.03\pm0.04$ &  $0.01\pm0.02$ &  $0.02\pm0.04$ 
                                        &  [$0.00\pm0.01$] &  $0.01\pm0.00$ &  $0.00\pm0.01$ \\

\bottomrule

\end{tabular}

}
\end{table}

\begin{figure}[t!]  
    \centering
    \subfloat[\ac{Task-IL} Results w/o Augmentation.\label{fig:ckp-task-il-noaug}]{\includegraphics[width=0.5\textwidth]{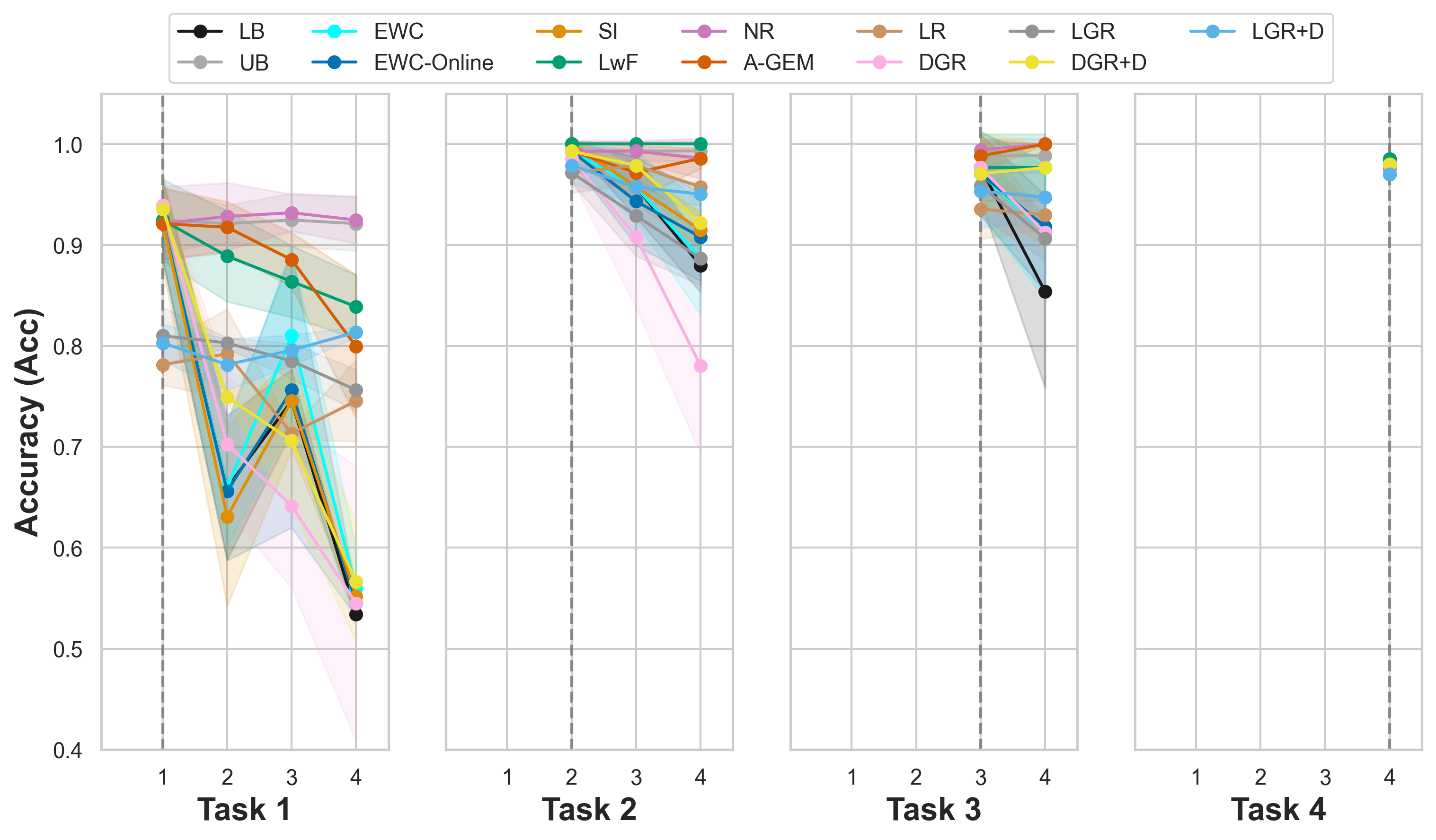}} \hfill
    \subfloat[\ac{Task-IL} Results w/ Augmentation.\label{fig:ckp-task-il-aug}]{\includegraphics[width=0.5\textwidth]{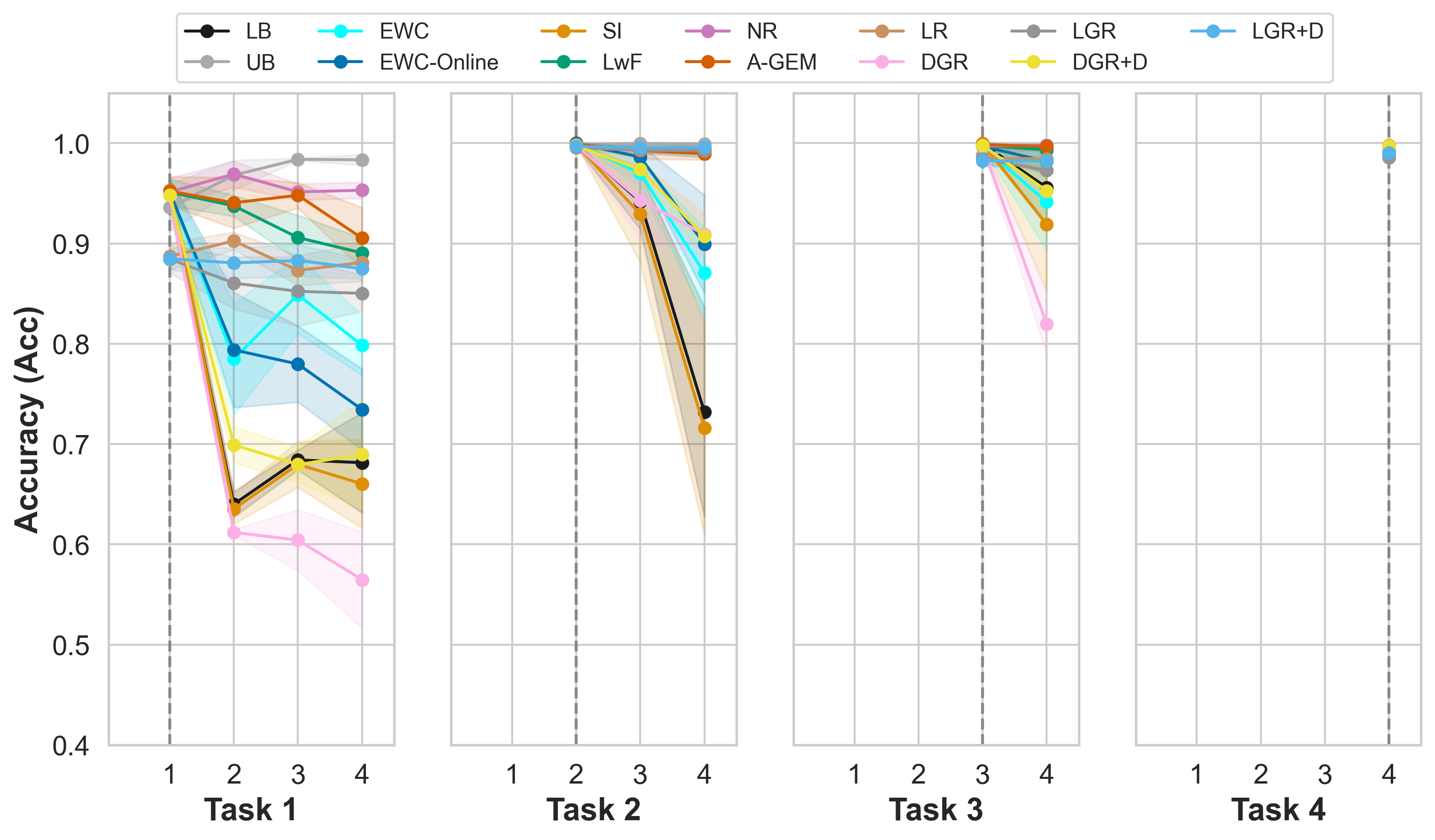}} \\
    \caption[Task-IL Results for CK+ W/ and W/O Augmentation.]{\ac{Task-IL} results for \ac{CK+} (a) without and (b) with augmentation. Test-set accuracy is shown as the learning progresses from when a task is introduced to the end of the training.}
    \label{fig:ckp-taskil}
    \vspace{-3mm}
\end{figure}

\paragraph{\acf{Task-IL}:} Table~\ref{tab:task-il-ck-acc} presents model performances, in terms of the \ac{Acc} under the \acs{Task-IL} settings with \acs{CK+}, both without and with data augmentation. \acs{UB} evaluations indeed represent the highest \acs{Acc} scores achievable, \textit{without data-augmentation}, with the \ac{NR} and A-\acs{GEM} approaches closing in on these scores. Both \acs{NR} and A-\acs{GEM} maintain a memory buffer large enough ($B_{size}=1500$ for \acs{NR} and $M_{size}=2000$ for A-\acs{GEM}) to store the entire \acs{CK+} dataset ($N\approx1.3K$), resulting in \acs{Acc} scores comparable to the \acs{UB} evaluations. \acs{NR} also achieves the \textit{lowest} \ac{CF} scores across all the methods (see Table~\ref{tab:task-il-ck-cf}). In general, replay-based methods are able to out-perform regularisation-based methods, with the exception \acs{LwF} which achieves the second-best results owing to its ability to learn task-specific parameters for both old and new tasks. \ac{DGR}, even though starting with the best model \acs{Acc}, is seen to perform the worse as the model learns more tasks. With the low number of samples for \acs{CK+}, the generator model for \acs{DGR} is not able to efficiently generate discriminative pseudo-samples of previously seen facial expression classes to efficiently rehearse past knowledge. This is in line with the suggestions made by \citet{van2019three} who conclude that generative replay becomes challenging on ``complicated input distributions''. The learning dynamics, that is, how each model's performance on each task evolves during the entire training process can be seen in Figure~\ref{fig:ckp-task-il-noaug}. \ac{DGR} experiences the most forgetting for Tasks~$1-3$ (also seen in the high \acs{CF} scores in Table~\ref{tab:task-il-ck-cf}) resulting in a low over-all model performance. 

Data-augmentation is seen to have a positive effect on the overall performance of the different models (Table~\ref{tab:task-il-ck-acc}). Similar relative model performances are witnessed \textit{with data-augmentation}, however, with reduced differences between the different approaches. \acs{UB} evaluations still show the highest achievable model \acs{Acc} scores across the different tasks with \acs{NR} and A-\acs{GEM} providing second-best evaluations. The additional data allows for other rehearsal-based approaches such as \ac{LR}, \ac{LGR} and \ac{LGR}+D to close the gap to a large extent, boosting their performance. This is not true for \ac{DGR} and \acs{DGR}+D who face similar problems in efficient pseudo-rehearsal of high-dimensional images. \ac{LGR} and \ac{LGR}+D avoid this problem by learning to generate low-dimensional \textit{latent features} for rehearsing past knowledge. This can also be seen in Table~\ref{tab:task-il-ck-cf} that shows these approaches achieving low \acs{CF} scores, comparable to \acs{NR} and A-\acs{GEM}. The learning dynamics showing the evolution of each task, with data-augmentation, can be seen in Figure~\ref{fig:ckp-task-il-aug}. 

\nl\textit{Task Ordering:} To evaluate whether the order in which the classes are ordered and split into tasks has an effect on model performance, $2$ additional class orders are explored. Given the positive impact of data-augmentation on model performance, both these orders are explored \textit{with data-augmentation}. Apart from minor differences in model performance for some of the rehearsal-based methods such as \ac{LR}, \ac{LGR} and \ac{LGR}+D, similar performance trends are witnessed across these orders. \ac{NR} and A-\ac{GEM} approaches perform the best, comparable to \acs{UB} evaluations, with no significant differences between model \acs{Acc} scores across the three different orders. 
More details on the results from the task-ordering experiments can be found in Appendix~\ref{app:taskorderCK}.

\begin{table}[t!]
\centering
\setlength\tabcolsep{3.5pt}

\caption[\acs{Class-IL} \acs{Acc} for \acs{CK+}  W/ and W/O Data-Augmentation.]{\acs{Class-IL} \acs{Acc} for \ac{CK+}  W/ and W/O Data-Augmentation. \textbf{Bold} values denote best (highest) while [\textit{bracketed}] denote second-best values for each column. 
}
\label{tab:class-il-ck-acc}

{
\scriptsize
\begin{tabular}{l|cccccccc}\toprule

\multirow{2}[2]{*}{\makecell[c]{\textbf{Method}}}           & \multicolumn{8}{c}{\textbf{\acs{Acc} W/O Data-Augmentation}} \\ \cmidrule{2-9}

\multicolumn{1}{c|}{ }                          & 
\multicolumn{1}{c|}{\textbf{Class 1}}            & \multicolumn{1}{c|}{\textbf{Class 2}}          &
\multicolumn{1}{c|}{\textbf{Class 3}}            & \multicolumn{1}{c|}{\textbf{Class 4}}           &
\multicolumn{1}{c|}{\textbf{Class 5}}            & \multicolumn{1}{c|}{\textbf{Class 6}}          &
\multicolumn{1}{c|}{\textbf{Class 7}}            & \multicolumn{1}{c}{\textbf{Class 8}}          \\ \midrule

\multicolumn{9}{c}{\textbf{Baseline Approaches}} \\ \midrule
                      LB & \cellcolor{gray!25}$\bm{1.00\pm0.00}$ & $0.50\pm0.00$ & $0.33\pm0.00$ & $0.25\pm0.00$ & $0.20\pm0.00$ & $0.17\pm0.00$ & $0.14\pm0.00$ & $0.12\pm0.00$ \\
                      UB & \cellcolor{gray!25}$\bm{1.00\pm0.00}$ & [$0.89\pm0.07$] & [$0.89\pm0.03$] & \cellcolor{gray!25}$\bm{0.94\pm0.01}$ & [$0.93\pm0.02$] & \cellcolor{gray!25}$\bm{0.95\pm0.00}$ & \cellcolor{gray!25}$\bm{0.92\pm0.01}$ & \cellcolor{gray!25}$\bm{0.92\pm0.01}$ \\ \midrule

\multicolumn{9}{c}{\textbf{Regularisation-Based Approaches}} \\ \midrule

 EWC  & \cellcolor{gray!25}$\bm{1.00\pm0.00}$ & $0.50\pm0.00$ & $0.33\pm0.00$ & $0.25\pm0.00$ & $0.20\pm0.00$ & $0.17\pm0.00$ & $0.14\pm0.00$ & $0.12\pm0.00$ \\
EWC-Online  & \cellcolor{gray!25}$\bm{1.00\pm0.00}$ & $0.50\pm0.00$ & $0.33\pm0.00$ & $0.25\pm0.00$ & $0.20\pm0.00$ & $0.17\pm0.00$ & $0.14\pm0.00$ & $0.12\pm0.00$ \\
         SI  & \cellcolor{gray!25}$\bm{1.00\pm0.00}$ & $0.50\pm0.00$ & $0.33\pm0.00$ & $0.25\pm0.00$ & $0.20\pm0.00$ & $0.17\pm0.00$ & $0.14\pm0.00$ & $0.12\pm0.00$ \\
                 LwF   & \cellcolor{gray!25}$\bm{1.00\pm0.00}$ & $0.50\pm0.00$ & $0.33\pm0.00$ & $0.25\pm0.00$ & $0.20\pm0.00$ & $0.17\pm0.00$ & $0.14\pm0.00$ & $0.12\pm0.00$ \\ \midrule
\multicolumn{9}{c}{\textbf{Replay-Based Approaches}} \\ \midrule

NR    & \cellcolor{gray!25}$\bm{1.00\pm0.00}$ & \cellcolor{gray!25}$\bm{0.93\pm0.01}$ & \cellcolor{gray!25}$\bm{0.93\pm0.03}$ & \cellcolor{gray!25}$\bm{0.94\pm0.02}$ & \cellcolor{gray!25}$\bm{0.94\pm0.01}$ & [$0.93\pm0.01$] & [$0.89\pm0.02$] & [$0.91\pm0.02$] \\

   A-GEM    & \cellcolor{gray!25}$\bm{1.00\pm0.00}$ & $0.59\pm0.06$ & $0.50\pm0.12$ & $0.25\pm0.00$ & $0.20\pm0.00$ & $0.17\pm0.00$ & $0.14\pm0.00$ & $0.12\pm0.00$ \\
       LR    & \cellcolor{gray!25}$\bm{1.00\pm0.00}$ & $0.77\pm0.02$ & $0.82\pm0.03$ & [$0.81\pm0.03$] & $0.77\pm0.02$ & $0.81\pm0.02$ & $0.79\pm0.04$ & $0.78\pm0.02$ \\
             DGR    & \cellcolor{gray!25}$\bm{1.00\pm0.00}$ & $0.50\pm0.00$ & $0.40\pm0.01$ & $0.26\pm0.01$ & $0.21\pm0.01$ & $0.18\pm0.01$ & $0.14\pm0.00$ & $0.13\pm0.01$ \\
           LGR    & \cellcolor{gray!25}$\bm{1.00\pm0.00}$ & $0.73\pm0.03$ & $0.72\pm0.02$ & $0.70\pm0.03$ & $0.55\pm0.01$ & $0.55\pm0.02$ & $0.39\pm0.03$ & $0.44\pm0.03$ \\
          DGR+D    & \cellcolor{gray!25}$\bm{1.00\pm0.00}$ & $0.51\pm0.01$ & $0.41\pm0.04$ & $0.26\pm0.01$ & $0.21\pm0.00$ & $0.18\pm0.00$ & $0.15\pm0.00$ & $0.12\pm0.00$ \\
         LGR+D    & \cellcolor{gray!25}$\bm{1.00\pm0.00}$ & $0.75\pm0.04$ & $0.74\pm0.06$ & $0.72\pm0.02$ & $0.66\pm0.05$ & $0.64\pm0.04$ & $0.58\pm0.01$ & $0.58\pm0.04$ \\

\midrule

\multirow{2}[2]{*}{\makecell[c]{\textbf{Method}}}           & \multicolumn{8}{c}{\textbf{\acs{Acc} W/ Data-Augmentation}} \\ \cmidrule{2-9}

\multicolumn{1}{c|}{ }                          & 
\multicolumn{1}{c|}{\textbf{Class 1}}            & \multicolumn{1}{c|}{\textbf{Class 2}}          &
\multicolumn{1}{c|}{\textbf{Class 3}}            & \multicolumn{1}{c|}{\textbf{Class 4}}           &
\multicolumn{1}{c|}{\textbf{Class 5}}            & \multicolumn{1}{c|}{\textbf{Class 6}}          &
\multicolumn{1}{c|}{\textbf{Class 7}}            & \multicolumn{1}{c}{\textbf{Class 8}}          \\ \midrule
\multicolumn{9}{c}{\textbf{Baseline Approaches}} \\ \midrule

                      LB & \cellcolor{gray!25}$\bm{1.00\pm0.00}$ & $0.50\pm0.00$ & $0.33\pm0.00$ & $0.25\pm0.00$ & $0.20\pm0.00$ & $0.17\pm0.00$ & $0.14\pm0.00$ & $0.12\pm0.00$ \\
                      UB & \cellcolor{gray!25}$\bm{1.00\pm0.00}$ & \cellcolor{gray!25}$\bm{0.99\pm0.00}$ & \cellcolor{gray!25}$\bm{0.99\pm0.00}$ & \cellcolor{gray!25}$\bm{0.99\pm0.00}$ & \cellcolor{gray!25}$\bm{1.00\pm0.00}$ & \cellcolor{gray!25}$\bm{1.00\pm0.00}$ & \cellcolor{gray!25}$\bm{0.99\pm0.00}$ & \cellcolor{gray!25}$\bm{0.99\pm0.00}$ \\ \midrule

\multicolumn{9}{c}{\textbf{Regularisation-Based Approaches}} \\ \midrule

 EWC  & \cellcolor{gray!25}$\bm{1.00\pm0.00}$ & $0.50\pm0.00$ & $0.33\pm0.00$ & $0.25\pm0.00$ & $0.20\pm0.00$ & $0.17\pm0.00$ & $0.14\pm0.00$ & $0.12\pm0.00$ \\
EWC-Online  & \cellcolor{gray!25}$\bm{1.00\pm0.00}$ & $0.50\pm0.00$ & $0.33\pm0.00$ & $0.25\pm0.00$ & $0.20\pm0.00$ & $0.17\pm0.00$ & $0.14\pm0.00$ & $0.12\pm0.00$ \\
         SI  & \cellcolor{gray!25}$\bm{1.00\pm0.00}$ & $0.50\pm0.00$ & $0.33\pm0.00$ & $0.25\pm0.00$ & $0.20\pm0.00$ & $0.17\pm0.00$ & $0.14\pm0.00$ & $0.12\pm0.00$ \\
                 LwF   & \cellcolor{gray!25}$\bm{1.00\pm0.00}$ & $0.50\pm0.00$ & $0.33\pm0.00$ & $0.25\pm0.00$ & $0.20\pm0.00$ & $0.17\pm0.00$ & $0.14\pm0.00$ & $0.12\pm0.00$ \\ \midrule
\multicolumn{9}{c}{\textbf{Replay-Based Approaches}} \\ \midrule

   NR    & \cellcolor{gray!25}$\bm{1.00\pm0.00}$ & \cellcolor{gray!25}$\bm{0.99\pm0.00}$ & [$0.98\pm0.00$] & [$0.97\pm0.00$] & [$0.96\pm0.00$] & [$0.96\pm0.00$] & [$0.93\pm0.00$] & $0.92\pm0.00$ \\
   A-GEM    & \cellcolor{gray!25}$\bm{1.00\pm0.00}$ & $0.57\pm0.10$ & $0.59\pm0.19$ & $0.26\pm0.01$ & $0.24\pm0.06$ & $0.33\pm0.02$ & $0.30\pm0.02$ & $0.18\pm0.05$ \\
    LR    & \cellcolor{gray!25}$\bm{1.00\pm0.00}$ & $0.94\pm0.02$ & $0.94\pm0.02$ & $0.93\pm0.02$ & $0.93\pm0.02$ & $0.94\pm0.01$ & [$0.93\pm0.02$] & [$0.93\pm0.02$]\\
             DGR    & \cellcolor{gray!25}$\bm{1.00\pm0.00}$ & $0.51\pm0.00$ & $0.40\pm0.02$ & $0.28\pm0.00$ & $0.23\pm0.01$ & $0.20\pm0.01$ & $0.16\pm0.02$ & $0.15\pm0.02$ \\
           LGR    & \cellcolor{gray!25}$\bm{1.00\pm0.00}$ & $0.85\pm0.05$ & $0.86\pm0.04$ & $0.85\pm0.05$ & $0.85\pm0.05$ & $0.84\pm0.07$ & $0.79\pm0.08$ & $0.79\pm0.07$ \\
          DGR+D    & \cellcolor{gray!25}$\bm{1.00\pm0.00}$ & $0.52\pm0.00$ & $0.42\pm0.02$ & $0.30\pm0.04$ & $0.22\pm0.01$ & $0.19\pm0.01$ & $0.16\pm0.00$ & $0.13\pm0.00$ \\
         LGR+D    & \cellcolor{gray!25}$\bm{1.00\pm0.00}$ & $0.88\pm0.04$ & $0.89\pm0.04$ & $0.87\pm0.05$ & $0.87\pm0.06$ & $0.87\pm0.07$ & $0.84\pm0.08$ & $0.84\pm0.08$ \\

\bottomrule

\end{tabular}

}
\end{table}
\begin{table}[ht!]
\centering
\caption[\acs{Class-IL} \acs{CF} scores for \acs{CK+}  W/ and W/O Data-Augmentation.]{\acs{Class-IL} \acs{CF} scores for \ac{CK+} W/ and W/O Data-Augmentation. \textbf{Bold} values denote best (lowest) while [\textit{bracketed}] denote second-best values for each column.}
\label{tab:class-il-ck-cf}

{
\scriptsize
\setlength\tabcolsep{3.5pt}

\begin{tabular}{l|ccccccc}\toprule

\multirow{2}[2]{*}{\makecell[c]{\textbf{Method}}}           & \multicolumn{7}{c}{\textbf{\acs{CF} W/O Data-Augmentation}} \\ \cmidrule{2-8}

\multicolumn{1}{c|}{ }                          & 
\multicolumn{1}{c|}{\textbf{Class 2}}          &
\multicolumn{1}{c|}{\textbf{Class 3}}            & \multicolumn{1}{c|}{\textbf{Class 4}}           &
\multicolumn{1}{c|}{\textbf{Class 5}}            & \multicolumn{1}{c|}{\textbf{Class 6}}          &
\multicolumn{1}{c|}{\textbf{Class 7}}            & \multicolumn{1}{c}{\textbf{Class 8}}          \\ \midrule
\multicolumn{8}{c}{\textbf{Baseline Approaches}} \\ \midrule

                      LB 
                      &  $1.00\pm0.00$ &  $1.00\pm0.00$ &  $1.00\pm0.00$ &  $1.00\pm0.00$ &  $1.00\pm0.00$ &  $1.00\pm0.00$ & $1.00\pm0.00$ \\
                      UB 
                      & \cellcolor{gray!25}$\bm{-0.03\pm0.09}$ & \cellcolor{gray!25}$\bm{-0.05\pm0.09}$ & \cellcolor{gray!25}$\bm{-0.03\pm0.08}$ & \cellcolor{gray!25}$\bm{-0.04\pm0.06}$ & \cellcolor{gray!25}$\bm{-0.02\pm0.05}$ & \cellcolor{gray!25}$\bm{-0.02\pm0.05}$ & \cellcolor{gray!25}$\bm{0.05\pm0.03}$ \\\midrule

\multicolumn{8}{c}{\textbf{Regularisation-Based Approaches}} \\ \midrule

 EWC  
 &  $1.00\pm0.00$ &  $1.00\pm0.00$ &  $1.00\pm0.00$ &  $1.00\pm0.00$ &  $1.00\pm0.00$ &  $1.00\pm0.00$ & $1.00\pm0.00$ \\
EWC-Online  
&  $1.00\pm0.00$ &  $1.00\pm0.00$ &  $1.00\pm0.00$ &  $1.00\pm0.00$ &  $1.00\pm0.00$ &  $1.00\pm0.00$ & $1.00\pm0.00$ \\
         SI  
         &  $1.00\pm0.00$ &  $1.00\pm0.00$ &  $1.00\pm0.00$ &  $1.00\pm0.00$ &  $1.00\pm0.00$ &  $1.00\pm0.00$ & $1.00\pm0.00$ \\
                 LwF   
                 &  $1.00\pm0.00$ &  $1.00\pm0.00$ &  $1.00\pm0.00$ &  $1.00\pm0.00$ &  $1.00\pm0.00$ &  $1.00\pm0.00$ & $1.00\pm0.00$ \\ \midrule

\multicolumn{8}{c}{\textbf{Replay-Based Approaches}} \\ \midrule

               NR    
               &  [$0.03\pm0.03$] &  [$0.03\pm0.01$] &  [$0.02\pm0.00$] &  [$0.04\pm0.01$] &  [$0.06\pm0.02$] &  [$0.04\pm0.01$] & [$0.06\pm0.01$] \\
   A-GEM    
   &  $0.76\pm0.17$ &  $1.00\pm0.00$ &  $1.00\pm0.00$ &  $1.00\pm0.00$ &  $1.00\pm0.00$ &  $1.00\pm0.00$ & $0.82\pm0.13$ \\
       LR    
       &  $0.16\pm0.03$ &  $0.12\pm0.04$ &  $0.10\pm0.02$ &  $0.07\pm0.02$ &  $0.07\pm0.03$ &  $0.07\pm0.03$ & $0.23\pm0.01$ \\
             DGR    
             &  $0.91\pm0.01$ &  $0.98\pm0.02$ &  $0.98\pm0.02$ &  $0.98\pm0.02$ &  $0.99\pm0.00$ &  $0.99\pm0.01$ & $1.00\pm0.00$ \\
           LGR    
           &  $0.42\pm0.03$ &  $0.35\pm0.05$ &  $0.52\pm0.01$ &  $0.49\pm0.01$ &  $0.66\pm0.03$ &  $0.60\pm0.03$ & $0.54\pm0.04$ \\
          DGR+D    
          &  $0.89\pm0.05$ &  $0.98\pm0.02$ &  $0.99\pm0.00$ &  $0.98\pm0.01$ &  $0.99\pm0.01$ &  $1.00\pm0.00$ & $0.98\pm0.01$ \\
         LGR+D    
         &  $0.32\pm0.08$ &  $0.28\pm0.03$ &  $0.32\pm0.04$ &  $0.35\pm0.04$ &  $0.41\pm0.02$ &  $0.40\pm0.05$ & $0.39\pm0.08$ \\ \midrule

\multirow{2}[2]{*}{\makecell[c]{\textbf{Method}}}          & \multicolumn{7}{c}{\textbf{\acs{CF} W/ Data-Augmentation}} \\ \cmidrule{2-8}

\multicolumn{1}{c|}{ }                          & 
\multicolumn{1}{c|}{\textbf{Class 2}}          &
\multicolumn{1}{c|}{\textbf{Class 3}}            & \multicolumn{1}{c|}{\textbf{Class 4}}           &
\multicolumn{1}{c|}{\textbf{Class 5}}            & \multicolumn{1}{c|}{\textbf{Class 6}}          &
\multicolumn{1}{c|}{\textbf{Class 7}}            & \multicolumn{1}{c}{\textbf{Class 8}}          \\ \midrule

\multicolumn{8}{c}{\textbf{Baseline Approaches}} \\ \midrule

                      LB 
                      &  $1.00\pm0.00$ &  $1.00\pm0.00$ &  $1.00\pm0.00$ &  $1.00\pm0.00$ &  $1.00\pm0.00$ &  $1.00\pm0.00$ & $1.00\pm0.00$ \\
                      UB 
                      &  \cellcolor{gray!25}$\bm{0.01\pm0.00}$ &  \cellcolor{gray!25}$\bm{0.01\pm0.00}$ &  \cellcolor{gray!25}$\bm{0.00\pm0.00}$ &  \cellcolor{gray!25}$\bm{0.00\pm0.00}$ &  \cellcolor{gray!25}$\bm{0.01\pm0.00}$ &  \cellcolor{gray!25}$\bm{0.00\pm0.00}$ & \cellcolor{gray!25}$\bm{0.01\pm0.00}$ \\ \midrule

\multicolumn{8}{c}{\textbf{Regularisation-Based Approaches}} \\ \midrule

 EWC  
 &  $1.00\pm0.00$ &  $1.00\pm0.00$ &  $1.00\pm0.00$ &  $1.00\pm0.00$ &  $1.00\pm0.00$ &  $1.00\pm0.00$ & $1.00\pm0.00$ \\
EWC-Online  
&  $1.00\pm0.00$ &  $1.00\pm0.00$ &  $1.00\pm0.00$ &  $1.00\pm0.00$ &  $1.00\pm0.00$ &  $1.00\pm0.00$ & $1.00\pm0.00$ \\
         SI  
         &  $1.00\pm0.00$ &  $1.00\pm0.00$ &  $1.00\pm0.00$ &  $1.00\pm0.00$ &  $1.00\pm0.00$ &  $1.00\pm0.00$ & $1.00\pm0.00$ \\
                 LwF   
                 &  $1.00\pm0.00$ &  $1.00\pm0.00$ &  $1.00\pm0.00$ &  $1.00\pm0.00$ &  $1.00\pm0.00$ &  $1.00\pm0.00$ & $1.00\pm0.00$ \\ \midrule

\multicolumn{8}{c}{\textbf{Replay-Based Approaches}} \\ \midrule

               NR    
               &  [$0.03\pm0.00$] &  [$0.04\pm0.01$] &  [$0.05\pm0.00$] &  [$0.04\pm0.00$] &  $0.08\pm0.00$ &  $0.10\pm0.01$ & [$0.03\pm0.01$] \\
   A-GEM    
   &  $0.61\pm0.28$ &  $0.99\pm0.01$ &  $0.95\pm0.07$ &  $0.80\pm0.03$ &  $0.82\pm0.02$ &  $0.94\pm0.05$ & $0.86\pm0.20$ \\
       LR    
       &  $0.05\pm0.02$ &  $0.05\pm0.01$ &  [$0.05\pm0.01$] &  [$0.04\pm0.01$] &  [$0.05\pm0.01$] &  [$0.04\pm0.01$] & $0.06\pm0.01$ \\
             DGR    
             &  $0.90\pm0.04$ &  $0.95\pm0.00$ &  $0.96\pm0.02$ &  $0.96\pm0.01$ &  $0.98\pm0.02$ &  $0.97\pm0.03$ & $0.98\pm0.00$ \\
           LGR    
           &  $0.21\pm0.06$ &  $0.19\pm0.06$ &  $0.18\pm0.06$ &  $0.18\pm0.08$ &  $0.22\pm0.09$ &  $0.22\pm0.08$ & $0.31\pm0.09$ \\
          DGR+D    
          &  $0.87\pm0.02$ &  $0.93\pm0.05$ &  $0.97\pm0.01$ &  $0.97\pm0.02$ &  $0.99\pm0.01$ &  $0.99\pm0.01$ & $0.96\pm0.01$ \\
         LGR+D    
         &  $0.16\pm0.06$ &  $0.15\pm0.06$ &  $0.14\pm0.07$ &  $0.14\pm0.08$ &  $0.17\pm0.09$ &  $0.16\pm0.08$ & $0.23\pm0.08$ \\

\bottomrule

\end{tabular}

}
\end{table}
\paragraph{\acf{Class-IL}:} Table~\ref{tab:class-il-ck-acc} presents \acs{Class-IL} results in terms of the \ac{Acc} achieved by the models, both with and without data augmentation. For the results \textit{without data-augmentation} \ac{LB} results show evidence of \textit{catastrophic forgetting} where, learning each new task completely \textit{overwrites} all past knowledge. \ac{UB}, on the other hand, allow complete access to the entire dataset for the model at each step of the learning, resulting in the least forgetting in the model.  

\begin{figure}  
    \centering
    \subfloat[\ac{Class-IL} Results W/o Augmentation.\label{fig:ckp-class-il-noaug}]{\includegraphics[width=0.5\textwidth]{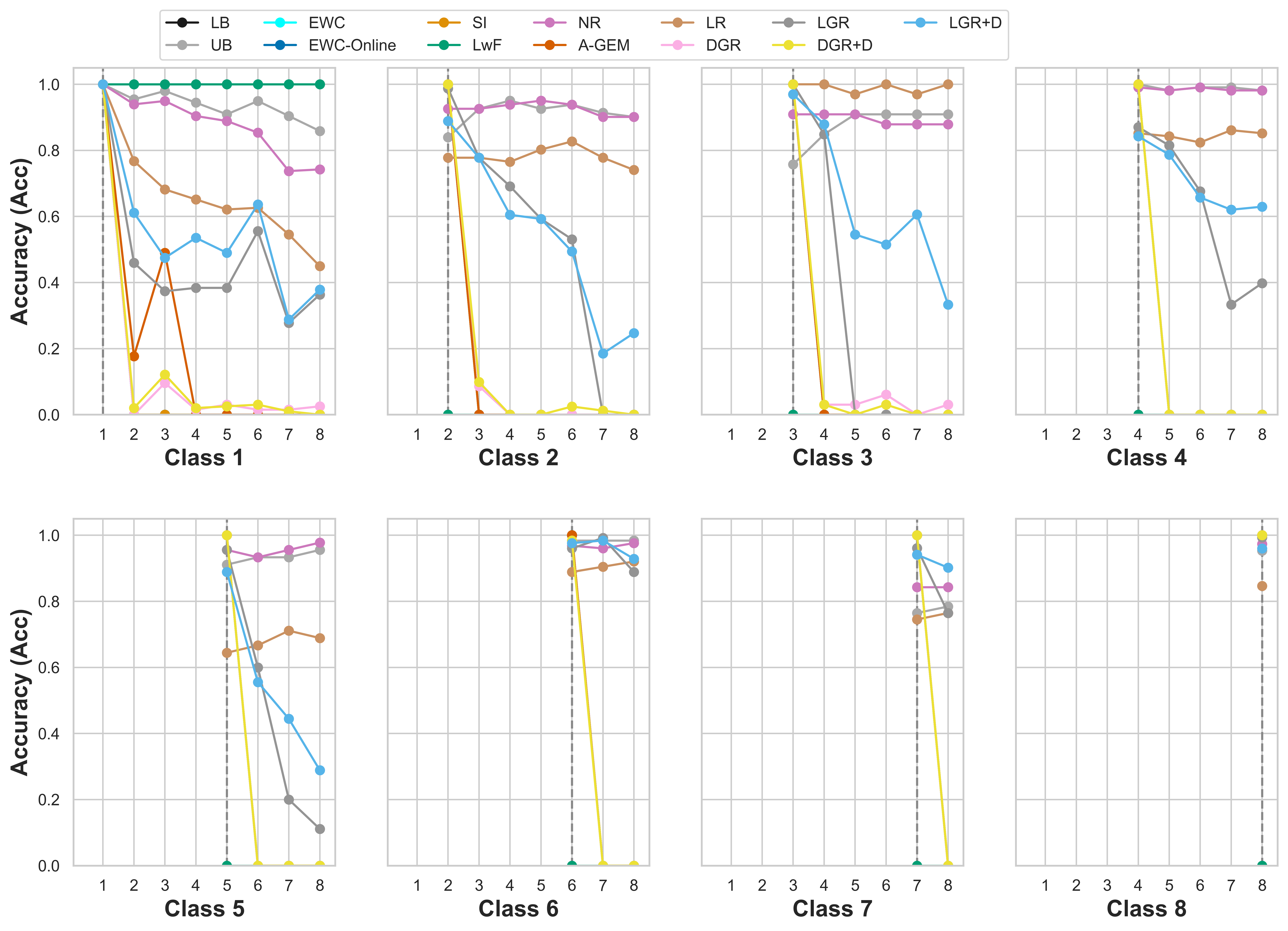}} \hfill
    \subfloat[\ac{Class-IL} Results W/ Augmentation.\label{fig:ckp-class-il-aug}]{\includegraphics[width=0.5\textwidth]{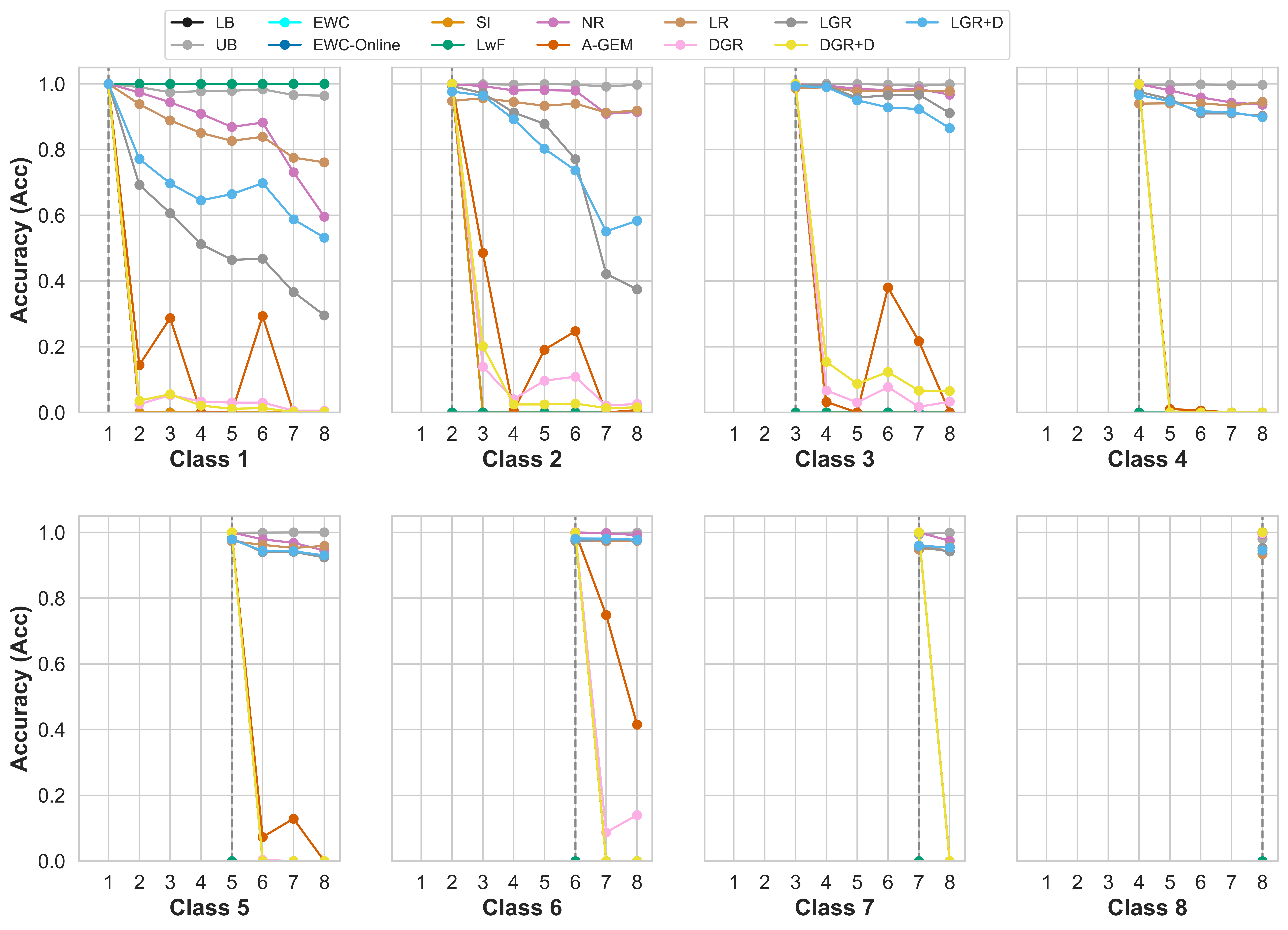}} \\
    \caption[Class-IL Results for CK+ W/ and W/O Augmentation.]{\ac{Class-IL} results for \ac{CK+} (a) without and (b) with augmentation. Test-set accuracy is shown as the learning progresses from when a class is introduced to the end of the training.}
    \label{fig:ckp-Class} 
    
\end{figure}

The \ac{NR} approach is able to match, if not better, \ac{UB} scores due to the memory buffer ($B_{size}=1000$) being able to accommodate almost the entire dataset ($N\approx1.3K$) and provide complete access to all the data during training. This results in an effective replay of past knowledge, mitigating forgetting in the model. This is also witnessed in Table~\ref{tab:class-il-ck-cf} where \ac{NR} achieves the second-best \acs{CF} scores, higher than only the \ac{UB} evaluations. Different from \ac{Task-IL} results, similar to other studies~\citep{van2019three}, regularisation-based approaches fail to preserve past knowledge completely, resulting in \textit{catastrophic forgetting}. Replay-based methods, on the other hand, are able to perform better at preserving information. The A-\ac{GEM} approach fails to adapt sufficiently to the dynamics of \ac{Class-IL} settings. After learning each class, the model tries to constrain the average episodic memory loss over all previous classes. This does not scale well with the increasing number of classes to be learnt, resulting in forgetting in the model. \ac{DGR} and \ac{DGR}+D rely on the generator to reconstruct high-dimensional pseudo-samples of facial images. This becomes even more difficult when learning one class at a time as limited data is available to train the generator~\citep{Liu2020Geneative, Lucic2018GANs} resulting in an inefficient pseudo-rehearsal of past knowledge. The learning dynamics for the different approaches can be seen in Figure~\ref{fig:ckp-class-il-noaug}.

Data-augmentation positively impacts model performance, however, only for replay-based approaches (Table~\ref{tab:class-il-ck-acc}), yet, no approach is able to match the near-perfect \ac{UB} scores. Regularisation-based methods, even with more training data, fail completely at mitigating forgetting in the model. Amongst the replay-based methods, \ac{NR} still performs the best, with \ac{LR} offering next best results. The additional data enables \ac{LGR} and \ac{LGR}+D to improve the quality of the \textit{latent features} generated for pseudo-rehearsal, positively impacting model performance. However, \ac{DGR} and \ac{DGR}+D continue to face the same issues of scalability as the models learns one class at a time. This is also witnessed in the high \ac{CF} scores (Table~\ref{tab:class-il-ck-cf}), indicating \textit{forgetting} of information. The learning dynamics showing the evolution of each class, with data-augmentation, can be seen in Figure~\ref{fig:ckp-class-il-aug}. 

\nl\textit{Class Ordering:} Similar to \acs{Task-IL}, the effect of class ordering is also evaluated. The $2$ class orders start from \textit{neutral}, either first learning \textit{positive} expressions first (O2=[C1:~\textit{Neutral}, C2:~\textit{Happiness}, C3:~\textit{Surprised}, C4:~\textit{Anger}, C5:~\textit{Fearful}, C6:~\textit{Sadness}, C7:~\textit{Disgust}, C8:~\textit{Contempt}]) or learn \textit{negative} expressions first 
(O3=[C1:~\textit{Neutral}, C2:~\textit{Contempt}, C3:~\textit{Sadness}, C4:~\textit{Anger}, C5:~\textit{Fearful}, C6:~\textit{Disgust}, C7:~\textit{Happiness}, C8:~\textit{Surprised}]) as presented in Table~\ref{tab:orders}. No significant differences in model performance are witnessed across the different class orderings. 
More details on the results from class-ordering experiments can be found in Appendix~\ref{app:classorderCK}.


\subsection{\acs{RAF-DB} Results}
The \acs{RAF-DB} dataset consists of facial images representing \textit{in-the-wild} settings split into $7$ expression classes. \acs{Task-IL} and \acs{Class-IL} experiments with \ac{RAF-DB} are discussed below.

\begin{figure} [t] 
    \centering
    \subfloat[\ac{Task-IL} Results W/O Augmentation.\label{fig:rafdb-task-il-noaug}]{\includegraphics[width=0.5\textwidth]{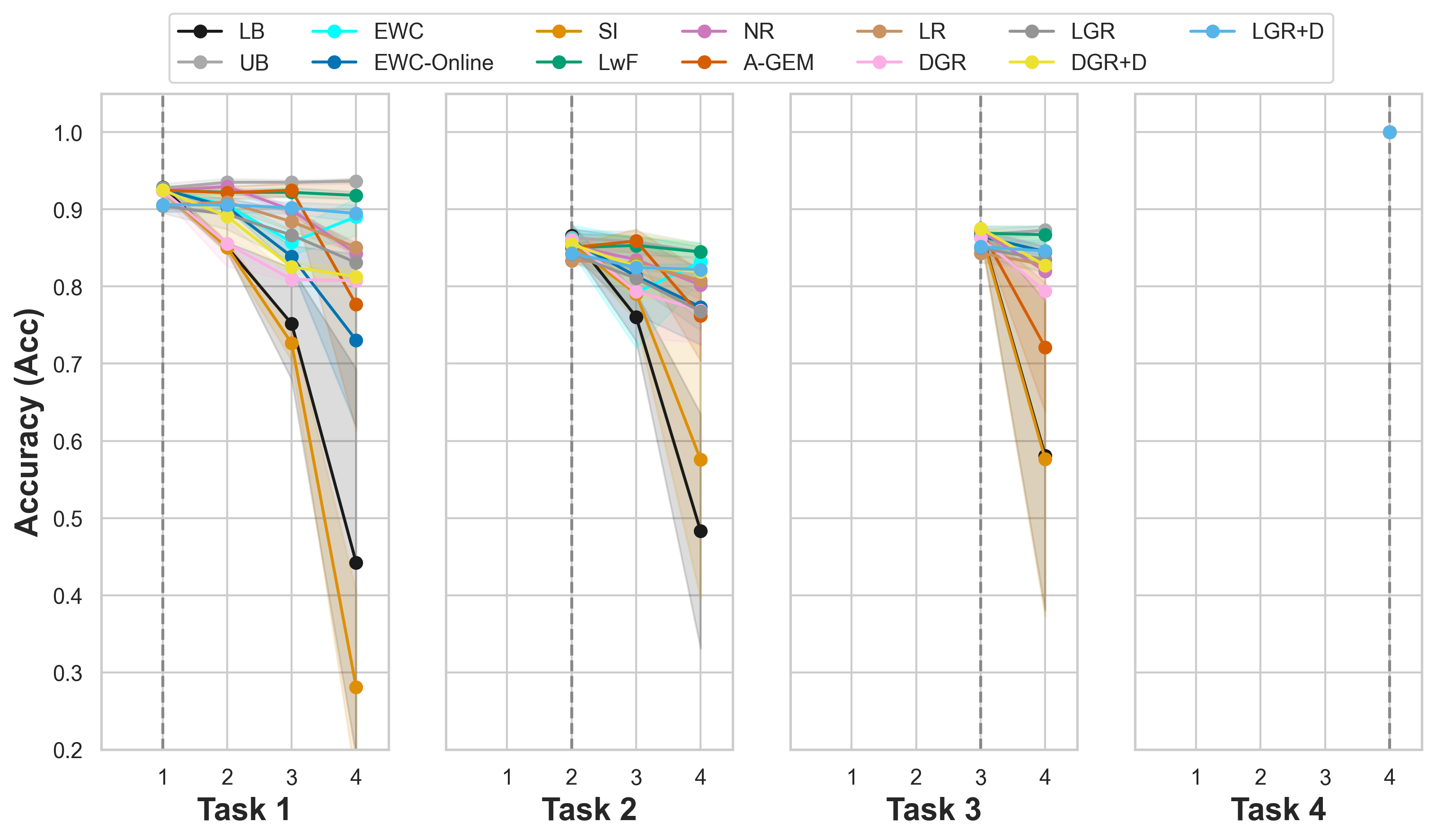}} \hfill
    \subfloat[\ac{Task-IL} Results W/ Augmentation.\label{fig:rafdb-taskil-aug}]{\includegraphics[width=0.5\textwidth]{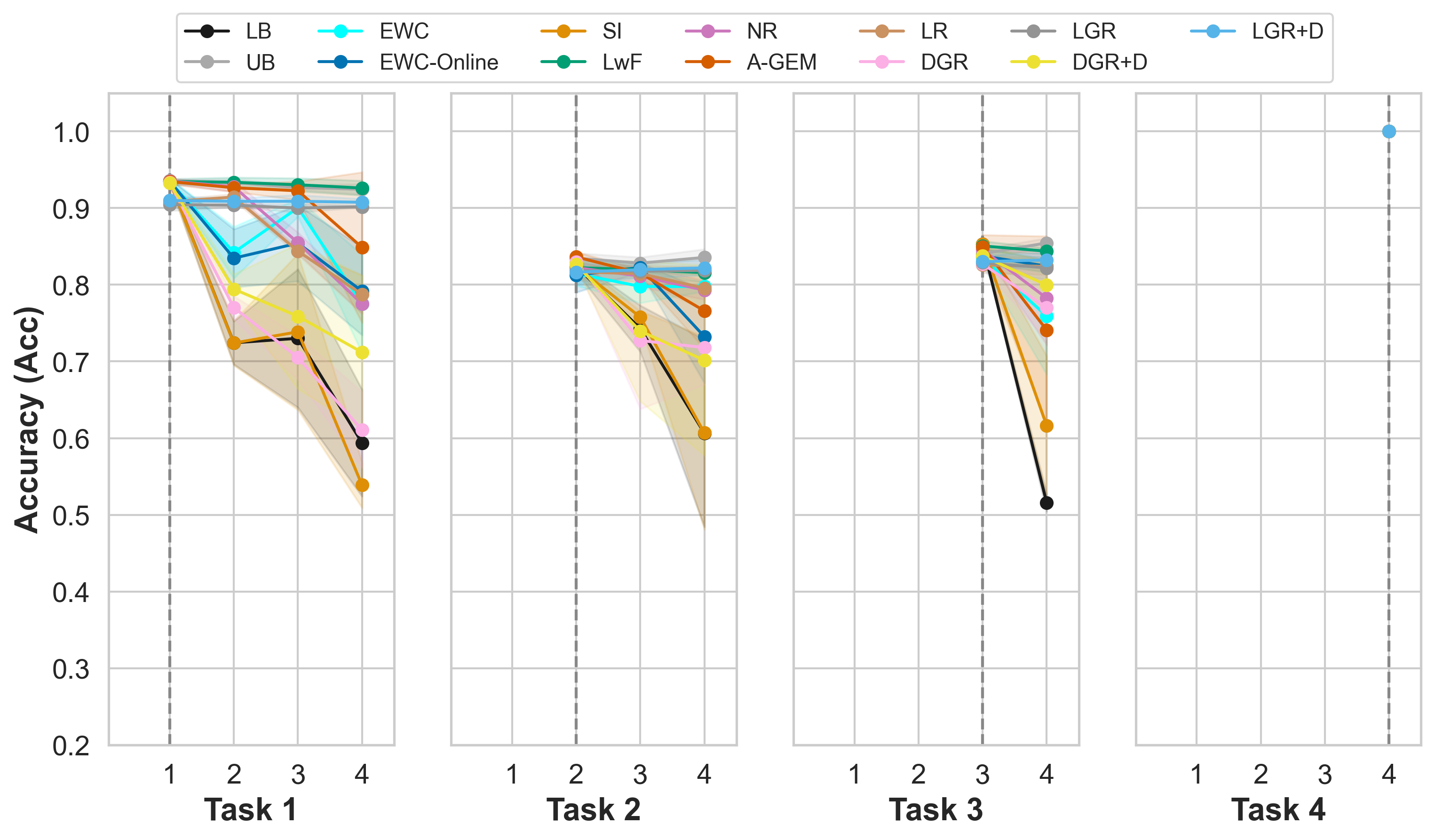}} \\
    \caption[Task-IL Results for \acs{RAF-DB}  W/ and W/O Augmentation.]{\ac{Task-IL} results for \ac{RAF-DB} (a)~without~and~(b)~with augmentation. Test-set accuracy is shown as learning progresses from when a task is introduced to the end of the training.}
    \label{fig:RafDB-taskil} 
    
\end{figure}

\begin{table}[t!]
\centering
\setlength\tabcolsep{3.5pt}

\caption[\acs{Task-IL} \acs{Acc} for \acs{RAF-DB} W/ and W/O Data-Augmentation.]{\acs{Task-IL} \acs{Acc} for \ac{RAF-DB} W/ and W/O Data-Augmentation. \textbf{Bold} values denote best (highest) while [\textit{bracketed}] denote second-best values for each column.}
\label{tab:task-il-rafdb-acc}

{
\scriptsize
\begin{tabular}{l|cccc|cccc}\toprule
\multicolumn{1}{c|}{\textbf{Method}}            & \multicolumn{4}{c|}{\textbf{\acs{Acc} W/O Data-Augmentation}} 
& \multicolumn{4}{c}{\textbf{\acs{Acc} W/ Data-Augmentation}} \\ \midrule

\multicolumn{1}{c|}{ }                          & 
\multicolumn{1}{c|}{\textbf{Task 1}}            & \multicolumn{1}{c|}{\textbf{Task 2}}          &
\multicolumn{1}{c|}{\textbf{Task 3}}            & \multicolumn{1}{c|}{\textbf{Task 4}}           &
\multicolumn{1}{c|}{\textbf{Task 1}}            & \multicolumn{1}{c|}{\textbf{Task 2}}          &
\multicolumn{1}{c|}{\textbf{Task 3}}            & \multicolumn{1}{c}{\textbf{Task 4}}          \\ \cmidrule{2-9}

\multicolumn{9}{c}{\textbf{Baseline Approaches}} \\ \midrule

                      LB & \cellcolor{gray!25}$\bm{0.93\pm0.01}$ & $0.86\pm0.01$ & $0.80\pm0.03$ & $0.63\pm0.05$ & [$0.93\pm0.00$] & $0.77\pm0.02$ & $0.78\pm0.03$ & $0.68\pm0.05$ \\
                      UB & \cellcolor{gray!25}$\bm{0.93\pm0.01}$ & \cellcolor{gray!25}$\bm{0.90\pm0.01}$ & \cellcolor{gray!25}$\bm{0.89\pm0.00}$ & \cellcolor{gray!25}$\bm{0.91\pm0.00}$ & \cellcolor{gray!25}$\bm{0.94\pm0.00}$ & \cellcolor{gray!25}$\bm{0.88\pm0.00}$ & \cellcolor{gray!25}$\bm{0.87\pm0.00}$ & \cellcolor{gray!25}$\bm{0.90\pm0.00}$ \\\midrule

\multicolumn{9}{c}{\textbf{Regularisation-Based Approaches}} \\ \midrule

 EWC  & \cellcolor{gray!25}$\bm{0.93\pm0.00}$ & $0.88\pm0.00$ & $0.84\pm0.03$ & [$0.89\pm0.01$] & [$0.93\pm0.00$] & $0.83\pm0.02$ & $0.84\pm0.01$ & $0.84\pm0.02$ \\
EWC-Online  & \cellcolor{gray!25}$\bm{0.93\pm0.00}$ & $0.88\pm0.01$ & $0.84\pm0.02$ & $0.84\pm0.03$ & [$0.93\pm0.00$] & $0.82\pm0.03$ & $0.84\pm0.02$ & $0.84\pm0.02$ \\
         SI  & \cellcolor{gray!25}$\bm{0.93\pm0.00}$ & $0.85\pm0.00$ & $0.80\pm0.01$ & $0.61\pm0.08$ & [$0.93\pm0.00$] & $0.77\pm0.02$ & $0.78\pm0.03$ & $0.69\pm0.05$ \\
                 LwF  & \cellcolor{gray!25}$\bm{0.93\pm0.00}$ & [$0.89\pm0.01$] & [$0.88\pm0.00$] & \cellcolor{gray!25}$\bm{0.91\pm0.00}$ & [$0.93\pm0.00$] & \cellcolor{gray!25}$\bm{0.88\pm0.00}$ & [$0.86\pm0.00$] & \cellcolor{gray!25}$\bm{0.90\pm0.00}$ \\ \midrule
                 
\multicolumn{9}{c}{\textbf{Replay-Based Approaches}} \\ \midrule

               NR  & \cellcolor{gray!25}$\bm{0.93\pm0.00}$ & [$0.89\pm0.00$] & $0.87\pm0.00$ & $0.87\pm0.01$ & [$0.93\pm0.00$] & [$0.87\pm0.00$] & $0.83\pm0.00$ & $0.84\pm0.01$ \\
   A-GEM  & \cellcolor{gray!25}$\bm{0.93\pm0.00}$ & [$0.89\pm0.00$] & [$0.88\pm0.00$] & $0.82\pm0.08$ & [$0.93\pm0.00$] & \cellcolor{gray!25}$\bm{0.88\pm0.00}$ & [$0.86\pm0.01$] & $0.84\pm0.05$ \\
       LR  & $0.91\pm0.01$ & $0.87\pm0.00$ & $0.85\pm0.01$ & $0.87\pm0.01$ & $0.91\pm0.00$ & $0.86\pm0.00$ & $0.83\pm0.00$ & $0.85\pm0.00$ \\
             DGR  & [$0.92\pm0.01$] & $0.86\pm0.02$ & $0.82\pm0.02$ & $0.84\pm0.01$ & \cellcolor{gray!25}$\bm{0.94\pm0.00}$ & $0.80\pm0.01$ & $0.75\pm0.02$ & $0.77\pm0.02$ \\
           LGR  & $0.91\pm0.01$ & $0.87\pm0.01$ & $0.84\pm0.02$ & $0.86\pm0.01$ & $0.91\pm0.01$ & $0.86\pm0.01$ & $0.85\pm0.00$ & [$0.89\pm0.00$] \\
          DGR+D  & [$0.92\pm0.01$] & $0.87\pm0.02$ & $0.84\pm0.01$ & $0.86\pm0.00$ & \cellcolor{gray!25}$\bm{0.94\pm0.00}$ & $0.81\pm0.01$ & $0.78\pm0.02$ & $0.80\pm0.02$ \\
         LGR+D  & $0.91\pm0.01$ & $0.87\pm0.00$ & $0.86\pm0.01$ & [$0.89\pm0.01$] & $0.91\pm0.00$ & $0.86\pm0.01$ & $0.85\pm0.00$ & [$0.89\pm0.00$] \\

\bottomrule

\end{tabular}

}
\end{table}
\paragraph{\acf{Task-IL}:} For \ac{Task-IL} evaluations, the $7$ expression classes from the \ac{RAF-DB} dataset are split into $4$ tasks, Tasks~$1-3$ consisting of $2$ classes each and Task~$4$ with just one class. Table~\ref{tab:task-il-rafdb-acc} presents model \ac{Acc} scores under \acs{Task-IL} settings for \acs{RAF-DB}, both without and with data augmentation. The larger size of the \acs{RAF-DB} dataset provides more data samples for regularisation-based methods to perform at par, if not better, with replay-based methods. The \ac{LwF} method, in particular, is able to learn task-specific parameters for old and new tasks, closing the gap to \ac{UB} scores. Since \ac{NR} and A-\ac{GEM} are able to store only a subset of samples in the memory buffer, their performance takes a slight hit. However, they are able to retain their performance across the tasks, achieving the lowest \ac{CF} scores (Table~\ref{tab:task-il-rafdb-cf}). The additional data also allows generative replay methods like \ac{DGR} and \ac{DGR}+D to improve performance, compared to \ac{CK+} evaluations. This also enables \ac{LGR} and \ac{LGR}+D to learn meaningful latent features enabling a successful pseudo-rehearsal. The model performances on each task, from the time it is introduced, can be seen in Figure~\ref{fig:rafdb-task-il-noaug}. It shows the evolution of task-specific \acs{Acc} scores for each of the compared approaches as the training progresses from Task~$1-4$.

\begin{table}[t!]
\centering
\caption[\acs{Task-IL} \acs{CF} scores for \acs{RAF-DB} W/ and W/O Data-Augmentation.]{\acs{Task-IL} \acs{CF} scores for \ac{RAF-DB} W/ and W/O Data-Augmentation. \textbf{Bold} values denote best (lowest) while [\textit{bracketed}] denote second-best values for each column.}
\label{tab:task-il-rafdb-cf}
{
\scriptsize
\setlength\tabcolsep{3.5pt}

\begin{tabular}{l|ccc|ccc}\toprule
\multicolumn{1}{c|}{\textbf{Method}}            & \multicolumn{3}{c|}{\textbf{\acs{CF} W/O Data-Augmentation}} 
& \multicolumn{3}{c}{\textbf{\acs{CF}  W/ Data-Augmentation}} \\ \cmidrule{2-7}

\multicolumn{1}{c|}{ }                          & 
\multicolumn{1}{c|}{\textbf{Task 2}}          &
\multicolumn{1}{c|}{\textbf{Task 3}}            & \multicolumn{1}{c|}{\textbf{Task 4}}           &
\multicolumn{1}{c|}{\textbf{Task 2}}          &
\multicolumn{1}{c|}{\textbf{Task 3}}            & \multicolumn{1}{c}{\textbf{Task 4}}          \\ \midrule
                      \multicolumn{7}{c}{\textbf{Baseline Approaches}} \\ \midrule

                      LB 
                      &  $0.28\pm0.09$ & $0.58\pm0.11$ &  $0.08\pm0.00$ 
                      &  $0.29\pm0.08$ &  $0.45\pm0.10$ &  $0.21\pm0.03$ \\
                      UB 
                      & [$-0.00\pm0.01$] & \cellcolor{gray!25}$\bm{0.00\pm0.01}$ & \cellcolor{gray!25}$\bm{-0.01\pm0.00}$ 
                      &  $0.02\pm0.00$ & \cellcolor{gray!25}$\bm{0.00\pm0.01}$ &  [$0.00\pm0.00$] \\\midrule

\multicolumn{7}{c}{\textbf{Regularisation-Based Approaches}} \\ \midrule
 EWC  
 &  $0.14\pm0.06$ & $0.04\pm0.02$ &  $0.02\pm0.01$ 
 &  $0.05\pm0.04$ &  $0.13\pm0.03$ &  $0.09\pm0.03$ \\
EWC-Online  
&  $0.13\pm0.04$ & $0.15\pm0.05$ &  $0.02\pm0.01$ 
&  $0.07\pm0.07$ &  $0.12\pm0.04$ &  $0.10\pm0.04$ \\
 SI  
         &  $0.26\pm0.04$ & $0.61\pm0.17$ &  $0.07\pm0.00$ 
         &  $0.26\pm0.10$ &  $0.42\pm0.10$ &  $0.21\pm0.03$ \\
LwF  
& [$-0.00\pm0.01$] & [$0.01\pm0.01$] &  $0.00\pm0.00$ 
&  [$0.01\pm0.00$] &  [$0.01\pm0.00$] & [$0.00\pm0.00$]\\ \midrule

\multicolumn{7}{c}{\textbf{Replay-Based Approaches}} \\ \midrule

NR  
&  $0.04\pm0.03$ & $0.09\pm0.02$ & \cellcolor{gray!25}$\bm{-0.01\pm0.00}$ 
&  $0.09\pm0.00$ &  $0.13\pm0.03$ &  $0.01\pm0.01$ \\
A-GEM  
& \cellcolor{gray!25}$\bm{-0.01\pm0.01}$ & $0.19\pm0.15$ &  [$0.00\pm0.00$] 
&  $0.03\pm0.01$ &  $0.13\pm0.10$ &  $0.01\pm0.00$ \\
LR  
&  $0.02\pm0.01$ & $0.05\pm0.01$ & [$0.00\pm0.01$] 
&  $0.07\pm0.01$ &  $0.07\pm0.00$ & \cellcolor{gray!25}$\bm{-0.01\pm0.00}$ \\
DGR  
&  $0.18\pm0.05$ & $0.14\pm0.01$ &  $0.07\pm0.02$ 
&  $0.33\pm0.06$ &  $0.25\pm0.04$ &  $0.16\pm0.02$ \\
LGR  
&  $0.07\pm0.03$ & $0.08\pm0.02$ &  $0.01\pm0.01$ 
&  \cellcolor{gray!25}$\bm{0.00\pm0.00}$ &  \cellcolor{gray!25}$\bm{0.00\pm0.01}$ &  [$0.00\pm0.01$] \\
DGR+D  
&  $0.13\pm0.03$ & $0.10\pm0.01$ &  $0.03\pm0.02$ 
&  $0.26\pm0.06$ &  $0.19\pm0.04$ &  $0.14\pm0.02$ \\
LGR+D  
&  $0.02\pm0.00$ & $0.02\pm0.01$ &  [$0.00\pm0.00$] 
& \cellcolor{gray!25}$\bm{0.00\pm0.00}$ & \cellcolor{gray!25}$\bm{0.00\pm0.00}$ &  [$0.00\pm0.01$] \\

\bottomrule
\end{tabular}
}
\end{table}

Data-augmentation does not seem to have a significant impact on the overall model performances on the \acs{RAF-DB} datasets. On the contrary, as seen in Table~\ref{tab:task-il-rafdb-acc}, the regularisation-based methods witness a decrease in model performance. This can be due to the fact that with the increased amount of data, with all its variation, the regularisation-based methods find it hard to assign task-specific priorities to weight updates, resulting in the models saturating their capacity to learn new tasks while preserving past knowledge. \ac{LwF}, despite this decrease, is able to match up to the \ac{UB} score. Replay-based methods, on the other hand, see more variance in how the different methods respond to this additional data. \ac{NR} finds it even harder to store representative data samples for each task in the memory buffer, resulting in an overall, albeit insignificant, decline in model performance. Similarly, \ac{DGR} and \ac{DGR}+D find it harder to deal with the variation, in terms of rotation and flipping, in the data samples to reconstruct high quality pseudo-samples. \ac{LGR} and \ac{LGR}+D, on the other hand are able to deal with such variation quite well, allowing the models to learn robust latent feature representations. This is also witnessed in Table~\ref{tab:task-il-rafdb-cf} where these methods are seen to achieve the lowest \ac{CF} scores.

\nl\textit{Task Ordering:} The two additional task orderings (O2 and O3) are also explored for \acs{RAF-DB} but no significant differences in model performances are witnessed. Similar to the original results without and with augmentation, \ac{LwF} is able match up to the \ac{UB} evaluations with \ac{NR} and A-\ac{GEM} performing the second-best. More details on the results from the task-ordering experiments on \acs{RAF-DB} can be found in Appendix~\ref{app:taskorderrafdb}.



\begin{table}[t!]
\centering
\setlength\tabcolsep{3.5pt}
\caption[\acs{Class-IL} \acs{Acc} for \acs{RAF-DB} W/ and W/O Data-Augmentation.]{\acs{Class-IL} \acs{Acc} for \ac{RAF-DB} W/ and W/O Data-Augmentation. \textbf{Bold} values denote best (highest) while [\textit{bracketed}] denote second-best values for each column.}
\label{tab:class-il-rafdb-acc}

{
\scriptsize
\begin{tabular}{l|ccccccc}\toprule

\multirow{2}[2]{*}{\makecell[c]{\textbf{Method}}}           & \multicolumn{7}{c}{\textbf{\acs{Acc} W/O Data-Augmentation}} \\ \cmidrule{2-8}

\multicolumn{1}{c|}{ }                          & 
\multicolumn{1}{c|}{\textbf{Class 1}}            & \multicolumn{1}{c|}{\textbf{Class 2}}          &
\multicolumn{1}{c|}{\textbf{Class 3}}            & \multicolumn{1}{c|}{\textbf{Class 4}}           &
\multicolumn{1}{c|}{\textbf{Class 5}}            & \multicolumn{1}{c|}{\textbf{Class 6}}          &
\multicolumn{1}{c}{\textbf{Class 7}}                     \\ \midrule

\multicolumn{8}{c}{\textbf{Baseline Approaches}} \\ \midrule
LB & \cellcolor{gray!25}$\bm{1.00\pm0.00}$ & $0.50\pm0.00$ & $0.33\pm0.00$ & $0.25\pm0.00$ & $0.20\pm0.00$ & $0.17\pm0.00$ & $0.14\pm0.00$ \\
UB & \cellcolor{gray!25}$\bm{1.00\pm0.00}$ & [$0.86\pm0.00$] & \cellcolor{gray!25}$\bm{0.69\pm0.00}$ & \cellcolor{gray!25}$\bm{0.64\pm0.01}$ & \cellcolor{gray!25}$\bm{0.62\pm0.01}$ & \cellcolor{gray!25}$\bm{0.54\pm0.02}$ & \cellcolor{gray!25}$\bm{0.55\pm0.02}$ \\\midrule

\multicolumn{8}{c}{\textbf{Regularisation-Based Approaches}} \\ \midrule

EWC  & \cellcolor{gray!25}$\bm{1.00\pm0.00}$ & $0.50\pm0.00$ & $0.33\pm0.00$ & $0.25\pm0.00$ & $0.20\pm0.00$ & $0.17\pm0.00$ & $0.14\pm0.00$ \\
EWC-Online  & \cellcolor{gray!25}$\bm{1.00\pm0.00}$ & $0.50\pm0.00$ & $0.33\pm0.00$ & $0.25\pm0.00$ & $0.20\pm0.00$ & $0.17\pm0.00$ & $0.14\pm0.00$ \\
SI  & \cellcolor{gray!25}$\bm{1.00\pm0.00}$ & $0.50\pm0.00$ & $0.33\pm0.00$ & $0.25\pm0.00$ & $0.20\pm0.00$ & $0.17\pm0.00$ & $0.14\pm0.00$ \\
LwF  & \cellcolor{gray!25}$\bm{1.00\pm0.00}$ & $0.50\pm0.00$ & $0.33\pm0.00$ & $0.25\pm0.00$ & $0.20\pm0.00$ & $0.17\pm0.00$ & $0.14\pm0.00$ \\ \midrule

\multicolumn{8}{c}{\textbf{Replay-Based Approaches}} \\ \midrule
NR  & \cellcolor{gray!25}$\bm{1.00\pm0.00}$ & \cellcolor{gray!25}$\bm{0.88\pm0.01}$ & [$0.67\pm0.01$] & [$0.62\pm0.01$] & $0.54\pm0.01$ & $0.46\pm0.01$ & $0.47\pm0.01$ \\
A-GEM  & \cellcolor{gray!25}$\bm{1.00\pm0.00}$ & $0.53\pm0.04$ & $0.33\pm0.00$ & $0.25\pm0.00$ & $0.20\pm0.00$ & $0.17\pm0.00$ & $0.14\pm0.00$ \\
LR  & \cellcolor{gray!25}$\bm{1.00\pm0.00}$ & $0.85\pm0.03$ & $0.62\pm0.04$ & $0.60\pm0.04$ & [$0.58\pm0.03$] & [$0.51\pm0.03$] & [$0.50\pm0.03$] \\
DGR  & \cellcolor{gray!25}$\bm{1.00\pm0.00}$ & $0.50\pm0.00$ & $0.33\pm0.00$ & $0.25\pm0.00$ & $0.20\pm0.00$ & $0.17\pm0.00$ & $0.15\pm0.00$ \\
LGR  & \cellcolor{gray!25}$\bm{1.00\pm0.00}$ & $0.72\pm0.02$ & $0.46\pm0.03$ & $0.39\pm0.08$ & $0.41\pm0.06$ & $0.36\pm0.06$ & $0.32\pm0.12$ \\
DGR+D  & \cellcolor{gray!25}$\bm{1.00\pm0.00}$ & $0.50\pm0.00$ & $0.33\pm0.00$ & $0.25\pm0.00$ & $0.20\pm0.00$ & $0.17\pm0.00$ & $0.15\pm0.00$ \\
LGR+D  & \cellcolor{gray!25}$\bm{1.00\pm0.00}$ & $0.80\pm0.02$ & $0.52\pm0.03$ & $0.44\pm0.05$ & $0.46\pm0.04$ & $0.39\pm0.04$ & $0.37\pm0.06$ \\

\midrule

\multirow{2}[2]{*}{\makecell[c]{\textbf{Method}}}           & \multicolumn{7}{c}{\textbf{\acs{Acc} W/ Data-Augmentation}} \\ \cmidrule{2-8}

\multicolumn{1}{c|}{ }                          & 
\multicolumn{1}{c|}{\textbf{Class 1}}            & \multicolumn{1}{c|}{\textbf{Class 2}}          &
\multicolumn{1}{c|}{\textbf{Class 3}}            & \multicolumn{1}{c|}{\textbf{Class 4}}           &
\multicolumn{1}{c|}{\textbf{Class 5}}            & \multicolumn{1}{c|}{\textbf{Class 6}}          &
\multicolumn{1}{c}{\textbf{Class 7}}                     \\ \midrule

\multicolumn{8}{c}{\textbf{Baseline Approaches}} \\ \midrule

LB & \cellcolor{gray!25}$\bm{1.00\pm0.00}$ & $0.50\pm0.00$ & $0.33\pm0.00$ & $0.25\pm0.00$ & $0.20\pm0.00$ & $0.17\pm0.00$ & $0.14\pm0.00$ \\
UB & \cellcolor{gray!25}$\bm{1.00\pm0.00}$ & \cellcolor{gray!25}$\bm{0.93\pm0.01}$ & \cellcolor{gray!25}$\bm{0.77\pm0.02}$ & \cellcolor{gray!25}$\bm{0.71\pm0.02}$ & \cellcolor{gray!25}$\bm{0.73\pm0.01}$ & \cellcolor{gray!25}$\bm{0.66\pm0.01}$ & \cellcolor{gray!25}$\bm{0.64\pm0.01}$ \\ \midrule

\multicolumn{8}{c}{\textbf{Regularisation-Based Approaches}} \\ \midrule

EWC  & \cellcolor{gray!25}$\bm{1.00\pm0.00}$ & $0.50\pm0.00$ & $0.33\pm0.00$ & $0.25\pm0.00$ & $0.20\pm0.00$ & $0.17\pm0.00$ & $0.14\pm0.00$ \\
EWC-Online  & \cellcolor{gray!25}$\bm{1.00\pm0.00}$ & $0.50\pm0.00$ & $0.33\pm0.00$ & $0.25\pm0.00$ & $0.20\pm0.00$ & $0.17\pm0.00$ & $0.14\pm0.00$ \\
SI  & \cellcolor{gray!25}$\bm{1.00\pm0.00}$ & $0.50\pm0.00$ & $0.33\pm0.00$ & $0.25\pm0.00$ & $0.20\pm0.00$ & $0.17\pm0.00$ & $0.14\pm0.00$ \\
LwF  & \cellcolor{gray!25}$\bm{1.00\pm0.00}$ & $0.50\pm0.00$ & $0.33\pm0.00$ & $0.25\pm0.00$ & $0.20\pm0.00$ & $0.17\pm0.00$ & $0.14\pm0.00$ \\ \midrule
\multicolumn{8}{c}{\textbf{Replay-Based Approaches}} \\ \midrule

NR  & \cellcolor{gray!25}$\bm{1.00\pm0.00}$ & \cellcolor{gray!25}$\bm{0.93\pm0.00}$ & [$0.71\pm0.02$] & $0.65\pm0.01$ & $0.63\pm0.01$ & $0.53\pm0.01$ & $0.48\pm0.01$ \\
A-GEM  & \cellcolor{gray!25}$\bm{1.00\pm0.00}$ & $0.51\pm0.01$ & $0.33\pm0.00$ & $0.32\pm0.09$ & $0.22\pm0.03$ & $0.19\pm0.03$ & $0.14\pm0.00$ \\
LR  & \cellcolor{gray!25}$\bm{1.00\pm0.00}$ & [$0.91\pm0.01$] & [$0.71\pm0.01$] & [$0.66\pm0.02$] & [$0.66\pm0.01$] & [$0.59\pm0.00$] & [$0.57\pm0.00$] \\
DGR~  & \cellcolor{gray!25}$\bm{1.00\pm0.00}$ & $0.50\pm0.00$ & $0.34\pm0.00$ & $0.26\pm0.00$ & $0.20\pm0.00$ & $0.17\pm0.00$ & $0.15\pm0.00$ \\
LGR  & \cellcolor{gray!25}$\bm{1.00\pm0.00}$ & $0.84\pm0.02$ & $0.61\pm0.01$ & $0.61\pm0.01$ & $0.60\pm0.01$ & $0.48\pm0.02$ & $0.47\pm0.01$ \\
DGR+D  & \cellcolor{gray!25}$\bm{1.00\pm0.00}$ & $0.50\pm0.00$ & $0.34\pm0.00$ & $0.25\pm0.00$ & $0.20\pm0.00$ & $0.17\pm0.00$ & $0.14\pm0.00$ \\
LGR+D  & \cellcolor{gray!25}$\bm{1.00\pm0.00}$ & $0.89\pm0.02$ & $0.66\pm0.02$ & $0.64\pm0.01$ & $0.63\pm0.01$ & $0.54\pm0.00$ & $0.51\pm0.00$ \\

\bottomrule

\end{tabular}
\vspace{-3mm}
}
\end{table}

\begin{figure}[t!]  
    \centering
    \subfloat[\ac{Class-IL} Results W/o Augmentation.\label{fig:rafdb-class-il-noaug}]{\includegraphics[width=0.5\textwidth]{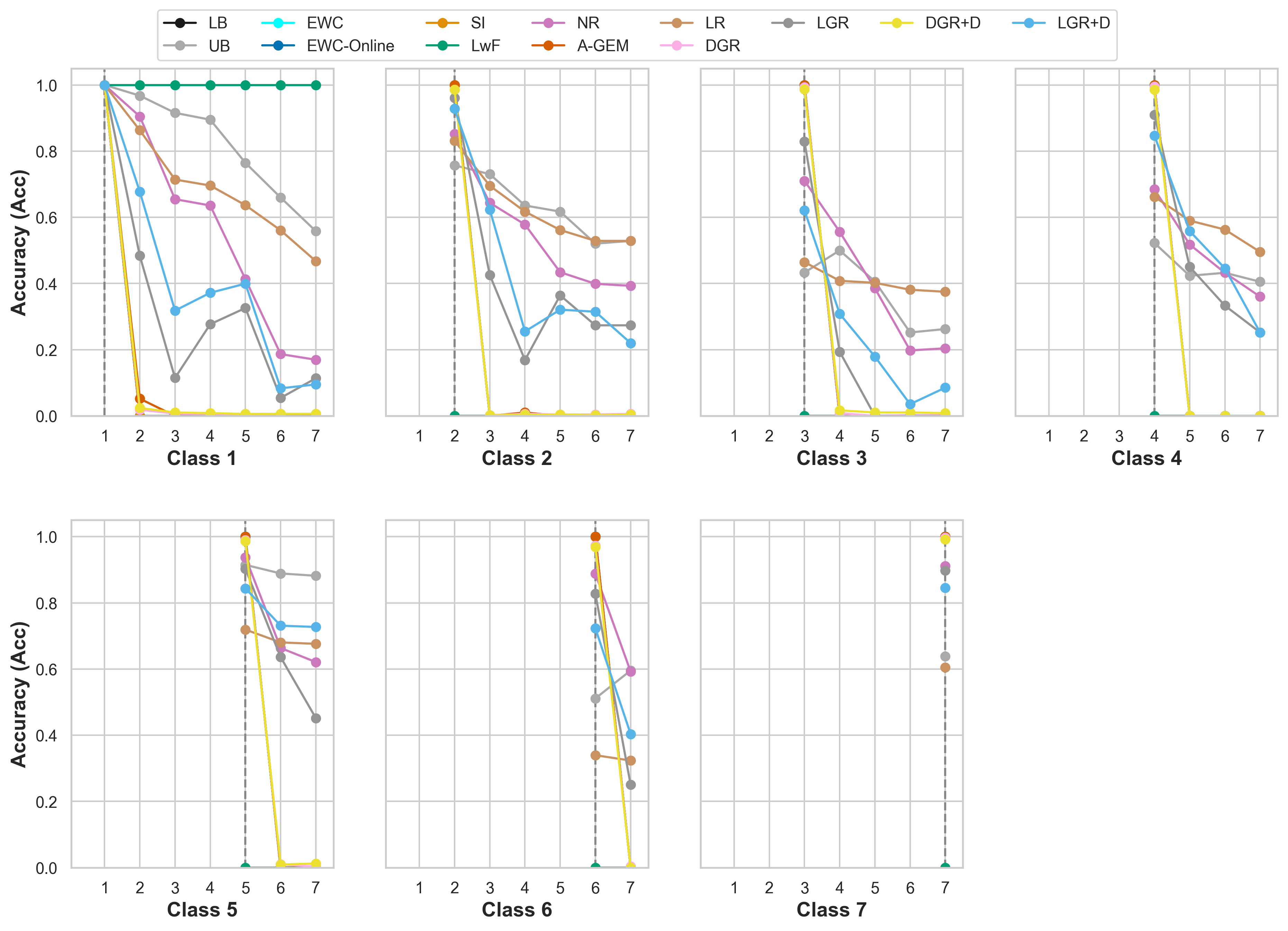}} \hfill
    \subfloat[\ac{Class-IL} Results W/ Augmentation.\label{fig:rafdb-class-il-aug}]{\includegraphics[width=0.5\textwidth]{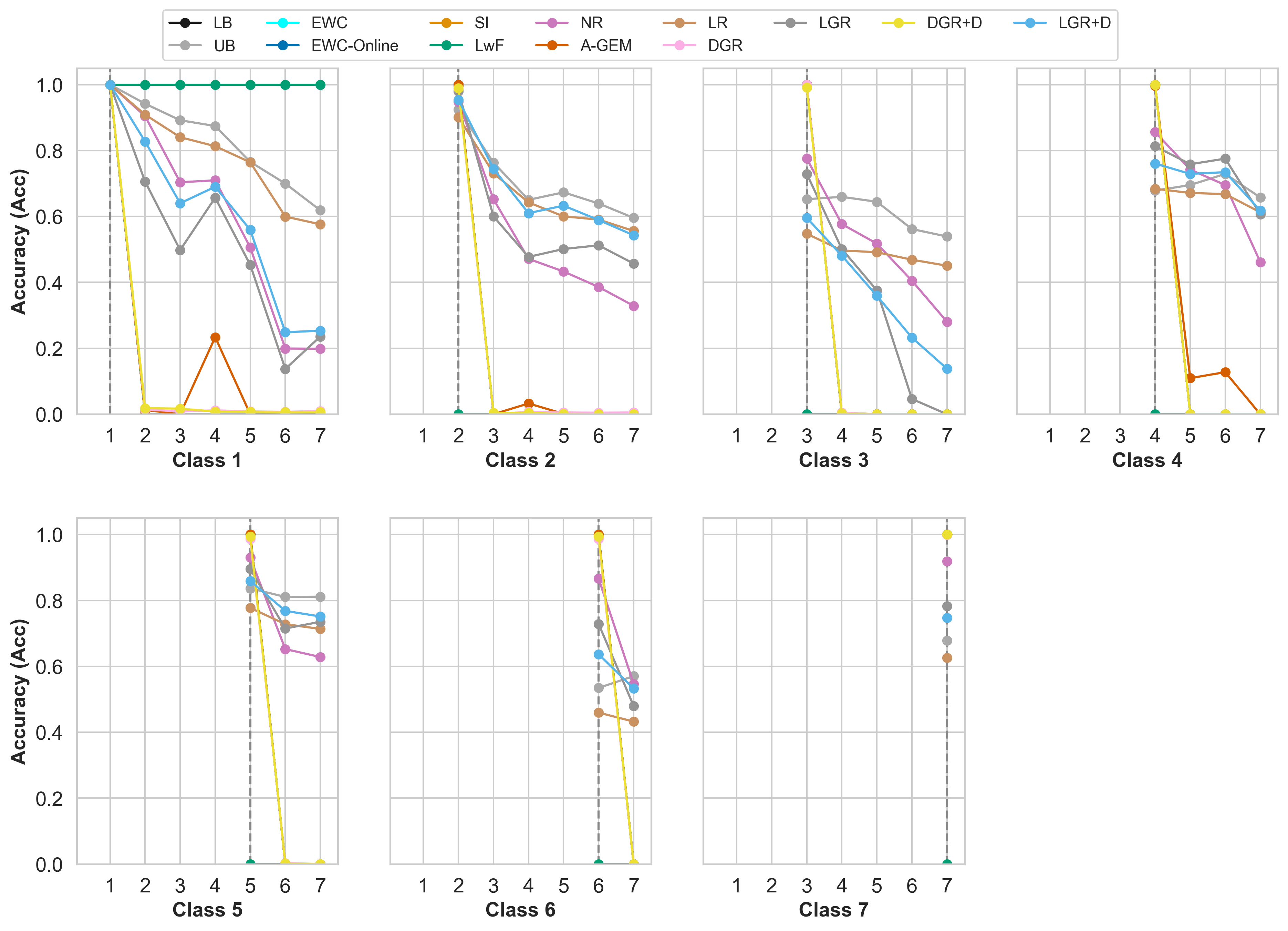}} \\
    \caption[Class-IL Results for \acs{RAF-DB}  W/ and W/O Augmentation.]{\ac{Class-IL} results for \ac{RAF-DB} (a)~without and (b)~with augmentation.~Test-set~accuracy is shown as the learning progresses from when a class is introduced to the end of the training.}
    \label{fig:RafDB-classil} 
    
\end{figure}

\begin{table}[t!]
\centering
\caption[\acs{Class-IL} \acs{CF} scores for \acs{RAF-DB}  W/ and W/O Data-Augmentation.]{\acs{Class-IL} \acs{CF} scores for \ac{RAF-DB}  W/ and W/O Data-Augmentation. \textbf{Bold} values denote best (lowest) while [\textit{bracketed}] denote second-best values for each column.}
\label{tab:class-il-rafdb-cf}

{
\scriptsize
\setlength\tabcolsep{3.5pt}

\begin{tabular}{l|cccccc}\toprule

\multirow{2}[2]{*}{\makecell[c]{\textbf{Method}}}           & \multicolumn{6}{c}{\textbf{\acs{CF} W/O Data-Augmentation}} \\ \cmidrule{2-7}

\multicolumn{1}{c|}{ }                          & 
\multicolumn{1}{c|}{\textbf{Class 2}}          &
\multicolumn{1}{c|}{\textbf{Class 3}}            & \multicolumn{1}{c|}{\textbf{Class 4}}           &
\multicolumn{1}{c|}{\textbf{Class 5}}            & \multicolumn{1}{c|}{\textbf{Class 6}}          &
\multicolumn{1}{c}{\textbf{Class 7}}                    \\ \midrule

\multicolumn{7}{c}{\textbf{Baseline Approaches}} \\ \midrule

                      LB & $1.00\pm0.00$ & $1.00\pm0.00$ & $1.00\pm0.00$ & $1.00\pm0.00$ & $1.00\pm0.00$ & $1.00\pm0.00$ \\
                      UB & \cellcolor{gray!25}$\bm{0.06\pm0.01}$ & \cellcolor{gray!25}$\bm{0.05\pm0.01}$ & \cellcolor{gray!25}$\bm{0.13\pm0.02}$ & \cellcolor{gray!25}$\bm{0.17\pm0.01}$ & \cellcolor{gray!25}$\bm{0.15\pm0.00}$ & \cellcolor{gray!25}$\bm{0.10\pm0.00}$ \\\midrule

\multicolumn{7}{c}{\textbf{Regularisation-Based Approaches}} \\ \midrule

 EWC  & $1.00\pm0.00$ & $1.00\pm0.00$ & $1.00\pm0.00$ & $1.00\pm0.00$ & $1.00\pm0.00$ & $1.00\pm0.00$ \\
EWC-Online  & $1.00\pm0.00$ & $1.00\pm0.00$ & $1.00\pm0.00$ & $1.00\pm0.00$ & $1.00\pm0.00$ & $1.00\pm0.00$ \\
         SI  & $1.00\pm0.00$ & $1.00\pm0.00$ & $1.00\pm0.00$ & $1.00\pm0.00$ & $1.00\pm0.00$ & $1.00\pm0.00$ \\
                 LwF  & $1.00\pm0.00$ & $1.00\pm0.00$ & $1.00\pm0.00$ & $1.00\pm0.00$ & $1.00\pm0.00$ & $1.00\pm0.00$ \\ \midrule

\multicolumn{7}{c}{\textbf{Replay-Based Approaches}} \\ \midrule

               NR  & $0.28\pm0.04$ & $0.26\pm0.02$ & $0.37\pm0.00$ & $0.46\pm0.01$ & $0.46\pm0.01$ & $0.42\pm0.00$ \\
   A-GEM  & $1.00\pm0.00$ & $0.99\pm0.00$ & $1.00\pm0.00$ & $1.00\pm0.00$ & $1.00\pm0.00$ & $1.00\pm0.00$ \\
       LR  & [$0.21\pm0.02$] & [$0.19\pm0.02$] & [$0.19\pm0.02$] & [$0.19\pm0.00$] & [$0.19\pm0.02$] & [$0.21\pm0.00$] \\
             DGR~  & $0.99\pm0.00$ & $0.99\pm0.00$ & $0.99\pm0.00$ & $0.99\pm0.01$ & $0.99\pm0.01$ & $0.99\pm0.00$ \\
           LGR  & $0.71\pm0.07$ & $0.72\pm0.12$ & $0.64\pm0.10$ & $0.66\pm0.10$ & $0.68\pm0.17$ & $0.50\pm0.00$ \\
          DGR+D  & $0.99\pm0.00$ & $0.98\pm0.01$ & $0.98\pm0.01$ & $0.98\pm0.01$ & $0.98\pm0.01$ & $0.99\pm0.00$ \\
         LGR+D  & $0.49\pm0.08$ & $0.54\pm0.09$ & $0.48\pm0.07$ & $0.53\pm0.07$ & $0.53\pm0.09$ & $0.49\pm0.00$ \\\midrule

\multirow{2}[2]{*}{\makecell[c]{\textbf{Method}}}          & \multicolumn{6}{c}{\textbf{\acs{CF} W/ Data-Augmentation}} \\ \cmidrule{2-7}

\multicolumn{1}{c|}{ }                          & 
\multicolumn{1}{c|}{\textbf{Class 2}}          &
\multicolumn{1}{c|}{\textbf{Class 3}}            & \multicolumn{1}{c|}{\textbf{Class 4}}           &
\multicolumn{1}{c|}{\textbf{Class 5}}            & \multicolumn{1}{c|}{\textbf{Class 6}}          &
\multicolumn{1}{c}{\textbf{Class 7}}                    \\ \midrule
\multicolumn{7}{c}{\textbf{Baseline Approaches}} \\ \midrule
LB & $1.00\pm0.00$ & $1.00\pm0.00$ & $1.00\pm0.00$ & $1.00\pm0.00$ & $1.00\pm0.00$ & $1.00\pm0.00$ \\
UB & \cellcolor{gray!25}$\bm{0.13\pm0.02}$ & \cellcolor{gray!25}$\bm{0.13\pm0.01}$ & \cellcolor{gray!25}$\bm{0.12\pm0.01}$ & \cellcolor{gray!25}$\bm{0.13\pm0.01}$ & \cellcolor{gray!25}$\bm{0.14\pm0.02}$ & \cellcolor{gray!25}$\bm{0.15\pm0.00}$ \\ \midrule

\multicolumn{7}{c}{\textbf{Regularisation-Based Approaches}} \\ \midrule

EWC  & $1.00\pm0.00$ & $1.00\pm0.00$ & $1.00\pm0.00$ & $1.00\pm0.00$ & $1.00\pm0.00$ & $1.00\pm0.00$ \\
EWC-Online  & $1.00\pm0.00$ & $1.00\pm0.00$ & $1.00\pm0.00$ & $1.00\pm0.00$ & $1.00\pm0.00$ & $1.00\pm0.00$ \\
SI  & $1.00\pm0.00$ & $1.00\pm0.00$ & $1.00\pm0.00$ & $1.00\pm0.00$ & $1.00\pm0.00$ & $1.00\pm0.00$ \\
LwF  & $1.00\pm0.00$ & $1.00\pm0.00$ & $1.00\pm0.00$ & $1.00\pm0.00$ & $1.00\pm0.00$ & $1.00\pm0.00$ \\ \midrule

\multicolumn{7}{c}{\textbf{Replay-Based Approaches}} \\ \midrule

NR  & $0.30\pm0.03$ & $0.32\pm0.01$ & $0.35\pm0.02$ & $0.43\pm0.02$ & $0.49\pm0.01$ & $0.42\pm0.00$ \\
A-GEM  & $1.00\pm0.00$ & $0.91\pm0.13$ & $0.97\pm0.04$ & $0.97\pm0.04$ & $1.00\pm0.00$ & $1.00\pm0.00$ \\
LR  & [$0.17\pm0.01$] & [$0.17\pm0.02$] & [$0.15\pm0.01$] & [$0.17\pm0.01$] & [$0.17\pm0.01$] & [$0.13\pm0.00$] \\
DGR~  & $0.99\pm0.00$ & $0.99\pm0.00$ & $0.99\pm0.00$ & $0.99\pm0.00$ & $0.99\pm0.00$ & $0.98\pm0.00$ \\
LGR  & $0.44\pm0.01$ & $0.36\pm0.01$ & $0.36\pm0.02$ & $0.45\pm0.04$ & $0.44\pm0.02$ & $0.44\pm0.00$ \\
DGR+D  & $0.98\pm0.01$ & $0.99\pm0.00$ & $0.99\pm0.00$ & $0.99\pm0.00$ & $0.99\pm0.00$ & $0.99\pm0.00$ \\
LGR+D  & $0.28\pm0.02$ & $0.26\pm0.00$ & $0.26\pm0.00$ & $0.32\pm0.01$ & $0.33\pm0.01$ & $0.33\pm0.00$ \\

\bottomrule

\end{tabular}

}
\end{table}
\paragraph{\acf{Class-IL}:} Table~\ref{tab:class-il-rafdb-acc} presents model performances, in terms of \ac{Acc} scores, incrementally learning the $7$ classes for \acs{RAF-DB} both without and with data augmentation. Similar to \acs{Class-IL} results on the \acs{CK+} dataset, regularisation-based methods completely fail at mitigating forgetting in the model for \acs{RAF-DB} as well while replay-based method such as \acs{NR}, \acs{LR}, \acs{LGR} and \acs{LGR}+D are able to preserve model performance to some extent. \ac{NR} starts of as the best performing approach but as the number of classes increase, the fraction of the memory buffer allocated to each class samples decreases. This results in the model not having enough data to efficiently rehearse past knowledge, causing a dip in \ac{NR} \acs{Acc} scores. \ac{LR}, on the other hand, despite having a similar replay mechanisms, albeit with \textit{latent features}, is able to preserve model performance throughout the learning (Table~\ref{tab:class-il-rafdb-cf}), although with only marginally better \acs{Acc} than \acs{NR}. \ac{DGR} and \ac{DGR}+D face a similar fate, as in the case of \acs{CK+} evaluations, with their generator models failing to efficiently reconstruct high-dimensional pseudo-samples, resulting in \textit{forgetting} in the models. \acs{LGR}-based methods are able to learn meaningful latent representations enabling an efficient pseudo-rehearsal of features to mitigate forgetting. The learning dynamics can be seen in Figure~\ref{fig:rafdb-class-il-noaug}. Replay-based methods are seen to preserve model performance as learning progresses across all classes while regularisation-based methods are not able to manage class priorities and are not able to balance learning across all classes.

Even though data-augmentation provides a performance boost for replay-based methods (Table~\ref{tab:class-il-rafdb-acc}), the relative results across the different methods do not change. Similar learning dynamics are witnessed with \acs{NR} starting off as a better performing approach with \ac{LR} closing the gap as the learning progresses across the $7$ classes, eventually resulting with the best performance, second only to \acs{UB} scores. Interestingly, the additional data aids \acs{LGR}-based methods to perform better than \acs{NR} as learning progresses, evidencing an efficient pseudo-rehearsal of latent features to mitigate model performance. This is also witnessed in the better (lower) \ac{CF} scores, especially for \ac{LGR}+D (Table~\ref{tab:class-il-rafdb-cf}) compared to \acs{NR}. The learning dynamics, with data-augmentation, can be seen in Figure~\ref{fig:rafdb-class-il-aug}. 

\nl \textit{Class Ordering:} Evaluating different class orders, similar to \acs{CK+} experiments, no significant differences in model performances are witnessed, with \acs{LR} still emerging as the best performing approach with similar \acs{Acc} and \acs{CF} scores. More details on the results from these class-ordering experiments can be found in Appendix~\ref{app:classorderrafdb}.

\subsection{AffectNet Results}
\label{subsec:affnetresults}
The AffectNet dataset, represents the most challenging of conditions with $\approx287K$ images across $8$ expression classes, representing \textit{in-the-wild} settings. For both \acs{Task-IL} and \acs{Class-IL} settings, instead of performing data-augmentation, the \textit{downsampling} method~\citep{2018Affectnet} is used to balance the dataset (see Section~\ref{sec:aug} for details). These results are presented below.

\begin{table}[t]
\centering
\setlength\tabcolsep{3.5pt}

\caption[\acs{Task-IL} \acs{Acc} for AffectNet using original and downsampled data splits.]{\acs{Task-IL} \acs{Acc} for AffectNet using original and \textit{downsampled} data splits. \textbf{Bold} values denote best (highest) while [\textit{bracketed}] denote second-best values for each column.}
\label{tab:task-il-AffectNet-acc}

{
\scriptsize
\begin{tabular}{l|cccc|cccc}\toprule
\multicolumn{1}{c|}{\textbf{Method}}            & \multicolumn{4}{c|}{\textbf{\acs{Acc} on Original Split}} 
& \multicolumn{4}{c}{\textbf{\acs{Acc} on Downsampled Split}} \\ \midrule

\multicolumn{1}{c|}{ }                          & 
\multicolumn{1}{c|}{\textbf{Task 1}}            & \multicolumn{1}{c|}{\textbf{Task 2}}          &
\multicolumn{1}{c|}{\textbf{Task 3}}            & \multicolumn{1}{c|}{\textbf{Task 4}}           &
\multicolumn{1}{c|}{\textbf{Task 1}}            & \multicolumn{1}{c|}{\textbf{Task 2}}          &
\multicolumn{1}{c|}{\textbf{Task 3}}            & \multicolumn{1}{c}{\textbf{Task 4}}          \\ \cmidrule{2-9}

\multicolumn{9}{c}{\textbf{Baseline Approaches}} \\ \midrule

                      LB                & $0.56\pm0.03$ & $0.65\pm0.01$ & $0.58\pm0.01$ & $0.50\pm0.02$ 
                                        & [$0.68\pm0.02$] & $0.68\pm0.00$ & $0.65\pm0.01$ & $0.60\pm0.01$ \\
                      UB                & [$0.57\pm0.01$] & \cellcolor{gray!25}$\bm{0.68\pm0.00}$ & [$0.61\pm0.01$] & [$0.62\pm0.00$] 
                                        & \cellcolor{gray!25}$\bm{0.69\pm0.00}$ & \cellcolor{gray!25}$\bm{0.71\pm0.00}$ & \cellcolor{gray!25}$\bm{0.75\pm0.00}$ & [$0.74\pm0.00$] \\\midrule

\multicolumn{9}{c}{\textbf{Regularisation-Based Approaches}} \\ \midrule
 EWC   & $0.55\pm0.02$ & [$0.67\pm0.01$] & $0.60\pm0.00$ & \cellcolor{gray!25}$\bm{0.63\pm0.01}$ 
                                        & [$0.68\pm0.02$] & $0.69\pm0.01$ & $0.73\pm0.02$ & $0.72\pm0.01$ \\
EWC-Online   & $0.55\pm0.02$ & [$0.67\pm0.01$] & $0.60\pm0.00$ & \cellcolor{gray!25}$\bm{0.63\pm0.01}$ 
                                        & [$0.68\pm0.02$] & [$0.70\pm0.01$] & $0.72\pm0.02$ & $0.71\pm0.01$ \\
         SI   & $0.56\pm0.03$ & \cellcolor{gray!25}$\bm{0.68\pm0.01}$ & $0.53\pm0.05$ & $0.54\pm0.06$ 
                                        & [$0.68\pm0.02$] & $0.68\pm0.01$ & $0.71\pm0.01$ & $0.70\pm0.00$ \\
                 LwF   & $0.55\pm0.02$ & $0.62\pm0.01$ & $0.58\pm0.00$ & $0.60\pm0.00$ 
                                        & [$0.68\pm0.02$] & [$0.70\pm0.01$] & [$0.74\pm0.00$] & [$0.74\pm0.01$] \\\midrule
                 
                 \multicolumn{9}{c}{\textbf{Replay-Based Approaches}} \\ \midrule

               NR   & $0.56\pm0.03$ & \cellcolor{gray!25}$\bm{0.68\pm0.00}$ & \cellcolor{gray!25}$\bm{0.62\pm0.01}$ & $0.59\pm0.01$ 
                                        & [$0.68\pm0.01$] & $0.69\pm0.00$ & $0.73\pm0.00$ & $0.69\pm0.00$ \\
   A-GEM   & $0.56\pm0.03$ & $0.66\pm0.00$ & $0.60\pm0.01$ & \cellcolor{gray!25}$\bm{0.63\pm0.01}$ 
                                        & [$0.68\pm0.02$] & \cellcolor{gray!25}$\bm{0.71\pm0.01}$ & \cellcolor{gray!25}$\bm{0.75\pm0.00}$ & \cellcolor{gray!25}$\bm{0.75\pm0.00}$ \\
       LR   & $0.51\pm0.01$ & $0.57\pm0.02$ & $0.56\pm0.01$ & $0.54\pm0.01$ 
                                        & $0.66\pm0.00$ & $0.65\pm0.01$ & $0.68\pm0.01$ & $0.66\pm0.01$ \\
             DGR   & \cellcolor{gray!25}$\bm{0.59\pm0.02}$ & $0.61\pm0.00$ & $0.52\pm0.03$ & $0.54\pm0.01$ 
                                        & \cellcolor{gray!25}$\bm{0.69\pm0.01}$ & $0.66\pm0.02$ & $0.67\pm0.03$ & $0.66\pm0.03$ \\
           LGR   & $0.51\pm0.01$ & $0.56\pm0.01$ & $0.53\pm0.01$ & $0.52\pm0.01$ 
                                        & $0.66\pm0.01$ & $0.63\pm0.01$ & $0.67\pm0.00$ & $0.65\pm0.00$ \\
          DGR+D   & \cellcolor{gray!25}$\bm{0.59\pm0.02}$ & $0.64\pm0.01$ & $0.56\pm0.00$ & $0.58\pm0.02$ 
                                        & \cellcolor{gray!25}$\bm{0.69\pm0.00}$ & [$0.70\pm0.01$] & $0.71\pm0.01$ & $0.72\pm0.02$ \\
         LGR+D   & $0.51\pm0.01$ & $0.57\pm0.01$ & $0.55\pm0.01$ & $0.53\pm0.00$ 
                                        & $0.65\pm0.00$ & $0.64\pm0.00$ & $0.69\pm0.00$ & $0.67\pm0.01$ \\

\bottomrule

\end{tabular}

}
\end{table}

\begin{figure}
    \centering
    \subfloat[\ac{Task-IL} Results on the original split.\label{fig:AffectNet-task-il-noaug}]{\includegraphics[width=0.5\textwidth]{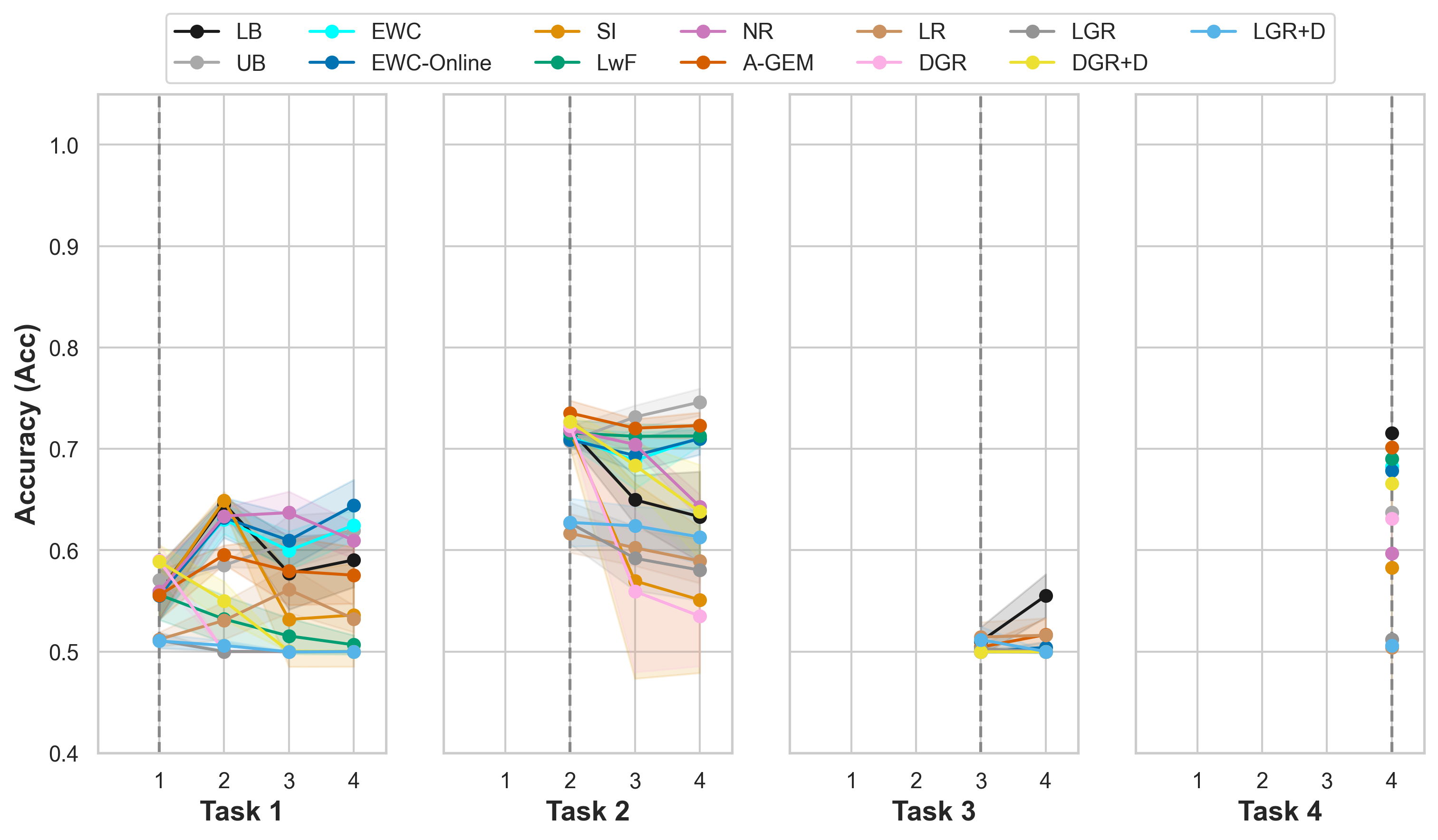}} \hfill
    \subfloat[\ac{Task-IL} Results on the downsampled split.\label{fig:AffectNet-task-il-aug}]{\includegraphics[width=0.5\textwidth]{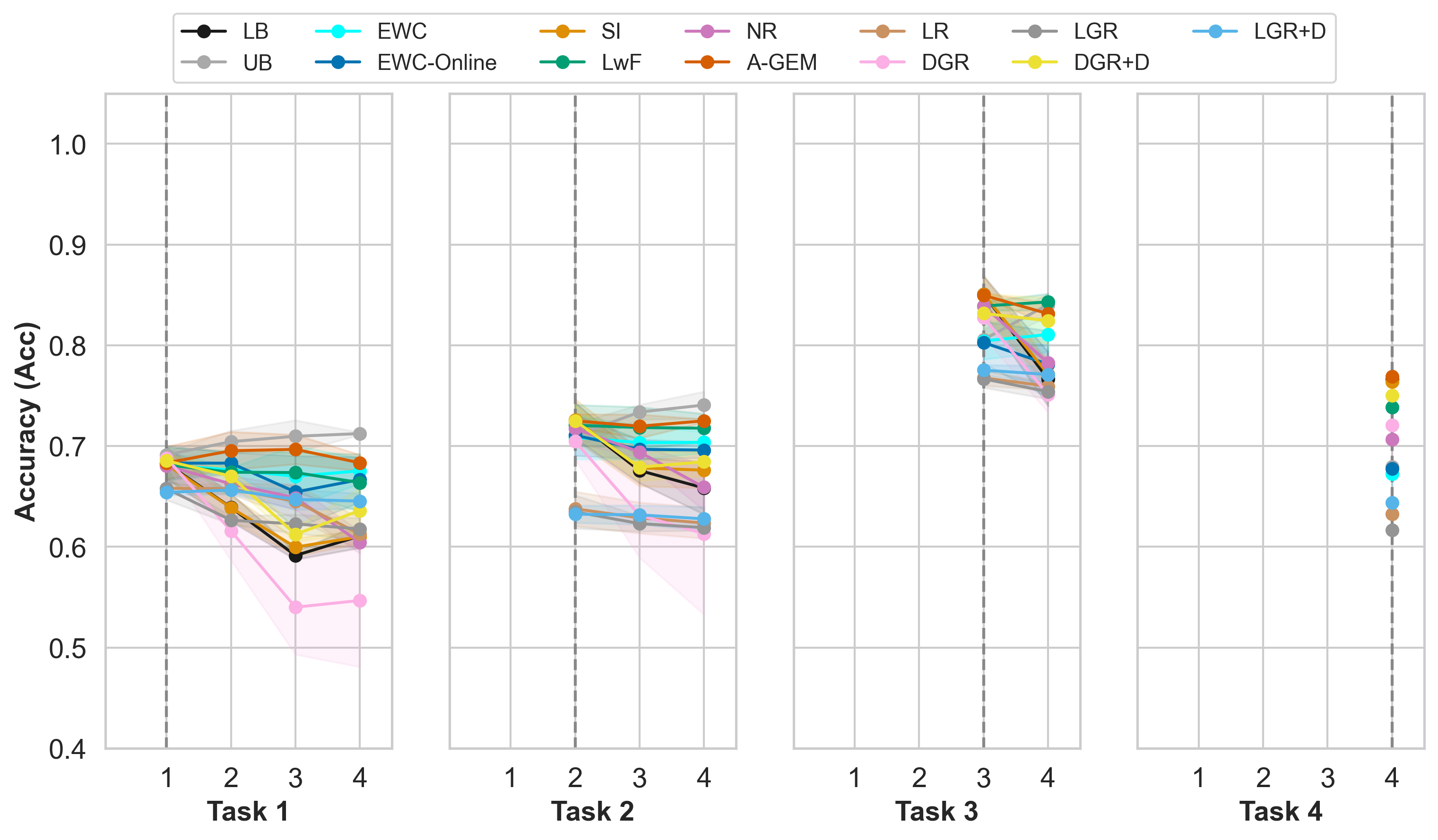}} \\
    \caption[Task-IL Results for AffectNet using original and downsampled data splits.]{\ac{Task-IL} results for AffectNet on the (a)~original and (b)~downsampled splits. Test-set~accuracy is shown as the learning progresses from when a task is introduced to the end of the training.}
    \label{fig:AffectNet-taskil} 
    
\end{figure}

\begin{table}[h]
\centering
\caption[\acs{Task-IL} \acs{CF} scores for AffectNet using original and downsampled data splits.]{\acs{Task-IL} \acs{CF} scores for AffectNet using original and \textit{downsampled} data splits. \textbf{Bold} values denote best (lowest) while [\textit{bracketed}] denote second-best values for each column.}
\label{tab:task-il-AffectNet-cf}
{
\scriptsize
\setlength\tabcolsep{3.5pt}

\begin{tabular}{l|ccc|ccc}\toprule
\multicolumn{1}{c|}{\textbf{Method}}            & \multicolumn{3}{c|}{\textbf{\acs{CF} on Original Split}} 
& \multicolumn{3}{c}{\textbf{\acs{CF}  on Downsampled Split}} \\ \cmidrule{2-7}

\multicolumn{1}{c|}{ }                          & 
\multicolumn{1}{c|}{\textbf{Task 2}}          &
\multicolumn{1}{c|}{\textbf{Task 3}}            & \multicolumn{1}{c|}{\textbf{Task 4}}           &
\multicolumn{1}{c|}{\textbf{Task 2}}          &
\multicolumn{1}{c|}{\textbf{Task 3}}            & \multicolumn{1}{c}{\textbf{Task 4}}          \\ \midrule

\multicolumn{7}{c}{\textbf{Baseline Approaches}} \\ \midrule
                      LB                &  $0.05\pm0.04$ &  $0.14\pm0.04$ & $0.09\pm0.03$ 
                                        &  $0.14\pm0.02$ &  $0.11\pm0.03$ &  $0.04\pm0.01$ \\
                      UB                & \cellcolor{gray!25}$\bm{-0.06\pm0.03}$ & [$-0.04\pm0.01$] & $-0.01\pm0.01$ 
                                        & \cellcolor{gray!25}$\bm{-0.04\pm0.02}$ & \cellcolor{gray!25}$\bm{-0.04\pm0.01}$ & \cellcolor{gray!25}$\bm{-0.01\pm0.01}$ \\\midrule

\multicolumn{7}{c}{\textbf{Regularisation-Based Approaches}} \\ \midrule
 EWC   & $-0.02\pm0.03$ & $-0.03\pm0.02$ & [$-0.07\pm0.02$] 
                                        &  $0.02\pm0.02$ &  [$0.00\pm0.00$] &  [$0.00\pm0.01$] \\
EWC-Online   & [$-0.04\pm0.03$] & \cellcolor{gray!25}$\bm{-0.05\pm0.04}$ & \cellcolor{gray!25}$\bm{-0.08\pm0.02}$ 
                                        &  $0.04\pm0.03$ &  $0.03\pm0.01$ &  $0.00\pm0.02$ \\
         SI   &  $0.17\pm0.14$ &  $0.09\pm0.06$ & $-0.05\pm0.03$ 
                                        &  $0.13\pm0.03$ &  $0.10\pm0.01$ &  $0.04\pm0.01$ \\
                 LwF   &  $0.04\pm0.02$ &  $0.03\pm0.01$ &  $0.02\pm0.02$ 
                                        &  $0.01\pm0.00$ &  $0.01\pm0.01$ &  $0.01\pm0.00$ \\\midrule

\multicolumn{7}{c}{\textbf{Replay-Based Approaches}} \\ \midrule
               NR   & \cellcolor{gray!25}$\bm{-0.06\pm0.03}$ &  $0.01\pm0.02$ & $0.07\pm0.03$ 
                                        &  $0.05\pm0.03$ &  $0.10\pm0.02$ &  $0.02\pm0.02$ \\
   A-GEM   & $-0.01\pm0.06$ & $-0.01\pm0.03$ & $-0.04\pm0.03$ 
                                        & [$-0.01\pm0.03$] &  $0.01\pm0.01$ & \cellcolor{gray!25}$\bm{-0.01\pm0.01}$ \\
       LR   & $-0.03\pm0.02$ &  $0.00\pm0.00$ & $-0.02\pm0.01$ 
                                        &  $0.02\pm0.01$ &  $0.03\pm0.01$ &  [$0.00\pm0.00$] \\
             DGR   &  $0.25\pm0.09$ &  $0.14\pm0.03$ &  $0.09\pm0.02$ 
                                        &  $0.22\pm0.07$ &  $0.15\pm0.05$ &  $0.07\pm0.04$ \\
           LGR   &  $0.05\pm0.01$ &  $0.03\pm0.00$ &  $0.01\pm0.01$ 
                                        &  $0.05\pm0.02$ &  $0.03\pm0.01$ &  $0.03\pm0.01$ \\
          DGR+D   &  $0.13\pm0.02$ &  $0.09\pm0.03$ &  $0.04\pm0.00$ 
                                        &  $0.12\pm0.01$ &  $0.05\pm0.01$ &  $0.02\pm0.01$ \\
         LGR+D   &  $0.01\pm0.01$ &  $0.02\pm0.01$ &  $0.00\pm0.00$ 
                                        &  $0.01\pm0.01$ &  $0.01\pm0.00$ &  [$0.00\pm0.00$] \\

\bottomrule
\end{tabular}

}
\end{table}
\paragraph{\acf{Task-IL}:} Table~\ref{tab:task-il-AffectNet-acc} presents the \acs{Task-IL} results both using the entire dataset as well as using \textit{downsampling}. For the evaluations with the original AffectNet split, regularisation-based approaches are able to outperform most replay-based methods. \acs{EWC} and \acs{EWC}-Online, in particular, are able to improve upon \acs{UB} scores, resulting in best model \acs{Acc} along with the A-\acs{GEM} method. Learning with large amounts of data from original data split ($\approx287K$ data samples), \acs{EWC} and \acs{EWC}-Online are able to efficiently learn Fischer Information Matrices that enable them to constraint model weights in a manner that reduces \textit{destructive interference} in the models. Similarly for A-\ac{GEM}, the model is able to constraint the average episodic memory loss across all the tasks, efficiently mitigating forgetting (Table~\ref{tab:task-il-AffectNet-cf}). The \acs{LwF} approach is also able to learn task-specific parameters, efficiently balancing learning novel tasks with preserving past knowledge. \acs{NR} also manages to mitigate forgetting to a large extent, however as the tasks increase, model performance suffers as the number of samples per task in the memory buffer decrease. The performance of the different models on each task individually can be seen in Figure~\ref{fig:AffectNet-task-il-noaug} showing the evolution of model \acs{Acc} scores throughout the learning.

The \textit{downsampled} split balances data distribution by restricting the maximum number of samples for each \textit{class} to $15000$. This results in a total of $\approx90K$ samples for the $8$ expression classes. The test-set, however, remains the same for both the splits. Learning with a balanced distribution of samples has a positive effect on model performance with all approaches seeing improvements in \acs{Acc} scores. Regularisation-based methods continue to outperform most replay-based methods with \acs{LwF} witnessing the biggest boost in model performance. The more balanced data split enables an efficient learning of task-specific parameters. A-\acs{GEM} still performs the best with the balanced data distribution allowing it to constraint the average loss across each task-specific episodic memory more efficiently, mitigating forgetting (Table~\ref{tab:task-il-AffectNet-cf}). Interestingly, \acs{DGR}-based methods experience a boost in model performance, despite the overall lower amount of data. A balanced data split across the tasks enables a more efficient generator that is able to reconstruct discriminative pseudo-samples, improving the overall performance of the models. \ac{LGR}-based methods also experience a boost in model performance but they are not able to match the performance on \ac{DGR}+D. Individual task \acs{Acc} scores throughout the learning process for the downsampled split can be seen in Figure~\ref{fig:AffectNet-task-il-aug}.

\nl \textit{Task Ordering:} Similar to \acs{Task-IL} experiments on other datasets, the order in which the classes (split in to tasks) are learnt has no significant impact on model performance. Different orders of the \textit{downsampled} split are compared where A-\acs{GEM} emerges as the best performing method across the $3$ task orderings. More details on the results from the task-ordering experiments can be found in Appendix~\ref{app:taskorderaffnet}.



\begin{table}[t]
\centering
\setlength\tabcolsep{3.5pt}

\caption[\acs{Class-IL} \acs{Acc} for AffectNet using original and downsampled data splits.]{\acs{Class-IL} \acs{Acc} for AffectNet using original and \textit{downsampled} data splits. \textbf{Bold} values denote best (highest) while [\textit{bracketed}] denote second-best values for each column.}

\label{tab:class-il-affnet-acc}

{
\scriptsize
\begin{tabular}{l|cccccccc}\toprule

\multirow{2}[2]{*}{\makecell[c]{\textbf{Method}}}           & \multicolumn{8}{c}{\textbf{\acs{Acc} on Original Split}} \\ \cmidrule{2-9}

\multicolumn{1}{c|}{ }                          & 
\multicolumn{1}{c|}{\textbf{Class 1}}            & \multicolumn{1}{c|}{\textbf{Class 2}}          &
\multicolumn{1}{c|}{\textbf{Class 3}}            & \multicolumn{1}{c|}{\textbf{Class 4}}           &
\multicolumn{1}{c|}{\textbf{Class 5}}            & \multicolumn{1}{c|}{\textbf{Class 6}}          &
\multicolumn{1}{c|}{\textbf{Class 7}}            & \multicolumn{1}{c}{\textbf{Class 8}}          \\ \midrule

\multicolumn{9}{c}{\textbf{Baseline Approaches}} \\ \midrule

                      LB & \cellcolor{gray!25}$\bm{1.00\pm0.00}$ & $0.50\pm0.00$ & $0.33\pm0.00$ & $0.25\pm0.00$ & $0.20\pm0.00$ & $0.17\pm0.00$ & $0.14\pm0.00$ & $0.12\pm0.00$ \\
                      UB & \cellcolor{gray!25}$\bm{1.00\pm0.00}$ & \cellcolor{gray!25}$\bm{0.68\pm0.01}$ & [$0.47\pm0.00$] & [$0.38\pm0.00$] & \cellcolor{gray!25}$\bm{0.37\pm0.01}$ & \cellcolor{gray!25}$\bm{0.41\pm0.01}$ & \cellcolor{gray!25}$\bm{0.36\pm0.00}$ & \cellcolor{gray!25}$\bm{0.34\pm0.00}$ \\ \midrule

\multicolumn{9}{c}{\textbf{Regularisation-Based Approaches}} \\ \midrule

 EWC  & \cellcolor{gray!25}$\bm{1.00\pm0.00}$ & $0.50\pm0.00$ & $0.33\pm0.00$ & $0.25\pm0.00$ & $0.20\pm0.00$ & $0.17\pm0.00$ & $0.14\pm0.00$ & $0.12\pm0.00$ \\
EWC-Online  & \cellcolor{gray!25}$\bm{1.00\pm0.00}$ & $0.50\pm0.00$ & $0.33\pm0.00$ & $0.25\pm0.00$ & $0.20\pm0.00$ & $0.17\pm0.00$ & $0.14\pm0.00$ & $0.12\pm0.00$ \\
         SI  & \cellcolor{gray!25}$\bm{1.00\pm0.00}$ & $0.50\pm0.00$ & $0.33\pm0.00$ & $0.25\pm0.00$ & $0.20\pm0.00$ & $0.17\pm0.00$ & $0.14\pm0.00$ & $0.12\pm0.00$ \\
                 LwF  & \cellcolor{gray!25}$\bm{1.00\pm0.00}$ & $0.50\pm0.00$ & $0.33\pm0.00$ & $0.25\pm0.00$ & $0.20\pm0.00$ & $0.17\pm0.00$ & $0.14\pm0.00$ & $0.12\pm0.00$ \\\midrule
\multicolumn{9}{c}{\textbf{Replay-Based Approaches}} \\ \midrule

               NR  & \cellcolor{gray!25}$\bm{1.00\pm0.00}$ & $0.66\pm0.01$ & $0.44\pm0.01$ & $0.32\pm0.01$ & $0.26\pm0.01$ & $0.23\pm0.01$ & $0.19\pm0.00$ & $0.15\pm0.01$ \\
   A-GEM  & \cellcolor{gray!25}$\bm{1.00\pm0.00}$ & $0.50\pm0.00$ & $0.33\pm0.00$ & $0.25\pm0.00$ & $0.20\pm0.00$ & $0.17\pm0.00$ & $0.14\pm0.00$ & $0.12\pm0.00$ \\
       LR  & \cellcolor{gray!25}$\bm{1.00\pm0.00}$ & [$0.67\pm0.01$] & \cellcolor{gray!25}$\bm{0.52\pm0.01}$ & \cellcolor{gray!25}$\bm{0.40\pm0.01}$ & \cellcolor{gray!25}$\bm{0.37\pm0.01}$ & [$0.36\pm0.01$] & [$0.31\pm0.01$] & [$0.27\pm0.00$] \\
             DGR  & \cellcolor{gray!25}$\bm{1.00\pm0.00}$ & $0.50\pm0.00$ & $0.33\pm0.00$ & $0.25\pm0.00$ & $0.20\pm0.00$ & $0.17\pm0.00$ & $0.14\pm0.00$ & $0.12\pm0.00$ \\
           LGR  & \cellcolor{gray!25}$\bm{1.00\pm0.00}$ & $0.55\pm0.02$ & $0.41\pm0.03$ & $0.29\pm0.02$ & $0.25\pm0.03$ & $0.26\pm0.03$ & $0.21\pm0.02$ & $0.17\pm0.03$ \\
          DGR+D  & \cellcolor{gray!25}$\bm{1.00\pm0.00}$ & $0.50\pm0.00$ & $0.33\pm0.00$ & $0.25\pm0.00$ & $0.20\pm0.00$ & $0.17\pm0.00$ & $0.14\pm0.00$ & $0.12\pm0.00$ \\
         LGR+D  & \cellcolor{gray!25}$\bm{1.00\pm0.00}$ & $0.64\pm0.02$ & $0.45\pm0.02$ & $0.30\pm0.03$ & [$0.28\pm0.02$] & $0.30\pm0.02$ & $0.23\pm0.01$ & $0.20\pm0.01$ \\

\midrule

\multirow{2}[2]{*}{\makecell[c]{\textbf{Method}}}           & \multicolumn{8}{c}{\textbf{\acs{Acc} on Downsampled Split}} \\ \cmidrule{2-9}

\multicolumn{1}{c|}{ }                          & 
\multicolumn{1}{c|}{\textbf{Class 1}}            & \multicolumn{1}{c|}{\textbf{Class 2}}          &
\multicolumn{1}{c|}{\textbf{Class 3}}            & \multicolumn{1}{c|}{\textbf{Class 4}}           &
\multicolumn{1}{c|}{\textbf{Class 5}}            & \multicolumn{1}{c|}{\textbf{Class 6}}          &
\multicolumn{1}{c|}{\textbf{Class 7}}            & \multicolumn{1}{c}{\textbf{Class 8}}          \\ \midrule
\multicolumn{9}{c}{\textbf{Baseline Approaches}} \\ \midrule

                      LB & \cellcolor{gray!25}$\bm{1.00\pm0.00}$ & $0.50\pm0.00$ & $0.33\pm0.00$ & $0.25\pm0.00$ & $0.20\pm0.00$ & $0.17\pm0.00$ & $0.14\pm0.00$ & $0.12\pm0.00$ \\
                      UB & \cellcolor{gray!25}$\bm{1.00\pm0.00}$ & $0.59\pm0.06$ & $0.41\pm0.01$ & $0.30\pm0.01$ & [$0.26\pm0.00$] & [$0.32\pm0.01$] & [$0.27\pm0.02$] & \cellcolor{gray!25}$\bm{0.25\pm0.01}$ \\\midrule

\multicolumn{9}{c}{\textbf{Regularisation-Based Approaches}} \\ \midrule

 EWC  & \cellcolor{gray!25}$\bm{1.00\pm0.00}$ & $0.50\pm0.00$ & $0.33\pm0.00$ & $0.25\pm0.00$ & $0.20\pm0.00$ & $0.17\pm0.00$ & $0.14\pm0.00$ & $0.12\pm0.00$ \\
EWC-Online  & \cellcolor{gray!25}$\bm{1.00\pm0.00}$ & $0.50\pm0.00$ & $0.33\pm0.00$ & $0.25\pm0.00$ & $0.20\pm0.00$ & $0.17\pm0.00$ & $0.14\pm0.00$ & $0.12\pm0.00$ \\
         SI  & \cellcolor{gray!25}$\bm{1.00\pm0.00}$ & $0.50\pm0.00$ & $0.33\pm0.00$ & $0.25\pm0.00$ & $0.20\pm0.00$ & $0.17\pm0.00$ & $0.14\pm0.00$ & $0.12\pm0.00$ \\
                 LwF  & \cellcolor{gray!25}$\bm{1.00\pm0.00}$ & $0.50\pm0.00$ & $0.33\pm0.00$ & $0.25\pm0.00$ & $0.20\pm0.00$ & $0.17\pm0.00$ & $0.14\pm0.00$ & $0.12\pm0.00$ \\ \midrule
\multicolumn{9}{c}{\textbf{Replay-Based Approaches}} \\ \midrule

               NR  & \cellcolor{gray!25}$\bm{1.00\pm0.00}$ & \cellcolor{gray!25}$\bm{0.66\pm0.01}$ & [$0.46\pm0.01$] & [$0.32\pm0.02$] & [$0.26\pm0.00$] & $0.23\pm0.00$ & $0.20\pm0.00$ & $0.16\pm0.00$ \\
   A-GEM  & \cellcolor{gray!25}$\bm{1.00\pm0.00}$ & $0.50\pm0.00$ & $0.33\pm0.00$ & $0.25\pm0.00$ & $0.20\pm0.00$ & $0.17\pm0.00$ & $0.14\pm0.00$ & $0.12\pm0.00$ \\
       LR  & \cellcolor{gray!25}$\bm{1.00\pm0.00}$ & [$0.64\pm0.00$] & \cellcolor{gray!25}$\bm{0.48\pm0.02}$ & \cellcolor{gray!25}$\bm{0.37\pm0.01}$ & \cellcolor{gray!25}$\bm{0.33\pm0.01}$ & \cellcolor{gray!25}$\bm{0.34\pm0.01}$ & \cellcolor{gray!25}$\bm{0.29\pm0.01}$ & [$0.24\pm0.01$] \\
             DGR  & \cellcolor{gray!25}$\bm{1.00\pm0.00}$ & $0.50\pm0.00$ & $0.34\pm0.00$ & $0.25\pm0.00$ & $0.19\pm0.00$ & $0.17\pm0.00$ & $0.14\pm0.00$ & $0.13\pm0.00$ \\
           LGR  & \cellcolor{gray!25}$\bm{1.00\pm0.00}$ & $0.53\pm0.02$ & $0.34\pm0.01$ & $0.25\pm0.00$ & $0.20\pm0.00$ & $0.21\pm0.03$ & $0.19\pm0.03$ & $0.14\pm0.02$ \\
          DGR+D  & \cellcolor{gray!25}$\bm{1.00\pm0.00}$ & $0.50\pm0.00$ & $0.34\pm0.00$ & $0.25\pm0.00$ & $0.20\pm0.00$ & $0.17\pm0.00$ & $0.14\pm0.00$ & $0.13\pm0.00$ \\
         LGR+D  & \cellcolor{gray!25}$\bm{1.00\pm0.00}$ & $0.58\pm0.01$ & $0.41\pm0.04$ & $0.28\pm0.02$ & $0.22\pm0.02$ & $0.26\pm0.02$ & $0.20\pm0.01$ & $0.14\pm0.02$ \\

\bottomrule

\end{tabular}

}
\end{table}

\begin{figure}[t]  
    \centering
    \subfloat[\ac{Class-IL} Results on the original split.\label{fig:affnet-class-il-noaug}]{\includegraphics[width=0.5\textwidth]{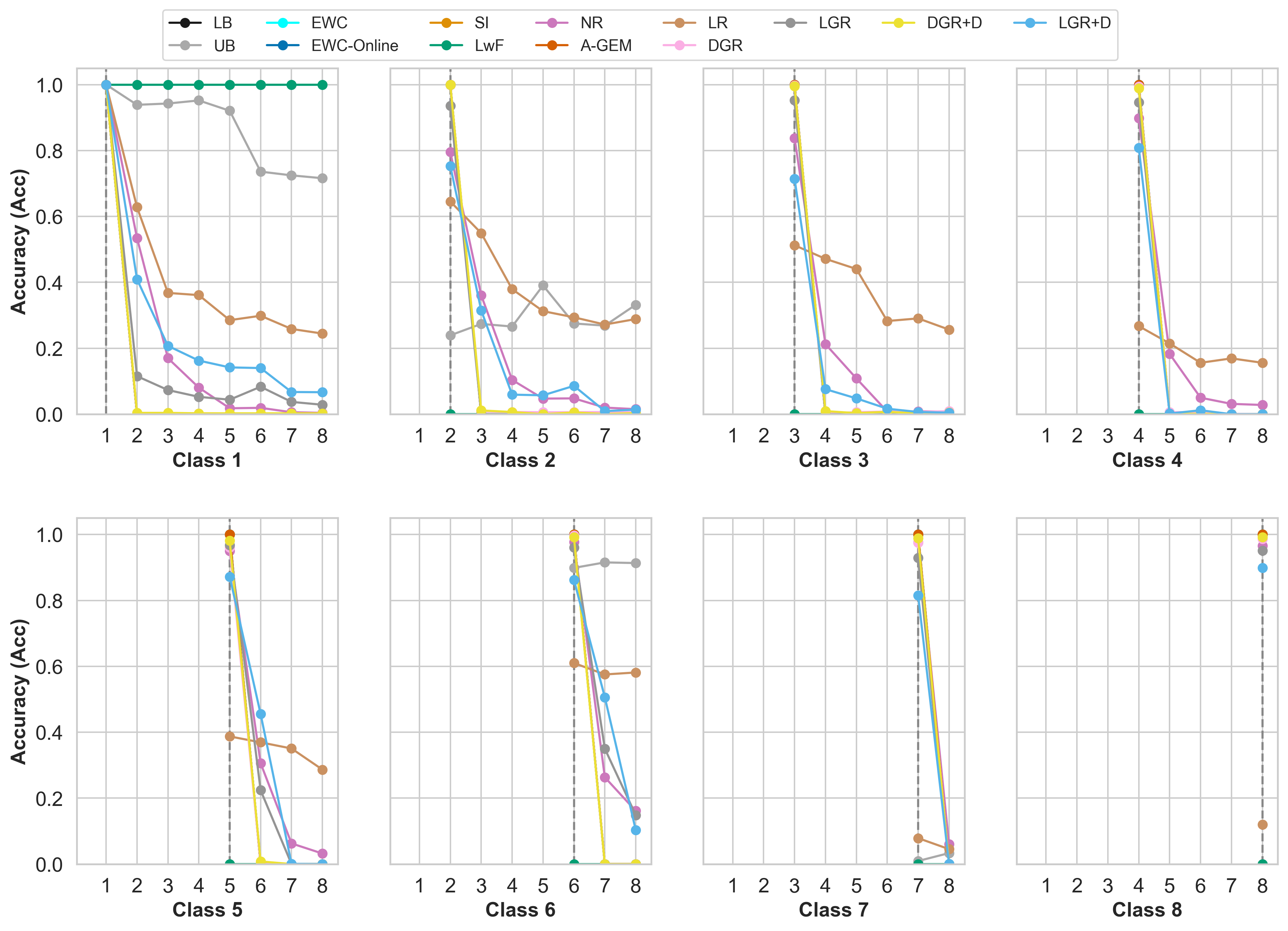}} 
    \hfill
    \subfloat[\ac{Class-IL} Results on the downsampled split.\label{fig:affnet-class-il-aug}]{\includegraphics[width=0.5\textwidth]{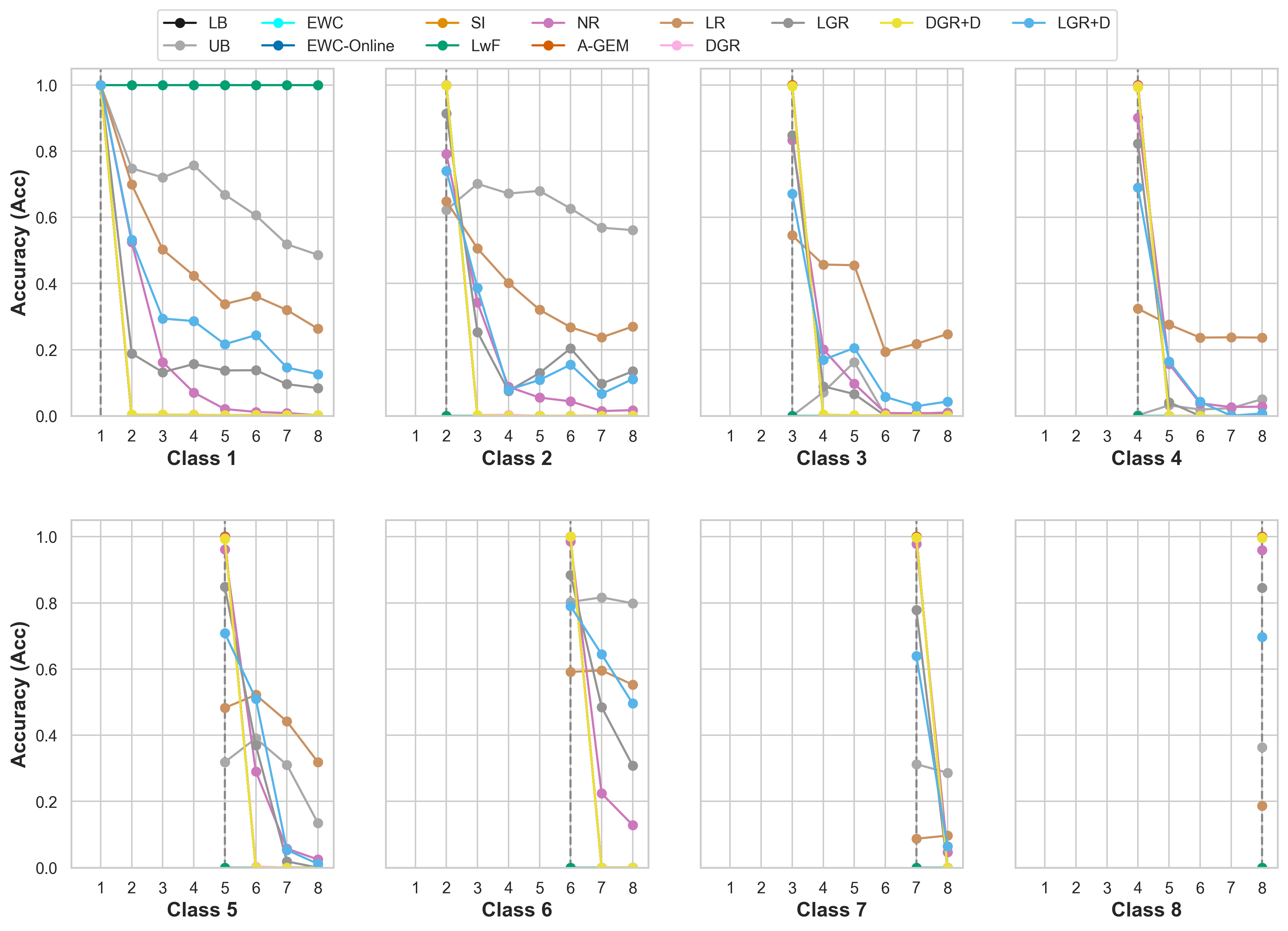}} \\
    \caption[Class-IL Results for AffectNet using original and downsampled data splits.]{\ac{Class-IL} results for AffectNet on the (a) original and (b) downsampled data splits.  Test-set accuracy is shown as the learning progresses from when a class is introduced to the end of the training.}
    \label{fig:affnet-classil} 
    
\end{figure}

\begin{table}[t]
\centering
\caption[\acs{Class-IL} \acs{CF} scores for AffectNet using original and downsampled data splits.]{\acs{Class-IL} \acs{CF} scores for AffectNet using original and \textit{downsampled} data splits. \textbf{Bold} values denote best (lowest) while [\textit{bracketed}] denote second-best values for each column.}
\label{tab:class-il-affnet-cf}

{
\scriptsize
\setlength\tabcolsep{3.5pt}

\begin{tabular}{l|ccccccc}\toprule

\multirow{2}[2]{*}{\makecell[c]{\textbf{Method}}}           & \multicolumn{7}{c}{\textbf{\acs{CF} on Original Split}} \\ \cmidrule{2-8}

\multicolumn{1}{c|}{ }                          & 
\multicolumn{1}{c|}{\textbf{Class 2}}          &
\multicolumn{1}{c|}{\textbf{Class 3}}            & \multicolumn{1}{c|}{\textbf{Class 4}}           &
\multicolumn{1}{c|}{\textbf{Class 5}}            & \multicolumn{1}{c|}{\textbf{Class 6}}          &
\multicolumn{1}{c|}{\textbf{Class 7}}            & \multicolumn{1}{c}{\textbf{Class 8}}          \\ \midrule
\multicolumn{8}{c}{\textbf{Baseline Approaches}} \\ \midrule

                      LB & $1.00\pm0.00$ & $1.00\pm0.00$ &  $1.00\pm0.00$ & $1.00\pm0.00$ & $1.00\pm0.00$ & $1.00\pm0.00$ & $0.00\pm0.00$ \\
                      UB & \cellcolor{gray!25}$\bm{0.10\pm0.05}$ & \cellcolor{gray!25}$\bm{0.04\pm0.03}$ &  \cellcolor{gray!25}$\bm{0.02\pm0.02}$ & \cellcolor{gray!25}$\bm{0.06\pm0.03}$ & \cellcolor{gray!25}$\bm{0.08\pm0.02}$ & \cellcolor{gray!25}$\bm{0.10\pm0.02}$ & \cellcolor{gray!25}$\bm{0.11\pm0.00}$ \\\midrule

\multicolumn{8}{c}{\textbf{Regularisation-Based Approaches}} \\ \midrule

EWC  & $1.00\pm0.00$ & $1.00\pm0.00$ &  $1.00\pm0.00$ & $1.00\pm0.00$ & $1.00\pm0.00$ & $1.00\pm0.00$ & $1.00\pm0.00$ \\
EWC-Online  & $1.00\pm0.00$ & $1.00\pm0.00$ &  $1.00\pm0.00$ & $1.00\pm0.00$ & $1.00\pm0.00$ & $1.00\pm0.00$ & $1.00\pm0.00$ \\
SI  & $1.00\pm0.00$ & $1.00\pm0.00$ &  $1.00\pm0.00$ & $1.00\pm0.00$ & $1.00\pm0.00$ & $1.00\pm0.00$ & $1.00\pm0.00$ \\
LwF  & $1.00\pm0.00$ & $1.00\pm0.00$ &  $1.00\pm0.00$ & $1.00\pm0.00$ & $1.00\pm0.00$ & $1.00\pm0.00$ & $1.00\pm0.00$ \\\midrule

\multicolumn{8}{c}{\textbf{Replay-Based Approaches}} \\ \midrule

NR  & $0.64\pm0.03$ & $0.76\pm0.02$ &  $0.80\pm0.02$ & $0.82\pm0.01$ & $0.86\pm0.01$ & $0.89\pm0.01$ & $0.87\pm0.00$ \\
A-GEM  & $1.00\pm0.00$ & $1.00\pm0.00$ &  $1.00\pm0.00$ & $1.00\pm0.00$ & $1.00\pm0.00$ & $1.00\pm0.00$ & $1.00\pm0.00$ \\
LR  & [$0.32\pm0.02$] & [$0.30\pm0.01$] &  [$0.28\pm0.01$] & [$0.28\pm0.01$] & [$0.26\pm0.01$] & [$0.24\pm0.02$] & [$0.26\pm0.00$] \\
DGR  & $1.00\pm0.00$ & $1.00\pm0.00$ &  $1.00\pm0.00$ & $1.00\pm0.00$ & $1.00\pm0.00$ & $1.00\pm0.00$ & $1.00\pm0.00$ \\
LGR  & $0.76\pm0.06$ & $0.81\pm0.07$ &  $0.80\pm0.09$ & $0.74\pm0.09$ & $0.77\pm0.09$ & $0.80\pm0.11$ & $0.81\pm0.07$ \\
DGR+D  & $1.00\pm0.00$ & $1.00\pm0.00$ &  $1.00\pm0.00$ & $1.00\pm0.00$ & $1.00\pm0.00$ & $1.00\pm0.00$ & $1.00\pm0.00$ \\
LGR+D  & $0.53\pm0.03$ & $0.63\pm0.07$ &  $0.60\pm0.07$ & $0.56\pm0.05$ & $0.61\pm0.06$ & $0.63\pm0.06$ & $0.65\pm0.00$ \\\midrule

\multirow{2}[2]{*}{\makecell[c]{\textbf{Method}}}          & \multicolumn{7}{c}{\textbf{\acs{CF} on Downsampled Split}} \\ \cmidrule{2-8}

\multicolumn{1}{c|}{ }                          & 
\multicolumn{1}{c|}{\textbf{Class 2}}          &
\multicolumn{1}{c|}{\textbf{Class 3}}            & \multicolumn{1}{c|}{\textbf{Class 4}}           &
\multicolumn{1}{c|}{\textbf{Class 5}}            & \multicolumn{1}{c|}{\textbf{Class 6}}          &
\multicolumn{1}{c|}{\textbf{Class 7}}            & \multicolumn{1}{c}{\textbf{Class 8}}          \\ \midrule

\multicolumn{8}{c}{\textbf{Baseline Approaches}} \\ \midrule

                      LB & $1.00\pm0.00$ & $1.00\pm0.00$ &  $1.00\pm0.00$ & $1.00\pm0.00$ & $1.00\pm0.00$ & $1.00\pm0.00$ & $0.00\pm0.00$ \\
                      UB & \cellcolor{gray!25}$\bm{0.01\pm0.07}$ & \cellcolor{gray!25}$\bm{0.01\pm0.06}$ & \cellcolor{gray!25}$\bm{-0.02\pm0.05}$ & \cellcolor{gray!25}$\bm{0.05\pm0.04}$ & \cellcolor{gray!25}$\bm{0.04\pm0.04}$ & \cellcolor{gray!25}$\bm{0.02\pm0.03}$ & \cellcolor{gray!25}$\bm{0.01\pm0.00}$ \\ \midrule

\multicolumn{8}{c}{\textbf{Regularisation-Based Approaches}} \\ \midrule

 EWC  & $1.00\pm0.00$ & $1.00\pm0.00$ &  $1.00\pm0.00$ & $1.00\pm0.00$ & $1.00\pm0.00$ & $1.00\pm0.00$ & $1.00\pm0.00$ \\
EWC-Online  & $1.00\pm0.00$ & $1.00\pm0.00$ &  $1.00\pm0.00$ & $1.00\pm0.00$ & $1.00\pm0.00$ & $1.00\pm0.00$ & $1.00\pm0.00$ \\
         SI  & $1.00\pm0.00$ & $1.00\pm0.00$ &  $1.00\pm0.00$ & $1.00\pm0.00$ & $1.00\pm0.00$ & $1.00\pm0.00$ & $1.00\pm0.00$ \\
                 LwF  & $1.00\pm0.00$ & $1.00\pm0.00$ &  $1.00\pm0.00$ & $1.00\pm0.00$ & $1.00\pm0.00$ & $1.00\pm0.00$ & $1.00\pm0.00$ \\\midrule

\multicolumn{8}{c}{\textbf{Replay-Based Approaches}} \\ \midrule
               NR  & $0.63\pm0.05$ & $0.75\pm0.04$ &  $0.79\pm0.02$ & $0.81\pm0.02$ & $0.84\pm0.01$ & $0.88\pm0.01$ & $0.89\pm0.00$ \\
   A-GEM  & $1.00\pm0.00$ & $1.00\pm0.00$ &  $1.00\pm0.00$ & $1.00\pm0.00$ & $1.00\pm0.00$ & $1.00\pm0.00$ & $1.00\pm0.00$ \\
       LR  & [$0.36\pm0.02$] & [$0.32\pm0.01$] &  [$0.29\pm0.01$] & [$0.28\pm0.00$] & [$0.25\pm0.00$] & [$0.23\pm0.01$] & [$0.26\pm0.00$] \\
             DGR  & $0.99\pm0.00$ & $1.00\pm0.00$ &  $0.99\pm0.00$ & $0.99\pm0.01$ & $0.99\pm0.01$ & $0.99\pm0.01$ & $1.00\pm0.00$ \\
           LGR  & $0.93\pm0.04$ & $0.95\pm0.04$ &  $0.95\pm0.03$ & $0.90\pm0.07$ & $0.90\pm0.07$ & $0.93\pm0.04$ & $0.90\pm0.00$ \\
          DGR+D  & $0.99\pm0.00$ & $0.99\pm0.01$ &  $0.99\pm0.00$ & $0.99\pm0.01$ & $0.99\pm0.01$ & $0.99\pm0.01$ & $1.00\pm0.00$ \\
         LGR+D  & $0.62\pm0.13$ & $0.72\pm0.10$ &  $0.76\pm0.09$ & $0.69\pm0.08$ & $0.74\pm0.07$ & $0.81\pm0.07$ & $0.91\pm0.00$ \\

\bottomrule

\end{tabular}

}
\end{table}
\paragraph{\acf{Class-IL}:} Similar to \acs{Class-IL} experiments on other datasets, regularisation-based methods fail completely at mitigating forgetting in the model for AffectNet as well resulting in the worst model \acs{Acc} (Table~\ref{tab:class-il-affnet-acc}) and \acs{CF} (Table~\ref{tab:class-il-affnet-cf}) scores. Rehearsal-based methods of \acs{NR} and \acs{LR} are able to perform reasonably well, trying to catch up to \ac{UB} scores. Yet, as the number of classes increase, they fail to maintain model performance as the fraction of the memory buffer allocated to samples of each class shrinks, resulting the model forgetting past knowledge. Regularisation-based methods struggle to prioritise weight-updates efficiently when learning one class at a time, also experiencing \textit{catastrophic forgetting}. A-\acs{GEM} is also unable to balance model loss across the class-specific episodic memories, failing to preserve past knowledge. \ac{DGR}-based approaches, again, fail to efficiently generate pseudo-samples, when learning one class at a time. \acs{LGR}-based methods persist for some classes, catching up to \acs{UB} scores but as the number of classes increase, they find it more and more difficult to learn to generate discrimintative pseudo-samples for latent features, succumbing to forgetting. The learning dynamics, with respect to how the different approaches manage learning across each class can be seen in Figure~\ref{fig:affnet-class-il-noaug}.

Using the \textit{downsampling} split does not seem to improve model performance for \acs{Class-IL} evaluations in the same way it impacted \acs{Task-IL} scores. Similar results are witnessed as the original data split with \acs{LR} emerging as the best performing method followed by \acs{NR}. Even with the balanced data splits, regularisation-based methods are not able to balance old \vs new learning, resulting in \textit{catastrophic forgetting} of information (Table~\ref{tab:class-il-affnet-cf}). The learning dynamics on the \textit{downsampled} data split can be seen in Figure~\ref{fig:affnet-class-il-aug}.

\nl \textit{Class Ordering:} Evaluating different class orders, similar to \acs{CK+} and \acs{RAF-DB} experiments, no significant differences in model performances are witnessed, with \acs{LR} still emerging as the best performing approach with similar \acs{Acc} and \acs{CF} scores. More details on the results from these class-ordering experiments can be found in Appendix~\ref{app:classorderaffnet}.

\section{Discussion}
\label{sec:discussion}
The benchmarking results across the $3$ \ac{FER} datasets successfully demonstrate the ability of \ac{CL}-based methods to incrementally learn facial expression classes. This is a positive result for affective computing where embedding such \ac{CL}-based affect perception models can help agents to \textit{continually} learn during interactions, learning novel expression classes while preserving past knowledge. These results provide the first evaluation of its kind, making a strong case for using \ac{CL} for \ac{FER}, here termed as \acf{ConFER}. Even though the datasets explored here evaluate only $8$ expression classes, the \ac{ConFER} benchmark can be extended to any number of classes whether they comprise of primary or secondary (compound) expressions~\citep{Picard:1997:AC:265013}. \acs{Task-IL} and \acs{Class-IL} learning settings present very difficult challenges for the models to overcome, even when exploring only $8$ basic expression classes. While \acs{Task-IL} settings require models to learn with $2$ classes at the same time, \acs{Class-IL} settings present a harder challenge to overcome, requiring models to learn with the data from only one class at a time. Under both settings, traditional \ac{ML}-based learning, here represented by \ac{LB} evaluations, struggle to incrementally learn facial expression classes and experience \textit{catastrophic forgetting}. As models under \acs{LB} `fine-tune' on incrementally made available data for a particular task or class, this results in past knowledge being overwritten. \ac{UB} evaluations, on the other hand, work with an unrealistic resource and computation-heavy assumption of \textit{joint-training} where models can access all the data at any given time. This provides a target for models to achieve in terms of model performance. As \acs{Task-IL} and \acs{Class-IL} settings present different challenges for the models to overcome, these results are discussed individually below.

\subsection{\acf{Task-IL}}

\acs{Task-IL} results, across the datasets, demonstrate that regularisation-based methods consistently perform well, outperforming replay-based methods in most of the evaluations. While methods like \acs{EWC}, \acs{EWC}-Online and \acs{SI} focus on penalising weight-updates in a manner that preserves past knowledge, \ac{LwF} learns task-specific parameters of \textit{old} and \textit{new} tasks and aim to minimise the loss on old task parameters using the new data. Each of these methods learn task-specific attributes (importance for model parameters) to mitigate forgetting. Under \acs{Task-IL} settings, learning with $2$ classes at a time allows models to balance the learnt parameters more efficiently, resulting in improved model performances. This is different from replay-based methods that either store past data samples in memory buffer (\acs{NR}, A-\ac{GEM} and \acs{LR}), splitting it into equal fractions for each learnt task, or train a generative model to reconstruct pseudo-samples for rehearsal (\acs{DGR}, \acs{DGR+D}, \acs{LGR} and \acs{LGR+D}). As the number of tasks increases, the fraction of data per task in the memory buffer shrinks, negatively impacting the models' ability to rehearse past knowledge. Pseudo-rehearsal also becomes difficult as it becomes harder for generative models to reconstruct discriminative pseudo-samples for each learnt task. 

For \acs{CK+}, \acs{Task-IL} evaluations, both without ($0.97$) and with ($0.98$) data-augmentation are able to achieve \acs{SOTA} scores (best scores of $0.968$~\citep{Ding2017FN2EN}) with \acs{NR}, while \ac{LwF} gets the second-best \acs{Acc} score of $0.95$. As the \acs{CK+} is a relatively small dataset, almost the entire dataset can be stored in the memory buffer for \acs{NR} resulting in similar model performance as the \acs{UB}. For larger datasets like \acs{RAF-DB}, the memory buffer for replay-based methods faces the challenge of maintaining only a fraction of the entire dataset in memory for rehearsal. This results in their performance being lower than regularisation-based methods. \acs{LwF} achieves the best model \acs{Acc} score of $0.91$, achieving \acs{SOTA} (best scores of $0.90$~\citep{Zhang2022EAC}). 
For AffectNet, \acs{EWC}, \acs{EWC}-Online and A-\acs{GEM} are able to match \acs{SOTA} (best scores of $0.63$~\citep{Savchenko2022Classifying}), albeit following the \acs{Task-IL} protocol with \acs{LwF} performing second-best only lower than \acs{UB} scores.

\subsection{\acf{Class-IL}}

\acs{Class-IL} settings represent a difficult problem for learning models to solve, having to incrementally learn facial expression classes, one at a time, while preserving past knowledge. Unlike \acs{Task-IL} where at each stage the model learns with batches representing $2$ classes at a time, \acs{Class-IL} imposes a stricter learning regime where the models need to incrementally learn with data from only one class. This is shown to cause regularisation-based methods to overfit, resulting in new data overwriting past knowledge~\citep{van2019three}. A similar effect is witnessed in the results presented here across the $3$ datasets with regularisation-based methods completely failing to mitigate \textit{catastrophic forgetting}. Pseudo-rehearsal-based methods also face a similar challenge where the generator models need to be trained incrementally, one class at a time, impacting the rehearsal of previously learnt classes. This is particularly true for \acs{DGR}-based methods that need to learn complex data distributions for the generator to reconstruct high-quality and discriminative facial images representing different expression classes. \acs{LGR}-based pseudo-rehearsal performs moderately better but faces similar challenges as their \acs{DGR}-based counterparts. Replay-based methods such as \acs{NR} and \acs{LR} are able to maintain model performance (relative to \acs{UB} scores) for longer but as the number of classes increase, similar to the issues faced under \acs{Task-IL} settings, the fraction of memory buffer reserved for each class shrinks, causing forgetting in the models due to an inefficient rehearsal of data.

\section{Conclusions}
\label{sec:conclusion}
The ability to incrementally learn and adapt during real-world interactions is crucial for \ac{FER} systems. This paper proposes and evaluates the novel use of \ac{CL} for \acf{ConFER}. It evaluates different learning settings under which \ac{CL} models may be expected to learn facial expression classes and presents the first benchmark of its kind that compares popular \acs{CL}-based approaches on their ability to incrementally \textit{adapt}, learning one (\acs{Class-IL}) or more (\acs{Task-IL}) classes at a time, on $3$ popular \acs{FER} datasets. Across the evaluations, different strategies are seen to be effective under different learning settings with no one strategy emerging as a `silver bullet' to solve all problems that comes with learning incrementally and sequentially. While regularisation-based methods perform better under \acs{Task-IL} settings owing to their ability to constraint weight updates to give importance to task-relevant parameters, none of the strategies are able to successfully mitigate forgetting under \acs{Class-IL} settings. Replay-based methods, however, perform better amongst all \acs{CL}-methods, able to maintain model performance for longer by storing data samples from previously learnt classes in memory buffers. 
\subsection{Limitations and Future Directions}
\label{sec:limitations}
The \acs{ConFER} benchmark successfully demonstrates the applicability of \ac{CL}-based methods for incrementally learning facial expression classes. This is particularly true under \acs{Task-IL} settings where these methods are able to match \acs{SOTA} evaluations. \acs{Class-IL} setting however present a major challenge for \acs{CL}-based methods to overcome with all the compared approaches suffering \textit{catastrophic forgetting} of information. Even though replay-based methods perform slightly better, these are much below \acs{SOTA} scores. To ensure a thorough evaluation of \ac{CL}-based methods, the LeNet-based architecture is trained, from scratch, for all the approaches, isolating any positive effects pre-training the model may have on its performance. However, recent studies~\citep{mehta2021empirical} have shown pre-training to positively impact the model's ability to retain past knowledge and mitigate forgetting. Thus, further experimentation is required using pre-trained backbones such as the VGG~\citep{VGG} or ResNet-18~\citep{He2016ResNet} as a starting point in learning for the different \ac{CL}-based methods. Additionally, \acs{ConFER} evaluations clearly define task-boundaries for the models' learning, under both \acs{Class-IL} and \acs{Task-IL} settings. This ensures that, at any given time, the model receives data for only one class or task. In real-world interactions, affective robots may be needed to learn with \textit{blurred} task boundaries as the robot may receive data from different classes. Thus, more sophisticated methods need to be explored that relaxes allow learning under \textit{task-free}~\citep{Aljundi2019TaskFree} or \textit{task-agnostic}~\citep{Zeno2021Task} settings.

\section{Acknowledgements}

\textbf{Funding:} N. Churamani's work was supported by EPSRC/UKRI under grant EP/R$513180$/$1$ (ref.~$2107412$) and grant ref. EP/R$030782$/$1$ (ARoEQ). H. Gunes is supported by the EPSRC/UKRI under grant ref. EP/R$030782$/$1$ \-- ARoEQ: Adaptive Robotic Emotional Intelligence for Well-being. T.~Dimlioglu contributed to this work while undertaking a summer research study at the Department of Computer Science and Technology, University of Cambridge.

\textbf{Data Access Statement:} This study involved secondary analyses of pre-existing datasets. All datasets are described in the text and cited accordingly. Licensing restrictions prevent sharing of the datasets.

\begin{acronym}
    \acro{AC}{Affective Computing}
    \acro{Acc}{Average Accuracy Score}
    \acro{AI}{Artificial Intelligence}
    \acro{ASD}{Autism Spectrum Disorder}
    \acro{ASR}{Automatic Speech Recognition}
    \acro{AU}{Action Unit}
    \acro{AUC}{Area Under The Curve}
    \acro{AULA-Caps}{Action Unit Lifecycle Aware Capsule Network}
    
    \acro{BLSTM}{Bidirectional Long Short Term Memory}
    \acro{BMU}{Best Matching Unit}
    \acro{BP4D}{Binghamton-Pittsburgh 3D Dynamic (4D) Spontaneous Facial Expression Database}
    \acro{BWT}{Backward Transfer}

    \acro{CAAE}{Conditional Adversarial Auto-Encoder}
    \acro{cGAN}{Conditional GAN}
    \acro{CHL}{Competitive Hebbian Learning}
    \acro{CCC}{Concordance Correlation Coefficient}
    \acro{CF}{Catastrophic Forgetting}
    \acro{CHL}{Competitive Hebbian Learning}
    \acro{CK+}{Extended Cohn-Kanade}
    \acro{CL}{Continual Learning}
    \acro{Class-IL}{Class-Incremental Learning}
    \acro{CLIFER}{Continual Learning with Imagination for Facial Expression Recognition}
    \acro{CLS}{Complementary Learning Systems}
    \acro{CNN}{Convolutional Neural Network}
    \acro{ConFER}{Continual Facial Expression Recognition}
    \acro{CRL}{Continual Reinforcement Learning}

    \acro{DA}{Disentangled Feature Learning}
    \acro{DDC}{Domain Discriminative Classification}
    \acro{DDPG}{Deep Deterministic Policy Gradients}
    \acro{DEN}{Dynamically Expanding Networks}
    \acro{DGM}{Dual Generative Memory}
    \acro{DGR}{Deep Generative Replay}
    \acro{DGR+D}{Deep Generative Replay with Distillation}
    \acro{DIC}{Domain Independent Classification}
    \acro{DMC}{Deep Model Consolidation}
    \acro{Domain-IL}{Domain-Incremental Learning}

    \acro{EDA}{Electrodermal Activity}
    \acro{EEG}{Electroencephalogram}
    \acro{EWC}{Elastic Weight Consolidation}
    
    \acro{FACS}{Facial Action Coding System}
    \acro{FC}{Fully Connected}
    \acro{FEL}{Fixed Expansion Layer}
    \acro{FER}{Facial Expression Recognition}
    \acro{FL}{Focal Loss}
    \acro{FoI}{Frame-of-Interest}
    \acro{FSM}{Finite-state Machine}
    \acro{FWT}{Forward Transfer}

    \acro{GAD7}{General Anxiety Disorder}
    \acro{GAN}{Generative Adversarial Network}
    \acro{GDM}{Growing Dual Memory}
    \acro{GEM}{Gradient Episodic Memory}
    \acro{GFT}{Group Formation Task}
    \acro{GWR}{Growing When Required}

    \acro{HCI}{Human-Computer Interaction}
    \acro{HHI}{Human-Human Interaction}
    \acro{HRI}{Human-Robot Interaction}
    \acro{HSR}{Humanoid Service Robot}
    
    \acro{iCaRL}{Incremental Classifier and Representation Learning}
    \acro{iCV-MEFED}{iCV Multi-Emotion Facial Expression Dataset}
    \acro{i.i.d}{\textit{independent and identically distributed}}
    \acro{IFL}{Incremental Feature Learning}
    \acro{IL-CAN}{Incremental Learning Using Conditional Adversarial Networks}
    \acro{IoT}{Internet of Things}
    \acro{IRL}{Interactive Reinforcement Learning}

    \acro{LB}{Lower Bound}
    \acro{LGR}{Latent Generative Replay}
    \acro{LGR+D}{Latent Generative Replay with Distillation}
    \acro{LML}{Lifelong Machine Learning}
    \acro{LSTM}{Long Short-Term Memory}
    \acro{LR}{Latent Replay}
    \acro{LwF}{Learning without Forgetting}
    
    \acro{MAS}{Memory Aware Synapses}
    \acro{MC-OCL}{Memory-Constrained Online Continual Learning}  
    \acro{MeRGAN}{Memory Replay Generative Adversarial Network}    
    \acro{MFCC}{Mel-frequency Cepstral Coefficients}
    \acro{ML}{Machine Learning}
    \acro{MLP}{Multilayer Perceptron}
    \acro{MRS}{Mobile Robot Systems}
    \acro{MTL}{Multi-Task Learning}

    \acro{NC}{New Concepts}
    \acro{NDL}{Neurogenesis Deep Learning}
    \acro{NI}{New Instances}
    \acro{NIC}{New Instances and Concepts}
    \acro{NLG}{Natural Language Generation}
    \acro{NLP}{Natural Language Processing}
    \acro{NR}{Naive Rehearsal}

    \acro{OMG-Emotion}{One Minute Gradual-Emotion}
    
    \acro{PAD}{Pleasure-Arousal-Dominance}
    \acro{PANAS}{Positive and Negative Affect Schedule}
    \acro{PD}{Participatory Design}
    \acro{PFC}{Pre-Frontal Cortex}
    \acro{PHQ9}{Participant Health Questionnaire}
    \acro{PNN}{Progressive Neural Nets}
    \acro{PP}{Positive Psychology}
    \acro{PPI}{Positive Psychology Intervention}
    \acro{PRS}{Partitioning Reservoir Sampling}

    \acro{RaaS}{Robotics as a Service}
    \acro{RAF-DB}{Real-world Affective Faces Database}
    \acro{RAVDESS}{Ryerson Audio-Visual Database of Emotional Speech and Song}
    \acro{RC}{Robotic Coach}
    \acro{R-CNN}{Recurrent Convolutional Neural Network}
    \acro{ReLU}{Rectified Linear Unit}
    \acro{REM}{Rapid Eye Movement}
    \acro{RL}{Reinforcement Learning}
    \acro{ROC}{Receiver Operating Characteristics}
    \acro{ROS}{Robot Operating System}
    \acro{RoSAS}{Robotic Social Attributes Scale}
    \acro{RtF}{Replay through Feedback}

    \acro{SAR}{Socially Assistive Robot}
    \acro{SAL-DB}{Sensitive Artificial Listener Database}
    \acro{SER}{Speech Emotion Recognition}
    \acro{SGD}{Stochastic Gradient Descent}
    \acro{SI}{Synaptic Intelligence}
    \acro{SOINN}{Self-Organizing Incremental Neural Network}
    \acro{SOM}{Self-Organising Map}
    \acro{SOTA}{state-of-the-art}
    \acro{SS}{Strategic Sampling}
    \acro{SSL}{Self Supervised Learning}
    \acro{SVM}{Support-Vector Machines}
    \acro{SVR}{Support-Vector Machines for Regression}
    
    \acro{TA}{Thematic Analysis}
    \acro{TD}{Temporal Difference}
    \acro{Task-IL}{Task-Incremental Learning}
    \acro{TL}{Transfer Learning}
    \acro{TTS}{Text-to-Speech}

    \acro{UB}{Upper Bound}

    \acro{VAE}{Variational Auto-Encoder}
    \acro{VCL}{Variational Continual Learning}
    \acro{VI}{Variational Inference}
    
    \acro{WEMWBS}{Warwick-Edinburgh Mental Well-being Scale}
    \acro{WoZ}{Wizard of Oz}   
\end{acronym}

\bibliographystyle{plainnat}

\bibliography{references}

\newpage
\appendix
{

\topskip0pt
\vspace*{\fill}
\Huge
\centering
\hfill\textit{\textbf{APPENDIX}}\hfill
\vspace*{\fill}
}
\newpage
\section{Task and Class-Ordering Results}
\subsection{\acs{CK+} Results}
\subsubsection{\acf{Task-IL} Results}
\label{app:taskorderCK}

\begin{table}[h!]
\centering
\setlength\tabcolsep{3.5pt}

\caption[\acs{Task-IL} \acs{Acc} for \acs{CK+} following Orderings O2 and O3.]{\acs{Task-IL} \acs{Acc} for \acs{CK+} W/ Data-Augmentation following Orderings O2 and O3. \textbf{Bold} values denote best (highest) while [\textit{bracketed}] denote second-best values for each column.}

\label{tab:task-il-ckp-orders-acc}

{
\scriptsize
\begin{tabular}{l|llll|llll}\toprule
\multicolumn{1}{c|}{\multirow{2}{*}{\textbf{Method}}}            & \multicolumn{4}{c|}{\textbf{Accuracy following O2}} 
& \multicolumn{4}{c}{\textbf{Accuracy following O3}} \\ \cmidrule{2-9}

\multicolumn{1}{c|}{ }                          & 
\multicolumn{1}{c|}{\textbf{Task 1}}            & \multicolumn{1}{c|}{\textbf{Task 2}}          &
\multicolumn{1}{c|}{\textbf{Task 3}}            & \multicolumn{1}{c|}{\textbf{Task 4}}           &
\multicolumn{1}{c|}{\textbf{Task 1}}            & \multicolumn{1}{c|}{\textbf{Task 2}}          &
\multicolumn{1}{c|}{\textbf{Task 3}}            & \multicolumn{1}{c}{\textbf{Task 4}}          \\ \midrule

\multicolumn{9}{c}{\textbf{Baseline Approaches}} \\ \midrule
LB & $0.88\pm0.04$ & $0.83\pm0.00$ & $0.82\pm0.02$ & $0.81\pm0.03$ & $0.89\pm0.02$ & $0.81\pm0.02$ & $0.81\pm0.02$ & $0.81\pm0.04$ \\
UB & \cellcolor{gray!25}$\bm{0.92\pm0.04}$ & \cellcolor{gray!25}$\bm{0.98\pm0.00}$ & \cellcolor{gray!25}$\bm{0.99\pm0.00}$ & \cellcolor{gray!25}$\bm{0.99\pm0.00}$ & $0.88\pm0.01$ & $0.95\pm0.02$ & [$0.96\pm0.02$] & \cellcolor{gray!25}$\bm{0.98\pm0.01}$ \\
 \midrule

\multicolumn{9}{c}{\textbf{Regularisation-Based Approaches}} \\ \midrule

EWC & $0.88\pm0.04$ & $0.87\pm0.01$ & $0.88\pm0.04$ & $0.86\pm0.03$ & [$0.90\pm0.02$] & $0.90\pm0.02$ & $0.90\pm0.02$ & $0.92\pm0.01$ \\
EWC-Online  & $0.88\pm0.04$ & $0.88\pm0.01$ & $0.87\pm0.04$ & $0.85\pm0.05$ & $0.89\pm0.02$ & $0.90\pm0.00$ & $0.90\pm0.02$ & $0.89\pm0.02$ \\
SI  & $0.87\pm0.05$ & $0.82\pm0.02$ & $0.83\pm0.02$ & $0.86\pm0.00$ & $0.89\pm0.03$ & $0.83\pm0.02$ & $0.83\pm0.02$ & $0.83\pm0.02$ \\
LwF  & $0.88\pm0.05$ & $0.93\pm0.03$ & $0.95\pm0.02$ & $0.95\pm0.01$ & $0.89\pm0.02$ & $0.94\pm0.01$ & $0.95\pm0.01$ & $0.94\pm0.01$ \\ \midrule
\multicolumn{9}{c}{\textbf{Replay-Based Approaches}} \\ \midrule

NR & $0.88\pm0.04$ & \cellcolor{gray!25}$\bm{0.98\pm0.01}$ & [$0.98\pm0.01$] & [$0.98\pm0.00$] & $0.89\pm0.02$ & \cellcolor{gray!25}$\bm{0.98\pm0.01}$ & \cellcolor{gray!25}$\bm{0.97\pm0.02}$ & \cellcolor{gray!25}$\bm{0.98\pm0.01}$ \\
A-GEM  & $0.88\pm0.05$ & $0.93\pm0.06$ & $0.96\pm0.02$ & $0.96\pm0.03$ & $0.89\pm0.02$ & [$0.96\pm0.02$] & \cellcolor{gray!25}$\bm{0.97\pm0.01}$ & $0.96\pm0.02$ \\
LR  & [$0.90\pm0.01$] & [$0.95\pm0.01$] & $0.96\pm0.01$ & $0.96\pm0.00$ & \cellcolor{gray!25}$\bm{0.91\pm0.01}$ & $0.95\pm0.01$ & [$0.96\pm0.00$] & $0.96\pm0.00$ \\

DGR  & \cellcolor{gray!25}$\bm{0.92\pm0.04}$ & $0.80\pm0.03$ & $0.81\pm0.01$ & $0.80\pm0.01$ & $0.87\pm0.03$ & $0.79\pm0.04$ & $0.77\pm0.04$ & $0.77\pm0.02$ \\
LGR  & [$0.90\pm0.02$] & $0.93\pm0.01$ & $0.94\pm0.01$ & $0.95\pm0.00$ & \cellcolor{gray!25}$\bm{0.91\pm0.01}$ & $0.93\pm0.02$ & $0.95\pm0.02$ & $0.95\pm0.01$ \\
DGR+D  & \cellcolor{gray!25}$\bm{0.92\pm0.05}$ & $0.86\pm0.03$ & $0.86\pm0.03$ & $0.87\pm0.02$ & $0.87\pm0.03$ & $0.85\pm0.02$ & $0.84\pm0.01$ & $0.85\pm0.02$ \\

LGR+D  & [$0.90\pm0.02$] & [$0.95\pm0.01$] & $0.96\pm0.00$ & $0.97\pm0.00$ & \cellcolor{gray!25}$\bm{0.91\pm0.01}$ & $0.95\pm0.00$ & [$0.96\pm0.00$] & [$0.97\pm0.00$] \\

\bottomrule

\end{tabular}

}
\end{table}

\begin{table}[h!]
\centering
\caption[\acs{Task-IL} \acs{CF} scores for \acs{CK+} following O2 and O3.]{\acs{Task-IL} \acs{CF} scores for \acs{CK+} W/ Data-Augmentation following Orderings O2 and O3. \textbf{Bold} values denote best (lowest) while [\textit{bracketed}] denote second-best values for each column.}

\label{tab:task-il-ckp-orders-cf}

{
\scriptsize
\setlength\tabcolsep{3.5pt}

\begin{tabular}{l|ccc|ccc}\toprule
\multicolumn{1}{c|}{\multirow{2}{*}{\textbf{Method}}}           & \multicolumn{3}{c|}{\textbf{CF following O2}} 
& \multicolumn{3}{c}{\textbf{CF following O3}} \\ \cmidrule{2-7}

\multicolumn{1}{c|}{ }                          & \multicolumn{1}{c|}{\textbf{Task 2}}          &
\multicolumn{1}{c|}{\textbf{Task 3}}            & \multicolumn{1}{c|}{\textbf{Task 4}}           &
\multicolumn{1}{c|}{\textbf{Task 2}}          &
\multicolumn{1}{c|}{\textbf{Task 3}}            & \multicolumn{1}{c}{\textbf{Task 4}}          \\ \midrule

\multicolumn{7}{c}{\textbf{Baseline Approaches}} \\ \midrule
LB &  $0.43\pm0.04$ &  $0.33\pm0.05$ &  $0.23\pm0.04$ &  $0.45\pm0.07$ &  $0.32\pm0.09$ &  $0.26\pm0.05$ \\
UB & \cellcolor{gray!25}$\bm{-0.06\pm0.03}$ & \cellcolor{gray!25}$\bm{-0.04\pm0.02}$ & [$-0.06\pm0.03$] & \cellcolor{gray!25}$\bm{-0.08\pm0.01}$ & \cellcolor{gray!25}$\bm{-0.08\pm0.02}$ & [$-0.05\pm0.02$] \\\midrule

\multicolumn{7}{c}{\textbf{Regularisation-Based Approaches}} \\ \midrule

EWC &  $0.23\pm0.08$ &  $0.19\pm0.04$ &  $0.12\pm0.06$ &  $0.19\pm0.07$ &  $0.09\pm0.01$ &  $0.09\pm0.06$ \\
EWC-Online  &  $0.24\pm0.08$ &  $0.22\pm0.07$ &  $0.12\pm0.05$ &  $0.16\pm0.09$ &  $0.16\pm0.04$ &  $0.09\pm0.03$ \\
SI  &  $0.38\pm0.03$ &  $0.22\pm0.02$ &  $0.23\pm0.01$ &  $0.41\pm0.08$ &  $0.29\pm0.05$ &  $0.23\pm0.07$ \\
LwF  &  $0.03\pm0.00$ &  $0.03\pm0.01$ &  $0.01\pm0.02$ &  $0.05\pm0.02$ &  $0.05\pm0.02$ &  $0.01\pm0.00$ \\ \midrule

\multicolumn{7}{c}{\textbf{Replay-Based Approaches}} \\ \midrule

NR & \cellcolor{gray!25}$\bm{-0.06\pm0.07}$ & [$-0.03\pm0.01$] & \cellcolor{gray!25}$\bm{-0.09\pm0.03}$ & [$-0.04\pm0.02$] & [$-0.01\pm0.01$] & \cellcolor{gray!25}$\bm{-0.07\pm0.01}$ \\
A-GEM  & [$-0.02\pm0.03$] &  $0.02\pm0.03$ &  $0.01\pm0.08$ & $-0.03\pm0.01$ &  $0.02\pm0.03$ & $-0.02\pm0.03$ \\
LR  &  $0.01\pm0.02$ &  $0.01\pm0.01$ & $-0.02\pm0.00$ &  $0.02\pm0.01$ &  $0.01\pm0.00$ & $-0.01\pm0.00$ \\
DGR  &  $0.50\pm0.06$ &  $0.36\pm0.05$ &  $0.32\pm0.06$ &  $0.53\pm0.06$ &  $0.38\pm0.04$ &  $0.27\pm0.08$ \\
LGR  &  $0.05\pm0.01$ &  $0.03\pm0.01$ &  $0.04\pm0.01$ &  $0.05\pm0.05$ &  $0.04\pm0.02$ &  $0.03\pm0.04$ \\
DGR+D  &  $0.32\pm0.09$ &  $0.22\pm0.06$ &  $0.20\pm0.05$ &  $0.34\pm0.05$ &  $0.23\pm0.04$ &  $0.17\pm0.05$ \\
LGR+D  &  $0.00\pm0.01$ &  $0.00\pm0.00$ & $-0.01\pm0.00$ &  $0.00\pm0.01$ &  $0.01\pm0.00$ &  $0.00\pm0.01$ \\

\bottomrule

\end{tabular}

}
\end{table}

\begin{figure}[t!] 
 \centering
 \subfloat[\acs{Task-IL} Results following O$2$.\label{fig:ckp-task-il-aug-order-1}]{\includegraphics[width=0.5\textwidth]{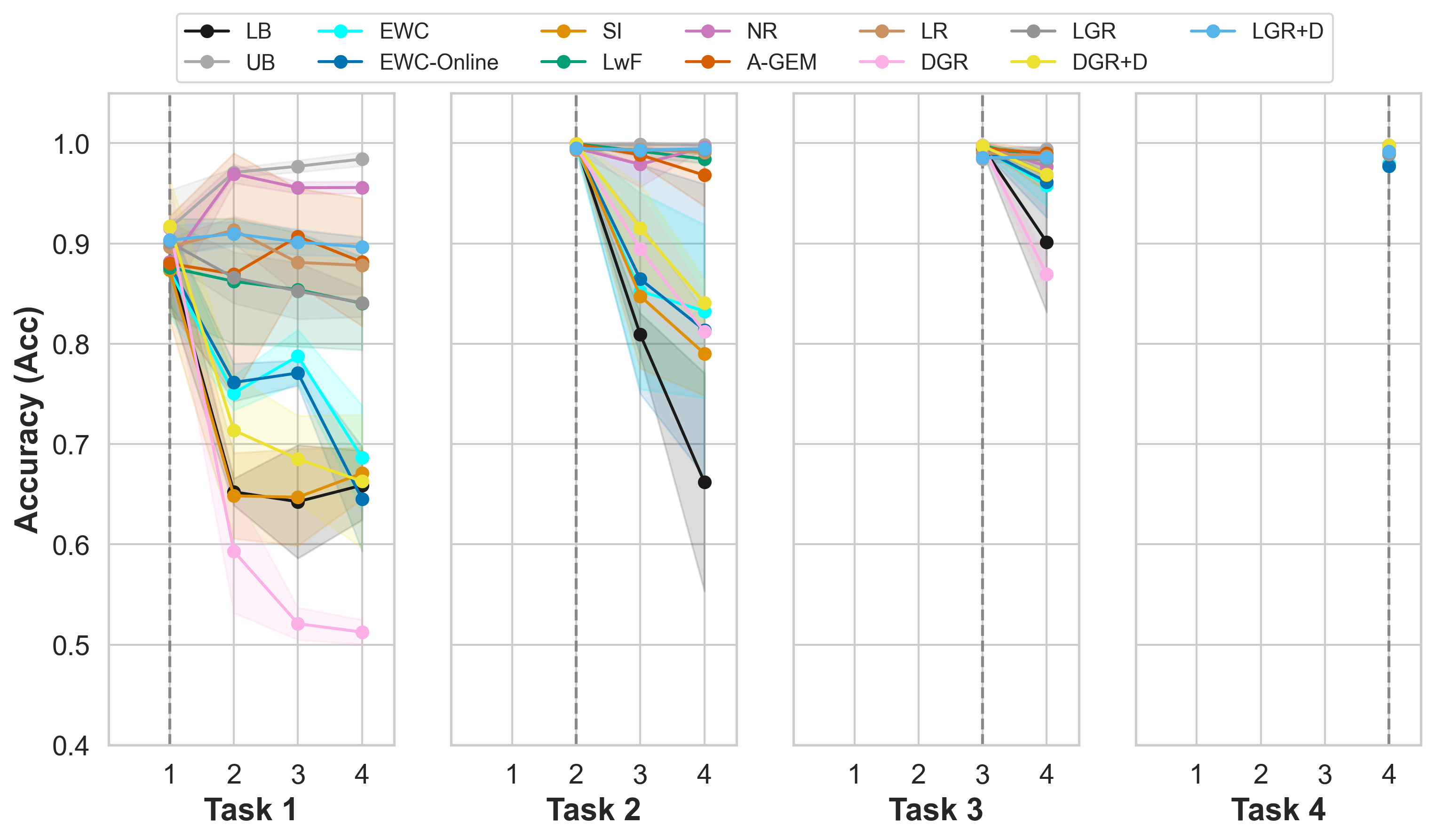}} \hfill
 \subfloat[\acs{Task-IL} Results following O$3$.\label{fig:ckp-task-il-aug-order-2}]{\includegraphics[width=0.5\textwidth]{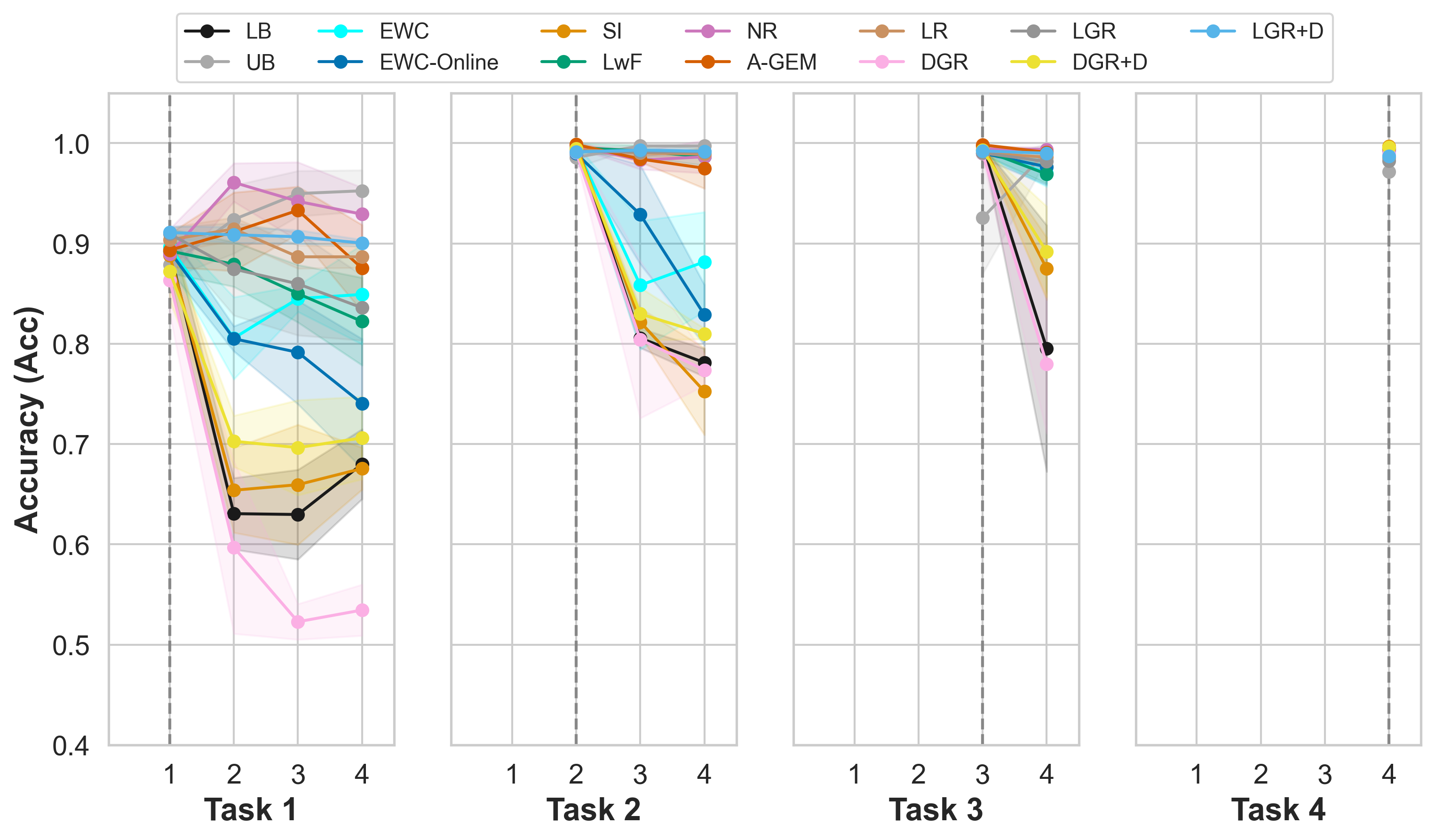}} \hfill
 \caption[Task-IL Results for CK+ following Order 2 and Order 3.]{\acs{Task-IL} results for \acs{CK+} following (a) Order $2$ and (b) Order $3$, respectively. For each task, test-set accuracy is shown as the learning progresses from when the task is introduced to the end of the overall training procedure.}
 \label{fig:ckp-taskil-orders} 
    
\end{figure}


\normalsize
\textit{}
\newpage

\subsubsection{\acf{Class-IL} Results}
\label{app:classorderCK}

\begin{table}[h!]
\centering
\setlength\tabcolsep{3.5pt}

\caption[\acs{Class-IL} \acs{Acc} for \acs{CK+} following Orderings O2 and O3.]{\acs{Class-IL} \acs{Acc} for \acs{CK+}  W/ Data-Augmentation following Orderings O2 and O3. \textbf{Bold} values denote best (highest) while [\textit{bracketed}] denote second-best values for each column.}
\label{tab:class-il-ck-acc-orders}

{
\scriptsize
\begin{tabular}{l|cccccccc}\toprule

\multirow{2}[2]{*}{\makecell[c]{\textbf{Method}}}           & \multicolumn{8}{c}{\textbf{Accuracy following O2}} \\ \cmidrule{2-9}

\multicolumn{1}{c|}{ }                          & 
\multicolumn{1}{c|}{\textbf{Class 1}}            & \multicolumn{1}{c|}{\textbf{Class 2}}          &
\multicolumn{1}{c|}{\textbf{Class 3}}            & \multicolumn{1}{c|}{\textbf{Class 4}}           &
\multicolumn{1}{c|}{\textbf{Class 5}}            & \multicolumn{1}{c|}{\textbf{Class 6}}          &
\multicolumn{1}{c|}{\textbf{Class 7}}            & \multicolumn{1}{c}{\textbf{Class 8}}          \\ \midrule

\multicolumn{9}{c}{\textbf{Baseline Approaches}} \\ \midrule

LB & \cellcolor{gray!25}$\bm{1.00\pm0.00}$ & $0.50\pm0.00$ & $0.33\pm0.00$ & $0.25\pm0.00$ & $0.20\pm0.00$ & $0.17\pm0.00$ & $0.14\pm0.00$ & $0.12\pm0.00$ \\
UB & \cellcolor{gray!25}$\bm{1.00\pm0.00}$ & \cellcolor{gray!25}$\bm{0.99\pm0.00}$ & \cellcolor{gray!25}$\bm{0.99\pm0.00}$ & \cellcolor{gray!25}$\bm{0.99\pm0.00}$ & \cellcolor{gray!25}$\bm{0.99\pm0.00}$ & \cellcolor{gray!25}$\bm{0.99\pm0.00}$ & \cellcolor{gray!25}$\bm{0.99\pm0.00}$ & \cellcolor{gray!25}$\bm{0.99\pm0.00}$ \\ \midrule

\multicolumn{9}{c}{\textbf{Regularisation-Based Approaches}} \\ \midrule
EWC  & \cellcolor{gray!25}$\bm{1.00\pm0.00}$ & $0.50\pm0.00$ & $0.33\pm0.00$ & $0.25\pm0.00$ & $0.20\pm0.00$ & $0.17\pm0.00$ & $0.14\pm0.00$ & $0.12\pm0.00$ \\
EWC-Online  & \cellcolor{gray!25}$\bm{1.00\pm0.00}$ & $0.50\pm0.00$ & $0.33\pm0.00$ & $0.25\pm0.00$ & $0.20\pm0.00$ & $0.17\pm0.00$ & $0.14\pm0.00$ & $0.12\pm0.00$ \\
SI  & \cellcolor{gray!25}$\bm{1.00\pm0.00}$ & $0.50\pm0.00$ & $0.33\pm0.00$ & $0.25\pm0.00$ & $0.20\pm0.00$ & $0.17\pm0.00$ & $0.14\pm0.00$ & $0.12\pm0.00$ \\
LwF  & \cellcolor{gray!25}$\bm{1.00\pm0.00}$ & $0.50\pm0.00$ & $0.33\pm0.00$ & $0.25\pm0.00$ & $0.20\pm0.00$ & $0.17\pm0.00$ & $0.14\pm0.00$ & $0.12\pm0.00$ \\\midrule
\multicolumn{9}{c}{\textbf{Replay-Based Approaches}} \\ \midrule

NR  & \cellcolor{gray!25}$\bm{1.00\pm0.00}$ & [$0.98\pm0.00$] & [$0.97\pm0.01$] & [$0.96\pm0.00$] & [$0.96\pm0.00$] & [$0.95\pm0.01$] & [$0.93\pm0.01$] & [$0.92\pm0.00$] \\
A-GEM  & \cellcolor{gray!25}$\bm{1.00\pm0.00}$ & $0.51\pm0.02$ & $0.67\pm0.18$ & $0.26\pm0.01$ & $0.21\pm0.01$ & $0.25\pm0.11$ & $0.19\pm0.07$ & $0.12\pm0.00$ \\
LR  & \cellcolor{gray!25}$\bm{1.00\pm0.00}$ & $0.93\pm0.03$ & $0.93\pm0.02$ & $0.92\pm0.03$ & $0.92\pm0.03$ & $0.93\pm0.03$ & $0.91\pm0.03$ & $0.91\pm0.03$ \\
DGR  & \cellcolor{gray!25}$\bm{1.00\pm0.00}$ & $0.51\pm0.00$ & $0.37\pm0.01$ & $0.26\pm0.00$ & $0.22\pm0.01$ & $0.18\pm0.01$ & $0.15\pm0.00$ & $0.13\pm0.00$ \\
LGR  & \cellcolor{gray!25}$\bm{1.00\pm0.00}$ & $0.83\pm0.02$ & $0.83\pm0.03$ & $0.82\pm0.05$ & $0.83\pm0.04$ & $0.83\pm0.06$ & $0.74\pm0.10$ & $0.70\pm0.12$ \\
DGR+D  & \cellcolor{gray!25}$\bm{1.00\pm0.00}$ & $0.52\pm0.00$ & $0.38\pm0.03$ & $0.26\pm0.01$ & $0.21\pm0.01$ & $0.18\pm0.00$ & $0.15\pm0.00$ & $0.13\pm0.00$ \\
LGR+D  & \cellcolor{gray!25}$\bm{1.00\pm0.00}$ & $0.86\pm0.03$ & $0.87\pm0.04$ & $0.85\pm0.05$ & $0.86\pm0.05$ & $0.87\pm0.05$ & $0.82\pm0.06$ & $0.82\pm0.06$\\\midrule

\multirow{2}[2]{*}{\makecell[c]{\textbf{Method}}}           & \multicolumn{8}{c}{\textbf{Accuracy following O3}} \\ \cmidrule{2-9}

\multicolumn{1}{c|}{ }                          & 
\multicolumn{1}{c|}{\textbf{Class 1}}            & \multicolumn{1}{c|}{\textbf{Class 2}}          &
\multicolumn{1}{c|}{\textbf{Class 3}}            & \multicolumn{1}{c|}{\textbf{Class 4}}           &
\multicolumn{1}{c|}{\textbf{Class 5}}            & \multicolumn{1}{c|}{\textbf{Class 6}}          &
\multicolumn{1}{c|}{\textbf{Class 7}}            & \multicolumn{1}{c}{\textbf{Class 8}}          \\ \midrule
\multicolumn{9}{c}{\textbf{Baseline Approaches}} \\ \midrule

LB & \cellcolor{gray!25}$\bm{1.00\pm0.00}$ & $0.50\pm0.00$ & $0.33\pm0.00$ & $0.25\pm0.00$ & $0.20\pm0.00$ & $0.17\pm0.00$ & $0.14\pm0.00$ & $0.12\pm0.00$ \\
UB & \cellcolor{gray!25}$\bm{1.00\pm0.00}$ & \cellcolor{gray!25}$\bm{0.99\pm0.00}$ & \cellcolor{gray!25}$\bm{0.99\pm0.00}$ & \cellcolor{gray!25}$\bm{0.99\pm0.00}$ & \cellcolor{gray!25}$\bm{0.99\pm0.00}$ & \cellcolor{gray!25}$\bm{0.99\pm0.00}$ & \cellcolor{gray!25}$\bm{0.99\pm0.00}$ & \cellcolor{gray!25}$\bm{0.99\pm0.00}$ \\\midrule

\multicolumn{9}{c}{\textbf{Regularisation-Based Approaches}} \\ \midrule

EWC  & \cellcolor{gray!25}$\bm{1.00\pm0.00}$ & $0.50\pm0.00$ & $0.33\pm0.00$ & $0.25\pm0.00$ & $0.20\pm0.00$ & $0.17\pm0.00$ & $0.14\pm0.00$ & $0.12\pm0.00$ \\
EWC-Online  & \cellcolor{gray!25}$\bm{1.00\pm0.00}$ & $0.50\pm0.00$ & $0.33\pm0.00$ & $0.25\pm0.00$ & $0.20\pm0.00$ & $0.17\pm0.00$ & $0.14\pm0.00$ & $0.12\pm0.00$ \\
SI  & \cellcolor{gray!25}$\bm{1.00\pm0.00}$ & $0.50\pm0.00$ & $0.33\pm0.00$ & $0.25\pm0.00$ & $0.20\pm0.00$ & $0.17\pm0.00$ & $0.14\pm0.00$ & $0.12\pm0.00$ \\
LwF  & \cellcolor{gray!25}$\bm{1.00\pm0.00}$ & $0.50\pm0.00$ & $0.33\pm0.00$ & $0.25\pm0.00$ & $0.20\pm0.00$ & $0.17\pm0.00$ & $0.14\pm0.00$ & $0.12\pm0.00$ \\\midrule
\multicolumn{9}{c}{\textbf{Replay-Based Approaches}} \\ \midrule

NR  & \cellcolor{gray!25}$\bm{1.00\pm0.00}$ & [$0.98\pm0.01$] & [$0.97\pm0.01$] & [$0.97\pm0.00$] & [$0.96\pm0.01$] & [$0.96\pm0.00$] & [$0.93\pm0.01$] & $0.91\pm0.00$ \\
A-GEM  & \cellcolor{gray!25}$\bm{1.00\pm0.00}$ & $0.57\pm0.05$ & $0.59\pm0.20$ & $0.25\pm0.00$ & $0.40\pm0.18$ & $0.22\pm0.08$ & $0.19\pm0.07$ & $0.12\pm0.00$ \\
LR  & \cellcolor{gray!25}$\bm{1.00\pm0.00}$ & $0.94\pm0.02$ & $0.94\pm0.01$ & $0.93\pm0.01$ & $0.93\pm0.01$ & $0.94\pm0.01$ & $0.92\pm0.01$ & [$0.92\pm0.01$] \\
DGR  & \cellcolor{gray!25}$\bm{1.00\pm0.00}$ & $0.51\pm0.00$ & $0.36\pm0.02$ & $0.26\pm0.02$ & $0.21\pm0.00$ & $0.18\pm0.01$ & $0.15\pm0.01$ & $0.13\pm0.01$ \\
LGR  & \cellcolor{gray!25}$\bm{1.00\pm0.00}$ & $0.83\pm0.02$ & $0.85\pm0.02$ & $0.83\pm0.01$ & $0.83\pm0.01$ & $0.84\pm0.00$ & $0.79\pm0.02$ & $0.76\pm0.04$ \\
DGR+D  & \cellcolor{gray!25}$\bm{1.00\pm0.00}$ & $0.51\pm0.00$ & $0.38\pm0.03$ & $0.28\pm0.02$ & $0.22\pm0.00$ & $0.18\pm0.01$ & $0.15\pm0.01$ & $0.13\pm0.00$ \\
LGR+D  & \cellcolor{gray!25}$\bm{1.00\pm0.00}$ & $0.86\pm0.02$ & $0.88\pm0.02$ & $0.87\pm0.02$ & $0.87\pm0.02$ & $0.87\pm0.03$ & $0.82\pm0.03$ & $0.82\pm0.03$ \\

\bottomrule

\end{tabular}

}
\end{table}

\begin{table}[h!]
\centering
\caption[\acs{Class-IL} \acs{CF} scores for \acs{CK+}  following O2 and O3.]{\acs{Class-IL} \acs{CF} scores for \acs{CK+} W/ Data-Augmentation following Orderings O2 and O3. \textbf{Bold} values denote best (lowest) while [\textit{bracketed}] denote second-best values for each column.}

\label{tab:class-il-ck-cf-orders}

{
\scriptsize
\setlength\tabcolsep{3.5pt}

\begin{tabular}{l|crrrrrrr}\toprule

\multirow{2}[2]{*}{\makecell[c]{\textbf{Method}}}           & \multicolumn{7}{c}{\textbf{\acs{CF} following O2}} \\ \cmidrule{2-8}

\multicolumn{1}{c|}{ }                          & \multicolumn{1}{c|}{\textbf{Class 2}}          &
\multicolumn{1}{c|}{\textbf{Class 3}}            & \multicolumn{1}{c|}{\textbf{Class 4}}           &
\multicolumn{1}{c|}{\textbf{Class 5}}            & \multicolumn{1}{c|}{\textbf{Class 6}}          &
\multicolumn{1}{c|}{\textbf{Class 7}}            & \multicolumn{1}{c}{\textbf{Class 8}}          \\ \midrule

\multicolumn{8}{c}{\textbf{Baseline Approaches}} \\ \midrule

LB &  $1.00\pm0.00$ &  $1.00\pm0.00$ &  $1.00\pm0.00$ &  $1.00\pm0.00$ &  $1.00\pm0.00$ &  $1.00\pm0.00$ & $1.00\pm0.00$ \\
UB & \cellcolor{gray!25}$\bm{-0.03\pm0.09}$ & \cellcolor{gray!25}$\bm{-0.05\pm0.09}$ & \cellcolor{gray!25}$\bm{-0.03\pm0.08}$ & \cellcolor{gray!25}$\bm{-0.04\pm0.06}$ & \cellcolor{gray!25}$\bm{-0.02\pm0.05}$ & \cellcolor{gray!25}$\bm{-0.02\pm0.05}$ & \cellcolor{gray!25}$\bm{0.05\pm0.03}$ \\\midrule

\multicolumn{8}{c}{\textbf{Regularisation-Based Approaches}} \\ \midrule

EWC  &  $1.00\pm0.00$ &  $1.00\pm0.00$ &  $1.00\pm0.00$ &  $1.00\pm0.00$ &  $1.00\pm0.00$ &  $1.00\pm0.00$ & $1.00\pm0.00$ \\
EWC-Online  &  $1.00\pm0.00$ &  $1.00\pm0.00$ &  $1.00\pm0.00$ &  $1.00\pm0.00$ &  $1.00\pm0.00$ &  $1.00\pm0.00$ & $1.00\pm0.00$ \\
SI  &  $1.00\pm0.00$ &  $1.00\pm0.00$ &  $1.00\pm0.00$ &  $1.00\pm0.00$ &  $1.00\pm0.00$ &  $1.00\pm0.00$ & $1.00\pm0.00$ \\
LwF  &  $1.00\pm0.00$ &  $1.00\pm0.00$ &  $1.00\pm0.00$ &  $1.00\pm0.00$ &  $1.00\pm0.00$ &  $1.00\pm0.00$ & $1.00\pm0.00$ \\ \midrule

\multicolumn{8}{c}{\textbf{Replay-Based Approaches}} \\ \midrule

NR  &  [$0.03\pm0.03$] &  [$0.03\pm0.01$] &  [$0.02\pm0.00$] &  [$0.04\pm0.01$] &  [$0.06\pm0.02$] &  [$0.04\pm0.01$] & [$0.06\pm0.01$] \\
A-GEM  &  $0.76\pm0.17$ &  $1.00\pm0.00$ &  $1.00\pm0.00$ &  $1.00\pm0.00$ &  $1.00\pm0.00$ &  $1.00\pm0.00$ & $0.82\pm0.13$ \\
LR  &  $0.16\pm0.03$ &  $0.12\pm0.04$ &  $0.10\pm0.02$ &  $0.07\pm0.02$ &  $0.07\pm0.03$ &  $0.07\pm0.03$ & $0.23\pm0.01$ \\
DGR  &  $0.91\pm0.01$ &  $0.98\pm0.02$ &  $0.98\pm0.02$ &  $0.98\pm0.02$ &  $0.99\pm0.00$ &  $0.99\pm0.01$ & $1.00\pm0.00$ \\
LGR  &  $0.42\pm0.03$ &  $0.35\pm0.05$ &  $0.52\pm0.01$ &  $0.49\pm0.01$ &  $0.66\pm0.03$ &  $0.60\pm0.03$ & $0.54\pm0.04$ \\
DGR+D  &  $0.89\pm0.05$ &  $0.98\pm0.02$ &  $0.99\pm0.00$ &  $0.98\pm0.01$ &  $0.99\pm0.01$ &  $1.00\pm0.00$ & $0.98\pm0.01$ \\
LGR+D  &  $0.32\pm0.08$ &  $0.28\pm0.03$ &  $0.32\pm0.04$ &  $0.35\pm0.04$ &  $0.41\pm0.02$ &  $0.40\pm0.05$ & $0.39\pm0.08$ \\ \midrule

\multirow{2}[2]{*}{\makecell[c]{\textbf{Method}}}           & \multicolumn{7}{c}{\textbf{\acs{CF} following O3}} \\ \cmidrule{2-8}

\multicolumn{1}{c|}{ }                          & \multicolumn{1}{c|}{\textbf{Class 2}}          &
\multicolumn{1}{c|}{\textbf{Class 3}}            & \multicolumn{1}{c|}{\textbf{Class 4}}           &
\multicolumn{1}{c|}{\textbf{Class 5}}            & \multicolumn{1}{c|}{\textbf{Class 6}}          &
\multicolumn{1}{c|}{\textbf{Class 7}}            & \multicolumn{1}{c}{\textbf{Class 8}}          \\ \midrule

\multicolumn{8}{c}{\textbf{Baseline Approaches}} \\ \midrule
LB &  $1.00\pm0.00$ &  $1.00\pm0.00$ &  $1.00\pm0.00$ &  $1.00\pm0.00$ &  $1.00\pm0.00$ &  $1.00\pm0.00$ & $1.00\pm0.00$ \\
UB &  \cellcolor{gray!25}$\bm{0.01\pm0.00}$ &  \cellcolor{gray!25}$\bm{0.01\pm0.00}$ &  \cellcolor{gray!25}$\bm{0.00\pm0.00}$ & \cellcolor{gray!25}$\bm{0.00\pm0.00}$ &  \cellcolor{gray!25}$\bm{0.01\pm0.00}$ &  \cellcolor{gray!25}$\bm{0.00\pm0.00}$ & \cellcolor{gray!25}$\bm{0.01\pm0.00}$ \\ \midrule

\multicolumn{8}{c}{\textbf{Regularisation-Based Approaches}} \\ \midrule

EWC  &  $1.00\pm0.00$ &  $1.00\pm0.00$ &  $1.00\pm0.00$ &  $1.00\pm0.00$ &  $1.00\pm0.00$ &  $1.00\pm0.00$ & $1.00\pm0.00$ \\
EWC-Online  &  $1.00\pm0.00$ &  $1.00\pm0.00$ &  $1.00\pm0.00$ &  $1.00\pm0.00$ &  $1.00\pm0.00$ &  $1.00\pm0.00$ & $1.00\pm0.00$ \\
SI  &  $1.00\pm0.00$ &  $1.00\pm0.00$ &  $1.00\pm0.00$ &  $1.00\pm0.00$ &  $1.00\pm0.00$ &  $1.00\pm0.00$ & $1.00\pm0.00$ \\
LwF  &  $1.00\pm0.00$ &  $1.00\pm0.00$ &  $1.00\pm0.00$ &  $1.00\pm0.00$ &  $1.00\pm0.00$ &  $1.00\pm0.00$ & $1.00\pm0.00$ \\ \midrule

\multicolumn{8}{c}{\textbf{Replay-Based Approaches}} \\ \midrule

NR  &  [$0.03\pm0.00$] &  [$0.04\pm0.01$] &  [$0.05\pm0.00$] &  [$0.04\pm0.00$] &  $0.08\pm0.00$ &  $0.10\pm0.01$ & [$0.03\pm0.01$] \\
A-GEM  &  $0.61\pm0.28$ &  $0.99\pm0.01$ &  $0.95\pm0.07$ &  $0.80\pm0.03$ &  $0.82\pm0.02$ &  $0.94\pm0.05$ & $0.86\pm0.20$ \\
LR  &  $0.05\pm0.02$ &  $0.05\pm0.01$ &  [$0.05\pm0.01$] &  [$0.04\pm0.01$] &  [$0.05\pm0.01$] &  [$0.04\pm0.01$] & $0.06\pm0.01$ \\
DGR  &  $0.90\pm0.04$ &  $0.95\pm0.00$ &  $0.96\pm0.02$ &  $0.96\pm0.01$ &  $0.98\pm0.02$ &  $0.97\pm0.03$ & $0.98\pm0.00$ \\
LGR  &  $0.21\pm0.06$ &  $0.19\pm0.06$ &  $0.18\pm0.06$ &  $0.18\pm0.08$ &  $0.22\pm0.09$ &  $0.22\pm0.08$ & $0.31\pm0.09$ \\
DGR+D  &  $0.87\pm0.02$ &  $0.93\pm0.05$ &  $0.97\pm0.01$ &  $0.97\pm0.02$ &  $0.99\pm0.01$ &  $0.99\pm0.01$ & $0.96\pm0.01$ \\
LGR+D  &  $0.16\pm0.06$ &  $0.15\pm0.06$ &  $0.14\pm0.07$ &  $0.14\pm0.08$ &  $0.17\pm0.09$ &  $0.16\pm0.08$ & $0.23\pm0.08$ \\

\bottomrule

\end{tabular}

}
\end{table}

\begin{figure}[h!]  
    \centering
    \subfloat[\acs{Class-IL} Results following O$2$.\label{fig:ckp-Class-il-aug-order-1}]{\includegraphics[width=0.5\textwidth]{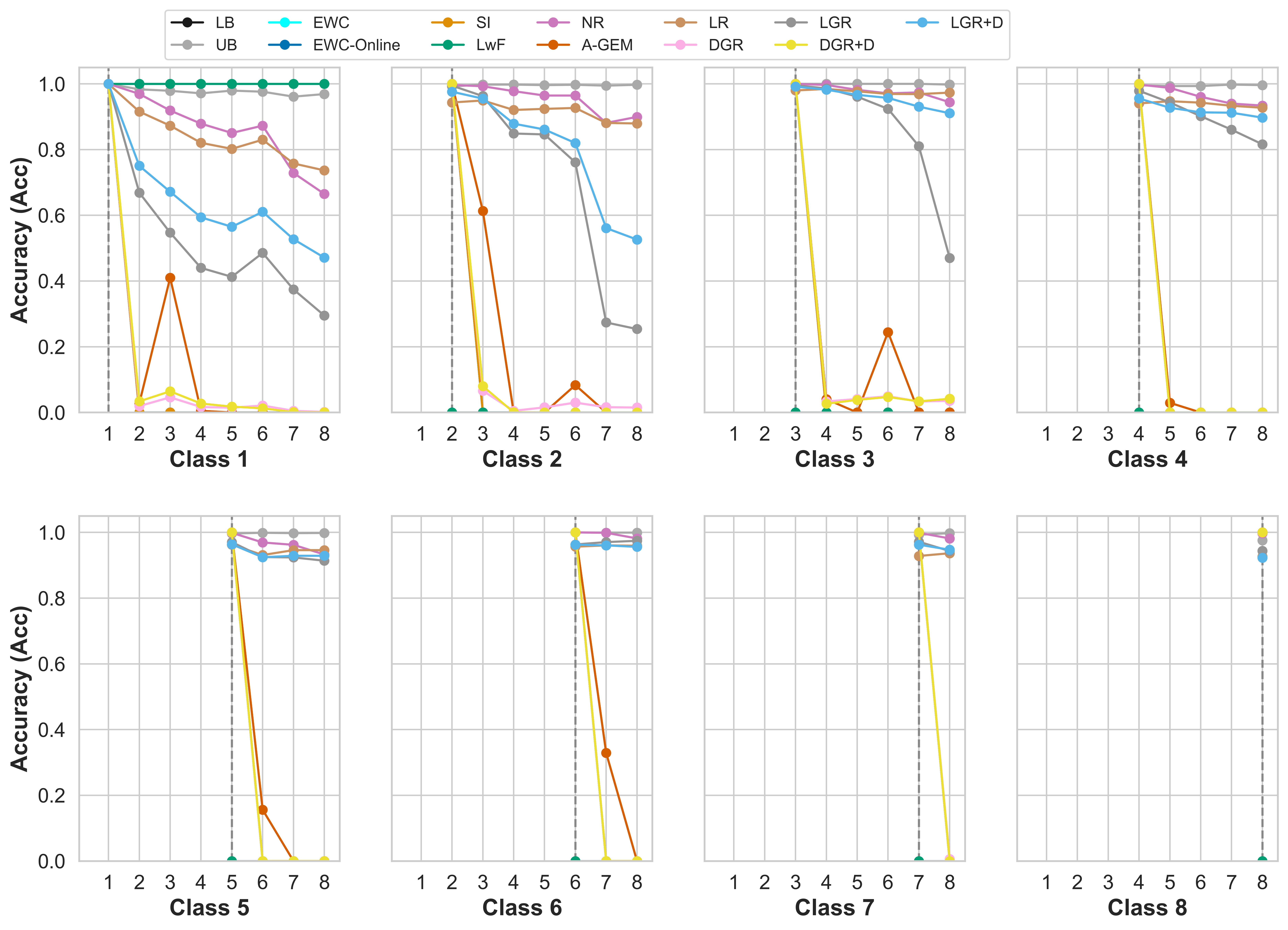}}
    \hfill
    \subfloat[\acs{Class-IL} Results following O$3$.\label{fig:ckp-Class-il-aug-order-2}]{\includegraphics[width=0.5\textwidth]{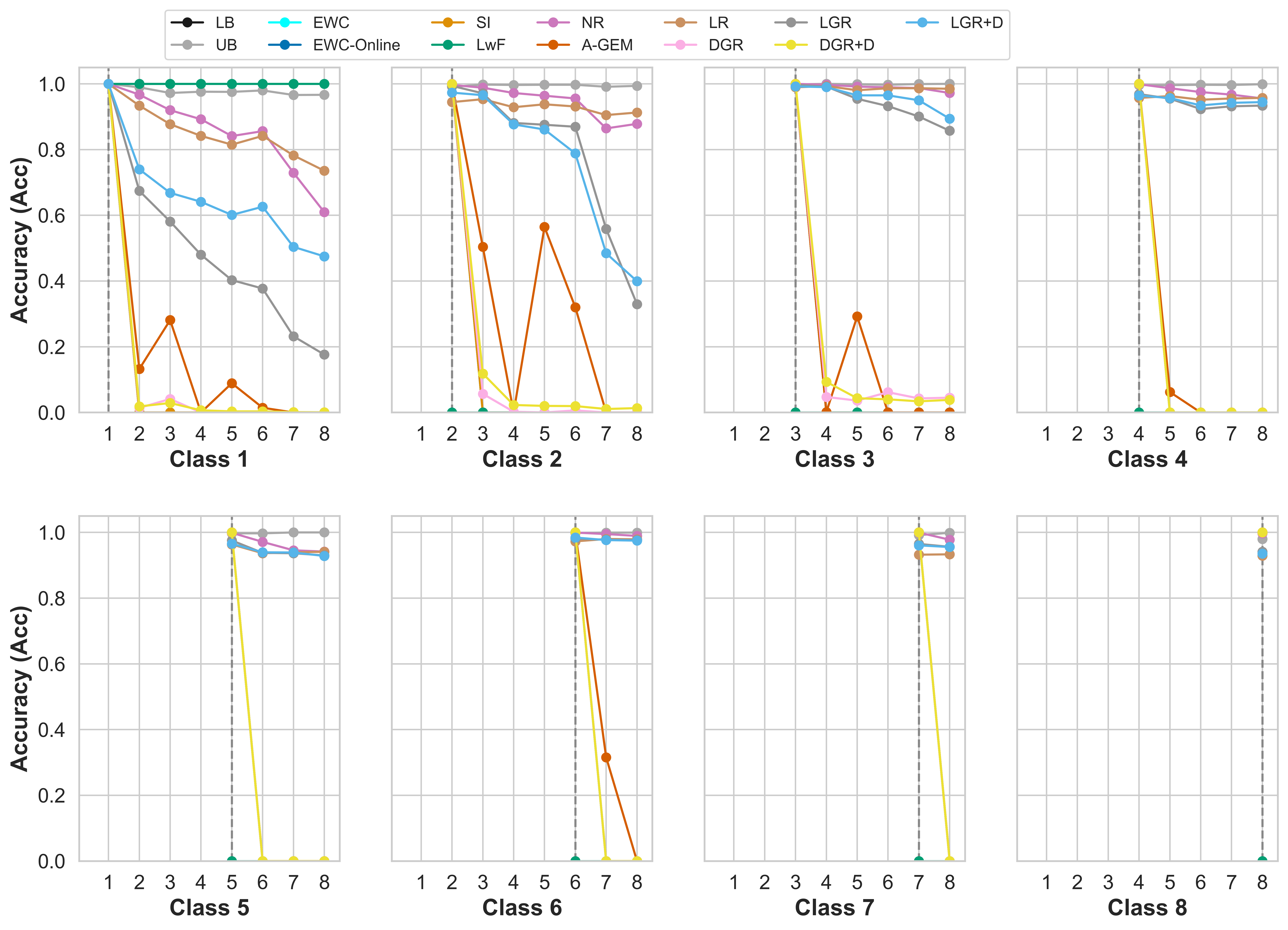}} \\
    \caption[Class-IL Results for \acs{CK+} following Order 2 and Order 3.]{\acs{Class-IL} results for \acs{CK+} following (a) Order $2$ and (b) Order $3$, respectively. For each class, test-set accuracy is shown as the learning progresses from when the class is introduced to the end of the overall training procedure.}
    \label{fig:ckp-classil-orders} 
    
\end{figure}

\FloatBarrier

\normalsize

\subsection{\acs{RAF-DB} Results}
\subsubsection{\acf{Task-IL} Results}
\label{app:taskorderrafdb}
\begin{table}[h!]
\centering
\setlength\tabcolsep{2.0pt}

\caption[\acs{Task-IL} \acs{Acc} for \acs{RAF-DB} following Orderings O2 and O3.]{\acs{Task-IL} \acs{Acc} for \acs{RAF-DB} W/ Data-Augmentation following Orderings O2 and O3. \textbf{Bold} values denote best (highest) while [\textit{bracketed}] denote second-best values for each column.}

\label{tab:task-il-rafdb-orders-acc}

{
\scriptsize
\begin{tabular}{l|cccc|cccc}\toprule
\multicolumn{1}{c|}{\textbf{Method}}            & \multicolumn{4}{c|}{\textbf{Accuracy following O2}} 
& \multicolumn{4}{c}{\textbf{Accuracy following O3}} \\ \cmidrule{2-9}

\multicolumn{1}{c|}{ }                          & 
\multicolumn{1}{c|}{\textbf{Task 1}}            & \multicolumn{1}{c|}{\textbf{Task 2}}          &
\multicolumn{1}{c|}{\textbf{Task 3}}            & \multicolumn{1}{c|}{\textbf{Task 4}}           &
\multicolumn{1}{c|}{\textbf{Task 1}}            & \multicolumn{1}{c|}{\textbf{Task 2}}          &
\multicolumn{1}{c|}{\textbf{Task 3}}            & \multicolumn{1}{c}{\textbf{Task 4}}          \\\midrule 

\multicolumn{9}{c}{\textbf{Baseline Approaches}} \\ \midrule
LB & \cellcolor{gray!25}$\bm{0.94\pm0.01}$ & $0.78\pm0.01$ & $0.79\pm0.05$ & $0.72\pm0.05$ 
& \cellcolor{gray!25}$\bm{0.93\pm0.00}$ & $0.79\pm0.01$ & $0.79\pm0.01$ & $0.67\pm0.03$ \\
UB & \cellcolor{gray!25}$\bm{0.94\pm0.00}$ & \cellcolor{gray!25}$\bm{0.88\pm0.00}$ & \cellcolor{gray!25}$\bm{0.87\pm0.00}$ & \cellcolor{gray!25}$\bm{0.90\pm0.00}$ 
& \cellcolor{gray!25}$\bm{0.93\pm0.00}$ & [$0.87\pm0.00$] & \cellcolor{gray!25}$\bm{0.87\pm0.00}$ & \cellcolor{gray!25}$\bm{0.91\pm0.00}$ \\
 \midrule

\multicolumn{9}{c}{\textbf{Regularisation-Based Approaches}} \\ \midrule

EWC  & \cellcolor{gray!25}$\bm{0.94\pm0.01}$ & $0.83\pm0.03$ & $0.85\pm0.02$ & $0.85\pm0.02$ 
& \cellcolor{gray!25}$\bm{0.93\pm0.00}$ & $0.84\pm0.01$ & $0.84\pm0.00$ & $0.82\pm0.01$ \\
EWC-Online  & \cellcolor{gray!25}$\bm{0.94\pm0.01}$ & $0.83\pm0.03$ & $0.85\pm0.01$ & $0.82\pm0.02$ 
& \cellcolor{gray!25}$\bm{0.93\pm0.00}$ & $0.84\pm0.01$ & $0.84\pm0.00$ & $0.87\pm0.00$ \\
SI  & \cellcolor{gray!25}$\bm{0.94\pm0.01}$ & $0.78\pm0.01$ & $0.79\pm0.05$ & $0.70\pm0.05$ 
& \cellcolor{gray!25}$\bm{0.93\pm0.00}$ & $0.79\pm0.01$ & $0.79\pm0.01$ & $0.72\pm0.01$ \\
LwF  & \cellcolor{gray!25}$\bm{0.94\pm0.01}$ & \cellcolor{gray!25}$\bm{0.88\pm0.00}$ & [$0.86\pm0.00$] & \cellcolor{gray!25}$\bm{0.90\pm0.00}$ 
& \cellcolor{gray!25}$\bm{0.93\pm0.00}$ & [$0.87\pm0.00$] & \cellcolor{gray!25}$\bm{0.87\pm0.00}$ & [$0.90\pm0.00$] \\\midrule

\multicolumn{9}{c}{\textbf{Replay-Based Approaches}} \\ \midrule

NR  & \cellcolor{gray!25}$\bm{0.94\pm0.01}$ & \cellcolor{gray!25}$\bm{0.88\pm0.00}$ & $0.84\pm0.00$ & $0.84\pm0.01$ 
& \cellcolor{gray!25}$\bm{0.93\pm0.00}$ & \cellcolor{gray!25}$\bm{0.88\pm0.00}$ & $0.84\pm0.00$ & $0.83\pm0.01$ \\
A-GEM  & \cellcolor{gray!25}$\bm{0.94\pm0.01}$ & \cellcolor{gray!25}$\bm{0.88\pm0.00}$ & \cellcolor{gray!25}$\bm{0.87\pm0.01}$ & $0.80\pm0.04$ 
& \cellcolor{gray!25}$\bm{0.93\pm0.00}$ & \cellcolor{gray!25}$\bm{0.88\pm0.00}$ & \cellcolor{gray!25}$\bm{0.87\pm0.00}$ & $0.86\pm0.00$ \\
LR  & $0.91\pm0.00$ & [$0.87\pm0.01$] & $0.82\pm0.00$ & $0.85\pm0.01$ 
& [$0.92\pm0.01$] & [$0.87\pm0.00$] & $0.83\pm0.00$ & $0.85\pm0.01$ \\
DGR  & [$0.93\pm0.01$] & $0.80\pm0.06$ & $0.78\pm0.05$ & $0.78\pm0.04$ 
& \cellcolor{gray!25}$\bm{0.93\pm0.01}$ & $0.83\pm0.02$ & $0.81\pm0.01$ & $0.78\pm0.04$ \\
LGR  & $0.92\pm0.01$ & $0.86\pm0.00$ & $0.85\pm0.00$ & $0.88\pm0.00$ 
& [$0.92\pm0.00$] & $0.86\pm0.00$ & [$0.85\pm0.00$] & $0.89\pm0.00$ \\
DGR+D  & [$0.93\pm0.01$] & $0.85\pm0.02$ & $0.80\pm0.01$ & $0.80\pm0.01$ 
& \cellcolor{gray!25}$\bm{0.93\pm0.00}$ & $0.85\pm0.02$ & $0.82\pm0.02$ & $0.83\pm0.01$ \\
LGR+D  & $0.91\pm0.00$ & [$0.87\pm0.01$] & $0.85\pm0.01$ & [$0.89\pm0.00$] 
     & [$0.92\pm0.00$] & [$0.87\pm0.00$] & [$0.85\pm0.00$] & $0.89\pm0.00$ \\

\bottomrule

\end{tabular}

}
\end{table}

\begin{table}[h!]
\centering
\caption[\acs{Task-IL} \acs{CF} scores for \acs{RAF-DB}  following O2 and O3.]{\acs{Task-IL} \acs{CF} scores for \acs{RAF-DB} W/ Data-Augmentation following Orderings O2 and O3. \textbf{Bold} values denote best (lowest) while [\textit{bracketed}] denote second-best values for each column.}

\label{tab:task-il-rafdb-orders-cf}

{
\scriptsize

\begin{tabular}{l|ccc|ccc}\toprule
\multicolumn{1}{c|}{\textbf{Method}}            & \multicolumn{3}{c|}{\textbf{CF following O2}} 
& \multicolumn{3}{c}{\textbf{CF following O3}} \\ \cmidrule{2-7}

\multicolumn{1}{c|}{ }                          & 
\multicolumn{1}{c|}{\textbf{Task 2}}          &
\multicolumn{1}{c|}{\textbf{Task 3}}            & \multicolumn{1}{c|}{\textbf{Task 4}}           &
\multicolumn{1}{c|}{\textbf{Task 2}}          &
\multicolumn{1}{c|}{\textbf{Task 3}}            & \multicolumn{1}{c}{\textbf{Task 4}}          \\ \midrule

\multicolumn{7}{c}{\textbf{Baseline Approaches}} \\ \midrule
LB 
& $0.25\pm0.13$ &  $0.37\pm0.09$ &  $0.20\pm0.03$ 
&  $0.24\pm0.03$ &  $0.46\pm0.08$ & $0.17\pm0.01$ \\
UB 
& \cellcolor{gray!25}$\bm{-0.01\pm0.02}$ & \cellcolor{gray!25}$\bm{-0.01\pm0.01}$ & \cellcolor{gray!25}$\bm{0.00\pm0.00}$ 
& \cellcolor{gray!25}$\bm{-0.01\pm0.01}$ & \cellcolor{gray!25}$\bm{-0.02\pm0.01}$ & \cellcolor{gray!25}$\bm{0.00\pm0.02}$ \\\midrule

\multicolumn{7}{c}{\textbf{Regularisation-Based Approaches}} \\ \midrule

EWC  &  $0.05\pm0.04$ &  $0.10\pm0.04$ &  $0.10\pm0.07$ &  $0.05\pm0.02$ &  $0.16\pm0.03$ & $0.05\pm0.02$ \\
EWC-Online  &  $0.05\pm0.02$ &  $0.15\pm0.03$ &  $0.09\pm0.05$ &  $0.05\pm0.00$ &  $0.06\pm0.02$ & $0.05\pm0.02$ \\
SI  &  $0.27\pm0.15$ &  $0.42\pm0.09$ &  $0.19\pm0.03$ &  $0.23\pm0.03$ &  $0.38\pm0.03$ & $0.18\pm0.02$ \\
LwF  &  $0.01\pm0.01$ &  $0.01\pm0.01$ &  \cellcolor{gray!25}$\bm{0.00\pm0.00}$ &  [$0.00\pm0.00$] &  [$0.00\pm0.00$] & \cellcolor{gray!25}$\bm{0.00\pm0.00}$ \\ \midrule

\multicolumn{7}{c}{\textbf{Replay-Based Approaches}} \\ \midrule

NR  &  $0.09\pm0.01$ &  $0.11\pm0.01$ &  \cellcolor{gray!25}$\bm{0.00\pm0.01}$ &  $0.08\pm0.01$ &  $0.15\pm0.02$ & \cellcolor{gray!25}$\bm{0.00\pm0.01}$ \\
A-GEM  &  [$0.00\pm0.01$] &  $0.21\pm0.07$ &  \cellcolor{gray!25}$\bm{0.00\pm0.01}$ &  $0.01\pm0.01$ &  $0.08\pm0.00$ & \cellcolor{gray!25}$\bm{0.00\pm0.00}$ \\
LR  &  $0.07\pm0.01$ &  $0.08\pm0.02$ &  \cellcolor{gray!25}$\bm{0.00\pm0.00}$ &  $0.08\pm0.01$ &  $0.09\pm0.02$ & \cellcolor{gray!25}$\bm{0.00\pm0.00}$ \\
DGR  &  $0.27\pm0.12$ &  $0.25\pm0.06$ &  $0.17\pm0.12$ &  $0.18\pm0.03$ &  $0.24\pm0.07$ & $0.10\pm0.03$ \\
LGR  &  $0.02\pm0.01$ &  $0.02\pm0.00$ &  [$0.01\pm0.01$] &  [$0.00\pm0.00$] &  $0.01\pm0.00$ & [$0.01\pm0.00$] \\
DGR+D  &  $0.20\pm0.02$ &  $0.21\pm0.01$ &  $0.08\pm0.02$ &  $0.16\pm0.06$ &  $0.14\pm0.02$ & $0.05\pm0.04$ \\
LGR+D  &  $0.01\pm0.00$ &  [$0.00\pm0.00$] &  \cellcolor{gray!25}$\bm{0.00\pm0.00}$ &  [$0.00\pm0.00$] &  [$0.00\pm0.00$] & \cellcolor{gray!25}$\bm{0.00\pm0.00}$ \\

\bottomrule
\end{tabular}
}
\end{table}

\begin{figure}  
    \centering
    \subfloat[\ac{Task-IL} Results following O$2$.\label{fig:rafdb-task-il-aug-order-1}]{\includegraphics[width=0.5\textwidth]{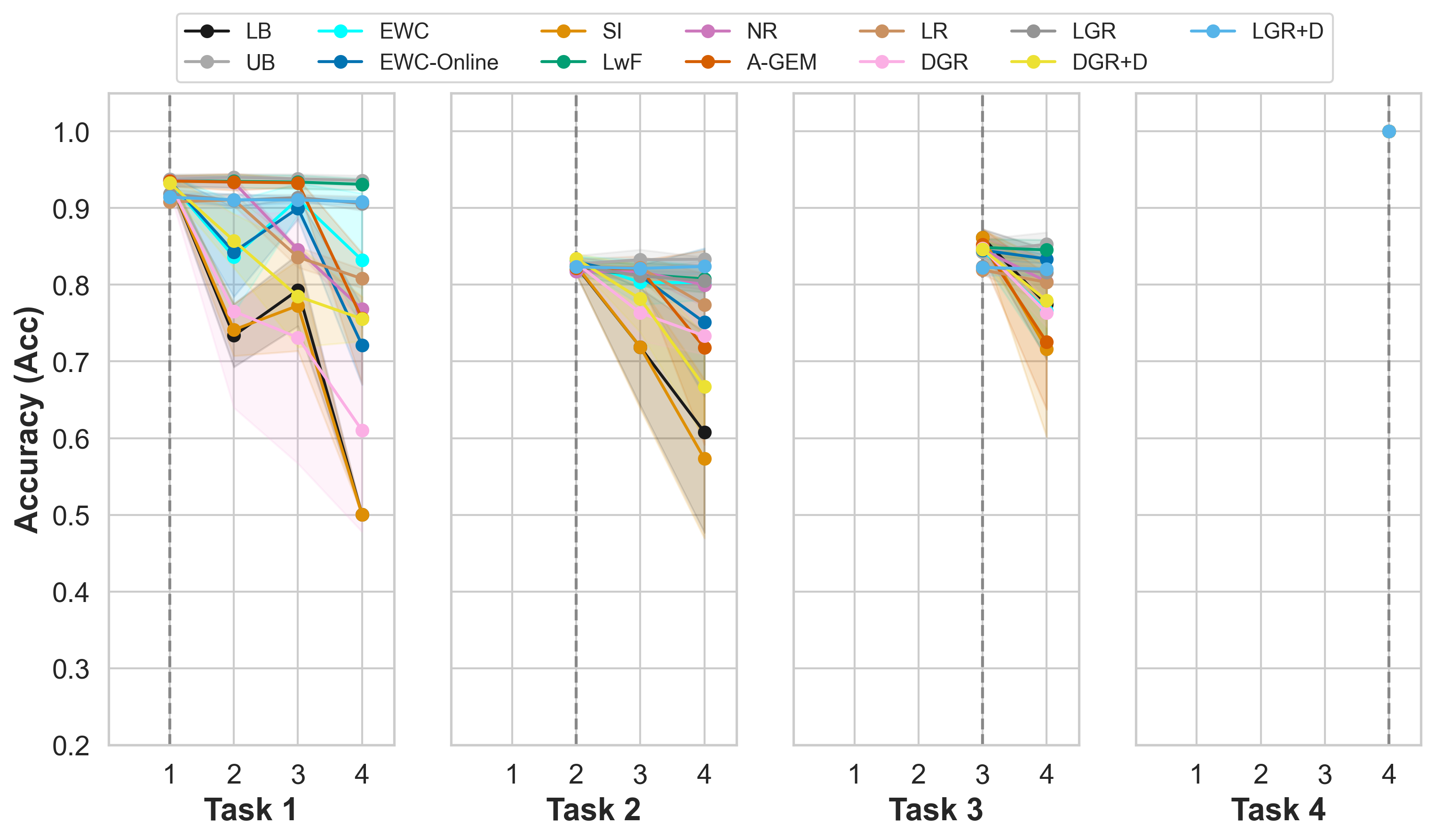}}     \hfill
    \subfloat[\ac{Task-IL} Results following O$3$.\label{fig:rafdb-task-il-aug-order-2}]{\includegraphics[width=0.5\textwidth]{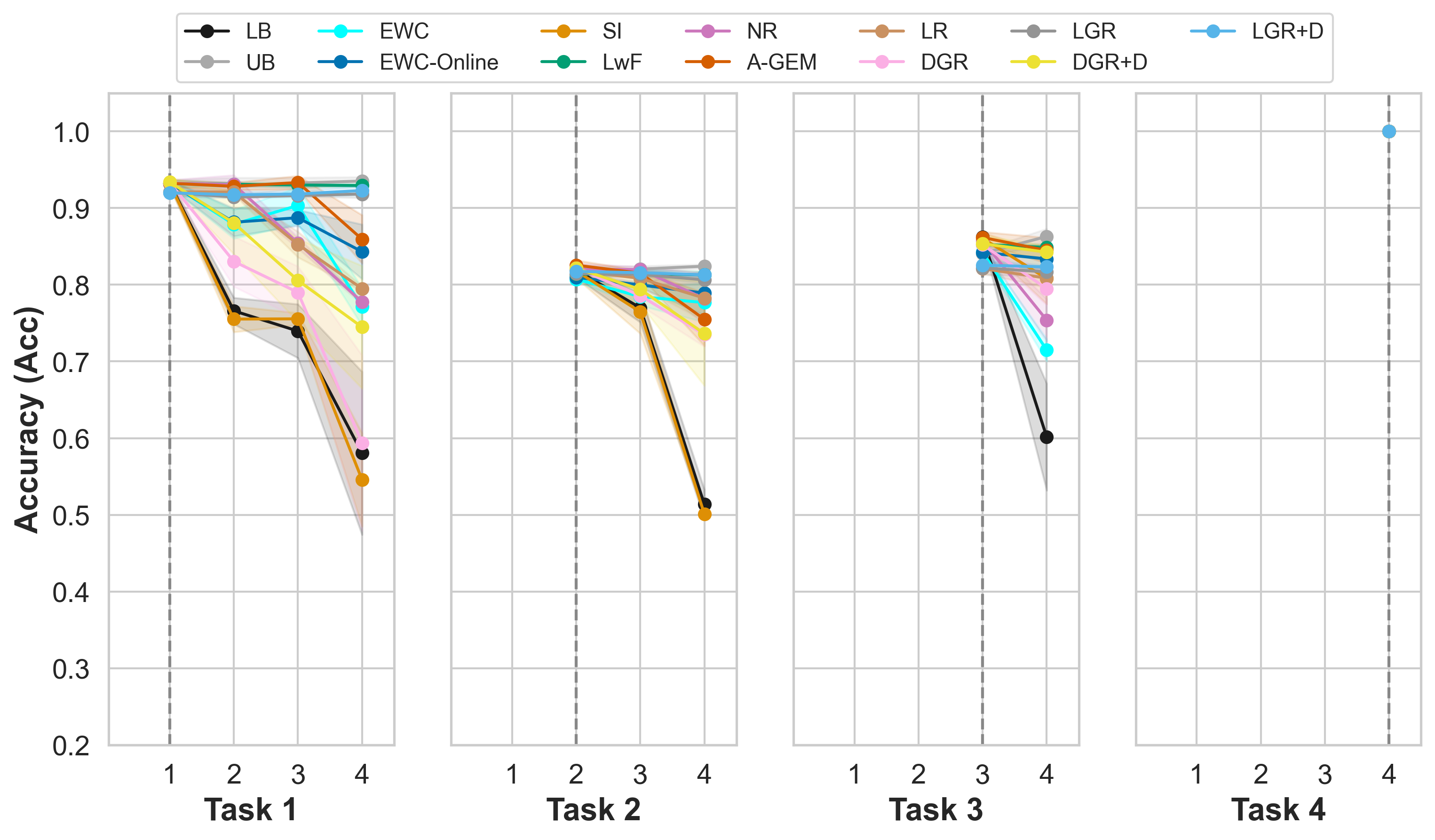}} \\
    \caption[Task-IL Results for \acs{RAF-DB}  following Order 2 and Order 3.]{\ac{Task-IL} results for \acs{RAF-DB} following (a) Order $2$ and (b) Order $3$, respectively. For each task, test-set accuracy is shown as learning progresses from when a task is introduced to the end of the overall training procedure.}
    \label{fig:rafdb-taskil-orders} 
\end{figure}

\normalsize
\textit{}
\newpage
\subsubsection{\acf{Class-IL} Results}
\label{app:classorderrafdb}
\begin{table}[h!]
\centering
\setlength\tabcolsep{3.5pt}

\caption[\acs{Class-IL} \acs{Acc} for \acs{RAF-DB} following Orderings O2 and O3.]{\acs{Class-IL} \acs{Acc} for \acs{RAF-DB} W/ Data-Augmentation following Orderings O2 and O3. \textbf{Bold} values denote best (highest) while [\textit{bracketed}] denote second-best values for each column.}

\label{tab:class-il-rafdb-acc-orders}

{
\scriptsize
\begin{tabular}{l|ccccccc}\toprule

\multirow{2}[2]{*}{\makecell[c]{\textbf{Method}}}           & \multicolumn{7}{c}{\textbf{Accuracy following O2}} \\ \cmidrule{2-8}

\multicolumn{1}{c|}{ }                          & 
\multicolumn{1}{c|}{\textbf{Class 1}}            & \multicolumn{1}{c|}{\textbf{Class 2}}          &
\multicolumn{1}{c|}{\textbf{Class 3}}            & \multicolumn{1}{c|}{\textbf{Class 4}}           &
\multicolumn{1}{c|}{\textbf{Class 5}}            & \multicolumn{1}{c|}{\textbf{Class 6}}          &
\multicolumn{1}{c}{\textbf{Class 7}}                      \\ \midrule

\multicolumn{8}{c}{\textbf{Baseline Approaches}} \\ \midrule
LB & \cellcolor{gray!25}$\bm{1.00\pm0.00}$ & $0.50\pm0.00$ & $0.33\pm0.00$ & $0.25\pm0.00$ & $0.20\pm0.00$ & $0.17\pm0.00$ & $0.14\pm0.00$ \\
UB & \cellcolor{gray!25}$\bm{1.00\pm0.00}$ & \cellcolor{gray!25}$\bm{0.93\pm0.01}$ & \cellcolor{gray!25}$\bm{0.77\pm0.00}$ & \cellcolor{gray!25}$\bm{0.73\pm0.01}$ & \cellcolor{gray!25}$\bm{0.72\pm0.01}$ & \cellcolor{gray!25}$\bm{0.65\pm0.00}$ & \cellcolor{gray!25}$\bm{0.64\pm0.01}$ \\\midrule

\multicolumn{8}{c}{\textbf{Regularisation-Based Approaches}} \\ \midrule

EWC  & \cellcolor{gray!25}$\bm{1.00\pm0.00}$ & $0.50\pm0.00$ & $0.33\pm0.00$ & $0.25\pm0.00$ & $0.20\pm0.00$ & $0.17\pm0.00$ & $0.14\pm0.00$ \\
EWC-Online  & \cellcolor{gray!25}$\bm{1.00\pm0.00}$ & $0.50\pm0.00$ & $0.33\pm0.00$ & $0.25\pm0.00$ & $0.20\pm0.00$ & $0.17\pm0.00$ & $0.14\pm0.00$ \\
SI  & \cellcolor{gray!25}$\bm{1.00\pm0.00}$ & $0.50\pm0.00$ & $0.33\pm0.00$ & $0.25\pm0.00$ & $0.20\pm0.00$ & $0.17\pm0.00$ & $0.14\pm0.00$ \\
LwF  & \cellcolor{gray!25}$\bm{1.00\pm0.00}$ & $0.50\pm0.00$ & $0.33\pm0.00$ & $0.25\pm0.00$ & $0.20\pm0.00$ & $0.17\pm0.00$ & $0.14\pm0.00$ \\\midrule
                 
\multicolumn{8}{c}{\textbf{Replay-Based Approaches}} \\ \midrule

NR  & \cellcolor{gray!25}$\bm{1.00\pm0.00}$ & \cellcolor{gray!25}$\bm{0.93\pm0.01}$ & [$0.72\pm0.02$] & [$0.66\pm0.01$] & $0.64\pm0.00$ & $0.53\pm0.01$ & $0.48\pm0.02$ \\
A-GEM  & \cellcolor{gray!25}$\bm{1.00\pm0.00}$ & $0.50\pm0.00$ & $0.33\pm0.00$ & $0.32\pm0.09$ & $0.20\pm0.00$ & $0.17\pm0.00$ & $0.14\pm0.00$ \\
LR  & \cellcolor{gray!25}$\bm{1.00\pm0.00}$ & [$0.92\pm0.01$] & $0.71\pm0.01$ & [$0.66\pm0.01$] & [$0.66\pm0.01$] & [$0.58\pm0.01$] & [$0.56\pm0.01$] \\
DGR  & \cellcolor{gray!25}$\bm{1.00\pm0.00}$ & $0.50\pm0.00$ & $0.33\pm0.00$ & $0.26\pm0.00$ & $0.20\pm0.00$ & $0.17\pm0.00$ & $0.14\pm0.00$ \\
LGR  & \cellcolor{gray!25}$\bm{1.00\pm0.00}$ & $0.83\pm0.02$ & $0.58\pm0.02$ & $0.61\pm0.01$ & $0.58\pm0.01$ & $0.48\pm0.02$ & $0.47\pm0.01$ \\
DGR+D  & \cellcolor{gray!25}$\bm{1.00\pm0.00}$ & $0.50\pm0.00$ & $0.33\pm0.00$ & $0.26\pm0.00$ & $0.20\pm0.00$ & $0.17\pm0.00$ & $0.14\pm0.00$ \\
LGR+D  & \cellcolor{gray!25}$\bm{1.00\pm0.00}$ & $0.89\pm0.00$ & $0.65\pm0.02$ & $0.62\pm0.01$ & $0.61\pm0.01$ & $0.51\pm0.00$ & $0.49\pm0.02$ \\\midrule

\multirow{2}[2]{*}{\makecell[c]{\textbf{Method}}}           & \multicolumn{7}{c}{\textbf{Accuracy following O3}} \\ \cmidrule{2-8}

\multicolumn{1}{c|}{ }                          & 
\multicolumn{1}{c|}{\textbf{Class 1}}            & \multicolumn{1}{c|}{\textbf{Class 2}}          &
\multicolumn{1}{c|}{\textbf{Class 3}}            & \multicolumn{1}{c|}{\textbf{Class 4}}           &
\multicolumn{1}{c|}{\textbf{Class 5}}            & \multicolumn{1}{c|}{\textbf{Class 6}}          &
\multicolumn{1}{c}{\textbf{Class 7}}                    \\ \midrule
\multicolumn{8}{c}{\textbf{Baseline Approaches}} \\ \midrule

LB & \cellcolor{gray!25}$\bm{1.00\pm0.00}$ & $0.50\pm0.00$ & $0.33\pm0.00$ & $0.25\pm0.00$ & $0.20\pm0.00$ & $0.17\pm0.00$ & $0.14\pm0.00$ \\
UB & \cellcolor{gray!25}$\bm{1.00\pm0.00}$ & \cellcolor{gray!25}$\bm{0.93\pm0.00}$ & \cellcolor{gray!25}$\bm{0.76\pm0.01}$ & \cellcolor{gray!25}$\bm{0.73\pm0.01}$ & \cellcolor{gray!25}$\bm{0.72\pm0.01}$ & \cellcolor{gray!25}$\bm{0.64\pm0.01}$ & \cellcolor{gray!25}$\bm{0.63\pm0.01}$ \\\midrule

\multicolumn{8}{c}{\textbf{Regularisation-Based Approaches}} \\ \midrule

EWC  & \cellcolor{gray!25}$\bm{1.00\pm0.00}$ & $0.50\pm0.00$ & $0.33\pm0.00$ & $0.25\pm0.00$ & $0.20\pm0.00$ & $0.17\pm0.00$ & $0.14\pm0.00$ \\
EWC-Online  & \cellcolor{gray!25}$\bm{1.00\pm0.00}$ & $0.50\pm0.00$ & $0.33\pm0.00$ & $0.25\pm0.00$ & $0.20\pm0.00$ & $0.17\pm0.00$ & $0.14\pm0.00$ \\
SI  & \cellcolor{gray!25}$\bm{1.00\pm0.00}$ & $0.50\pm0.00$ & $0.33\pm0.00$ & $0.25\pm0.00$ & $0.20\pm0.00$ & $0.17\pm0.00$ & $0.14\pm0.00$ \\
LwF  & \cellcolor{gray!25}$\bm{1.00\pm0.00}$ & $0.50\pm0.00$ & $0.33\pm0.00$ & $0.25\pm0.00$ & $0.20\pm0.00$ & $0.17\pm0.00$ & $0.14\pm0.00$ \\\midrule

\multicolumn{8}{c}{\textbf{Replay-Based Approaches}} \\ \midrule

NR  & \cellcolor{gray!25}$\bm{1.00\pm0.00}$ & [$0.92\pm0.00$] & $0.70\pm0.01$ & $0.64\pm0.02$ & $0.63\pm0.02$ & $0.52\pm0.03$ & $0.48\pm0.03$ \\
A-GEM  & \cellcolor{gray!25}$\bm{1.00\pm0.00}$ & $0.50\pm0.00$ & $0.33\pm0.00$ & $0.25\pm0.00$ & $0.20\pm0.00$ & $0.17\pm0.00$ & $0.14\pm0.00$ \\
LR  & \cellcolor{gray!25}$\bm{1.00\pm0.00}$ & $0.91\pm0.00$ & [$0.71\pm0.01$] & [$0.66\pm0.01$] & [$0.66\pm0.01$] & [$0.58\pm0.01$] & [$0.56\pm0.02$] \\
DGR  & \cellcolor{gray!25}$\bm{1.00\pm0.00}$ & $0.50\pm0.01$ & $0.33\pm0.00$ & $0.25\pm0.00$ & $0.20\pm0.00$ & $0.17\pm0.00$ & $0.14\pm0.00$ \\
LGR  & \cellcolor{gray!25}$\bm{1.00\pm0.00}$ & $0.81\pm0.01$ & $0.58\pm0.02$ & $0.60\pm0.01$ & $0.58\pm0.02$ & $0.47\pm0.02$ & $0.44\pm0.02$ \\
DGR+D  & \cellcolor{gray!25}$\bm{1.00\pm0.00}$ & $0.50\pm0.01$ & $0.34\pm0.00$ & $0.26\pm0.00$ & $0.20\pm0.00$ & $0.17\pm0.00$ & $0.14\pm0.00$ \\
LGR+D  & \cellcolor{gray!25}$\bm{1.00\pm0.00}$ & $0.87\pm0.01$ & $0.64\pm0.01$ & $0.62\pm0.01$ & $0.61\pm0.02$ & $0.51\pm0.03$ & $0.48\pm0.02$ \\

\bottomrule

\end{tabular}

}
\end{table}

\begin{table}[h!]
\centering
\caption[\acs{Class-IL} \acs{CF} scores for \acs{RAF-DB}  following O2 and O3.]{\acs{Class-IL} \acs{CF} scores for \acs{RAF-DB} W/ Data-Augmentation following Orderings O2 and O3. \textbf{Bold} values denote best (lowest) while [\textit{bracketed}] denote second-best values for each column.}

\label{tab:class-il-rafdb-cf-orders}

{
\scriptsize

\begin{tabular}{l|cccccc}\toprule

\multirow{2}[2]{*}{\makecell[c]{\textbf{Method}}}           & \multicolumn{6}{c}{\textbf{\acs{CF} following O2}} \\ \cmidrule{2-7}

\multicolumn{1}{c|}{ }                          & 
\multicolumn{1}{c|}{\textbf{Class 2}}          &
\multicolumn{1}{c|}{\textbf{Class 3}}            & \multicolumn{1}{c|}{\textbf{Class 4}}           &
\multicolumn{1}{c|}{\textbf{Class 5}}            & \multicolumn{1}{c|}{\textbf{Class 6}}          &
\multicolumn{1}{c}{\textbf{Class 7}}                    \\ \midrule

\multicolumn{7}{c}{\textbf{Baseline Approaches}} \\ \midrule

LB & $1.00\pm0.00$ & $1.00\pm0.00$ & $1.00\pm0.00$ & $1.00\pm0.00$ & $1.00\pm0.00$ & $1.00\pm0.00$ \\
UB & \cellcolor{gray!25}$\bm{0.13\pm0.04}$ & \cellcolor{gray!25}$\bm{0.12\pm0.02}$ & \cellcolor{gray!25}$\bm{0.12\pm0.01}$ & \cellcolor{gray!25}$\bm{0.13\pm0.02}$ & \cellcolor{gray!25}$\bm{0.12\pm0.01}$ & \cellcolor{gray!25}$\bm{0.10\pm0.00}$ \\\midrule

\multicolumn{7}{c}{\textbf{Regularisation-Based Approaches}} \\ \midrule

EWC  & $1.00\pm0.00$ & $1.00\pm0.00$ & $1.00\pm0.00$ & $1.00\pm0.00$ & $1.00\pm0.00$ & $1.00\pm0.00$ \\
EWC-Online  & $1.00\pm0.00$ & $1.00\pm0.00$ & $1.00\pm0.00$ & $1.00\pm0.00$ & $1.00\pm0.00$ & $1.00\pm0.00$ \\
SI  & $1.00\pm0.00$ & $1.00\pm0.00$ & $1.00\pm0.00$ & $1.00\pm0.00$ & $1.00\pm0.00$ & $1.00\pm0.00$ \\
LwF  & $1.00\pm0.00$ & $1.00\pm0.00$ & $1.00\pm0.00$ & $1.00\pm0.00$ & $1.00\pm0.00$ & $1.00\pm0.00$ \\ \midrule

\multicolumn{7}{c}{\textbf{Replay-Based Approaches}} \\ \midrule

NR  & $0.27\pm0.02$ & $0.31\pm0.02$ & $0.32\pm0.01$ & $0.43\pm0.02$ & $0.48\pm0.03$ & $0.49\pm0.00$ \\
A-GEM  & $1.00\pm0.00$ & $0.90\pm0.13$ & $1.00\pm0.01$ & $0.99\pm0.01$ & $1.00\pm0.00$ & $1.00\pm0.00$ \\
LR  & [$0.17\pm0.01$] & [$0.18\pm0.01$] & [$0.16\pm0.01$] & [$0.18\pm0.02$] & [$0.17\pm0.01$] & [$0.19\pm0.00$] \\
DGR  & $0.98\pm0.00$ & $0.98\pm0.01$ & $0.99\pm0.00$ & $0.99\pm0.00$ & $0.99\pm0.00$ & $1.00\pm0.00$ \\
LGR  & $0.49\pm0.05$ & $0.37\pm0.02$ & $0.39\pm0.02$ & $0.45\pm0.01$ & $0.44\pm0.01$ & $0.35\pm0.00$ \\
DGR+D  & $0.98\pm0.00$ & $0.98\pm0.01$ & $0.99\pm0.01$ & $0.99\pm0.00$ & $0.99\pm0.01$ & $1.00\pm0.00$ \\
LGR+D  & $0.30\pm0.04$ & $0.29\pm0.03$ & $0.28\pm0.02$ & $0.34\pm0.01$ & $0.36\pm0.03$ & $0.29\pm0.00$ \\\midrule

\multirow{2}[2]{*}{\makecell[c]{\textbf{Method}}}           & \multicolumn{6}{c}{\textbf{\acs{CF} following O3}} \\ \cmidrule{2-7}

\multicolumn{1}{c|}{ }                          & 
\multicolumn{1}{c|}{\textbf{Class 2}}          &
\multicolumn{1}{c|}{\textbf{Class 3}}            & \multicolumn{1}{c|}{\textbf{Class 4}}           &
\multicolumn{1}{c|}{\textbf{Class 5}}            & \multicolumn{1}{c|}{\textbf{Class 6}}          &
\multicolumn{1}{c}{\textbf{Class 7}}                     \\ \midrule

\multicolumn{7}{c}{\textbf{Baseline Approaches}} \\ \midrule

LB & $1.00\pm0.00$ & $1.00\pm0.00$ & $1.00\pm0.00$ & $1.00\pm0.00$ & $1.00\pm0.00$ & $1.00\pm0.00$ \\
UB & \cellcolor{gray!25}$\bm{0.15\pm0.01}$ & \cellcolor{gray!25}$\bm{0.13\pm0.01}$ & \cellcolor{gray!25}$\bm{0.12\pm0.00}$ & \cellcolor{gray!25}$\bm{0.14\pm0.01}$ & \cellcolor{gray!25}$\bm{0.13\pm0.01}$ & \cellcolor{gray!25}$\bm{0.10\pm0.00}$ \\\midrule

\multicolumn{7}{c}{\textbf{Regularisation-Based Approaches}} \\ \midrule

EWC  & $1.00\pm0.00$ & $1.00\pm0.00$ & $1.00\pm0.00$ & $1.00\pm0.00$ & $1.00\pm0.00$ & $1.00\pm0.00$ \\
EWC-Online  & $1.00\pm0.00$ & $1.00\pm0.00$ & $1.00\pm0.00$ & $1.00\pm0.00$ & $1.00\pm0.00$ & $1.00\pm0.00$ \\
SI  & $1.00\pm0.00$ & $1.00\pm0.00$ & $1.00\pm0.00$ & $1.00\pm0.00$ & $1.00\pm0.00$ & $1.00\pm0.00$ \\
LwF  & $1.00\pm0.00$ & $1.00\pm0.00$ & $1.00\pm0.00$ & $1.00\pm0.00$ & $1.00\pm0.00$ & $1.00\pm0.00$ \\ \midrule

\multicolumn{7}{c}{\textbf{Replay-Based Approaches}} \\ \midrule

NR  & $0.33\pm0.02$ & $0.35\pm0.01$ & $0.36\pm0.03$ & $0.46\pm0.04$ & $0.49\pm0.04$ & $0.48\pm0.00$ \\
A-GEM  & $1.00\pm0.00$ & $1.00\pm0.00$ & $1.00\pm0.00$ & $1.00\pm0.00$ & $1.00\pm0.00$ & $1.00\pm0.00$ \\
LR  & [$0.18\pm0.00$] & [$0.17\pm0.01$] & [$0.16\pm0.01$] & [$0.17\pm0.01$] & [$0.17\pm0.01$] & [$0.19\pm0.00$] \\
DGR  & $0.99\pm0.00$ & $0.99\pm0.00$ & $0.99\pm0.00$ & $0.99\pm0.00$ & $0.99\pm0.00$ & $1.00\pm0.00$ \\
LGR  & $0.50\pm0.03$ & $0.39\pm0.02$ & $0.39\pm0.03$ & $0.47\pm0.04$ & $0.50\pm0.04$ & $0.45\pm0.00$ \\
DGR+D  & $0.99\pm0.00$ & $0.98\pm0.00$ & $0.99\pm0.00$ & $0.99\pm0.00$ & $0.99\pm0.00$ & $1.00\pm0.00$ \\
LGR+D  & $0.33\pm0.03$ & $0.29\pm0.02$ & $0.30\pm0.03$ & $0.37\pm0.04$ & $0.38\pm0.02$ & $0.31\pm0.00$ \\

\bottomrule
\end{tabular}
}
\end{table}
\begin{figure}[t!]
    \centering
    \subfloat[\acs{Class-IL} Results following O$2$.\label{fig:rafdb-class-il-aug-order-1}]{\includegraphics[width=0.5\textwidth]{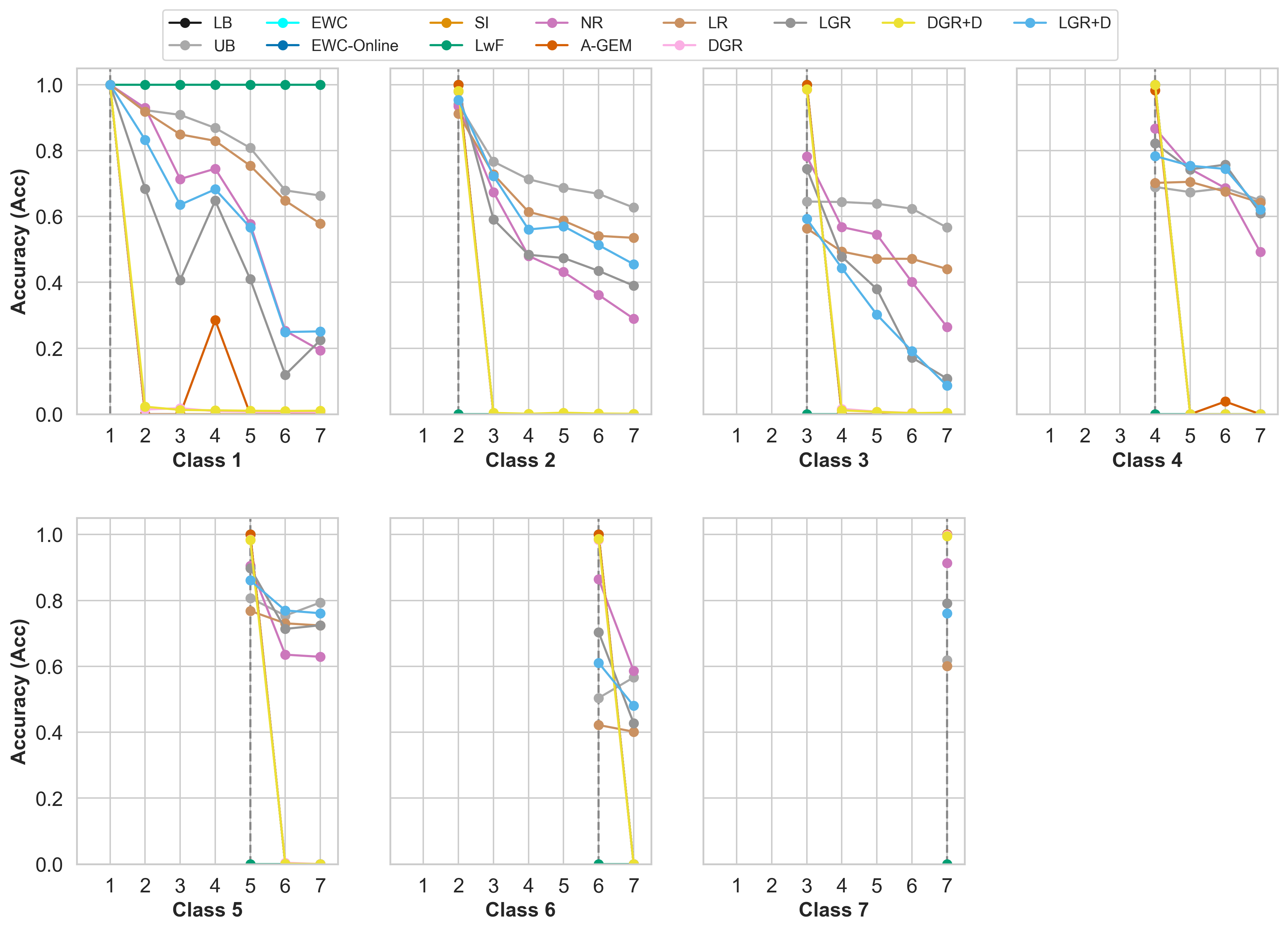}} \hfill
    \subfloat[\acs{Class-IL} Results following O$3$.\label{fig:rafdb-class-aug-order-2}]{\includegraphics[width=0.5\textwidth]{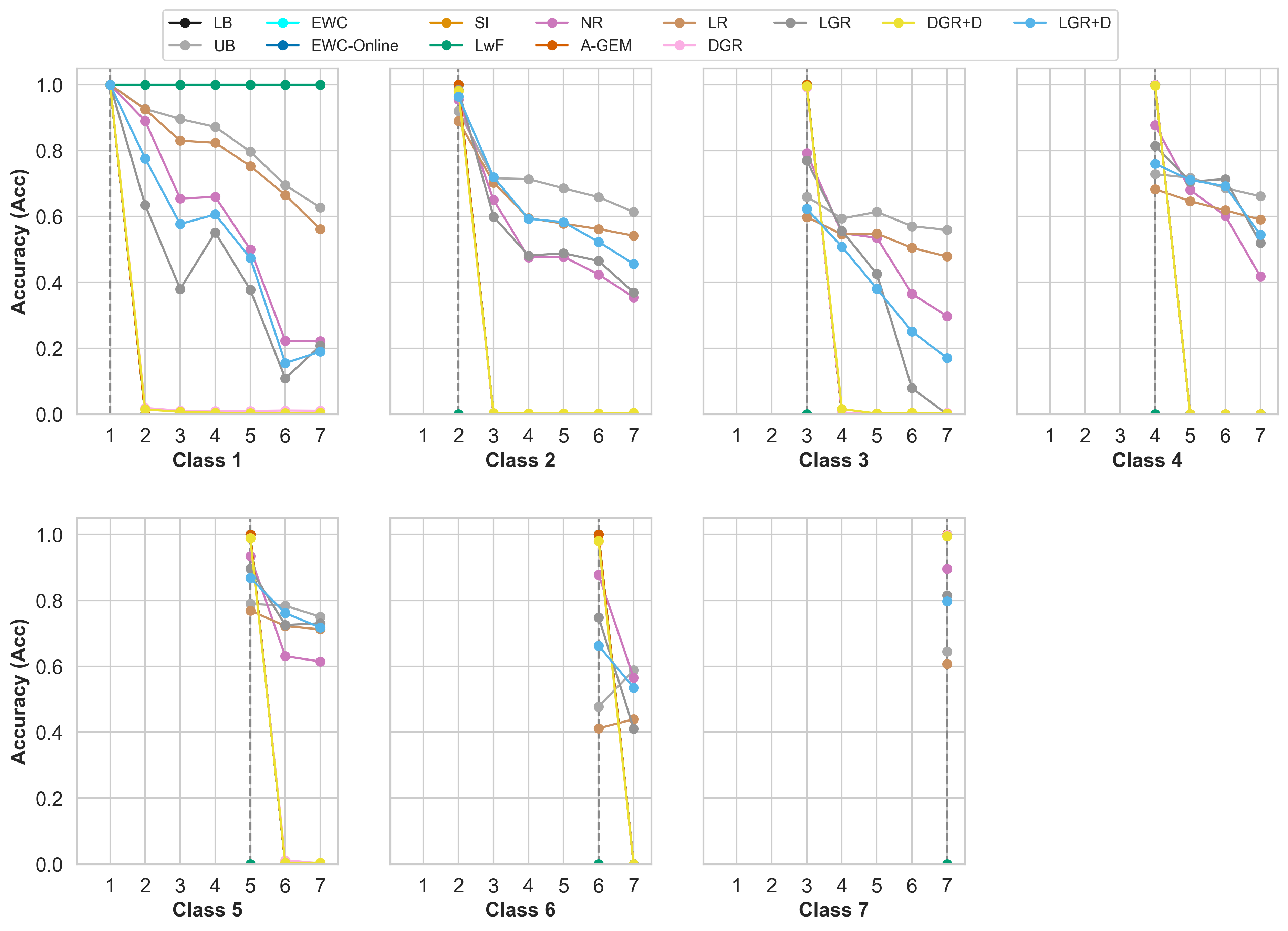}} \\
    \caption[Class-IL Results for \acs{RAF-DB}  following Order 2 and Order 3.]{\acs{Class-IL} results for \acs{RAF-DB} following (a) Order $2$ and (b) Order $3$, respectively. For each class, test-set accuracy is shown as the learning progresses from when the class is introduced to the end of the overall training procedure.}
    \label{fig:RafDB-classil-orders} 
\end{figure}
\FloatBarrier
\normalsize





\subsection{AffectNet Results}

\subsubsection{\acf{Task-IL} Results}
\label{app:taskorderaffnet}
\begin{table}[h!]
\centering
\setlength\tabcolsep{2.0pt}

\caption[\acs{Task-IL} \acs{Acc} for AffectNet following Orderings O2 and O3.]{\acs{Task-IL} \acs{Acc} for AffectNet using the Downsampled split following Orderings O2 and O3. \textbf{Bold} values denote best (highest) while [\textit{bracketed}] denote second-best values for each column.}

\label{tab:task-il-AffectNet-orders-acc}

{
\scriptsize
\begin{tabular}{l|cccc|cccc}\toprule
\multicolumn{1}{c|}{\textbf{Method}}            & \multicolumn{4}{c|}{\textbf{Accuracy following O2}} 
& \multicolumn{4}{c}{\textbf{Accuracy following O3}} \\  \cmidrule{2-9}

\multicolumn{1}{c|}{ }                          & 
\multicolumn{1}{c|}{\textbf{Task 1}}            & \multicolumn{1}{c|}{\textbf{Task 2}}          &
\multicolumn{1}{c|}{\textbf{Task 3}}            & \multicolumn{1}{c|}{\textbf{Task 4}}           &
\multicolumn{1}{c|}{\textbf{Task 1}}            & \multicolumn{1}{c|}{\textbf{Task 2}}          &
\multicolumn{1}{c|}{\textbf{Task 3}}            & \multicolumn{1}{c}{\textbf{Task 4}}          \\ \midrule

\multicolumn{9}{c}{\textbf{Baseline Approaches}} \\ \midrule
LB                & $0.67\pm0.00$ & $0.68\pm0.01$ & $0.65\pm0.01$ & $0.60\pm0.01$ 
        & [$0.68\pm0.01$] & $0.68\pm0.01$ & $0.64\pm0.01$ & $0.61\pm0.01$ \\
UB                & [$0.69\pm0.00$] & $0.70\pm0.00$ & [$0.75\pm0.00$] & [$0.74\pm0.00$]
                & \cellcolor{gray!25}$\bm{0.69\pm0.01}$ & [$0.69\pm0.01$] & [$0.75\pm0.00$] & $0.73\pm0.01$ \\
 \midrule

\multicolumn{9}{c}{\textbf{Regularisation-Based Approaches}} \\ \midrule

 EWC   & $0.68\pm0.00$ & $0.69\pm0.02$ & $0.72\pm0.01$ & $0.70\pm0.00$ 
                                        & [$0.68\pm0.01$] & $0.68\pm0.02$ & $0.72\pm0.01$ & $0.69\pm0.00$ \\
EWC-Online    & [$0.69\pm0.00$] & $0.70\pm0.01$ & $0.72\pm0.01$ & $0.69\pm0.00$ 
                                        & [$0.68\pm0.00$] & $0.68\pm0.02$ & $0.72\pm0.01$ & $0.69\pm0.01$ \\
         SI    & $0.68\pm0.01$ & $0.68\pm0.02$ & $0.70\pm0.02$ & $0.69\pm0.01$ 
                                        & $0.67\pm0.02$ & $0.68\pm0.01$ & $0.70\pm0.01$ & $0.69\pm0.00$ \\
                 LwF    & $0.68\pm0.00$ & [$0.71\pm0.00$] & [$0.75\pm0.01$] & [$0.74\pm0.00$] 
                                        & [$0.68\pm0.01$] & [$0.69\pm0.01]$ & $0.74\pm0.01$ & [$0.74\pm0.01$] \\ \midrule
\multicolumn{9}{c}{\textbf{Replay-Based Approaches}} \\ \midrule

               NR    & [$0.69\pm0.00$] & $0.69\pm0.00$ & $0.73\pm0.01$ & $0.69\pm0.01$ 
                                        & [$0.68\pm0.01$] & [$0.69\pm0.01$] & $0.72\pm0.00$ & $0.69\pm0.00$ \\
   A-GEM    & $0.68\pm0.00$ & \cellcolor{gray!25}$\bm{0.72\pm0.00}$ & \cellcolor{gray!25}$\bm{0.76\pm0.00}$ & \cellcolor{gray!25}$\bm{0.74\pm0.00}$ 
                                        & [$0.68\pm0.01$] & \cellcolor{gray!25}$\bm{0.71\pm0.01}$ & \cellcolor{gray!25}$\bm{0.76\pm0.00}$ & \cellcolor{gray!25}$\bm{0.75\pm0.01}$ \\
       LR    & $0.65\pm0.01$ & $0.65\pm0.01$ & $0.68\pm0.01$ & $0.66\pm0.02$ 
                                        & $0.66\pm0.01$ & $0.66\pm0.00$ & $0.69\pm0.01$ & $0.66\pm0.00$ \\
             DGR    & [$0.69\pm0.00$] & $0.63\pm0.01$ & $0.66\pm0.02$ & $0.63\pm0.02$ 
                                        & \cellcolor{gray!25}$\bm{0.69\pm0.00}$ & $0.64\pm0.03$ & $0.65\pm0.02$ & $0.63\pm0.02$ \\
           LGR    & $0.66\pm0.01$ & $0.63\pm0.02$ & $0.67\pm0.01$ & $0.65\pm0.01$ 
                                        & $0.66\pm0.00$ & $0.65\pm0.01$ & $0.68\pm0.00$ & $0.66\pm0.00$ \\
          DGR+D    & \cellcolor{gray!25}$\bm{0.70\pm0.00}$ & $0.69\pm0.01$ & $0.70\pm0.02$ & $0.71\pm0.02$ 
                                        & [$0.68\pm0.00$] & $0.68\pm0.01$ & $0.69\pm0.03$ & $0.71\pm0.00$ \\
         LGR+D    & $0.66\pm0.01$ & $0.65\pm0.02$ & $0.69\pm0.02$ & $0.68\pm0.01$ 
                                        & $0.66\pm0.00$ & $0.66\pm0.00$ & $0.69\pm0.00$ & $0.68\pm0.00$ \\

\bottomrule

\end{tabular}

}
\end{table}

\begin{table}[h!]
\centering
\caption[\acs{Task-IL} \acs{CF} scores for AffectNet following O2 and O3.]{\acs{Task-IL} \acs{CF} scores for AffectNet using the Downsampled split following Orderings O2 and O3. \textbf{Bold} values denote best (lowest) while [\textit{bracketed}] denote second-best values for each column.}

\label{tab:task-il-AffectNet-orders-cf}

{
\scriptsize
\setlength\tabcolsep{4.5pt}

\begin{tabular}{l|ccc|ccc}\toprule
\multicolumn{1}{c|}{\textbf{Method}}            & \multicolumn{3}{c|}{\textbf{CF following O2}} 
& \multicolumn{3}{c}{\textbf{CF following O3}} \\ \cmidrule{2-7}
\multicolumn{1}{c|}{ }                          & 
\multicolumn{1}{c|}{\textbf{Task 2}}          &
\multicolumn{1}{c|}{\textbf{Task 3}}            & \multicolumn{1}{c|}{\textbf{Task 4}}           &
\multicolumn{1}{c|}{\textbf{Task 2}}          &
\multicolumn{1}{c|}{\textbf{Task 3}}            & \multicolumn{1}{c}{\textbf{Task 4}}          \\ \midrule

\multicolumn{7}{c}{\textbf{Baseline Approaches}} \\ \midrule
                      LB                &  $0.15\pm0.01$ &  $0.11\pm0.04$ &  $0.06\pm0.01$ 
                                        &  $0.17\pm0.01$ &  $0.12\pm0.03$ &  $0.04\pm0.01$ \\
                      UB                & \cellcolor{gray!25}$\bm{-0.03\pm0.02}$ & \cellcolor{gray!25}$\bm{-0.03\pm0.02}$ & \cellcolor{gray!25}$\bm{-0.01\pm0.00}$ 
                                        & \cellcolor{gray!25}$\bm{-0.04\pm0.02}$ & \cellcolor{gray!25}$\bm{-0.03\pm0.03}$ &  $0.01\pm0.02$ \\\midrule

\multicolumn{7}{c}{\textbf{Regularisation-Based Approaches}} \\ \midrule

 EWC   &  $0.02\pm0.02$ &  $0.01\pm0.02$ &  $0.02\pm0.03$ 
                                        &  $0.01\pm0.02$ &  $0.02\pm0.02$ &  $0.01\pm0.02$ \\
EWC-Online    &  $0.04\pm0.02$ &  $0.03\pm0.01$ &  $0.02\pm0.02$ 
                                        &  $0.00\pm0.04$ &  $0.03\pm0.01$ &  $0.01\pm0.03$ \\
         SI    &  $0.15\pm0.02$ &  $0.10\pm0.03$ &  $0.06\pm0.01$ 
                                        &  $0.15\pm0.01$ &  $0.12\pm0.03$ &  $0.03\pm0.00$ \\
                 LwF    & [$-0.01\pm0.01$] & [$-0.01\pm0.01$] & [$0.00\pm0.00$] 
                                        &  $0.00\pm0.01$ &  $0.01\pm0.01$ &  $0.00\pm0.00$ \\ \midrule

\multicolumn{7}{c}{\textbf{Replay-Based Approaches}} \\ \midrule

               NR    &  $0.05\pm0.01$ &  $0.07\pm0.02$ &  $0.02\pm0.01$ 
                                        &  $0.06\pm0.02$ &  $0.08\pm0.01$ &  $0.00\pm0.01$ \\
   A-GEM    &  $0.01\pm0.01$ &  $0.02\pm0.02$ &  \cellcolor{gray!25}$\bm{-0.01\pm0.01}$ 
                                        & [$-0.01\pm0.00$] &  [$0.00\pm0.01$] &  \cellcolor{gray!25}$\bm{-0.02\pm0.00}$ \\
       LR    &  $0.02\pm0.02$ &  $0.04\pm0.02$ &  [$0.00\pm0.00$] 
                                        &  $0.02\pm0.02$ &  $0.03\pm0.01$ &  [$-0.01\pm0.01$] \\
             DGR    &  $0.24\pm0.02$ &  $0.21\pm0.02$ &  $0.15\pm0.03$ 
                                        &  $0.22\pm0.01$ &  $0.15\pm0.01$ &  $0.09\pm0.06$ \\
           LGR    &  $0.06\pm0.01$ &  $0.04\pm0.01$ &  $0.03\pm0.00$ 
                                        &  $0.05\pm0.01$ &  $0.03\pm0.01$ &  $0.02\pm0.01$ \\
          DGR+D    &  $0.14\pm0.02$ &  $0.07\pm0.01$ &  $0.06\pm0.02$ 
                                        &  $0.14\pm0.07$ &  $0.07\pm0.01$ &  $0.03\pm0.02$ \\
         LGR+D    &  $0.00\pm0.00$ &  $0.00\pm0.00$ &  [$0.00\pm0.00$] 
                                        &  $0.02\pm0.01$ &  $0.01\pm0.01$ &  $0.00\pm0.00$ \\

\bottomrule

\end{tabular}

}
\end{table}
\begin{figure}  
    \centering
    \subfloat[\ac{Task-IL} Results following O$2$.\label{fig:AffectNet-task-il-aug-order-1}]{\includegraphics[width=0.5\textwidth]{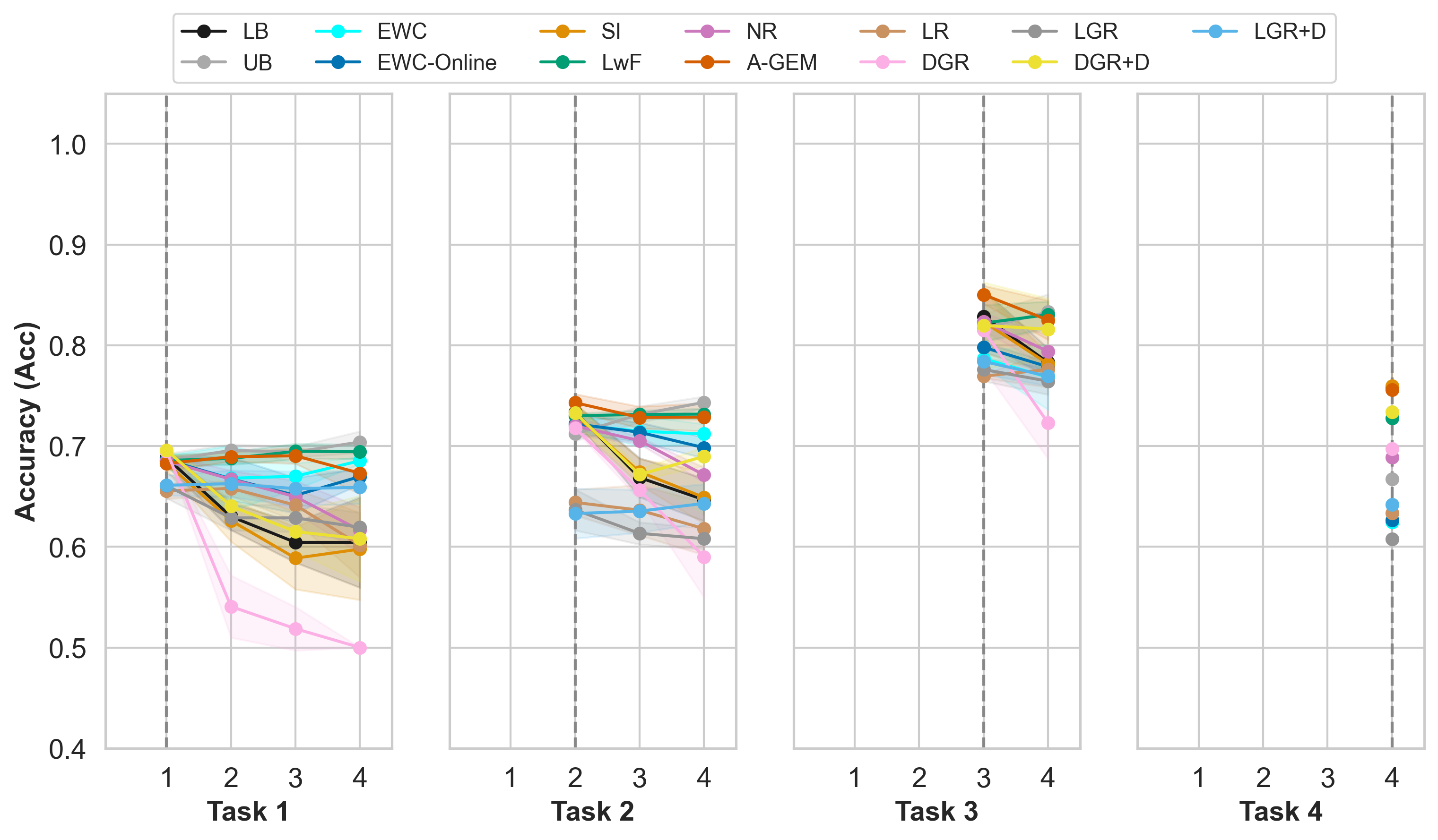}} \hfill
    \subfloat[\ac{Task-IL} Results following O$3$.\label{fig:AffectNet-task-il-aug-order-2}]{\includegraphics[width=0.5\textwidth]{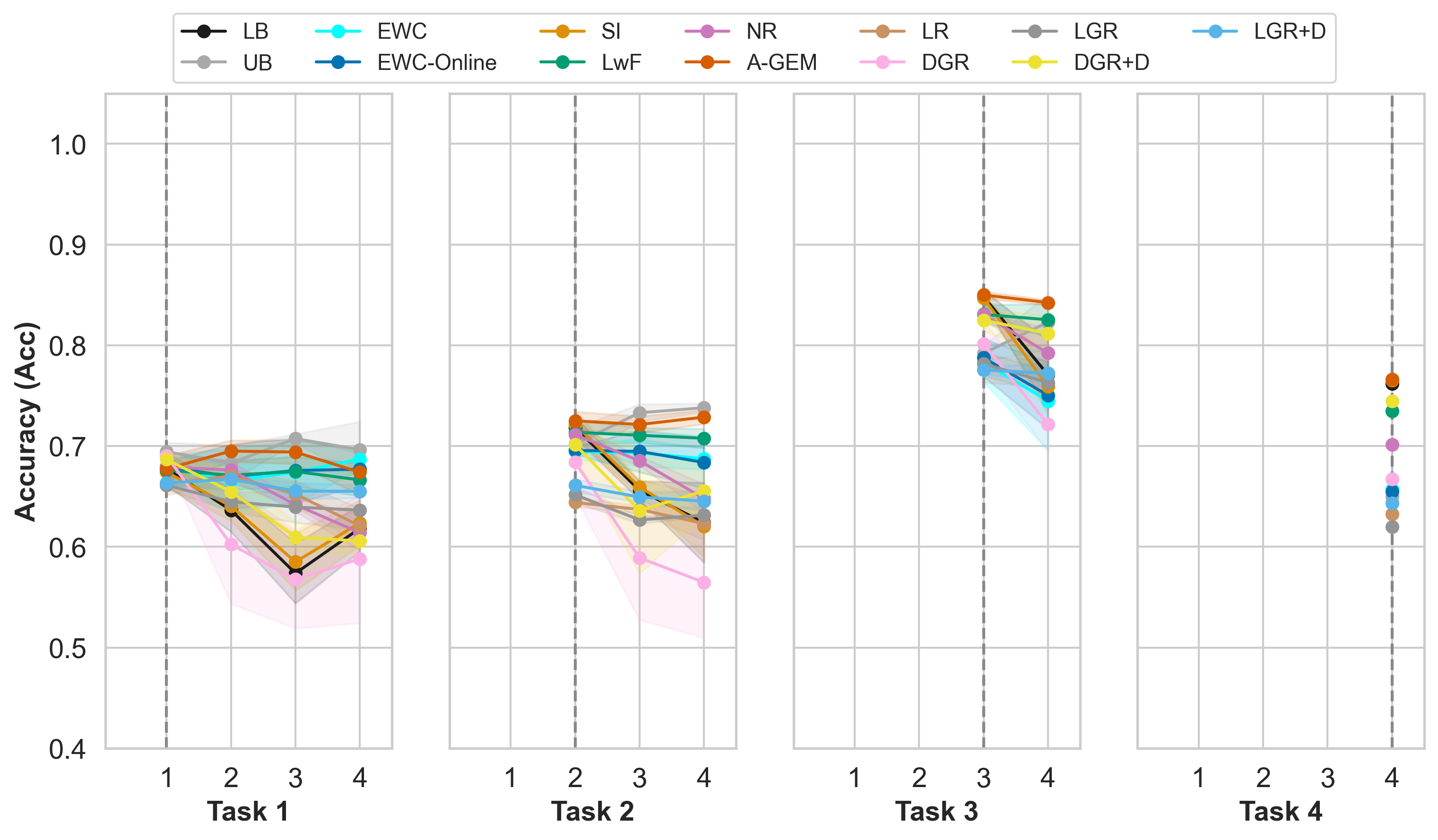}} \\
    \caption[Task-IL Results for AffectNet following Order 2 and Order 3.]{\ac{Task-IL} results for AffectNet following (a) Order $2$ and (b) Order $3$, respectively. For each task, test-set accuracy is shown as the learning progresses from when the task is introduced to the end of the overall training procedure.}
    \label{fig:AffectNet-taskil-orders} 
\end{figure}

\normalsize
\textit{}
\newpage
\subsubsection{\acf{Class-IL} Results}
\label{app:classorderaffnet}

\begin{table}[h!]
\centering
\setlength\tabcolsep{1.0pt}

\caption[\acs{Class-IL} \acs{Acc} for AffectNet following Orderings O2 and O3.]{\acs{Class-IL} \acs{Acc} for AffectNet using the Downsampled split following Orderings O2 and O3. \textbf{Bold} values denote best (highest) while [\textit{bracketed}] denote second-best values for each column.}

\label{tab:class-il-affnet-acc-orders}

{
\scriptsize
\begin{tabular}{l|cccccccc}\toprule

\multirow{2}[2]{*}{\makecell[c]{\textbf{Method}}}           & \multicolumn{8}{c}{\textbf{Accuracy following O2}} \\ \cmidrule{2-9}

\multicolumn{1}{c|}{ }                          & 
\multicolumn{1}{c|}{\textbf{Class 1}}            & \multicolumn{1}{c|}{\textbf{Class 2}}          &
\multicolumn{1}{c|}{\textbf{Class 3}}            & \multicolumn{1}{c|}{\textbf{Class 4}}           &
\multicolumn{1}{c|}{\textbf{Class 5}}            & \multicolumn{1}{c|}{\textbf{Class 6}}          &
\multicolumn{1}{c|}{\textbf{Class 7}}            & \multicolumn{1}{c}{\textbf{Class 8}}          \\ \midrule

\multicolumn{9}{c}{\textbf{Baseline Approaches}} \\ \midrule
LB &  \cellcolor{gray!25}$\bm{1.00\pm0.00}$ & $0.50\pm0.00$ & $0.33\pm0.00$ & $0.25\pm0.00$ & $0.20\pm0.00$ & $0.17\pm0.00$ & $0.14\pm0.00$ & $0.12\pm0.00$ \\
UB &  \cellcolor{gray!25}$\bm{1.00\pm0.00}$ & \cellcolor{gray!25}$\bm{0.71\pm0.01}$ & \cellcolor{gray!25}$\bm{0.48\pm0.01}$ & \cellcolor{gray!25}$\bm{0.37\pm0.00}$ & \cellcolor{gray!25}$\bm{0.37\pm0.01}$ & \cellcolor{gray!25}$\bm{0.40\pm0.00}$ & \cellcolor{gray!25}$\bm{0.37\pm0.00}$ & \cellcolor{gray!25}$\bm{0.33\pm0.00}$ \\ \midrule

\multicolumn{9}{c}{\textbf{Regularisation-Based Approaches}} \\ \midrule
EWC   &  \cellcolor{gray!25}$\bm{1.00\pm0.00}$ & $0.50\pm0.00$ & $0.33\pm0.00$ & $0.25\pm0.00$ & $0.20\pm0.00$ & $0.17\pm0.00$ & $0.14\pm0.00$ & $0.12\pm0.00$ \\
EWC-Online   &  \cellcolor{gray!25}$\bm{1.00\pm0.00}$ & $0.50\pm0.00$ & $0.33\pm0.00$ & $0.25\pm0.00$ & $0.20\pm0.00$ & $0.17\pm0.00$ & $0.14\pm0.00$ & $0.12\pm0.00$ \\
SI   &  \cellcolor{gray!25}$\bm{1.00\pm0.00}$ & $0.50\pm0.00$ & $0.33\pm0.00$ & $0.25\pm0.00$ & $0.20\pm0.00$ & $0.17\pm0.00$ & $0.14\pm0.00$ & $0.12\pm0.00$ \\
LwF   &  \cellcolor{gray!25}$\bm{1.00\pm0.00}$ & $0.50\pm0.00$ & $0.33\pm0.00$ & $0.25\pm0.00$ & $0.20\pm0.00$ & $0.17\pm0.00$ & $0.14\pm0.00$ & $0.12\pm0.00$ \\\midrule
\multicolumn{9}{c}{\textbf{Replay-Based Approaches}} \\ \midrule

NR   &  \cellcolor{gray!25}$\bm{1.00\pm0.00}$ & [$0.65\pm0.00$] & [$0.47\pm0.02$] & [$0.34\pm0.03$] & $0.28\pm0.03$ & $0.25\pm0.02$ & $0.20\pm0.01$ & $0.16\pm0.01$ \\
A-GEM   &  \cellcolor{gray!25}$\bm{1.00\pm0.00}$ & $0.50\pm0.00$ & $0.33\pm0.00$ & $0.25\pm0.00$ & $0.20\pm0.00$ & $0.18\pm0.01$ & $0.14\pm0.00$ & $0.12\pm0.00$ \\
LR   &  \cellcolor{gray!25}$\bm{1.00\pm0.00}$ & $0.64\pm0.01$ & [$0.47\pm0.03$] & \cellcolor{gray!25}$\bm{0.37\pm0.02}$ & [$0.34\pm0.02$] & [$0.33\pm0.02$] & [$0.28\pm0.02$] & [$0.24\pm0.02$] \\
DGR   &  \cellcolor{gray!25}$\bm{1.00\pm0.00}$ & $0.50\pm0.00$ & $0.33\pm0.00$ & $0.25\pm0.00$ & $0.20\pm0.00$ & $0.17\pm0.00$ & $0.14\pm0.00$ & $0.12\pm0.00$ \\
LGR   &  \cellcolor{gray!25}$\bm{1.00\pm0.00}$ & $0.51\pm0.00$ & $0.36\pm0.02$ & $0.27\pm0.02$ & $0.21\pm0.02$ & $0.21\pm0.04$ & $0.16\pm0.02$ & $0.13\pm0.00$ \\
DGR+D   &  \cellcolor{gray!25}$\bm{1.00\pm0.00}$ & $0.50\pm0.00$ & $0.33\pm0.00$ & $0.25\pm0.00$ & $0.20\pm0.00$ & $0.17\pm0.00$ & $0.14\pm0.00$ & $0.12\pm0.00$ \\
LGR+D   &  \cellcolor{gray!25}$\bm{1.00\pm0.00}$ & $0.57\pm0.01$ & $0.43\pm0.01$ & $0.28\pm0.03$ & $0.26\pm0.02$ & $0.27\pm0.02$ & $0.20\pm0.01$ & $0.14\pm0.02$ \\ \midrule

\multirow{2}[2]{*}{\makecell[c]{\textbf{Method}}}           & \multicolumn{8}{c}{\textbf{Accuracy following O3}} \\ \cmidrule{2-9}

\multicolumn{1}{c|}{ }                          & 
\multicolumn{1}{c|}{\textbf{Class 1}}            & \multicolumn{1}{c|}{\textbf{Class 2}}          &
\multicolumn{1}{c|}{\textbf{Class 3}}            & \multicolumn{1}{c|}{\textbf{Class 4}}           &
\multicolumn{1}{c|}{\textbf{Class 5}}            & \multicolumn{1}{c|}{\textbf{Class 6}}          &
\multicolumn{1}{c|}{\textbf{Class 7}}            & \multicolumn{1}{c}{\textbf{Class 8}}          \\ \midrule
\multicolumn{9}{c}{\textbf{Baseline Approaches}} \\ \midrule
LB &  \cellcolor{gray!25}$\bm{1.00\pm0.00}$ & $0.50\pm0.00$ & $0.33\pm0.00$ & $0.25\pm0.00$ & $0.20\pm0.00$ & $0.17\pm0.00$ & $0.14\pm0.00$ & $0.12\pm0.00$ \\
UB &  \cellcolor{gray!25}$\bm{1.00\pm0.00}$ & \cellcolor{gray!25}$\bm{0.70\pm0.00}$ & [$0.47\pm0.01$] & [$0.37\pm0.01$] & [$0.34\pm0.01$] & \cellcolor{gray!25}$\bm{0.40\pm0.01}$ & \cellcolor{gray!25}$\bm{0.35\pm0.00}$ & \cellcolor{gray!25}$\bm{0.32\pm0.01}$ \\\midrule

\multicolumn{9}{c}{\textbf{Regularisation-Based Approaches}} \\ \midrule

EWC   &  \cellcolor{gray!25}$\bm{1.00\pm0.00}$ & $0.50\pm0.00$ & $0.33\pm0.00$ & $0.25\pm0.00$ & $0.20\pm0.00$ & $0.17\pm0.00$ & $0.14\pm0.00$ & $0.12\pm0.00$ \\
EWC-Online   &  \cellcolor{gray!25}$\bm{1.00\pm0.00}$ & $0.50\pm0.00$ & $0.33\pm0.00$ & $0.25\pm0.00$ & $0.20\pm0.00$ & $0.17\pm0.00$ & $0.14\pm0.00$ & $0.12\pm0.00$ \\
SI   &  \cellcolor{gray!25}$\bm{1.00\pm0.00}$ & $0.50\pm0.00$ & $0.33\pm0.00$ & $0.25\pm0.00$ & $0.20\pm0.00$ & $0.17\pm0.00$ & $0.14\pm0.00$ & $0.12\pm0.00$ \\
LwF   &  \cellcolor{gray!25}$\bm{1.00\pm0.00}$ & $0.50\pm0.00$ & $0.33\pm0.00$ & $0.25\pm0.00$ & $0.20\pm0.00$ & $0.17\pm0.00$ & $0.14\pm0.00$ & $0.12\pm0.00$ \\\midrule
\multicolumn{9}{c}{\textbf{Replay-Based Approaches}} \\ \midrule

NR   &  \cellcolor{gray!25}$\bm{1.00\pm0.00}$ & [$0.67\pm0.01$] & [$0.47\pm0.01$] & $0.34\pm0.01$ & $0.27\pm0.01$ & $0.24\pm0.01$ & $0.20\pm0.00$ & $0.16\pm0.00$ \\
A-GEM   &  \cellcolor{gray!25}$\bm{1.00\pm0.00}$ & $0.50\pm0.00$ & $0.33\pm0.00$ & $0.25\pm0.00$ & $0.20\pm0.00$ & $0.17\pm0.00$ & $0.14\pm0.00$ & $0.12\pm0.00$ \\
LR   &  \cellcolor{gray!25}$\bm{1.00\pm0.00}$ & [$0.67\pm0.02$] & \cellcolor{gray!25}$\bm{0.52\pm0.02}$ & \cellcolor{gray!25}$\bm{0.41\pm0.02}$ & \cellcolor{gray!25}$\bm{0.37\pm0.02}$ & [$0.36\pm0.02$] & [$0.31\pm0.02$] & [$0.27\pm0.01$] \\
DGR   &  \cellcolor{gray!25}$\bm{1.00\pm0.00}$ & $0.50\pm0.00$ & $0.33\pm0.00$ & $0.25\pm0.00$ & $0.20\pm0.00$ & $0.17\pm0.00$ & $0.14\pm0.00$ & $0.12\pm0.00$ \\
LGR   &  \cellcolor{gray!25}$\bm{1.00\pm0.00}$ & $0.54\pm0.03$ & $0.39\pm0.04$ & $0.28\pm0.01$ & $0.25\pm0.03$ & $0.27\pm0.04$ & $0.22\pm0.01$ & $0.17\pm0.02$ \\
DGR+D   &  \cellcolor{gray!25}$\bm{1.00\pm0.00}$ & $0.50\pm0.00$ & $0.33\pm0.00$ & $0.25\pm0.00$ & $0.20\pm0.00$ & $0.17\pm0.00$ & $0.14\pm0.00$ & $0.12\pm0.00$ \\
LGR+D   &  \cellcolor{gray!25}$\bm{1.00\pm0.00}$ & $0.61\pm0.02$ & $0.43\pm0.03$ & $0.30\pm0.02$ & $0.27\pm0.02$ & $0.29\pm0.02$ & $0.23\pm0.01$ & $0.19\pm0.03$ \\
\bottomrule

\end{tabular}

}
\end{table}

\begin{table}[h!]
\centering
\caption[\acs{Class-IL} \acs{CF} scores for AffectNet following O2 and O3.]{\acs{Class-IL} \acs{CF} scores for AffectNet using the Downsampled split following Orderings O2 and O3. \textbf{Bold} values denote best (lowest) while [\textit{bracketed}] denote second-best values for each column.}

\label{tab:class-il-affnet-cf-orders}

{
\scriptsize
\setlength\tabcolsep{3.5pt}

\begin{tabular}{l|ccccccc}\toprule

\multirow{2}[2]{*}{\makecell[c]{\textbf{Method}}}           & \multicolumn{7}{c}{\textbf{\acs{CF} following O2}} \\ \cmidrule{2-8}

\multicolumn{1}{c|}{ }                          & 
\multicolumn{1}{c|}{\textbf{Class 2}}          &
\multicolumn{1}{c|}{\textbf{Class 3}}            & \multicolumn{1}{c|}{\textbf{Class 4}}           &
\multicolumn{1}{c|}{\textbf{Class 5}}            & \multicolumn{1}{c|}{\textbf{Class 6}}          &
\multicolumn{1}{c|}{\textbf{Class 7}}            & \multicolumn{1}{c}{\textbf{Class 8}}          \\ \midrule

\multicolumn{8}{c}{\textbf{Baseline Approaches}} \\ \midrule

LB & $1.00\pm0.00$ & $1.00\pm0.00$ & $1.00\pm0.00$ & $1.00\pm0.00$ & $1.00\pm0.00$ & $1.00\pm0.00$ & $1.00\pm0.00$ \\
UB & \cellcolor{gray!25}$\bm{0.12\pm0.02}$ & \cellcolor{gray!25}$\bm{0.06\pm0.02}$ & \cellcolor{gray!25}$\bm{0.02\pm0.01}$ & \cellcolor{gray!25}$\bm{0.06\pm0.01}$ & \cellcolor{gray!25}$\bm{0.08\pm0.01}$ & \cellcolor{gray!25}$\bm{0.13\pm0.02}$ & \cellcolor{gray!25}$\bm{0.12\pm0.03}$ \\\midrule

\multicolumn{8}{c}{\textbf{Regularisation-Based Approaches}} \\ \midrule

EWC   & $1.00\pm0.00$ & $1.00\pm0.00$ & $1.00\pm0.00$ & $1.00\pm0.00$ & $1.00\pm0.00$ & $1.00\pm0.00$ & $1.00\pm0.00$ \\
EWC-Online   & $1.00\pm0.00$ & $1.00\pm0.00$ & $1.00\pm0.00$ & $1.00\pm0.00$ & $1.00\pm0.00$ & $1.00\pm0.00$ & $1.00\pm0.00$ \\
SI   & $1.00\pm0.00$ & $1.00\pm0.00$ & $1.00\pm0.00$ & $1.00\pm0.00$ & $1.00\pm0.00$ & $1.00\pm0.00$ & $1.00\pm0.00$ \\
LwF   & $1.00\pm0.00$ & $1.00\pm0.00$ & $1.00\pm0.00$ & $1.00\pm0.00$ & $1.00\pm0.00$ & $1.00\pm0.00$ & $1.00\pm0.00$ \\\midrule

\multicolumn{8}{c}{\textbf{Replay-Based Approaches}} \\ \midrule

NR   & $0.58\pm0.06$ & $0.68\pm0.07$ & $0.73\pm0.07$ & $0.76\pm0.06$ & $0.81\pm0.04$ & $0.85\pm0.04$ & $0.82\pm0.01$ \\
A-GEM   & $1.00\pm0.00$ & $1.00\pm0.00$ & $1.00\pm0.00$ & $0.99\pm0.02$ & $1.00\pm0.00$ & $1.00\pm0.00$ & $1.00\pm0.00$ \\
LR   & [$0.36\pm0.03$] & [$0.29\pm0.02$] & [$0.27\pm0.01$] & [$0.28\pm0.02$] & [$0.25\pm0.02$] & [$0.22\pm0.00$] & [$0.19\pm0.00$] \\
DGR   & $1.00\pm0.00$ & $1.00\pm0.00$ & $1.00\pm0.00$ & $1.00\pm0.00$ & $1.00\pm0.00$ & $1.00\pm0.00$ & $1.00\pm0.00$ \\
LGR   & $0.89\pm0.08$ & $0.90\pm0.07$ & $0.90\pm0.07$ & $0.86\pm0.09$ & $0.90\pm0.07$ & $0.92\pm0.05$ & $0.90\pm0.02$ \\
DGR+D   & $1.00\pm0.00$ & $1.00\pm0.00$ & $1.00\pm0.00$ & $1.00\pm0.00$ & $1.00\pm0.00$ & $1.00\pm0.00$ & $1.00\pm0.00$ \\
LGR+D   & $0.57\pm0.09$ & $0.67\pm0.09$ & $0.64\pm0.09$ & $0.61\pm0.06$ & $0.67\pm0.06$ & $0.74\pm0.07$ & $0.88\pm0.01$ \\ \midrule

\multirow{2}[2]{*}{\makecell[c]{\textbf{Method}}}           & \multicolumn{7}{c}{\textbf{\acs{CF} following O3}} \\ \cmidrule{2-8}

\multicolumn{1}{c|}{ }                          & 
\multicolumn{1}{c|}{\textbf{Class 2}}          &
\multicolumn{1}{c|}{\textbf{Class 3}}            & \multicolumn{1}{c|}{\textbf{Class 4}}           &
\multicolumn{1}{c|}{\textbf{Class 5}}            & \multicolumn{1}{c|}{\textbf{Class 6}}          &
\multicolumn{1}{c|}{\textbf{Class 7}}            & \multicolumn{1}{c}{\textbf{Class 8}}          \\ \midrule

\multicolumn{8}{c}{\textbf{Baseline Approaches}} \\ \midrule
LB & $1.00\pm0.00$ & $1.00\pm0.00$ & $1.00\pm0.00$ & $1.00\pm0.00$ & $1.00\pm0.00$ & $1.00\pm0.00$ & $1.00\pm0.00$ \\
UB & \cellcolor{gray!25}$\bm{0.15\pm0.02}$ & \cellcolor{gray!25}$\bm{0.08\pm0.02}$ & \cellcolor{gray!25}$\bm{0.07\pm0.01}$ & \cellcolor{gray!25}$\bm{0.09\pm0.01}$ & \cellcolor{gray!25}$\bm{0.10\pm0.01}$ & \cellcolor{gray!25}$\bm{0.13\pm0.02}$ & \cellcolor{gray!25}$\bm{0.11\pm0.02}$ \\ \midrule

\multicolumn{8}{c}{\textbf{Regularisation-Based Approaches}} \\ \midrule

EWC   & $1.00\pm0.00$ & $1.00\pm0.00$ & $1.00\pm0.00$ & $1.00\pm0.00$ & $1.00\pm0.00$ & $1.00\pm0.00$ & $1.00\pm0.00$ \\
EWC-Online   & $1.00\pm0.00$ & $1.00\pm0.00$ & $1.00\pm0.00$ & $1.00\pm0.00$ & $1.00\pm0.00$ & $1.00\pm0.00$ & $1.00\pm0.00$ \\
SI   & $1.00\pm0.00$ & $1.00\pm0.00$ & $1.00\pm0.00$ & $1.00\pm0.00$ & $1.00\pm0.00$ & $1.00\pm0.00$ & $1.00\pm0.00$ \\
LwF   & $1.00\pm0.00$ & $1.00\pm0.00$ & $1.00\pm0.00$ & $1.00\pm0.00$ & $1.00\pm0.00$ & $1.00\pm0.00$ & $1.00\pm0.00$ \\\midrule

\multicolumn{8}{c}{\textbf{Replay-Based Approaches}} \\ \midrule

NR   & $0.56\pm0.06$ & $0.68\pm0.05$ & $0.74\pm0.04$ & $0.77\pm0.03$ & $0.80\pm0.03$ & $0.85\pm0.03$ & $0.80\pm0.00$ \\
A-GEM   & $1.00\pm0.00$ & $1.00\pm0.00$ & $1.00\pm0.00$ & $1.00\pm0.00$ & $1.00\pm0.00$ & $1.00\pm0.00$ & $1.00\pm0.00$ \\
LR   & [$0.33\pm0.03$] & [$0.29\pm0.01$] & [$0.29\pm0.00$] & [$0.29\pm0.02$] & [$0.27\pm0.01$] & [$0.25\pm0.00$] & [$0.26\pm0.00$] \\
DGR   & $1.00\pm0.00$ & $1.00\pm0.00$ & $1.00\pm0.00$ & $1.00\pm0.00$ & $1.00\pm0.00$ & $1.00\pm0.00$ & $1.00\pm0.00$ \\
LGR   & $0.80\pm0.09$ & $0.84\pm0.06$ & $0.81\pm0.08$ & $0.75\pm0.09$ & $0.77\pm0.05$ & $0.80\pm0.07$ & $0.81\pm0.05$ \\
DGR+D   & $1.00\pm0.00$ & $1.00\pm0.00$ & $1.00\pm0.00$ & $1.00\pm0.00$ & $1.00\pm0.00$ & $1.00\pm0.00$ & $1.00\pm0.00$ \\
LGR+D   & $0.61\pm0.07$ & $0.70\pm0.07$ & $0.67\pm0.06$ & $0.62\pm0.06$ & $0.66\pm0.04$ & $0.69\pm0.06$ & $0.68\pm0.07$ \\

\bottomrule

\end{tabular}
\vspace{-5mm}
}
\end{table}
\begin{figure}[h!]  
    \centering
    \subfloat[\ac{Class-IL} Results following O$2$.\label{fig:affnet-Class-il-aug-order-1}]{\includegraphics[width=0.5\textwidth]{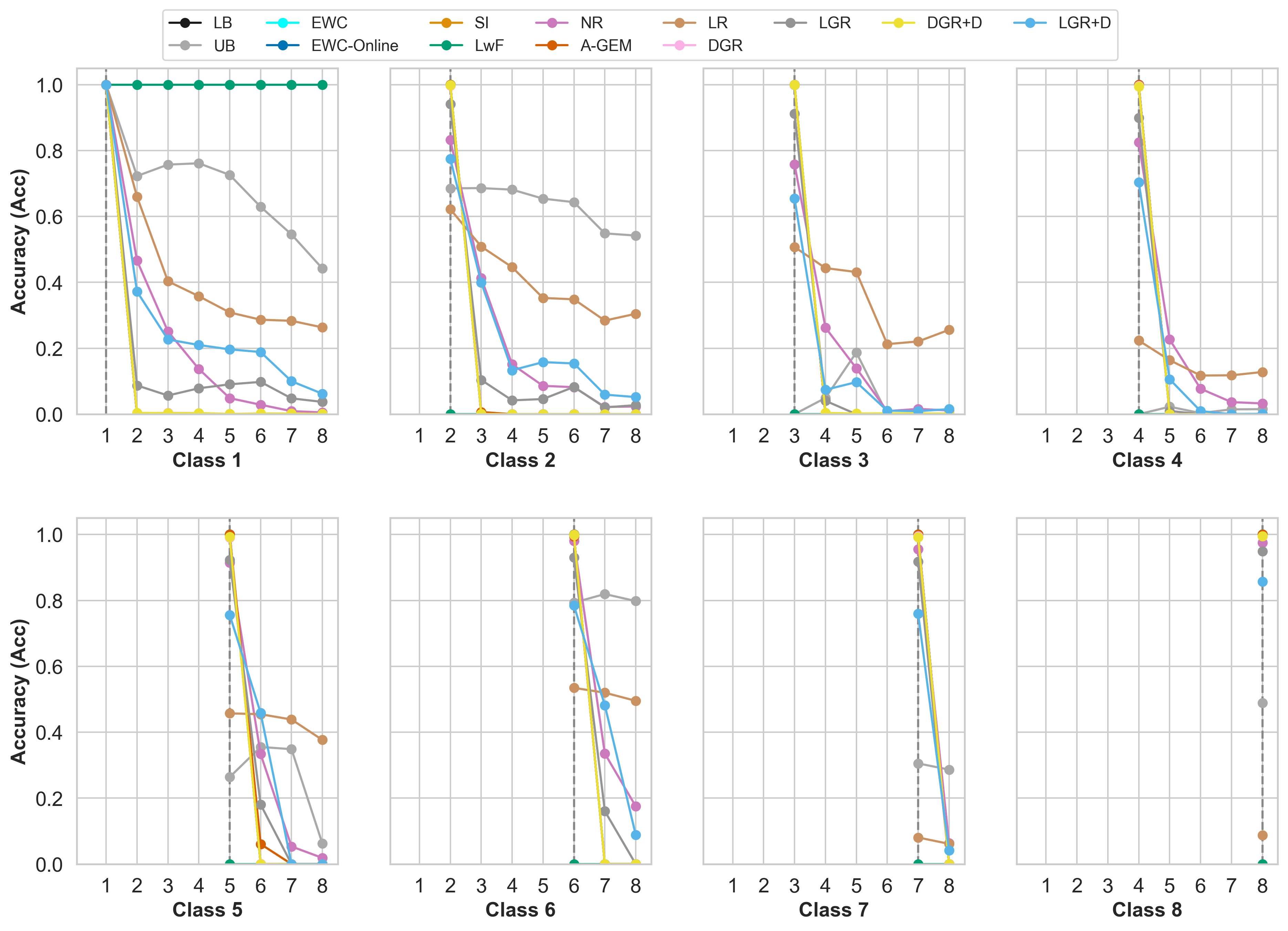}}
    \hfill
    \subfloat[\ac{Class-IL} Results following O$3$.\label{fig:affnet-Class-il-aug-order-2}]{\includegraphics[width=0.5\textwidth]{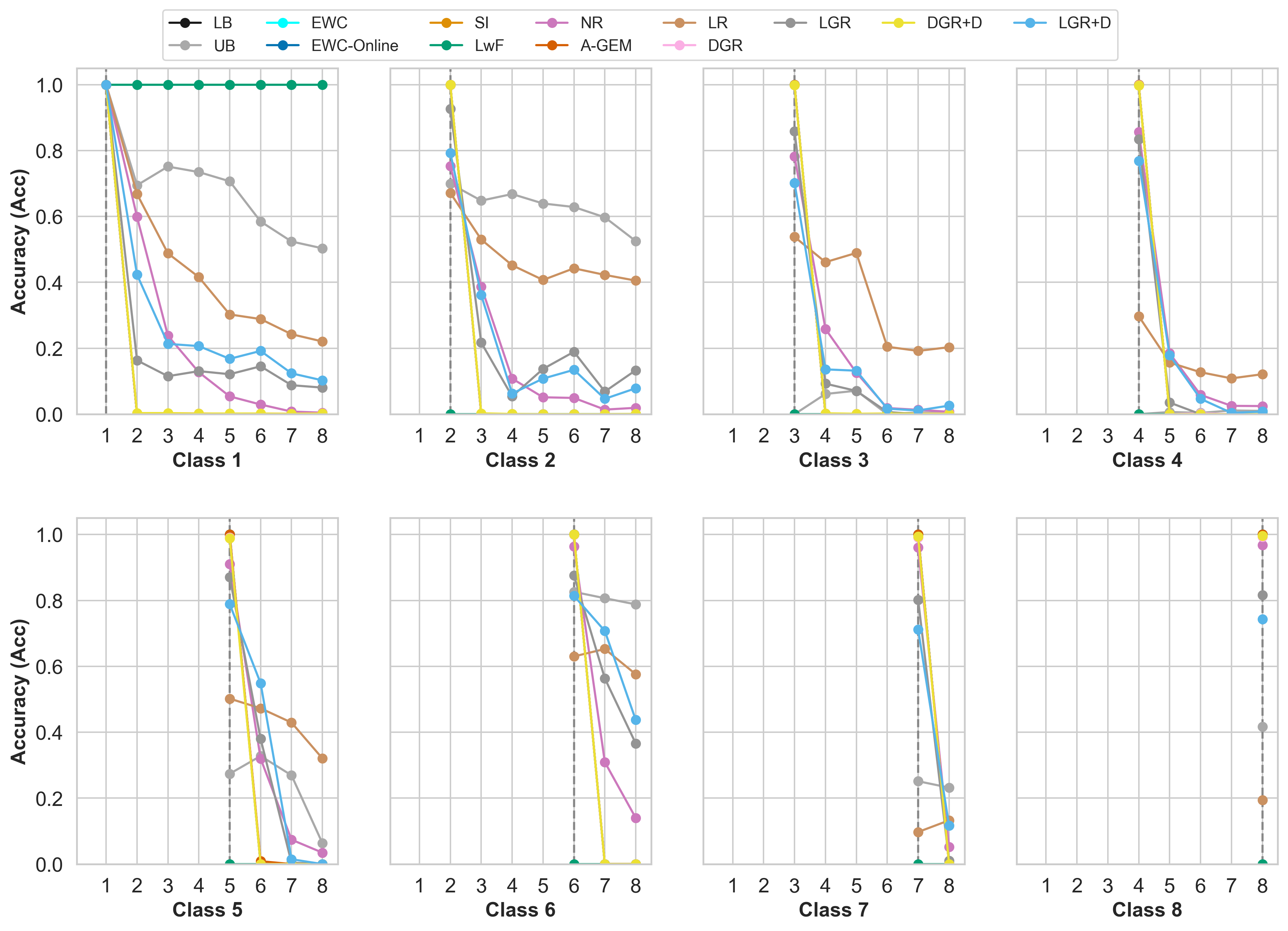}} \\
    \caption[Class-IL Results for AffectNet following Order 2 and Order 3.]{\ac{Class-IL} results for AffectNet following (a) Order $2$ and (b) Order $3$, respectively. For each class, test-set accuracy is shown as learning progresses from when a class is introduced to the end of the overall training procedure.}
    \label{fig:affnet-classil-orders} 
    
\end{figure}

\FloatBarrier

\normalsize

\section{Additional Results}
\label{app:additional}

\subsection{\acf{FER}-2013}
The \ac{FER}-2013 dataset~\citep{Goodfellow2013Challenges} consists of $\approx 36K$ facial images downloaded using the Google image search API by searching 184 different keywords. The dataset is annotated for $7$ different expression classes, namely, \textit{Anger, Surprise, Fearful, Disgust, Happy, Sad} and \textit{Neutral}. The dataset was curated for the \ac{FER}-2013 challenge collecting images representing \textit{in-the-wild} settings with great diversity with respect to gender, age or ethnicity. The training-set consists of $\approx28K$ images while the publicly available test-set consists of $\approx 3.5K$ images. We use the same train \textit{vs.} test-set split for our experiments. 


\subsubsection{\acf{Task-IL} Results}
\label{app:taskilFER}
\begin{table}[h!]
\centering
\setlength\tabcolsep{2.0pt}

\caption[\acs{Task-IL} \acs{Acc} for \acs{FER}-2013  W/ and W/O Data-Augmentation.]{\acs{Task-IL} \acs{Acc} for \ac{FER}-2013  W/ and W/O Data-Augmentation. \textbf{Bold} values denote best (highest) while [\textit{bracketed}] denote second-best values for each column.}
\label{tab:task-il-fer-acc}

{
\scriptsize
\begin{tabular}{l|cccc|cccc}\toprule
\multicolumn{1}{c|}{\textbf{Method}}            & \multicolumn{4}{c|}{\textbf{Accuracy W/O Data-Augmentation}} 
& \multicolumn{4}{c}{\textbf{Accuracy W/ Data-Augmentation}} \\ \cmidrule{2-9}

\multicolumn{1}{c|}{ }                          & 
\multicolumn{1}{c|}{\textbf{Task 1}}            & \multicolumn{1}{c|}{\textbf{Task 2}}          &
\multicolumn{1}{c|}{\textbf{Task 3}}            & \multicolumn{1}{c|}{\textbf{Task 4}}           &
\multicolumn{1}{c|}{\textbf{Task 1}}            & \multicolumn{1}{c|}{\textbf{Task 2}}          &
\multicolumn{1}{c|}{\textbf{Task 3}}            & \multicolumn{1}{c}{\textbf{Task 4}}          \\ \midrule

\multicolumn{9}{c}{\textbf{Baseline Approaches}} \\ \midrule

LB & \cellcolor{gray!25}$\bm{0.68\pm0.00}$ & $0.77\pm0.02$ & $0.75\pm0.01$ & $0.75\pm0.03$ 
& [$0.65\pm0.00$] & $0.76\pm0.01$ & [$0.75\pm0.00$] & $0.75\pm0.03$ \\

UB & \cellcolor{gray!25}$\bm{0.68\pm0.01}$ & \cellcolor{gray!25}$\bm{0.80\pm0.00}$ & [$0.78\pm0.00$] & \cellcolor{gray!25}$\bm{0.84\pm0.00}$ 
& [$0.65\pm0.01$] & \cellcolor{gray!25}$\bm{0.78\pm0.00}$ & [$0.75\pm0.00$] & \cellcolor{gray!25}$\bm{0.82\pm0.00}$ \\ \midrule

\multicolumn{9}{c}{\textbf{Regularisation-Based Approaches}} \\ \midrule

EWC  & [$0.67\pm0.00$] & [$0.79\pm0.00$] & $0.77\pm0.00$ & $0.81\pm0.00$ 
& \cellcolor{gray!25}$\bm{0.66\pm0.00}$ & \cellcolor{gray!25}$\bm{0.78\pm0.00}$ & [$0.75\pm0.01$] & $0.78\pm0.04$ \\

EWC-Online & [$0.67\pm0.00$] & [$0.79\pm0.00$] & $0.77\pm0.00$ & $0.82\pm0.00$ 
& [$0.65\pm0.00$] & [$0.77\pm0.00$] & [$0.75\pm0.00$] & [$0.81\pm0.00$] \\

SI  & \cellcolor{gray!25}$\bm{0.68\pm0.00}$ & $0.78\pm0.00$ & $0.75\pm0.00$ & $0.74\pm0.02$ 
& [$0.65\pm0.00$] & $0.74\pm0.02$ & $0.74\pm0.01$ & $0.75\pm0.04$ \\

LwF  & [$0.67\pm0.00$] & [$0.79\pm0.00$] & [$0.78\pm0.00$] & [$0.83\pm0.00$] 
& \cellcolor{gray!25}$\bm{0.66\pm0.00}$ & \cellcolor{gray!25}$\bm{0.78\pm0.00}$ & \cellcolor{gray!25}$\bm{0.76\pm0.00}$ & \cellcolor{gray!25}$\bm{0.82\pm0.00}$ \\ \midrule
                 
\multicolumn{9}{c}{\textbf{Replay-Based Approaches}} \\ \midrule

NR  & \cellcolor{gray!25}$\bm{0.68\pm0.00}$ & $0.78\pm0.00$ & $0.75\pm0.00$ & $0.79\pm0.00$ 
& [$0.65\pm0.00$] & $0.76\pm0.01$ & $0.72\pm0.00$ & $0.79\pm0.00$ \\

A-GEM  & [$0.67\pm0.00$] & [$0.79\pm0.00$] & \cellcolor{gray!25}$\bm{0.79\pm0.00}$ & $0.79\pm0.03$ 
& [$0.65\pm0.01$] & \cellcolor{gray!25}$\bm{0.78\pm0.00}$ & \cellcolor{gray!25}$\bm{0.76\pm0.00}$ & $0.78\pm0.02$ \\

LR  & $0.65\pm0.01$ & $0.78\pm0.01$ & $0.70\pm0.02$ & $0.77\pm0.03$ 
& $0.63\pm0.00$ & [$0.77\pm0.00$] & $0.68\pm0.01$ & $0.77\pm0.00$ \\

DGR  & \cellcolor{gray!25}$\bm{0.68\pm0.00}$ & $0.75\pm0.02$ & $0.73\pm0.00$ & $0.79\pm0.00$  
& \cellcolor{gray!25}$\bm{0.66\pm0.00}$ & $0.73\pm0.00$ & $0.73\pm0.00$ & $0.77\pm0.01$ \\

LGR  & $0.65\pm0.01$ & $0.77\pm0.01$ & $0.75\pm0.01$ & $0.80\pm0.01$ 
& $0.63\pm0.01$ & $0.75\pm0.00$ & $0.73\pm0.00$ & $0.78\pm0.00$ \\

DGR+D  & [$0.67\pm0.00$] & $0.78\pm0.00$ & $0.77\pm0.00$ & $0.80\pm0.01$ 
& $0.65\pm0.00$ & [$0.77\pm0.01$] & [$0.75\pm0.01$] & $0.78\pm0.00$ \\

LGR+D  & $0.65\pm0.01$ & $0.78\pm0.01$ & $0.75\pm0.00$ & $0.81\pm0.00$ 
& $0.63\pm0.00$ & [$0.77\pm0.00$] & $0.74\pm0.00$ & $0.80\pm0.00$ \\
\bottomrule

\end{tabular}

}
\end{table}

\begin{table}[h!]
\centering
\caption[\acs{Task-IL} \acs{CF} scores \acs{FER}-2013  W/ and W/O Data-Augmentation.]{\acs{Task-IL} \acs{CF} scores \ac{FER}-2013  W/ and W/O Data-Augmentation. \textbf{Bold} values denote best (lowest) while [\textit{bracketed}] denote second-best values for each column.}
\label{tab:task-il-fer-cf}

{
\scriptsize
\setlength\tabcolsep{4.5pt}

\begin{tabular}{l|ccc|ccc}\toprule
\multicolumn{1}{c|}{\textbf{Method}}            & \multicolumn{3}{c|}{\textbf{\acs{CF} W/O Data-Augmentation}} 
& \multicolumn{3}{c}{\textbf{\acs{CF}  W/ Data-Augmentation}} \\ \cmidrule{2-7}

\multicolumn{1}{c|}{ }                          & 
\multicolumn{1}{c|}{\textbf{Task 2}}          &
\multicolumn{1}{c|}{\textbf{Task 3}}            & \multicolumn{1}{c|}{\textbf{Task 4}}           &
\multicolumn{1}{c|}{\textbf{Task 2}}          &
\multicolumn{1}{c|}{\textbf{Task 3}}            & \multicolumn{1}{c}{\textbf{Task 4}}          \\ \midrule

\multicolumn{7}{c}{\textbf{Baseline Approaches}} \\ \midrule

LB &  $0.09\pm0.02$ &  $0.16\pm0.06$ &  $0.04\pm0.02$ 
&  $0.03\pm0.01$ &  $0.14\pm0.05$ &  $0.03\pm0.03$ \\
UB & \cellcolor{gray!25}$\bm{-0.02\pm0.00}$ & \cellcolor{gray!25}$\bm{-0.03\pm0.01}$ & \cellcolor{gray!25}$\bm{-0.01\pm0.01}$ 
& \cellcolor{gray!25}$\bm{-0.02\pm0.01}$ & \cellcolor{gray!25}$\bm{-0.02\pm0.01}$ & \cellcolor{gray!25}$\bm{-0.02\pm0.01}$ \\\midrule

\multicolumn{7}{c}{\textbf{Regularisation-Based Approaches}} \\ \midrule
EWC  &  $0.01\pm0.00$ &  $0.03\pm0.01$ &  [$0.00\pm0.01$] 
&  $0.01\pm0.01$ &  $0.07\pm0.07$ &  $0.01\pm0.00$ \\

EWC-Online &  $0.01\pm0.00$ &  $0.02\pm0.01$ &  [$0.00\pm0.01$] 
&  $0.01\pm0.00$ &  $0.02\pm0.01$ &  $0.01\pm0.01$ \\

SI  &  $0.09\pm0.00$ &  $0.18\pm0.04$ &  $0.04\pm0.01$ 
&  $0.05\pm0.02$ &  $0.13\pm0.07$ &  $0.08\pm0.04$ \\

LwF  &  $0.00\pm0.00$ &  [$0.01\pm0.00$] &  [$0.00\pm0.00$] 
&  $0.01\pm0.01$ &  [$0.01\pm0.00$] &  [$0.00\pm0.00$] \\\midrule

\multicolumn{7}{c}{\textbf{Replay-Based Approaches}} \\ \midrule

NR  &  $0.11\pm0.01$ &  $0.07\pm0.00$ &  $0.04\pm0.01$ 
&  $0.09\pm0.00$ &  $0.06\pm0.00$ &  $0.04\pm0.01$ \\

A-GEM  & [$-0.01\pm0.00$] &  $0.09\pm0.05$ &  [$0.00\pm0.01$] 
& [$0.00\pm0.01$] &  $0.07\pm0.03$ & [$0.00\pm0.01$] \\

LR  &  $0.15\pm0.04$ &  $0.09\pm0.04$ & [$0.00\pm0.01$] 
&  $0.16\pm0.02$ &  $0.07\pm0.01$ & [$0.00\pm0.00$] \\

DGR  &  $0.10\pm0.01$ &  $0.08\pm0.00$ &  $0.08\pm0.04$ 
&  $0.09\pm0.01$ &  $0.09\pm0.02$ &  $0.09\pm0.01$ \\

LGR  &  $0.02\pm0.01$ &  $0.02\pm0.01$ &  $0.01\pm0.01$ 
&  $0.02\pm0.00$ &  $0.04\pm0.01$ &  $0.02\pm0.00$ \\

DGR+D  &  $0.02\pm0.02$ &  $0.05\pm0.02$ &  $0.01\pm0.01$ 
&  $0.03\pm0.01$ &  $0.06\pm0.01$ &  $0.02\pm0.01$ \\
LGR+D  &  $0.00\pm0.01$ &  [$0.01\pm0.01$] & [$0.00\pm0.00$] 
&  [$0.00\pm0.01$] & [$0.01\pm0.01$] &  [$0.00\pm0.00$] \\

\bottomrule
\end{tabular}
}
\vfill
\end{table}

\begin{figure}[h!]
    \centering
    \subfloat[\ac{Task-IL} Results w/o Augmentation.\label{fig:fer-task-il-noaug}]{\includegraphics[width=0.5\textwidth]{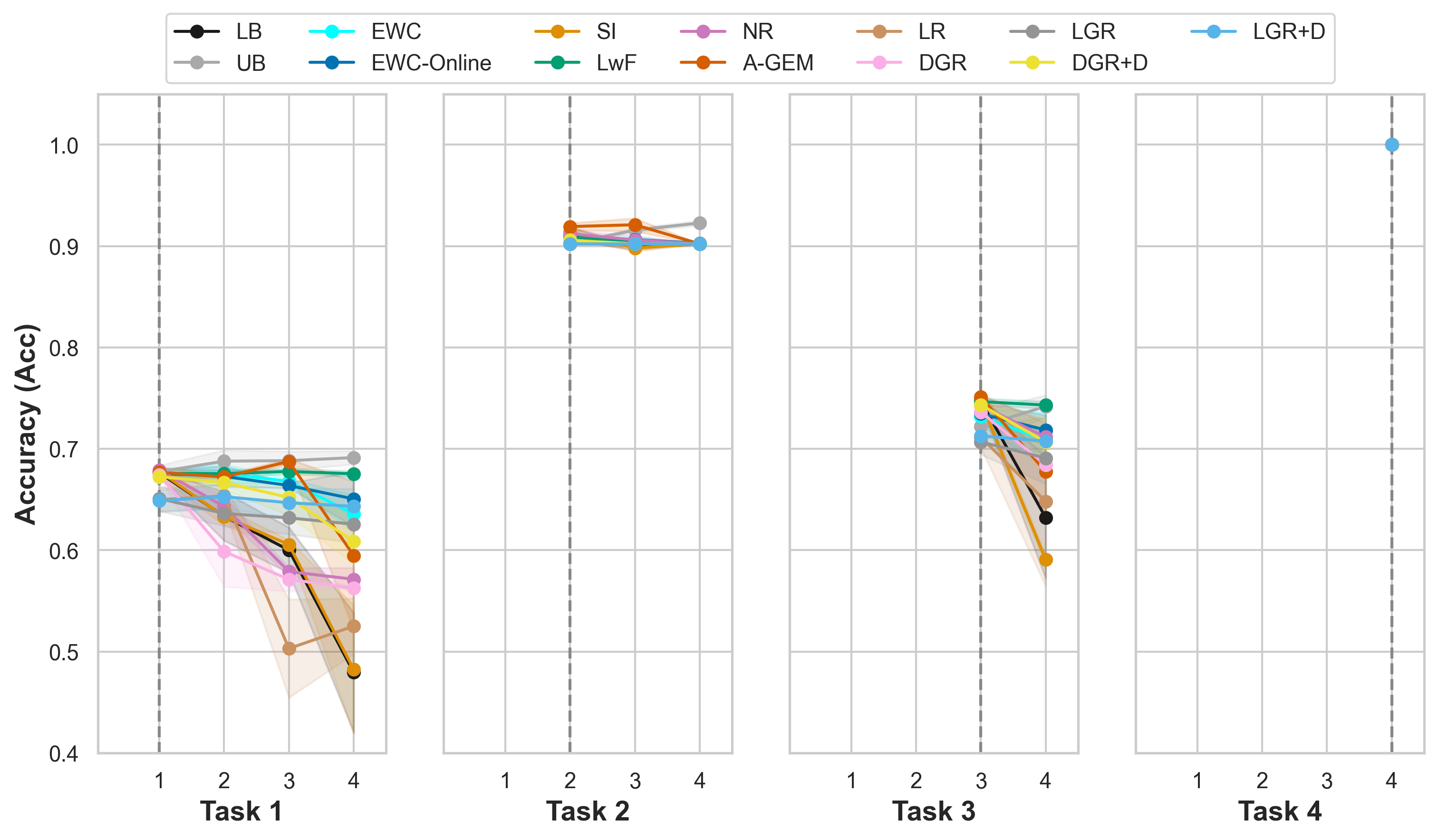}} \hfill
    \subfloat[\ac{Task-IL} Results w/ Augmentation.\label{fig:fer-task-il-aug}]{\includegraphics[width=0.5\textwidth]{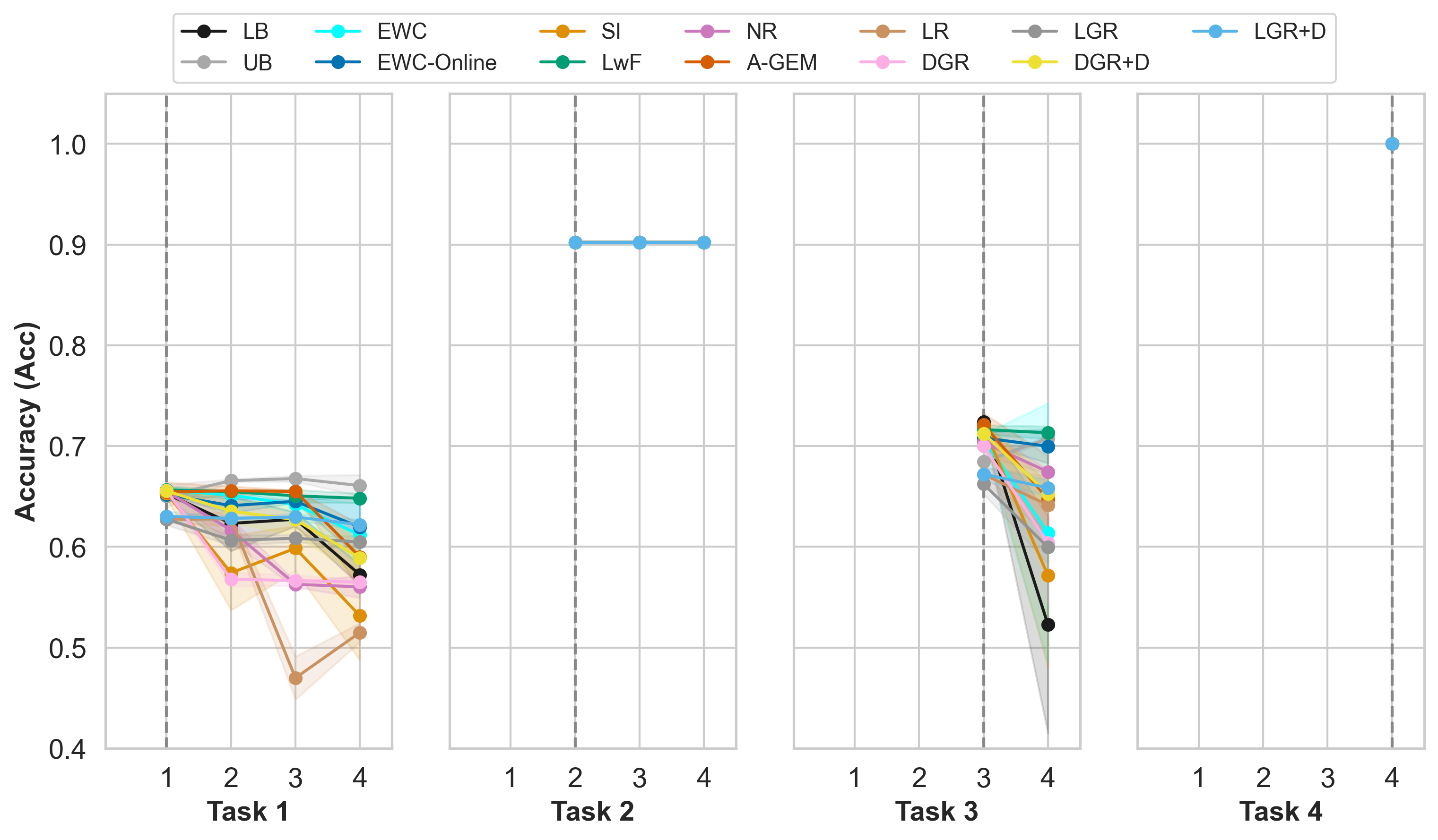}} \\
    \caption[Task-IL Results for FER W/ and W/O Augmentation.]{\ac{Task-IL} results for \ac{FER}-2013 (a) without and (b) with augmentation. For each task, test-set accuracy is shown as the learning progresses from when the task is introduced to the end of the overall training procedure.}
    \label{fig:fer-taskil} 
\end{figure}


\FloatBarrier

\subsubsection{Task-Ordering}
\label{app:taskorderFER}
\begin{table}[h!]
\centering
\setlength\tabcolsep{1.0pt}

\caption[\acs{Task-IL} \acs{Acc} for \ac{FER}-2013 following Orderings O2 and O3.]{\acs{Task-IL} \acs{Acc} for \ac{FER}-2013 W/ Data-Augmentation following Orderings O2 and O3. \textbf{Bold} values denote best (highest) while [\textit{bracketed}] denote second-best values for each column.}

\label{tab:task-il-fer-orders-acc}

{
\scriptsize
\begin{tabular}{l|cccc|cccc}\toprule
\multicolumn{1}{c|}{\textbf{Method}}            & \multicolumn{4}{c|}{\textbf{Accuracy following O2}} 
& \multicolumn{4}{c}{\textbf{Accuracy following O3}} \\ \midrule

\multicolumn{1}{c|}{ }                          & 
\multicolumn{1}{c|}{\textbf{Task 1}}            & \multicolumn{1}{c|}{\textbf{Task 2}}          &
\multicolumn{1}{c|}{\textbf{Task 3}}            & \multicolumn{1}{c|}{\textbf{Task 4}}           &
\multicolumn{1}{c|}{\textbf{Task 1}}            & \multicolumn{1}{c|}{\textbf{Task 2}}          &
\multicolumn{1}{c|}{\textbf{Task 3}}            & \multicolumn{1}{c}{\textbf{Task 4}}          \\ \cmidrule{2-9}

\multicolumn{9}{c}{\textbf{Baseline Approaches}} \\ \midrule
LB & [$0.64\pm0.00$] & $0.75\pm0.01$ & $0.74\pm0.01$ & $0.72\pm0.00$ 
& [$0.66\pm0.00$] & $0.76\pm0.01$ & $0.75\pm0.00$ & $0.73\pm0.01$ \\

UB & \cellcolor{gray!25}$\bm{0.65\pm0.00}$ & \cellcolor{gray!25}$\bm{0.78\pm0.00}$ & [$0.75\pm0.00$] & \cellcolor{gray!25}$\bm{0.82\pm0.00}$ 
& [$0.66\pm0.00$] & \cellcolor{gray!25}$\bm{0.79\pm0.00}$ & [$0.76\pm0.00$] & \cellcolor{gray!25}$\bm{0.82\pm0.00}$ \\
 \midrule

\multicolumn{9}{c}{\textbf{Regularisation-Based Approaches}} \\ \midrule

EWC  & \cellcolor{gray!25}$\bm{0.65\pm0.00}$ & [$0.77\pm0.00$] & [$0.75\pm0.00$] & $0.77\pm0.01$ 
& $0.65\pm0.00$ & [$0.78\pm0.00$] & [$0.76\pm0.00$] & [$0.79\pm0.02$] \\

EWC-Online  & \cellcolor{gray!25}$\bm{0.65\pm0.00}$ & [$0.77\pm0.00$] & [$0.75\pm0.00$] & $0.78\pm0.01$ 
& [$0.66\pm0.01$] & [$0.78\pm0.00$] & $0.75\pm0.00$ & $0.78\pm0.02$ \\

SI  & [$0.64\pm0.00$] & $0.75\pm0.01$ & $0.74\pm0.01$ & $0.75\pm0.04$ & 
$0.65\pm0.00$ & $0.75\pm0.02$ & $0.75\pm0.00$ & $0.74\pm0.03$ \\

LwF  & [$0.64\pm0.00$] & [$0.77\pm0.00$] & $0.75\pm0.00$ & [$0.81\pm0.00$]
& \cellcolor{gray!25}$\bm{0.67\pm0.00}$ & [$0.78\pm0.00$] & [$0.76\pm0.00$] & \cellcolor{gray!25}$\bm{0.82\pm0.00}$ \\ \midrule

\multicolumn{9}{c}{\textbf{Replay-Based Approaches}} \\ \midrule

NR  & [$0.64\pm0.00$] & $0.76\pm0.00$ & $0.73\pm0.00$ & $0.79\pm0.00$ 
& [$0.66\pm0.01$] & $0.77\pm0.00$ & $0.73\pm0.01$ & [$0.79\pm0.00$] \\

A-GEM  & [$0.64\pm0.00$] & [$0.77\pm0.00$] & \cellcolor{gray!25}$\bm{0.76\pm0.00}$ & $0.79\pm0.00$ 
& [$0.66\pm0.00$] & [$0.78\pm0.00$] & \cellcolor{gray!25}$\bm{0.77\pm0.00}$ & $0.78\pm0.00$ \\

LR  & $0.63\pm0.00$ & $0.76\pm0.00$ & $0.67\pm0.00$ & $0.74\pm0.02$ 
& $0.64\pm0.00$ & $0.76\pm0.00$ & $0.68\pm0.01$ & $0.76\pm0.01$ \\

DGR  & \cellcolor{gray!25}$\bm{0.65\pm0.01}$ & $0.73\pm0.00$ & $0.72\pm0.00$ & $0.78\pm0.00$ 
& [$0.66\pm0.00$] & $0.74\pm0.01$ & $0.72\pm0.01$ & $0.78\pm0.00$ \\

LGR  & $0.63\pm0.01$ & $0.76\pm0.00$ & $0.72\pm0.00$ & $0.77\pm0.00$ 
& $0.63\pm0.00$ & $0.75\pm0.01$ & $0.72\pm0.01$ & $0.77\pm0.01$ \\

DGR+D  & \cellcolor{gray!25}$\bm{0.65\pm0.00}$ & $0.75\pm0.00$ & $0.74\pm0.00$ & $0.78\pm0.00$ 
& [$0.66\pm0.00$] & $0.77\pm0.01$ & $0.75\pm0.00$ & $0.78\pm0.01$ \\

LGR+D  & $0.63\pm0.01$ & $0.76\pm0.00$ & $0.73\pm0.00$ & $0.80\pm0.00$ 
& $0.63\pm0.00$ & $0.76\pm0.00$ & $0.73\pm0.00$ & [$0.79\pm0.00$] \\
\bottomrule
\end{tabular}
\vspace{-3mm}
}
\end{table}

\begin{table}[h!]
\centering
\caption[\acs{Task-IL} \acs{CF} scores for \ac{FER}-2013 following O2 and O3.]{\acs{Task-IL} \acs{CF} scores for \ac{FER}-2013 W/ Data-Augmentation following Orderings O2 and O3. \textbf{Bold} values denote best (lowest) while [\textit{bracketed}] denote second-best values for each column.}

\label{tab:task-il-fer-orders-cf}

{
\scriptsize

\begin{tabular}{l|ccc|ccc}\toprule
\multicolumn{1}{c|}{\textbf{Method}}            & \multicolumn{3}{c|}{\textbf{CF following O2}} 
& \multicolumn{3}{c}{\textbf{CF following O3}} \\ \cmidrule{2-7}

\multicolumn{1}{c|}{ }                          & 
\multicolumn{1}{c|}{\textbf{Task 2}}          &
\multicolumn{1}{c|}{\textbf{Task 3}}            & \multicolumn{1}{c|}{\textbf{Task 4}}           &
\multicolumn{1}{c|}{\textbf{Task 2}}          &
\multicolumn{1}{c|}{\textbf{Task 3}}            & \multicolumn{1}{c}{\textbf{Task 4}}          \\ \midrule

\multicolumn{7}{c}{\textbf{Baseline Approaches}} \\ \midrule

LB                                      &  $0.04\pm0.01$ &  $0.19\pm0.01$ &  $0.04\pm0.02$ 
                                        &  $0.03\pm0.01$ &  $0.17\pm0.03$ &  $0.04\pm0.01$ \\
UB                                      & \cellcolor{gray!25}$\bm{-0.02\pm0.00}$ & \cellcolor{gray!25}$\bm{-0.02\pm0.00}$ & \cellcolor{gray!25}$\bm{0.00\pm0.00}$ 
                                        & \cellcolor{gray!25}$\bm{-0.01\pm0.00}$ & \cellcolor{gray!25}$\bm{-0.02\pm0.01}$ & \cellcolor{gray!25}$\bm{-0.01\pm0.01}$ \\\midrule

\multicolumn{7}{c}{\textbf{Regularisation-Based Approaches}} \\ \midrule

EWC    &  $0.01\pm0.00$ &  $0.08\pm0.02$ &  $0.02\pm0.01$ 
                                        &  [$0.00\pm0.00$] &  $0.05\pm0.03$ &  [$0.00\pm0.00$] \\
EWC-Online   &  $0.02\pm0.01$ &  $0.07\pm0.02$ &  [$0.01\pm0.00$] 
                                        &  $0.02\pm0.01$ &  $0.08\pm0.04$ &  [$0.00\pm0.00$] \\
SI            &  $0.04\pm0.02$ &  $0.14\pm0.08$ &  $0.04\pm0.03$ 
                                        &  $0.02\pm0.00$ &  $0.17\pm0.07$ &  $0.05\pm0.03$ \\
LwF                    &  [$0.00\pm0.01$] &  [$0.00\pm0.00$] &  \cellcolor{gray!25}$\bm{0.00\pm0.00}$ 
                                        &  [$0.00\pm0.00$] &  [$0.00\pm0.00$] &  [$0.00\pm0.00$] \\ \midrule

\multicolumn{7}{c}{\textbf{Replay-Based Approaches}} \\ \midrule

               NR   &  $0.09\pm0.02$ &  $0.06\pm0.01$ &  $0.03\pm0.01$
                                        &  $0.09\pm0.02$ &  $0.06\pm0.01$ &  $0.01\pm0.01$ \\
   A-GEM   &  $0.00\pm0.00$ &  $0.06\pm0.01$ &  [$0.01\pm0.01$] 
                                        &  [$0.00\pm0.00$] &  $0.10\pm0.00$ &  $0.01\pm0.01$ \\
       LR   &  $0.17\pm0.01$ &  $0.11\pm0.05$ &  \cellcolor{gray!25}$\bm{0.00\pm0.00}$ 
                                        &  $0.17\pm0.01$ &  $0.08\pm0.02$ &  $0.01\pm0.01$ \\
             DGR   &  $0.09\pm0.01$ &  $0.07\pm0.00$ &  $0.09\pm0.01$ 
                                        &  $0.10\pm0.01$ &  $0.07\pm0.01$ &  $0.08\pm0.01$ \\
           LGR   &  $0.03\pm0.01$ &  $0.04\pm0.01$ &  $0.02\pm0.01$ 
                                        &  $0.05\pm0.02$ &  $0.05\pm0.02$ &  $0.04\pm0.02$ \\
          DGR+D   &  $0.06\pm0.02$ &  $0.07\pm0.00$ &  $0.07\pm0.01$ 
                                        &  $0.02\pm0.01$ &  $0.08\pm0.02$ &  $0.02\pm0.02$ \\
         LGR+D   &  [$0.00\pm0.01$] &  $0.01\pm0.00$ &  \cellcolor{gray!25}$\bm{0.00\pm0.00}$ 
                                        &  $0.01\pm0.01$ &  $0.02\pm0.01$ &  [$0.00\pm0.01$] \\

\bottomrule

\end{tabular}

}
\end{table}

\begin{figure}[h!]
    \centering
    \subfloat[\ac{Task-IL} Results following O$2$.\label{fig:fer-task-il-aug-order-1}]{\includegraphics[width=0.5\textwidth]{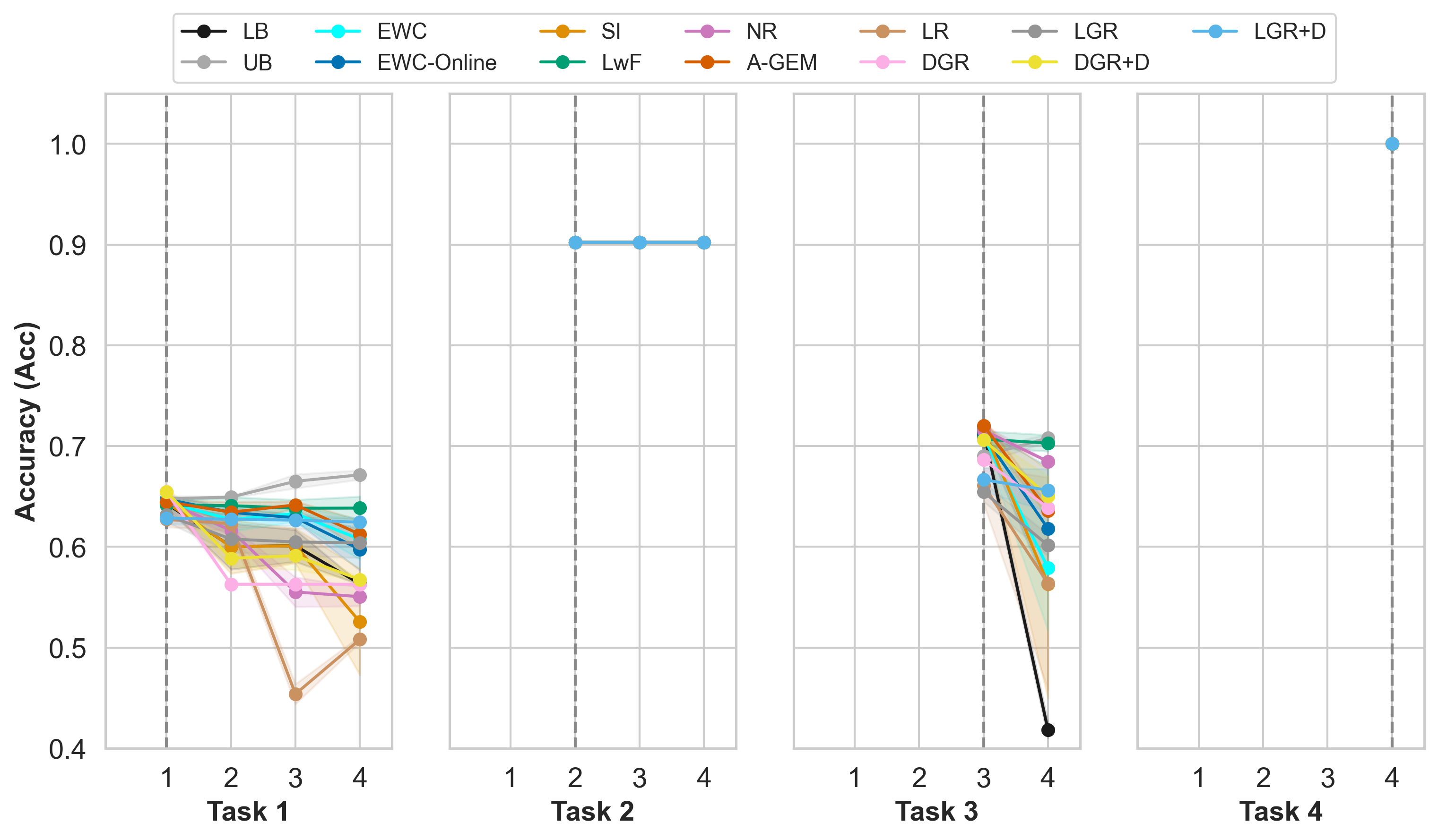}}
    \hfill
    \subfloat[\ac{Task-IL} Results following O$3$.\label{fig:fer-task-il-aug-order-2}]{\includegraphics[width=0.5\textwidth]{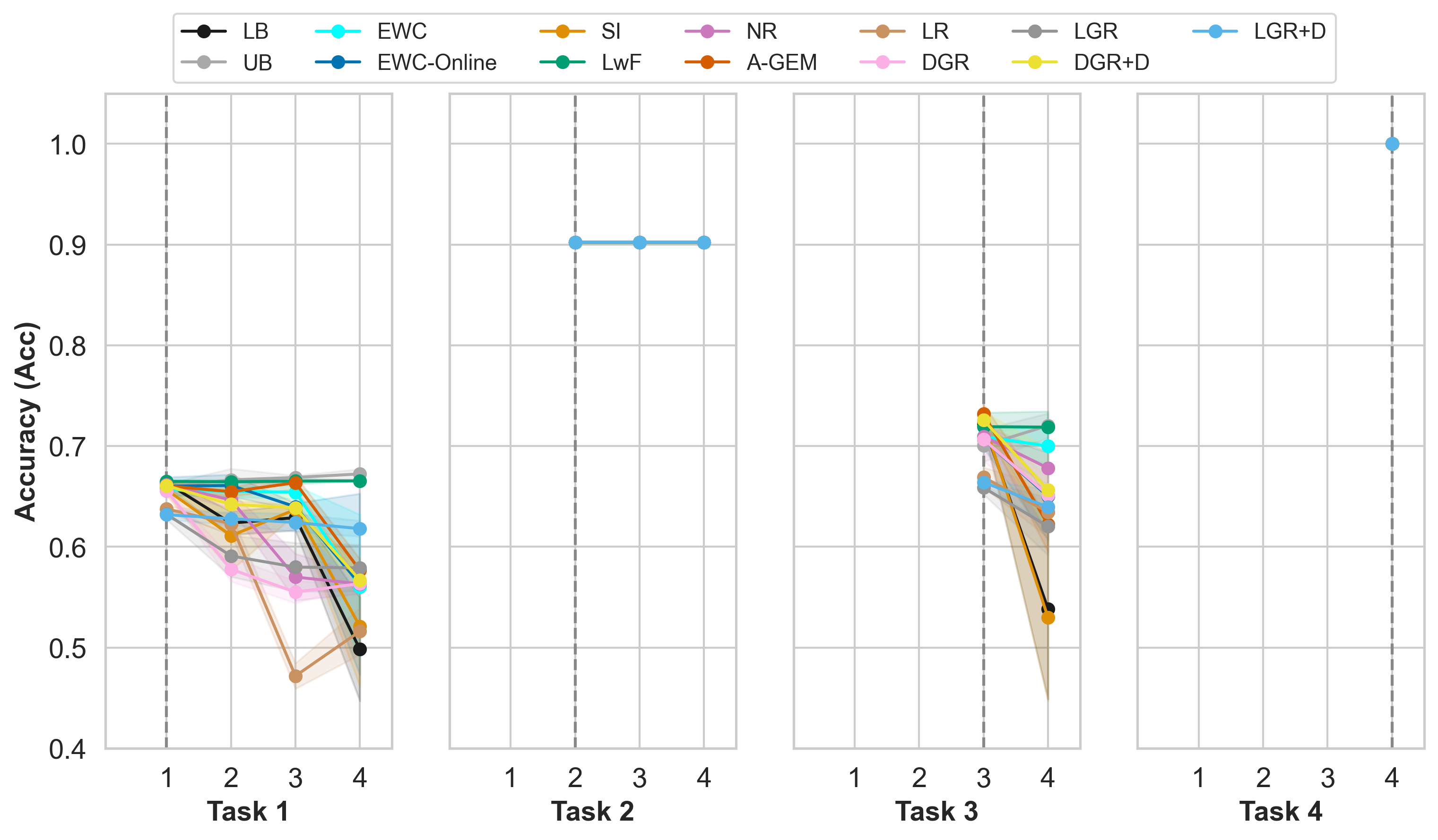}} \\
    \caption[Task-IL Results for FER following Order 2 and Order 3.]{\ac{Task-IL} results for \ac{FER}-2013 following (a) Order $2$ and (b) Order $3$, respectively. For each task, test-set accuracy is shown as learning progresses from when a task is introduced to the end of the overall training procedure.}
    \label{fig:fer-taskil-orders} 
    
\end{figure}
\FloatBarrier
\textit{}
\newpage
\subsubsection{\acf{Class-IL} Results}
\label{app:classilFER}
\begin{table}[h!]
\centering
\setlength\tabcolsep{4.5pt}
\caption[\acs{Class-IL} \acs{Acc} for \acs{FER}-2013  W/ and W/O Data-Augmentation.]{\acs{Class-IL} \acs{Acc} for \acs{FER}-2013  W/ and W/O Data-Augmentation. \textbf{Bold} values denote best (highest) while [\textit{bracketed}] denote second-best values for each column.}
\label{tab:class-il-fer-acc}

{
\scriptsize
\begin{tabular}{l|ccccccc}\toprule

\multirow{2}[2]{*}{\makecell[c]{\textbf{Method}}}           & \multicolumn{7}{c}{\textbf{Accuracy W/O Data-Augmentation}} \\ \cmidrule{2-8}

\multicolumn{1}{c|}{ }                          & 
\multicolumn{1}{c|}{\textbf{Class 1}}            & \multicolumn{1}{c|}{\textbf{Class 2}}          &
\multicolumn{1}{c|}{\textbf{Class 3}}            & \multicolumn{1}{c|}{\textbf{Class 4}}           &
\multicolumn{1}{c|}{\textbf{Class 5}}            & \multicolumn{1}{c|}{\textbf{Class 6}}          &
\multicolumn{1}{c}{\textbf{Class 7}}                     \\ \midrule

\multicolumn{8}{c}{\textbf{Baseline Approaches}} \\ \midrule

                      LB                & \cellcolor{gray!25}$\bm{1.00\pm0.00}$ & $0.50\pm0.00$ & $0.33\pm0.00$ & $0.25\pm0.00$ & $0.20\pm0.00$ & $0.17\pm0.00$ & $0.14\pm0.00$ \\
                      UB                & \cellcolor{gray!25}$\bm{1.00\pm0.00}$ & \cellcolor{gray!25}$\bm{0.67\pm0.01}$ & $0.45\pm0.01$ & [$0.40\pm0.02$] & \cellcolor{gray!25}$\bm{0.40\pm0.02}$ & \cellcolor{gray!25}$\bm{0.36\pm0.03}$ & \cellcolor{gray!25}$\bm{0.37\pm0.04}$ \\\midrule

\multicolumn{8}{c}{\textbf{Regularisation-Based Approaches}} \\ \midrule

 EWC  & \cellcolor{gray!25}$\bm{1.00\pm0.00}$ & $0.50\pm0.00$ & $0.33\pm0.00$ & $0.25\pm0.00$ & $0.20\pm0.00$ & $0.17\pm0.00$ & $0.14\pm0.00$ \\
EWC-Online  & \cellcolor{gray!25}$\bm{1.00\pm0.00}$ & $0.50\pm0.00$ & $0.33\pm0.00$ & $0.25\pm0.00$ & $0.20\pm0.00$ & $0.17\pm0.00$ & $0.14\pm0.00$ \\
         SI  & \cellcolor{gray!25}$\bm{1.00\pm0.00}$ & $0.50\pm0.00$ & $0.33\pm0.00$ & $0.25\pm0.00$ & $0.20\pm0.00$ & $0.17\pm0.00$ & $0.14\pm0.00$ \\
                 LwF  & \cellcolor{gray!25}$\bm{1.00\pm0.00}$ & $0.50\pm0.00$ & $0.33\pm0.00$ & $0.25\pm0.00$ & $0.20\pm0.00$ & $0.17\pm0.00$ & $0.14\pm0.00$ \\ \midrule
\multicolumn{8}{c}{\textbf{Replay-Based Approaches}} \\ \midrule

               NR  & \cellcolor{gray!25}$\bm{1.00\pm0.00}$ & $0.64\pm0.01$ & \cellcolor{gray!25}$\bm{0.58\pm0.01}$ & $0.39\pm0.01$ & $0.31\pm0.02$ & $0.26\pm0.00$ & $0.24\pm0.01$ \\
   A-GEM  & \cellcolor{gray!25}$\bm{1.00\pm0.00}$ & $0.51\pm0.00$ & $0.33\pm0.00$ & $0.25\pm0.00$ & $0.20\pm0.00$ & $0.17\pm0.00$ & $0.14\pm0.00$ \\
       LR  & \cellcolor{gray!25}$\bm{1.00\pm0.00}$ & [$0.65\pm0.01$] & [$0.51\pm0.02$] & \cellcolor{gray!25}$\bm{0.41\pm0.02}$ & [$0.39\pm0.01$] & [$0.34\pm0.01$] & [$0.34\pm0.02$] \\
             DGR  & \cellcolor{gray!25}$\bm{1.00\pm0.00}$ & $0.50\pm0.00$ & $0.34\pm0.00$ & $0.25\pm0.00$ & $0.20\pm0.00$ & $0.17\pm0.00$ & $0.14\pm0.00$ \\
           LGR  & \cellcolor{gray!25}$\bm{1.00\pm0.00}$ & $0.53\pm0.00$ & $0.37\pm0.02$ & $0.26\pm0.01$ & $0.27\pm0.02$ & $0.21\pm0.02$ & $0.20\pm0.02$ \\
          DGR+D & \cellcolor{gray!25}$\bm{1.00\pm0.00}$ & $0.50\pm0.00$ & $0.34\pm0.00$ & $0.25\pm0.00$ & $0.20\pm0.00$ & $0.17\pm0.00$ & $0.14\pm0.00$ \\
         LGR+D  & \cellcolor{gray!25}$\bm{1.00\pm0.00}$ & $0.58\pm0.02$ & $0.43\pm0.03$ & $0.28\pm0.00$ & $0.29\pm0.01$ & $0.22\pm0.03$ & $0.24\pm0.04$ \\

\midrule

\multirow{2}[2]{*}{\makecell[c]{\textbf{Method}}}           & \multicolumn{7}{c}{\textbf{Accuracy W/ Data-Augmentation}} \\ \cmidrule{2-8}

\multicolumn{1}{c|}{ }                          & 
\multicolumn{1}{c|}{\textbf{Class 1}}            & \multicolumn{1}{c|}{\textbf{Class 2}}          &
\multicolumn{1}{c|}{\textbf{Class 3}}            & \multicolumn{1}{c|}{\textbf{Class 4}}           &
\multicolumn{1}{c|}{\textbf{Class 5}}            & \multicolumn{1}{c|}{\textbf{Class 6}}          &
\multicolumn{1}{c}{\textbf{Class 7}}                     \\ \midrule

\multicolumn{8}{c}{\textbf{Baseline Approaches}} \\ \midrule

                      LB                & \cellcolor{gray!25}$\bm{1.00\pm0.00}$ & $0.50\pm0.00$ & $0.33\pm0.00$ & $0.25\pm0.00$ & $0.20\pm0.00$ & $0.17\pm0.00$ & $0.14\pm0.00$ \\
                      UB                & \cellcolor{gray!25}$\bm{1.00\pm0.00}$ & \cellcolor{gray!25}$\bm{0.65\pm0.01}$ & $0.44\pm0.01$ & \cellcolor{gray!25}$\bm{0.36\pm0.01}$ & \cellcolor{gray!25}$\bm{0.35\pm0.00}$ & \cellcolor{gray!25}$\bm{0.32\pm0.00}$ & \cellcolor{gray!25}$\bm{0.32\pm0.01}$ \\ \midrule

\multicolumn{8}{c}{\textbf{Regularisation-Based Approaches}} \\ \midrule

 EWC  & \cellcolor{gray!25}$\bm{1.00\pm0.00}$ & $0.50\pm0.00$ & $0.33\pm0.00$ & $0.25\pm0.00$ & $0.20\pm0.00$ & $0.17\pm0.00$ & $0.14\pm0.00$ \\
EWC-Online  & \cellcolor{gray!25}$\bm{1.00\pm0.00}$ & $0.50\pm0.00$ & $0.33\pm0.00$ & $0.25\pm0.00$ & $0.20\pm0.00$ & $0.17\pm0.00$ & $0.14\pm0.00$ \\
         SI  & \cellcolor{gray!25}$\bm{1.00\pm0.00}$ & $0.50\pm0.00$ & $0.33\pm0.00$ & $0.25\pm0.00$ & $0.20\pm0.00$ & $0.17\pm0.00$ & $0.14\pm0.00$ \\
                 LwF  & \cellcolor{gray!25}$\bm{1.00\pm0.00}$ & $0.50\pm0.00$ & $0.33\pm0.00$ & $0.25\pm0.00$ & $0.20\pm0.00$ & $0.17\pm0.00$ & $0.14\pm0.00$ \\ \midrule
\multicolumn{8}{c}{\textbf{Replay-Based Approaches}} \\ \midrule

               NR  & \cellcolor{gray!25}$\bm{1.00\pm0.00}$ & [$0.63\pm0.01$] & \cellcolor{gray!25}$\bm{0.47\pm0.02}$ & [$0.32\pm0.00$] & $0.24\pm0.01$ & $0.20\pm0.00$ & $0.18\pm0.00$ \\
   A-GEM  & \cellcolor{gray!25}$\bm{1.00\pm0.00}$ & $0.50\pm0.00$ & $0.35\pm0.01$ & $0.25\pm0.00$ & $0.20\pm0.00$ & $0.17\pm0.00$ & $0.14\pm0.00$ \\
       LR  & \cellcolor{gray!25}$\bm{1.00\pm0.00}$ & [$0.63\pm0.00$] & [$0.45\pm0.01$] & \cellcolor{gray!25}$\bm{0.36\pm0.00}$ & [$0.33\pm0.01$] & [$0.28\pm0.01$] & [$0.27\pm0.00$] \\
             DGR  & \cellcolor{gray!25}$\bm{1.00\pm0.00}$ & $0.50\pm0.00$ & $0.34\pm0.00$ & $0.25\pm0.00$ & $0.20\pm0.00$ & $0.17\pm0.00$ & $0.14\pm0.00$ \\
           LGR  & \cellcolor{gray!25}$\bm{1.00\pm0.00}$ & $0.51\pm0.00$ & $0.35\pm0.00$ & $0.26\pm0.01$ & $0.24\pm0.00$ & $0.19\pm0.00$ & $0.17\pm0.01$ \\
          DGR+D & \cellcolor{gray!25}$\bm{1.00\pm0.00}$ & $0.50\pm0.00$ & $0.34\pm0.00$ & $0.25\pm0.00$ & $0.20\pm0.00$ & $0.17\pm0.00$ & $0.14\pm0.00$ \\
         LGR+D  & \cellcolor{gray!25}$\bm{1.00\pm0.00}$ & $0.55\pm0.01$ & $0.37\pm0.00$ & $0.28\pm0.00$ & $0.28\pm0.01$ & $0.20\pm0.01$ & $0.20\pm0.01$ \\

\bottomrule

\end{tabular}
}
\end{table}

\begin{table}[h!]
\centering
\caption[\acs{Class-IL} \acs{CF} scores for \acs{FER}-2013  W/ and W/O Data-Augmentation.]{\acs{CF} scores for \acs{Class-IL} on \ac{FER}-2013  W/ and W/O Data-Augmentation. \textbf{Bold} values denote best (lowest) while [\textit{bracketed}] denote second-best values for each column.}
\label{tab:class-il-fer-cf}

{
\scriptsize

\begin{tabular}{l|cccccc}\toprule

\multirow{2}[2]{*}{\makecell[c]{\textbf{Method}}}           & \multicolumn{6}{c}{\textbf{\acs{CF} W/O Data-Augmentation}} \\ \cmidrule{2-7}

\multicolumn{1}{c|}{ }                          & 
\multicolumn{1}{c|}{\textbf{Class 2}}          &
\multicolumn{1}{c|}{\textbf{Class 3}}            & \multicolumn{1}{c|}{\textbf{Class 4}}           &
\multicolumn{1}{c|}{\textbf{Class 5}}            & \multicolumn{1}{c|}{\textbf{Class 6}}          &
\multicolumn{1}{c}{\textbf{Class 7}}                    \\ \midrule
\multicolumn{7}{c}{\textbf{Baseline Approaches}} \\ \midrule

                LB                      & $1.00\pm0.00$ & $1.00\pm0.00$ & $1.00\pm0.00$ & $1.00\pm0.00$ & $1.00\pm0.00$ & $1.00\pm0.00$ \\
                UB                      & \cellcolor{gray!25}$\bm{0.11\pm0.03}$ & \cellcolor{gray!25}$\bm{0.14\pm0.04}$ & \cellcolor{gray!25}$\bm{0.18\pm0.04}$ & \cellcolor{gray!25}$\bm{0.16\pm0.04}$ & \cellcolor{gray!25}$\bm{0.14\pm0.04}$ & \cellcolor{gray!25}$\bm{0.15\pm0.00}$ \\\midrule

\multicolumn{7}{c}{\textbf{Regularisation-Based Approaches}} \\ \midrule

 EWC  & $1.00\pm0.00$ & $1.00\pm0.00$ & $1.00\pm0.00$ & $1.00\pm0.00$ & $1.00\pm0.00$ & $1.00\pm0.00$ \\
EWC-Online  & $1.00\pm0.00$ & $1.00\pm0.00$ & $1.00\pm0.00$ & $1.00\pm0.00$ & $1.00\pm0.00$ & $1.00\pm0.00$ \\
         SI  & $1.00\pm0.00$ & $1.00\pm0.00$ & $1.00\pm0.00$ & $1.00\pm0.00$ & $1.00\pm0.00$ & $1.00\pm0.00$ \\
                 LwF  & $1.00\pm0.00$ & $1.00\pm0.00$ & $1.00\pm0.00$ & $1.00\pm0.00$ & $1.00\pm0.00$ & $1.00\pm0.00$ \\ \midrule

\multicolumn{7}{c}{\textbf{Replay-Based Approaches}} \\ \midrule

               NR  & $0.36\pm0.04$ & $0.60\pm0.05$ & $0.69\pm0.05$ & $0.74\pm0.03$ & $0.76\pm0.03$ & $0.68\pm0.00$ \\
   A-GEM  & $1.00\pm0.00$ & $0.99\pm0.01$ & $1.00\pm0.00$ & $1.00\pm0.00$ & $1.00\pm0.00$ & $1.00\pm0.00$ \\
       LR  & [$0.35\pm0.01$] & [$0.27\pm0.02$] & [$0.26\pm0.01$] & [$0.23\pm0.01$] & [$0.23\pm0.02$] & [$0.19\pm0.00$] \\
             DGR  & $0.99\pm0.01$ & $0.99\pm0.00$ & $0.99\pm0.00$ & $1.00\pm0.00$ & $1.00\pm0.00$ & $1.00\pm0.00$ \\
           LGR  & $0.88\pm0.02$ & $0.93\pm0.00$ & $0.84\pm0.03$ & $0.86\pm0.04$ & $0.85\pm0.04$ & $0.84\pm0.00$ \\
          DGR+D & $0.99\pm0.00$ & $0.99\pm0.00$ & $0.99\pm0.00$ & $1.00\pm0.00$ & $1.00\pm0.00$ & $1.00\pm0.00$ \\
         LGR+D  & $0.67\pm0.06$ & $0.79\pm0.04$ & $0.70\pm0.03$ & $0.74\pm0.06$ & $0.69\pm0.06$ & $0.65\pm0.00$ \\\midrule

\multirow{2}[2]{*}{\makecell[c]{\textbf{Method}}}          & \multicolumn{6}{c}{\textbf{\acs{CF} W/ Data-Augmentation}} \\ \cmidrule{2-7}

\multicolumn{1}{c|}{ }                          & 
\multicolumn{1}{c|}{\textbf{Class 2}}          &
\multicolumn{1}{c|}{\textbf{Class 3}}            & \multicolumn{1}{c|}{\textbf{Class 4}}           &
\multicolumn{1}{c|}{\textbf{Class 5}}            & \multicolumn{1}{c|}{\textbf{Class 6}}          &
\multicolumn{1}{c}{\textbf{Class 7}}                    \\ \midrule

\multicolumn{7}{c}{\textbf{Baseline Approaches}} \\ \midrule

                      LB                      & $1.00\pm0.00$ & $1.00\pm0.00$ & $1.00\pm0.00$ & $1.00\pm0.00$ & $1.00\pm0.00$ & $1.00\pm0.00$ \\
                      UB                      & \cellcolor{gray!25}$\bm{0.09\pm0.02}$ & \cellcolor{gray!25}$\bm{0.15\pm0.01}$ & \cellcolor{gray!25}$\bm{0.21\pm0.01}$ & \cellcolor{gray!25}$\bm{0.18\pm0.01}$ & \cellcolor{gray!25}$\bm{0.18\pm0.01}$ & \cellcolor{gray!25}$\bm{0.15\pm0.00}$ \\\midrule

\multicolumn{7}{c}{\textbf{Regularisation-Based Approaches}} \\ \midrule

 EWC  & $1.00\pm0.00$ & $1.00\pm0.00$ & $1.00\pm0.00$ & $1.00\pm0.00$ & $1.00\pm0.00$ & $1.00\pm0.00$ \\
EWC-Online  & $1.00\pm0.00$ & $1.00\pm0.00$ & $1.00\pm0.00$ & $1.00\pm0.00$ & $1.00\pm0.00$ & $1.00\pm0.00$ \\
         SI  & $1.00\pm0.00$ & $1.00\pm0.00$ & $1.00\pm0.00$ & $1.00\pm0.00$ & $1.00\pm0.00$ & $1.00\pm0.00$ \\
                 LwF  & $1.00\pm0.00$ & $1.00\pm0.00$ & $1.00\pm0.00$ & $1.00\pm0.00$ & $1.00\pm0.00$ & $1.00\pm0.00$ \\ \midrule

\multicolumn{7}{c}{\textbf{Replay-Based Approaches}} \\ \midrule

               NR  & $0.62\pm0.03$ & $0.78\pm0.01$ & $0.84\pm0.02$ & $0.86\pm0.01$ & $0.87\pm0.01$ & $0.80\pm0.00$ \\
   A-GEM  & $0.97\pm0.02$ & $1.00\pm0.00$ & $1.00\pm0.00$ & $1.00\pm0.00$ & $1.00\pm0.00$ & $1.00\pm0.00$ \\
       LR  & [$0.34\pm0.03$] & [$0.30\pm0.02$] & [$0.27\pm0.01$] & [$0.24\pm0.01$] & [$0.25\pm0.03$] & [$0.22\pm0.00$] \\
             DGR  & $1.00\pm0.00$ & $0.99\pm0.00$ & $1.00\pm0.00$ & $1.00\pm0.00$ & $0.99\pm0.00$ & $1.00\pm0.00$ \\
           LGR  & $0.88\pm0.02$ & $0.87\pm0.04$ & $0.82\pm0.04$ & $0.82\pm0.04$ & $0.81\pm0.05$ & $0.80\pm0.00$ \\
          DGR+D & $0.99\pm0.01$ & $0.99\pm0.02$ & $0.99\pm0.01$ & $0.99\pm0.01$ & $0.99\pm0.01$ & $1.00\pm0.00$ \\
         LGR+D  & $0.66\pm0.03$ & $0.68\pm0.03$ & $0.61\pm0.02$ & $0.65\pm0.03$ & $0.63\pm0.03$ & $0.60\pm0.00$ \\

\bottomrule
\end{tabular}

}
\end{table}

\begin{figure}[h!]
    \centering
    \subfloat[\ac{Class-IL} Results w/o Augmentation.\label{fig:fer-class-il-noaug}]{\includegraphics[width=0.5\textwidth]{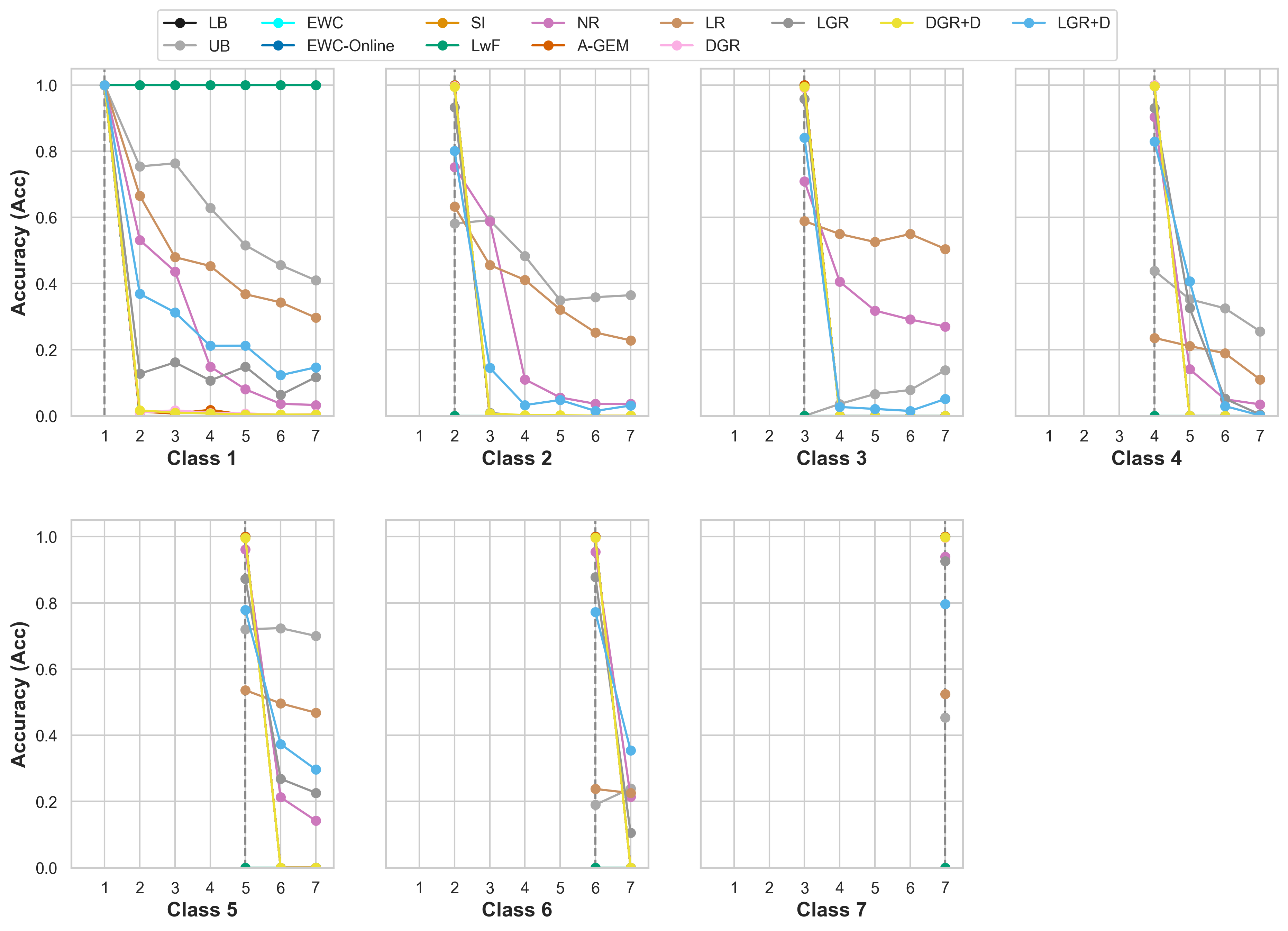}} 
    \hfill
    \subfloat[\ac{Class-IL} Results w/ Augmentation.\label{fig:fer-class-il-aug}]{\includegraphics[width=0.5\textwidth]{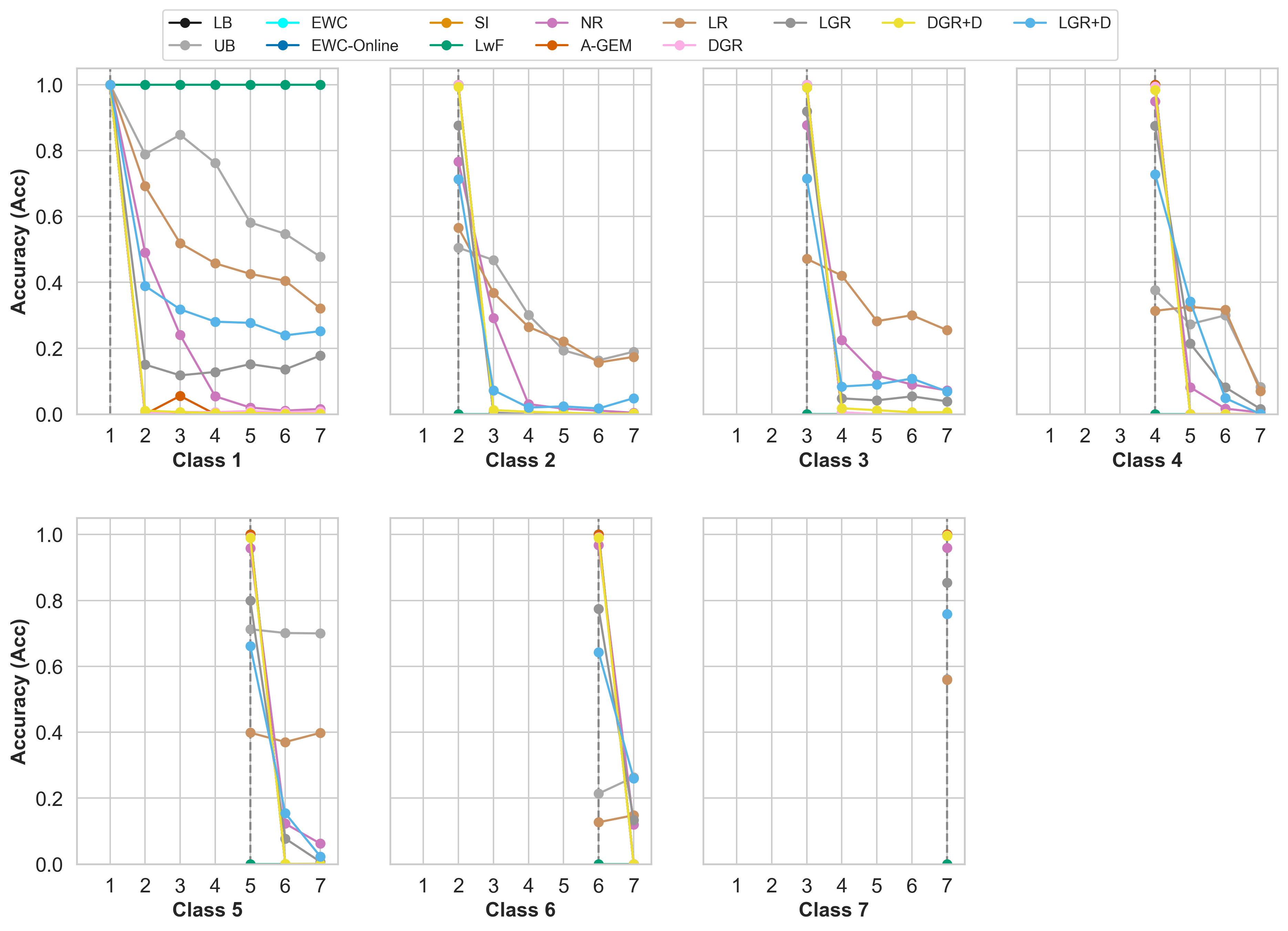}} 
    \caption[Class-IL Results for FER W/ and W/O Augmentation.]{\acs{Class-IL} results for \ac{FER}-2013 (a) without and (b) with augmentation.  For each class, test-set accuracy is shown as the learning progresses from when the class is introduced to the end of the overall training procedure.}
    \label{fig:fer-classil}

\end{figure}

\FloatBarrier
\normalsize
\subsubsection{Class-Ordering}
\FloatBarrier
\label{app:classorderFER}

\begin{table}[h!]
\centering
\setlength\tabcolsep{3.5pt}

\caption[\acs{Class-IL} \acs{Acc} for \acs{FER}-2013 following Orderings O2 and O3.]{\acs{Class-IL} \acs{Acc} for \acs{FER}-2013 W/ Data-Augmentation following Orderings O2 and O3. \textbf{Bold} values denote best (highest) while [\textit{bracketed}] denote second-best values for each column.}

\label{tab:class-il-fer-acc-orders}

{
\scriptsize
\begin{tabular}{l|ccccccc}\toprule

\multirow{2}[2]{*}{\makecell[c]{\textbf{Method}}}           & \multicolumn{7}{c}{\textbf{Accuracy following O2}} \\ \cmidrule{2-8}

\multicolumn{1}{c|}{ }                          & 
\multicolumn{1}{c|}{\textbf{Class 1}}            & \multicolumn{1}{c|}{\textbf{Class 2}}          &
\multicolumn{1}{c|}{\textbf{Class 3}}            & \multicolumn{1}{c|}{\textbf{Class 4}}           &
\multicolumn{1}{c|}{\textbf{Class 5}}            & \multicolumn{1}{c|}{\textbf{Class 6}}          &
\multicolumn{1}{c|}{\textbf{Class 7}}                      \\ \midrule
\multicolumn{8}{c}{\textbf{Baseline Approaches}} \\ \midrule

LB                & \cellcolor{gray!25}$\bm{1.00\pm0.00}$ & $0.50\pm0.00$ & $0.33\pm0.00$ & $0.25\pm0.00$ & $0.20\pm0.00$ & $0.17\pm0.00$ & $0.14\pm0.00$ \\
UB                & \cellcolor{gray!25}$\bm{1.00\pm0.00}$ & \cellcolor{gray!25}$\bm{0.64\pm0.01}$ & $0.43\pm0.01$ & [$0.35\pm0.02$] & \cellcolor{gray!25}$\bm{0.34\pm0.02}$ & \cellcolor{gray!25}$\bm{0.30\pm0.02}$ & \cellcolor{gray!25}$\bm{0.30\pm0.03}$ \\\midrule

\multicolumn{8}{c}{\textbf{Regularisation-Based Approaches}} \\ \midrule

EWC  & \cellcolor{gray!25}$\bm{1.00\pm0.00}$ & $0.50\pm0.00$ & $0.33\pm0.00$ & $0.25\pm0.00$ & $0.20\pm0.00$ & $0.17\pm0.00$ & $0.14\pm0.00$ \\
EWC-Online  & \cellcolor{gray!25}$\bm{1.00\pm0.00}$ & $0.50\pm0.00$ & $0.33\pm0.00$ & $0.25\pm0.00$ & $0.20\pm0.00$ & $0.17\pm0.00$ & $0.14\pm0.00$ \\
SI  & \cellcolor{gray!25}$\bm{1.00\pm0.00}$ & $0.50\pm0.00$ & $0.33\pm0.00$ & $0.25\pm0.00$ & $0.20\pm0.00$ & $0.17\pm0.00$ & $0.14\pm0.00$ \\
LwF  & \cellcolor{gray!25}$\bm{1.00\pm0.00}$ & $0.50\pm0.00$ & $0.33\pm0.00$ & $0.25\pm0.00$ & $0.20\pm0.00$ & $0.17\pm0.00$ & $0.14\pm0.00$ \\\midrule

\multicolumn{8}{c}{\textbf{Replay-Based Approaches}} \\ \midrule

NR  & \cellcolor{gray!25}$\bm{1.00\pm0.00}$ & $0.62\pm0.02$ & \cellcolor{gray!25}$\bm{0.47\pm0.00}$ & $0.32\pm0.00$ & $0.25\pm0.01$ & $0.22\pm0.00$ & $0.19\pm0.00$ \\
A-GEM  & \cellcolor{gray!25}$\bm{1.00\pm0.00}$ & $0.50\pm0.00$ & $0.36\pm0.02$ & $0.25\pm0.00$ & $0.20\pm0.00$ & $0.17\pm0.00$ & $0.14\pm0.00$ \\
LR  & \cellcolor{gray!25}$\bm{1.00\pm0.00}$ & [$0.63\pm0.00$] & [$0.46\pm0.00$] & \cellcolor{gray!25}$\bm{0.36\pm0.00}$ & [$0.32\pm0.01$] & [$0.28\pm0.01$] & [$0.28\pm0.01$] \\
DGR  & \cellcolor{gray!25}$\bm{1.00\pm0.00}$ & $0.50\pm0.00$ & $0.34\pm0.00$ & $0.25\pm0.00$ & $0.20\pm0.00$ & $0.17\pm0.00$ & $0.15\pm0.00$ \\
LGR  & \cellcolor{gray!25}$\bm{1.00\pm0.00}$ & $0.51\pm0.01$ & $0.35\pm0.00$ & $0.26\pm0.00$ & $0.25\pm0.00$ & $0.19\pm0.01$ & $0.17\pm0.01$ \\
DGR+D  & \cellcolor{gray!25}$\bm{1.00\pm0.00}$ & $0.50\pm0.00$ & $0.34\pm0.00$ & $0.25\pm0.00$ & $0.20\pm0.00$ & $0.17\pm0.00$ & $0.15\pm0.00$ \\
LGR+D  & \cellcolor{gray!25}$\bm{1.00\pm0.00}$ & $0.56\pm0.01$ & $0.37\pm0.02$ & $0.27\pm0.00$ & $0.26\pm0.02$ & $0.20\pm0.02$ & $0.20\pm0.01$ \\\midrule

\multirow{2}[2]{*}{\makecell[c]{\textbf{Method}}}           & \multicolumn{7}{c}{\textbf{Accuracy following O3}} \\ \cmidrule{2-8}

\multicolumn{1}{c|}{ }                          & 
\multicolumn{1}{c|}{\textbf{Class 1}}            & \multicolumn{1}{c|}{\textbf{Class 2}}          &
\multicolumn{1}{c|}{\textbf{Class 3}}            & \multicolumn{1}{c|}{\textbf{Class 4}}           &
\multicolumn{1}{c|}{\textbf{Class 5}}            & \multicolumn{1}{c|}{\textbf{Class 6}}          &
\multicolumn{1}{c|}{\textbf{Class 7}}                    \\ \midrule
\multicolumn{8}{c}{\textbf{Baseline Approaches}} \\ \midrule

LB                & \cellcolor{gray!25}$\bm{1.00\pm0.00}$ & $0.50\pm0.00$ & $0.33\pm0.00$ & $0.25\pm0.00$ & $0.20\pm0.00$ & $0.17\pm0.00$ & $0.14\pm0.00$ \\
UB                & \cellcolor{gray!25}$\bm{1.00\pm0.00}$ & \cellcolor{gray!25}$\bm{0.65\pm0.00}$ & $0.44\pm0.01$ & \cellcolor{gray!25}$\bm{0.36\pm0.00}$ & \cellcolor{gray!25}$\bm{0.33\pm0.00}$ & \cellcolor{gray!25}$\bm{0.30\pm0.00}$ & \cellcolor{gray!25}$\bm{0.33\pm0.00}$ \\\midrule

\multicolumn{8}{c}{\textbf{Regularisation-Based Approaches}} \\ \midrule

EWC  & \cellcolor{gray!25}$\bm{1.00\pm0.00}$ & $0.50\pm0.00$ & $0.33\pm0.00$ & $0.25\pm0.00$ & $0.20\pm0.00$ & $0.17\pm0.00$ & $0.14\pm0.00$ \\
EWC-Online  & \cellcolor{gray!25}$\bm{1.00\pm0.00}$ & $0.50\pm0.00$ & $0.33\pm0.00$ & $0.25\pm0.00$ & $0.20\pm0.00$ & $0.17\pm0.00$ & $0.14\pm0.00$ \\
SI  & \cellcolor{gray!25}$\bm{1.00\pm0.00}$ & $0.50\pm0.00$ & $0.33\pm0.00$ & $0.25\pm0.00$ & $0.20\pm0.00$ & $0.17\pm0.00$ & $0.14\pm0.00$ \\
LwF  & \cellcolor{gray!25}$\bm{1.00\pm0.00}$ & $0.50\pm0.00$ & $0.33\pm0.00$ & $0.25\pm0.00$ & $0.20\pm0.00$ & $0.17\pm0.00$ & $0.14\pm0.00$ \\\midrule
\multicolumn{8}{c}{\textbf{Replay-Based Approaches}} \\ \midrule

NR  & \cellcolor{gray!25}$\bm{1.00\pm0.00}$ & [$0.63\pm0.01$] & \cellcolor{gray!25}$\bm{0.47\pm0.01}$ & [$0.31\pm0.02$] & $0.23\pm0.01$ & $0.19\pm0.01$ & $0.18\pm0.00$ \\
A-GEM  & \cellcolor{gray!25}$\bm{1.00\pm0.00}$ & $0.50\pm0.00$ & $0.35\pm0.00$ & $0.25\pm0.00$ & $0.20\pm0.00$ & $0.17\pm0.00$ & $0.14\pm0.00$ \\
LR  & \cellcolor{gray!25}$\bm{1.00\pm0.00}$ & [$0.63\pm0.01$] & [$0.45\pm0.03$] & \cellcolor{gray!25}$\bm{0.36\pm0.02}$ & \cellcolor{gray!25}$\bm{0.33\pm0.03}$ & [$0.29\pm0.02$] & [$0.28\pm0.02$] \\
DGR  & \cellcolor{gray!25}$\bm{1.00\pm0.00}$ & $0.50\pm0.00$ & $0.34\pm0.00$ & $0.25\pm0.00$ & $0.20\pm0.00$ & $0.17\pm0.00$ & $0.15\pm0.00$ \\
LGR  & \cellcolor{gray!25}$\bm{1.00\pm0.00}$ & $0.52\pm0.01$ & $0.34\pm0.01$ & $0.25\pm0.00$ & $0.23\pm0.02$ & $0.18\pm0.01$ & $0.16\pm0.02$ \\
DGR+D  & \cellcolor{gray!25}$\bm{1.00\pm0.00}$ & $0.50\pm0.00$ & $0.34\pm0.00$ & $0.25\pm0.00$ & $0.20\pm0.00$ & $0.17\pm0.00$ & $0.14\pm0.00$ \\
LGR+D  & \cellcolor{gray!25}$\bm{1.00\pm0.00}$ & $0.55\pm0.01$ & $0.35\pm0.02$ & $0.27\pm0.02$ & [$0.25\pm0.03$] & $0.19\pm0.01$ & $0.19\pm0.02$ \\

\bottomrule

\end{tabular}

}
\end{table}

\begin{table}[h!]
\centering
\caption[\acs{Class-IL} \acs{CF} scores for \ac{FER}-2013  following O2 and O3.]{\acs{Class-IL} \acs{CF} scores for \ac{FER}-2013 W/ Data-Augmentation following Orderings O2 and O3. \textbf{Bold} values denote best (lowest) while [\textit{bracketed}] denote second-best values for each column.}

\label{tab:class-il-fer-cf-orders}

{
\scriptsize

\begin{tabular}{l|cccccc}\toprule

\multirow{2}[2]{*}{\makecell[c]{\textbf{Method}}}           & \multicolumn{6}{c}{\textbf{\acs{CF} following O2}} \\ \cmidrule{2-7}

\multicolumn{1}{c|}{ }                          & 
\multicolumn{1}{c|}{\textbf{Class 2}}          &
\multicolumn{1}{c|}{\textbf{Class 3}}            & \multicolumn{1}{c|}{\textbf{Class 4}}           &
\multicolumn{1}{c|}{\textbf{Class 5}}            & \multicolumn{1}{c|}{\textbf{Class 6}}          &
\multicolumn{1}{c}{\textbf{Class 7}}                    \\ \midrule

\multicolumn{7}{c}{\textbf{Baseline Approaches}} \\ \midrule

                      LB & $1.00\pm0.00$ & $1.00\pm0.00$ & $1.00\pm0.00$ & $1.00\pm0.00$ & $1.00\pm0.00$ & $1.00\pm0.00$ \\
                      UB & \cellcolor{gray!25}$\bm{0.12\pm0.04}$ & \cellcolor{gray!25}$\bm{0.16\pm0.05}$ & \cellcolor{gray!25}$\bm{0.23\pm0.06}$ & \cellcolor{gray!25}$\bm{0.20\pm0.05}$ & \cellcolor{gray!25}$\bm{0.19\pm0.04}$ & \cellcolor{gray!25}$\bm{0.15\pm0.00}$ \\\midrule

\multicolumn{7}{c}{\textbf{Regularisation-Based Approaches}} \\ \midrule

 EWC  & $1.00\pm0.00$ & $1.00\pm0.00$ & $1.00\pm0.00$ & $1.00\pm0.00$ & $1.00\pm0.00$ & $1.00\pm0.00$ \\
EWC-Online  & $1.00\pm0.00$ & $1.00\pm0.00$ & $1.00\pm0.00$ & $1.00\pm0.00$ & $1.00\pm0.00$ & $1.00\pm0.00$ \\
         SI  & $1.00\pm0.00$ & $1.00\pm0.00$ & $1.00\pm0.00$ & $1.00\pm0.00$ & $1.00\pm0.00$ & $1.00\pm0.00$ \\
                 LwF  & $1.00\pm0.00$ & $1.00\pm0.00$ & $1.00\pm0.00$ & $1.00\pm0.00$ & $1.00\pm0.00$ & $1.00\pm0.00$ \\\midrule

\multicolumn{7}{c}{\textbf{Replay-Based Approaches}} \\ \midrule

               NR  & $0.60\pm0.05$ & $0.73\pm0.05$ & $0.76\pm0.08$ & $0.77\pm0.09$ & $0.78\pm0.09$ & $0.80\pm0.00$ \\
   A-GEM  & $0.96\pm0.04$ & $1.00\pm0.00$ & $1.00\pm0.00$ & $1.00\pm0.00$ & $1.00\pm0.00$ & $1.00\pm0.00$ \\
       LR  & [$0.34\pm0.05$] & [$0.30\pm0.05$] & [$0.30\pm0.04$] & [$0.25\pm0.03$] & [$0.25\pm0.03$] & [$0.25\pm0.00$] \\
             DGR  & $0.99\pm0.01$ & $0.99\pm0.01$ & $0.99\pm0.01$ & $0.99\pm0.01$ & $0.98\pm0.01$ & $1.00\pm0.00$ \\
           LGR  & $0.89\pm0.04$ & $0.89\pm0.04$ & $0.80\pm0.03$ & $0.82\pm0.04$ & $0.82\pm0.05$ & $0.85\pm0.00$ \\
          DGR+D  & $0.98\pm0.01$ & $0.99\pm0.01$ & $0.98\pm0.01$ & $0.99\pm0.01$ & $0.98\pm0.01$ & $1.00\pm0.00$ \\
         LGR+D  & $0.69\pm0.06$ & $0.72\pm0.04$ & $0.65\pm0.03$ & $0.68\pm0.03$ & $0.67\pm0.03$ & $0.62\pm0.00$ \\\midrule

\multirow{2}[2]{*}{\makecell[c]{\textbf{Method}}}           & \multicolumn{6}{c}{\textbf{\acs{CF} following O3}} \\ \cmidrule{2-7}

\multicolumn{1}{c|}{ }                          & 
\multicolumn{1}{c|}{\textbf{Class 2}}          &
\multicolumn{1}{c|}{\textbf{Class 3}}            & \multicolumn{1}{c|}{\textbf{Class 4}}           &
\multicolumn{1}{c|}{\textbf{Class 5}}            & \multicolumn{1}{c|}{\textbf{Class 6}}          &
\multicolumn{1}{c}{\textbf{Class 7}}                     \\ \midrule

\multicolumn{7}{c}{\textbf{Baseline Approaches}} \\ \midrule

                      LB & $1.00\pm0.00$ & $1.00\pm0.00$ & $1.00\pm0.00$ & $1.00\pm0.00$ & $1.00\pm0.00$ & $0.00\pm0.00$ \\
                      UB & \cellcolor{gray!25}$\bm{0.09\pm0.04}$ & \cellcolor{gray!25}$\bm{0.13\pm0.03}$ & \cellcolor{gray!25}$\bm{0.24\pm0.03}$ & \cellcolor{gray!25}$\bm{0.20\pm0.02}$ & \cellcolor{gray!25}$\bm{0.18\pm0.02}$ & \cellcolor{gray!25}$\bm{0.15\pm0.00}$ \\\midrule

\multicolumn{7}{c}{\textbf{Regularisation-Based Approaches}} \\ \midrule

 EWC  & $1.00\pm0.00$ & $1.00\pm0.00$ & $1.00\pm0.00$ & $1.00\pm0.00$ & $1.00\pm0.00$ & $1.00\pm0.00$ \\
EWC-Online  & $1.00\pm0.00$ & $1.00\pm0.00$ & $1.00\pm0.00$ & $1.00\pm0.00$ & $1.00\pm0.00$ & $1.00\pm0.00$ \\
         SI  & $1.00\pm0.00$ & $1.00\pm0.00$ & $1.00\pm0.00$ & $1.00\pm0.00$ & $1.00\pm0.00$ & $1.00\pm0.00$ \\
                 LwF  & $1.00\pm0.00$ & $1.00\pm0.00$ & $1.00\pm0.00$ & $1.00\pm0.00$ & $1.00\pm0.00$ & $1.00\pm0.00$ \\\midrule

\multicolumn{7}{c}{\textbf{Replay-Based Approaches}} \\ \midrule

               NR  & $0.60\pm0.05$ & $0.77\pm0.04$ & $0.83\pm0.02$ & $0.86\pm0.03$ & $0.87\pm0.02$ & $0.85\pm0.00$ \\
   A-GEM  & $0.98\pm0.01$ & $1.00\pm0.00$ & $1.00\pm0.00$ & $1.00\pm0.00$ & $1.00\pm0.00$ & $1.00\pm0.00$ \\
       LR  & [$0.38\pm0.05$] & [$0.30\pm0.01$] & [$0.29\pm0.02$] & [$0.24\pm0.01$] & [$0.25\pm0.02$] & [$0.24\pm0.00$] \\
             DGR  & $0.99\pm0.00$ & $0.99\pm0.00$ & $0.99\pm0.01$ & $0.99\pm0.00$ & $0.99\pm0.01$ & $1.00\pm0.00$ \\
           LGR  & $0.92\pm0.04$ & $0.92\pm0.04$ & $0.88\pm0.07$ & $0.89\pm0.07$ & $0.88\pm0.08$ & $0.85\pm0.00$ \\
          DGR+D  & $0.98\pm0.01$ & $0.99\pm0.01$ & $0.99\pm0.01$ & $0.99\pm0.01$ & $0.99\pm0.01$ & $1.00\pm0.00$ \\
         LGR+D  & $0.74\pm0.11$ & $0.75\pm0.12$ & $0.72\pm0.13$ & $0.74\pm0.12$ & $0.73\pm0.12$ & $0.70\pm0.00$ \\

\bottomrule

\end{tabular}

}
\end{table}

\begin{figure}[h!]
    \centering
    \subfloat[\ac{Class-IL} Results following O$2$.\label{fig:fer-Class-il-aug-order-1}]{\includegraphics[width=0.5\textwidth]{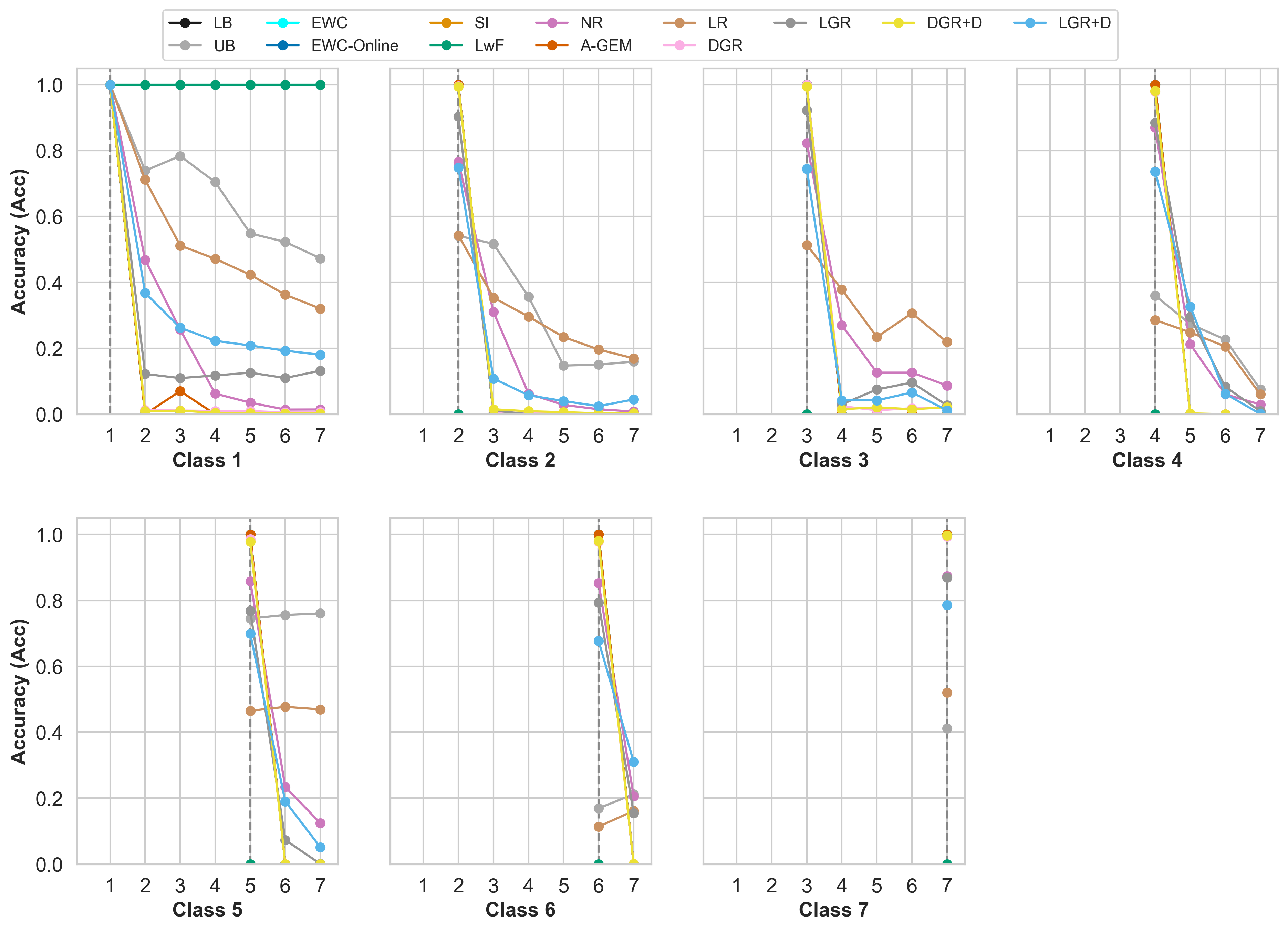}}
    \hfill
    \subfloat[\ac{Class-IL} Results following O$3$.\label{fig:fer-Class-il-aug-order-2}]{\includegraphics[width=0.5\textwidth]{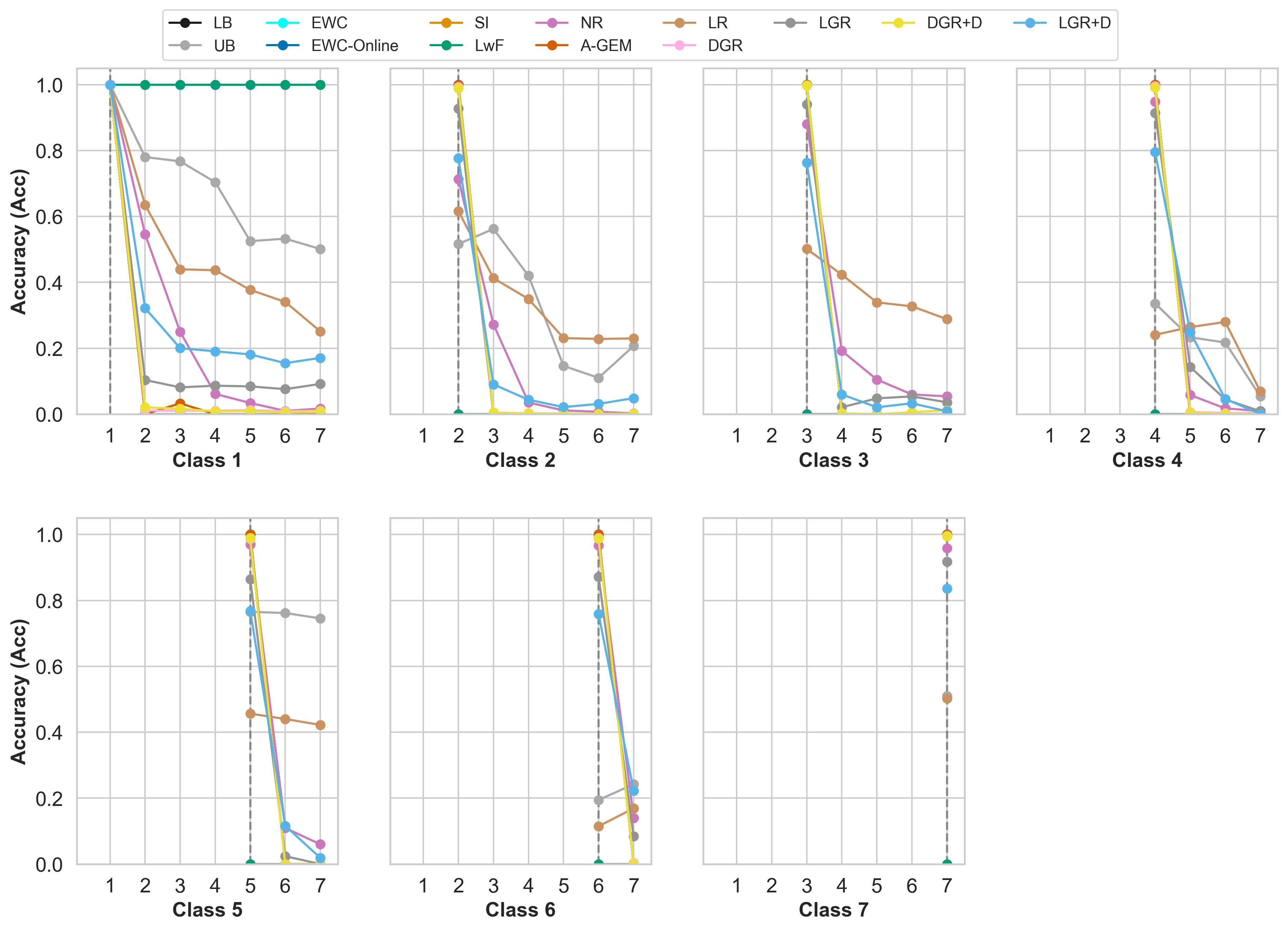}} \\
    \caption[Class-IL Results for FER following Order 2 and Order 3.]{\ac{Class-IL} results for \ac{FER}-2013 following (a) Order $2$ and (b) Order $3$, respectively. For each class, test-set accuracy is shown as the learning progresses from when the class is introduced to the end of the overall training procedure.}
    \label{fig:fer-classil-orders} 
    
\end{figure}
\FloatBarrier
\normalsize

\subsection{iCV Multi-Emotion Facial Expression Dataset}
The \acs{iCV-MEFED} dataset~\citep{Guo2018icv} consists of $\approx 31K$ facial images from $125$ subjects, captured in lab-controlled settings. The dataset is annotated for $8$ different expression classes, namely, \textit{Anger, Surprise, Fearful, Disgust, Happy, Sad, Contempt} and \textit{Neutral}. For each image, annotations are provided for \textit{dominant} (primary) and \textit{complementary} (secondary) expressions. We split the data into the $8$ classes on the basis of dominant class labels, combining the complementary labels to enhance data variability under each class. The training and test-split provided by the authors of the dataset is followed. 

\subsubsection{\acf{Task-IL} Results}
\label{app:taskilicv}
\begin{table}[h!]
\centering
\setlength\tabcolsep{4.5pt}

\caption[\acs{Task-IL} \acs{Acc} for \acs{iCV-MEFED}  W/ and W/O Data-Augmentation.]{\acs{Task-IL} \acs{Acc} for \acs{iCV-MEFED} W/ and W/O Data-Augmentation. \textbf{Bold} values denote best (highest) while [\textit{bracketed}] denote second-best values for each column.}
\label{tab:task-il-icvMefed-acc}

{
\scriptsize
\begin{tabular}{l|cccc|cccc}\toprule
\multicolumn{1}{c|}{\textbf{Method}}            & \multicolumn{4}{c|}{\textbf{Accuracy W/O Data-Augmentation}} 
& \multicolumn{4}{c}{\textbf{Accuracy W/ Data-Augmentation}} \\ \midrule

\multicolumn{1}{c|}{ }                          & 
\multicolumn{1}{c|}{\textbf{Task 1}}            & \multicolumn{1}{c|}{\textbf{Task 2}}          &
\multicolumn{1}{c|}{\textbf{Task 3}}            & \multicolumn{1}{c|}{\textbf{Task 4}}           &
\multicolumn{1}{c|}{\textbf{Task 1}}            & \multicolumn{1}{c|}{\textbf{Task 2}}          &
\multicolumn{1}{c|}{\textbf{Task 3}}            & \multicolumn{1}{c}{\textbf{Task 4}}          \\ \cmidrule{2-9}

\multicolumn{9}{c}{\textbf{Baselines Approaches}} \\ \midrule

LB & $0.85\pm0.01$ & $0.54\pm0.04$ & $0.57\pm0.00$ & $0.55\pm0.02$ & \cellcolor{gray!25}$\bm{0.88\pm0.00}$ & $0.68\pm0.01$ & $0.65\pm0.01$ & $0.60\pm0.03$ \\
UB & $0.84\pm0.02$ & [$0.70\pm0.01$] & \cellcolor{gray!25}$\bm{0.70\pm0.01}$ & $0.68\pm0.01$ & \cellcolor{gray!25}$\bm{0.88\pm0.00}$ & [$0.70\pm0.00$] & \cellcolor{gray!25}$\bm{0.71\pm0.00}$ & $0.68\pm0.01$ \\ \midrule

\multicolumn{9}{c}{\textbf{Regularisation-Based Approaches}} \\ \midrule

EWC  & $0.86\pm0.01$ & $0.64\pm0.03$ & $0.62\pm0.02$ & $0.66\pm0.02$ & \cellcolor{gray!25}$\bm{0.88\pm0.00}$ & $0.68\pm0.02$ & $0.67\pm0.04$ & $0.66\pm0.01$ \\
EWC-Online  & $0.86\pm0.01$ & $0.63\pm0.03$ & $0.59\pm0.01$ & $0.66\pm0.01$ & \cellcolor{gray!25}$\bm{0.88\pm0.00}$ & $0.68\pm0.02$ & $0.67\pm0.03$ & $0.64\pm0.03$ \\
SI  & $0.86\pm0.00$ & $0.54\pm0.04$ & $0.56\pm0.02$ & $0.63\pm0.00$ & \cellcolor{gray!25}$\bm{0.88\pm0.00}$ & [$0.70\pm0.01$] & $0.68\pm0.02$ & $0.64\pm0.04$ \\
LwF  & $0.86\pm0.00$ & [$0.70\pm0.00$] & \cellcolor{gray!25}$\bm{0.70\pm0.01}$ & [$0.69\pm0.00$] & \cellcolor{gray!25}$\bm{0.88\pm0.00}$ & [$0.70\pm0.00$] & \cellcolor{gray!25}$\bm{0.71\pm0.00}$ & [$0.69\pm0.00$] \\\midrule
\multicolumn{9}{c}{\textbf{Replay-Based Approaches}} \\ \midrule

NR  & $0.86\pm0.00$ & [$0.70\pm0.01$] & $0.67\pm0.02$ & $0.62\pm0.01$ & \cellcolor{gray!25}$\bm{0.88\pm0.00}$ & [$0.70\pm0.00$] & $0.65\pm0.01$ & $0.61\pm0.02$ \\
A-GEM  & $0.86\pm0.01$ & [$0.70\pm0.00$] & \cellcolor{gray!25}$\bm{0.70\pm0.00}$ & \cellcolor{gray!25}$\bm{0.70\pm0.00}$ & \cellcolor{gray!25}$\bm{0.88\pm0.00}$ & \cellcolor{gray!25}$\bm{0.71\pm0.00}$ & \cellcolor{gray!25}$\bm{0.71\pm0.00}$ & \cellcolor{gray!25}$\bm{0.70\pm0.01}$ \\
LR  & $0.87\pm0.00$ & [$0.70\pm0.01$] & $0.67\pm0.00$ & $0.63\pm0.01$ & \cellcolor{gray!25}$\bm{0.88\pm0.00}$ & [$0.70\pm0.02$] & $0.68\pm0.01$ & $0.61\pm0.02$ \\
DGR  & [$0.87\pm0.00$] & \cellcolor{gray!25}$\bm{0.71\pm0.00}$ & [$0.69\pm0.00$] & $0.67\pm0.00$ & \cellcolor{gray!25}$\bm{0.88\pm0.00}$ & [$0.70\pm0.01$] & [$0.70\pm0.00$] & $0.64\pm0.01$ \\
LGR  & \cellcolor{gray!25}$\bm{0.88\pm0.00}$ & [$0.70\pm0.00$] & \cellcolor{gray!25}$\bm{0.70\pm0.01}$ & $0.68\pm0.00$ & \cellcolor{gray!25}$\bm{0.88\pm0.00}$ & $0.69\pm0.00$ & $0.62\pm0.00$ & $0.59\pm0.00$ \\
DGR+D  & [$0.87\pm0.00$] & \cellcolor{gray!25}$\bm{0.71\pm0.00}$ & [$0.69\pm0.01$] & $0.68\pm0.00$ & \cellcolor{gray!25}$\bm{0.88\pm0.00}$ & \cellcolor{gray!25}$\bm{0.71\pm0.01}$ & [$0.70\pm0.00$] & $0.68\pm0.02$ \\
LGR+D  & \cellcolor{gray!25}$\bm{0.88\pm0.00}$ & [$0.70\pm0.00$] & \cellcolor{gray!25}$\bm{0.70\pm0.00}$ & $0.68\pm0.00$ & \cellcolor{gray!25}$\bm{0.88\pm0.00}$ & $0.69\pm0.01$ & $0.65\pm0.02$ & $0.62\pm0.02$ \\
\bottomrule

\end{tabular}

}
\end{table}

\begin{table}[h!]
\centering
\caption[\acs{Task-IL} \acs{CF} scores for \acs{iCV-MEFED}  W/ and W/O Data-Augmentation.]{\acs{Task-IL} \acs{CF} scores for \acs{iCV-MEFED} W/ and W/O Data-Augmentation. \textbf{Bold} values denote best (lowest) while [\textit{bracketed}] denote second-best values for each column.}
\label{tab:task-il-icvMefed-cf}

{
\scriptsize

\begin{tabular}{l|ccc|ccc}\toprule
\multicolumn{1}{c|}{\textbf{Method}}            & \multicolumn{3}{c|}{\textbf{\acs{CF} W/O Data-Augmentation}} 
& \multicolumn{3}{c}{\textbf{\acs{CF}  W/ Data-Augmentation}} \\ \midrule

\multicolumn{1}{c|}{ }                          & 
\multicolumn{1}{c|}{\textbf{Task 2}}          &
\multicolumn{1}{c|}{\textbf{Task 3}}            & \multicolumn{1}{c|}{\textbf{Task 4}}           &
\multicolumn{1}{c|}{\textbf{Task 2}}          &
\multicolumn{1}{c|}{\textbf{Task 3}}            & \multicolumn{1}{c}{\textbf{Task 4}}          \\ \cmidrule{2-7}
\multicolumn{7}{c}{\textbf{Baselines Approaches}} \\ \midrule

LB &  $0.38\pm0.02$ &  $0.14\pm0.04$ &  $0.32\pm0.07$ &  $0.09\pm0.05$ &  $0.10\pm0.05$ & $0.10\pm0.01$ \\
UB & \cellcolor{gray!25}$\bm{-0.03\pm0.02}$ & [$-0.01\pm0.02$] & \cellcolor{gray!25}$\bm{-0.03\pm0.02}$ & \cellcolor{gray!25}$\bm{-0.02\pm0.01}$ & \cellcolor{gray!25}$\bm{-0.03\pm0.02}$ & \cellcolor{gray!25}$\bm{0.01\pm0.00}$ \\\midrule

\multicolumn{7}{c}{\textbf{Regularisation-Based Approaches}} \\ \midrule
EWC  &  $0.22\pm0.04$ &  $0.05\pm0.03$ &  $0.12\pm0.05$ &  $0.08\pm0.10$ &  $0.02\pm0.02$ & $0.04\pm0.03$ \\
EWC-Online  &  $0.28\pm0.02$ &  $0.05\pm0.02$ &  $0.14\pm0.05$ &  $0.10\pm0.08$ &  $0.05\pm0.04$ & $0.05\pm0.03$ \\
SI  &  $0.42\pm0.05$ &  $0.14\pm0.01$ &  $0.32\pm0.07$ &  $0.10\pm0.07$ &  $0.13\pm0.07$ & $0.03\pm0.03$ \\
LwF  &  $0.01\pm0.01$ & $-0.00\pm0.01$ & [$0.00\pm0.00$] & $0.00\pm0.00$ & [$-0.01\pm0.00$] & \cellcolor{gray!25}$\bm{0.01\pm0.01}$ \\\midrule

\multicolumn{7}{c}{\textbf{Replay-Based Approaches}} \\ \midrule

NR  &  $0.11\pm0.04$ &  $0.13\pm0.01$ &  $0.00\pm0.02$ &  $0.15\pm0.03$ &  $0.11\pm0.02$ & $0.03\pm0.01$ \\
A-GEM  & [$-0.02\pm0.02$] & \cellcolor{gray!25}$\bm{-0.02\pm0.01}$ & $0.00\pm0.01$ & [$-0.01\pm0.00$] &  $0.00\pm0.00$ & \cellcolor{gray!25}$\bm{0.01\pm0.00}$ \\
LR  &  $0.10\pm0.01$ &  $0.11\pm0.02$ &  $0.01\pm0.00$ &  $0.02\pm0.02$ &  $0.06\pm0.01$ & $0.03\pm0.00$ \\
DGR  &  $0.04\pm0.01$ &  $0.04\pm0.01$ & [$-0.01\pm0.01$] &  $0.01\pm0.02$ &  $0.09\pm0.03$ & $0.03\pm0.00$ \\
LGR  &  $0.00\pm0.02$ &  $0.00\pm0.01$ &  $0.00\pm0.01$ &  $0.00\pm0.00$ &  $0.00\pm0.01$ & $0.03\pm0.00$ \\
DGR+D  &  $0.00\pm0.01$ &  $0.02\pm0.01$ & [$-0.01\pm0.01$] &  $0.01\pm0.01$ &  $0.03\pm0.02$ & [$0.02\pm0.00$] \\
LGR+D  & $0.00\pm0.01$ & $0.00\pm0.00$ &  $0.00\pm0.01$ &  $0.01\pm0.01$ &  $0.01\pm0.01$ & $0.03\pm0.00$ \\

\bottomrule

\end{tabular}

}
\end{table}
\begin{figure}[h!]
    \centering
    \subfloat[\ac{Task-IL} Results w/o Augmentation.\label{fig:icvMefed-task-il-noaug}]{\includegraphics[width=0.5\textwidth]{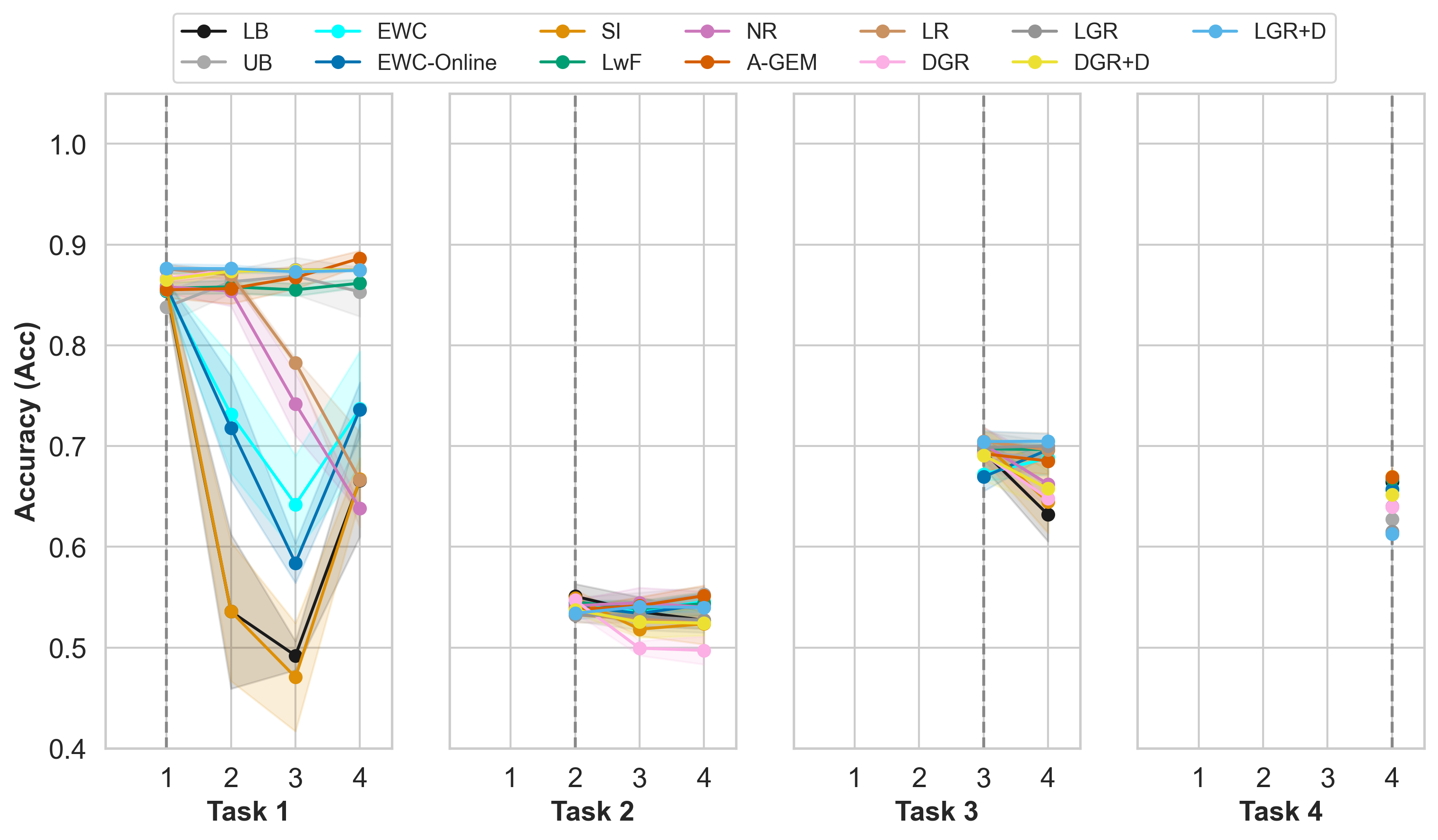}}     \hfill
    \subfloat[\ac{Task-IL} Results w/ Augmentation.\label{fig:icvMefed-taskil-aug}]{\includegraphics[width=0.5\textwidth]{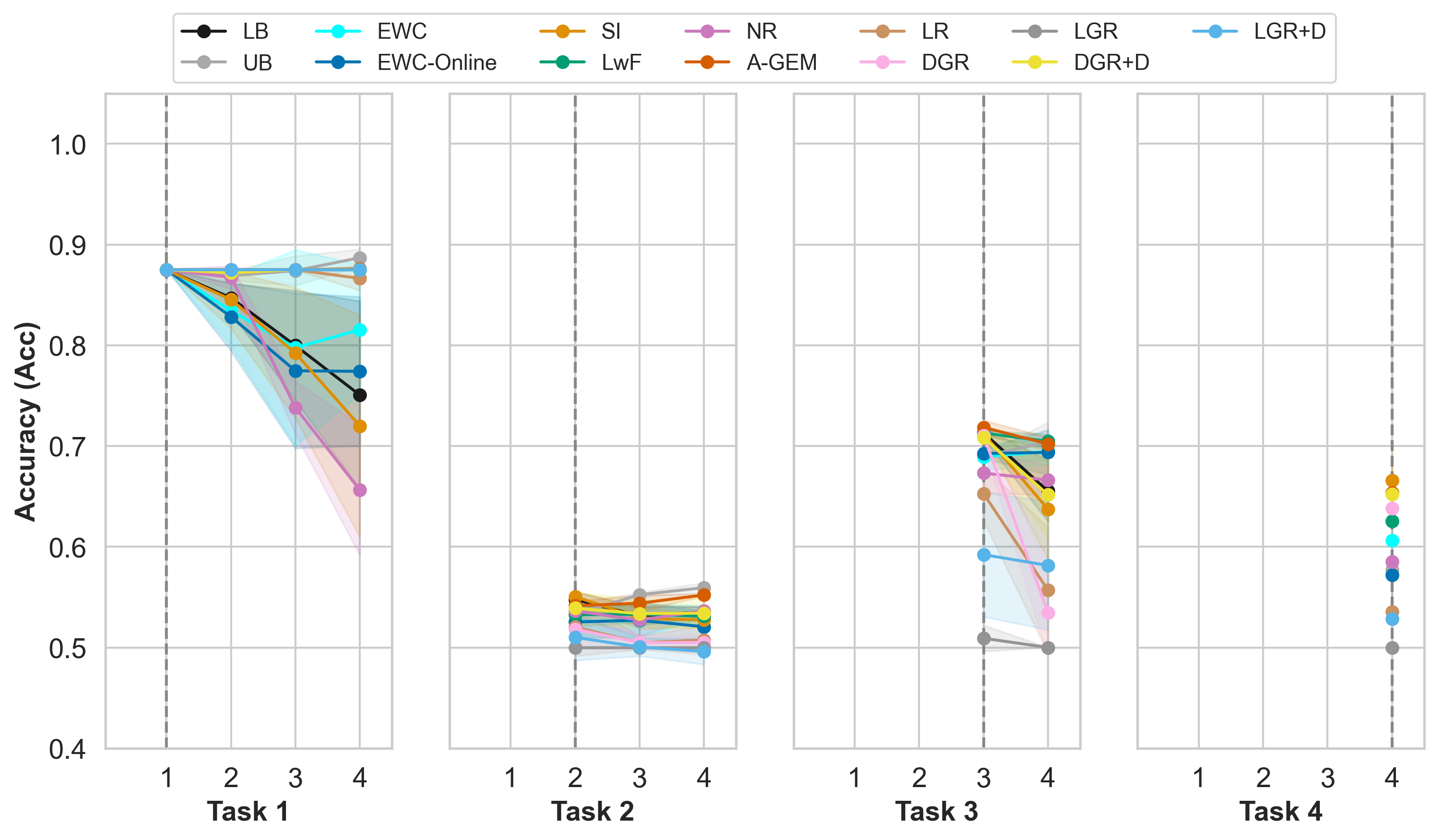}} \\
    \caption[Task-IL Results for \acs{iCV-MEFED}  W/ and W/O Augmentation.]{\ac{Task-IL} results for \acs{iCV-MEFED} (a) without and (b) with augmentation. For each task, test-set accuracy is shown as the learning progresses from when the task is introduced to the end of the overall training procedure.}
    \label{fig:icvMefed-taskil} 
    
\end{figure}
\FloatBarrier

\subsubsection{Task-Ordering}
\label{app:taskordericv}

\begin{table}[h!]
\centering
\setlength\tabcolsep{3.5pt}

\caption[\acs{Task-IL} \acs{Acc} for \acs{iCV-MEFED} following Orderings O2 and O3.]{\acs{Task-IL} \acs{Acc} for \acs{iCV-MEFED} W/ Data-Augmentation following Orderings O2 and O3. \textbf{Bold} values denote best (highest) while [\textit{bracketed}] denote second-best values for each column.}

\label{tab:task-il-icvMefed-orders-acc}

{
\scriptsize
\begin{tabular}{l|cccc|cccc}\toprule
\multicolumn{1}{c|}{\textbf{Method}}            & \multicolumn{4}{c|}{\textbf{Accuracy following O2}} 
& \multicolumn{4}{c}{\textbf{Accuracy following O3}} \\ \midrule

\multicolumn{1}{c|}{ }                          & 
\multicolumn{1}{c|}{\textbf{Task 1}}            & \multicolumn{1}{c|}{\textbf{Task 2}}          &
\multicolumn{1}{c|}{\textbf{Task 3}}            & \multicolumn{1}{c|}{\textbf{Task 4}}           &
\multicolumn{1}{c|}{\textbf{Task 1}}            & \multicolumn{1}{c|}{\textbf{Task 2}}          &
\multicolumn{1}{c|}{\textbf{Task 3}}            & \multicolumn{1}{c}{\textbf{Task 4}}          \\ \cmidrule{2-9}
\makecell[c]{}                            & \multicolumn{8}{c}{\textbf{Baselines Approaches}} \\ \midrule

LB & \cellcolor{gray!25}$\bm{0.88\pm0.00}$ & $0.65\pm0.05$ & $0.59\pm0.04$ & $0.54\pm0.05$ & \cellcolor{gray!25}$\bm{0.88\pm0.00}$ & $0.68\pm0.03$ & $0.65\pm0.04$ & $0.60\pm0.01$ \\
UB & \cellcolor{gray!25}$\bm{0.88\pm0.00}$ & \cellcolor{gray!25}$\bm{0.71\pm0.00}$ & $0.69\pm0.01$ & $0.67\pm0.00$ & \cellcolor{gray!25}$\bm{0.88\pm0.00}$ & [$0.70\pm0.00$] & [$0.69\pm0.01$] & $0.68\pm0.02$ \\
 \midrule

{}                                     & \multicolumn{8}{c}{\textbf{Regularisation-Based Approaches}} \\ \midrule

EWC  & \cellcolor{gray!25}$\bm{0.88\pm0.00}$ & $0.68\pm0.02$ & $0.68\pm0.00$ & $0.65\pm0.01$ & \cellcolor{gray!25}$\bm{0.88\pm0.00}$ & [$0.70\pm0.01$] & $0.66\pm0.05$ & $0.67\pm0.00$ \\
EWC-Online  & \cellcolor{gray!25}$\bm{0.88\pm0.00}$ & $0.69\pm0.01$ & $0.67\pm0.02$ & $0.63\pm0.02$ & \cellcolor{gray!25}$\bm{0.88\pm0.00}$ & [$0.70\pm0.00$] & [$0.69\pm0.02$] & $0.67\pm0.01$ \\
SI  & \cellcolor{gray!25}$\bm{0.88\pm0.00}$ & $0.63\pm0.06$ & $0.60\pm0.07$ & $0.62\pm0.05$ & \cellcolor{gray!25}$\bm{0.88\pm0.00}$ & [$0.70\pm0.01$] & $0.67\pm0.04$ & $0.65\pm0.03$ \\
LwF  & \cellcolor{gray!25}$\bm{0.88\pm0.00}$ & [$0.70\pm0.00$] & $0.69\pm0.00$ & [$0.68\pm0.01$] & \cellcolor{gray!25}$\bm{0.88\pm0.00}$ & \cellcolor{gray!25}$\bm{0.71\pm0.00}$ & \cellcolor{gray!25}$\bm{0.71\pm0.01}$ & [$0.69\pm0.01$] \\ \midrule
\makecell[c]{}                           & \multicolumn{8}{c}{\textbf{Replay-Based Approaches}} \\ \midrule

NR  & \cellcolor{gray!25}$\bm{0.88\pm0.00}$ & [$0.70\pm0.02$] & $0.64\pm0.02$ & $0.59\pm0.00$ & \cellcolor{gray!25}$\bm{0.88\pm0.00}$ & [$0.70\pm0.01$] & $0.64\pm0.01$ & $0.60\pm0.02$ \\
A-GEM  & \cellcolor{gray!25}$\bm{0.88\pm0.00}$ & \cellcolor{gray!25}$\bm{0.71\pm0.00}$ & \cellcolor{gray!25}$\bm{0.71\pm0.01}$ & \cellcolor{gray!25}$\bm{0.69\pm0.00}$ & \cellcolor{gray!25}$\bm{0.88\pm0.00}$ & \cellcolor{gray!25}$\bm{0.71\pm0.00}$ & \cellcolor{gray!25}$\bm{0.71\pm0.00}$ & \cellcolor{gray!25}$\bm{0.70\pm0.00}$ \\
LR  & \cellcolor{gray!25}$\bm{0.88\pm0.00}$ & $0.69\pm0.00$ & $0.67\pm0.01$ & $0.62\pm0.02$ & \cellcolor{gray!25}$\bm{0.88\pm0.00}$ & $0.69\pm0.00$ & $0.65\pm0.00$ & $0.61\pm0.01$ \\
DGR  & \cellcolor{gray!25}$\bm{0.88\pm0.00}$ & [$0.70\pm0.00$] & $0.68\pm0.01$ & $0.60\pm0.00$ & \cellcolor{gray!25}$\bm{0.88\pm0.00}$ & [$0.70\pm0.01$] & $0.64\pm0.03$ & $0.61\pm0.03$ \\
LGR  & \cellcolor{gray!25}$\bm{0.88\pm0.00}$ & [$0.70\pm0.00$] & $0.67\pm0.01$ & $0.64\pm0.02$ & \cellcolor{gray!25}$\bm{0.88\pm0.00}$ & $0.69\pm0.00$ & $0.65\pm0.02$ & $0.60\pm0.00$ \\
DGR+D & \cellcolor{gray!25}$\bm{0.88\pm0.00}$ & $0.69\pm0.00$ & [$0.70\pm0.01$] & [$0.68\pm0.01$] & \cellcolor{gray!25}$\bm{0.88\pm0.00}$ & [$0.70\pm0.01$] & $0.68\pm0.04$ & $0.65\pm0.02$ \\
LGR+D  & \cellcolor{gray!25}$\bm{0.88\pm0.00}$ & [$0.70\pm0.00$] & $0.68\pm0.01$ & $0.66\pm0.02$ & \cellcolor{gray!25}$\bm{0.88\pm0.00}$ & $0.69\pm0.00$ & $0.64\pm0.03$ & $0.61\pm0.03$ \\

\bottomrule

\end{tabular}

}
\end{table}

\begin{table}[h!]
\centering
\caption[\acs{Task-IL} \acs{CF} scores for \acs{iCV-MEFED} following O2 and O3.]{\acs{Task-IL} \acs{CF} scores for \acs{iCV-MEFED} W/ Data-Augmentation following Orderings O2 and O3. \textbf{Bold} values denote best (lowest) while [\textit{bracketed}] denote second-best values for each column.}

\label{tab:task-il-icvMefed-orders-cf}

{
\scriptsize

\begin{tabular}{l|ccc|ccc}\toprule
\multicolumn{1}{c|}{\textbf{Method}}            & \multicolumn{3}{c|}{\textbf{CF following O2}} 
& \multicolumn{3}{c}{\textbf{CF following O3}} \\ \midrule

\multicolumn{1}{c|}{ }                          & 
\multicolumn{1}{c|}{\textbf{Task 2}}          &
\multicolumn{1}{c|}{\textbf{Task 3}}            & \multicolumn{1}{c|}{\textbf{Task 4}}           &
\multicolumn{1}{c|}{\textbf{Task 2}}          &
\multicolumn{1}{c|}{\textbf{Task 3}}            & \multicolumn{1}{c}{\textbf{Task 4}}          \\ \cmidrule{2-7}

\makecell[c]{}                            & \multicolumn{6}{c}{\textbf{Baselines Approaches}} \\ \midrule

LB &  $0.34\pm0.12$ &  $0.12\pm0.11$ &  $0.10\pm0.09$ &  $0.16\pm0.12$ &  $0.08\pm0.03$ & $0.08\pm0.07$ \\
UB &  [$0.00\pm0.02$] & \cellcolor{gray!25}$\bm{-0.03\pm0.01}$ &  [$0.02\pm0.01$] & \cellcolor{gray!25}$\bm{-0.02\pm0.01}$ & \cellcolor{gray!25}$\bm{-0.03\pm0.02}$ & [$0.01\pm0.00$] \\\midrule

{}                                     & \multicolumn{6}{c}{\textbf{Regularisation-Based Approaches}} \\ \midrule

EWC  &  $0.03\pm0.02$ &  $0.03\pm0.01$ &  $0.03\pm0.04$ &  $0.12\pm0.13$ &  $0.02\pm0.01$ & [$0.01\pm0.01$] \\
EWC-Online  &  $0.06\pm0.04$ &  $0.07\pm0.05$ &  [$0.02\pm0.03$] &  $0.04\pm0.05$ &  $0.02\pm0.02$ & [$0.01\pm0.01$] \\
SI  &  $0.31\pm0.22$ &  $0.14\pm0.11$ &  $0.13\pm0.11$ &  $0.11\pm0.13$ &  $0.08\pm0.07$ & $0.02\pm0.02$ \\
LwF  &  $0.01\pm0.00$ & $0.00\pm0.01$ &  \cellcolor{gray!25}$\bm{0.01\pm0.01}$ &  $0.01\pm0.00$ &  $0.01\pm0.01$ & [$0.01\pm0.00$] \\\midrule

\makecell[c]{}                            & \multicolumn{6}{c}{\textbf{Replay-Based Approaches}} \\ \midrule

NR  &  $0.19\pm0.03$ &  $0.17\pm0.02$ &  $0.03\pm0.02$ &  $0.21\pm0.02$ &  $0.16\pm0.03$ & $0.02\pm0.01$ \\
A-GEM  &  $0.01\pm0.00$ &  $0.00\pm0.02$ & \cellcolor{gray!25}$\bm{0.01\pm0.01}$ &  [$-0.01\pm0.01$] &  [$0.00\pm0.00$] & \cellcolor{gray!25}$\bm{0.00\pm0.01}$ \\
LR  & \cellcolor{gray!25}$\bm{-0.01\pm0.01}$ &  $0.04\pm0.00$ &  $0.03\pm0.00$ & $0.01\pm0.01$ &  $0.01\pm0.02$ & $0.02\pm0.00$ \\
DGR  &  $0.01\pm0.01$ &  $0.09\pm0.01$ & $0.05\pm0.00$ &  $0.02\pm0.02$ &  $0.03\pm0.05$ & $0.03\pm0.00$ \\
LGR  &  $0.02\pm0.01$ &  $0.02\pm0.02$ &  $0.03\pm0.00$ &  $0.01\pm0.01$ &  $0.05\pm0.04$ & $0.03\pm0.00$ \\
DGR+D & [$0.00\pm0.02$] &  $0.01\pm0.02$ &  [$0.02\pm0.01$] &  $0.01\pm0.01$ &  $0.04\pm0.03$ & $0.03\pm0.00$ \\
LGR+D  & \cellcolor{gray!25}$\bm{-0.01\pm0.01}$ & [$-0.01\pm0.00$] &  $0.03\pm0.00$ &  $0.01\pm0.01$ &  $0.02\pm0.00$ & $0.03\pm0.00$ \\

\bottomrule

\end{tabular}

}
\end{table}

\begin{figure}[h!]
    \centering
    \subfloat[\ac{Task-IL} Results following O$2$.\label{fig:icvmefed-task-il-aug-order-1}]{\includegraphics[width=0.5\textwidth]{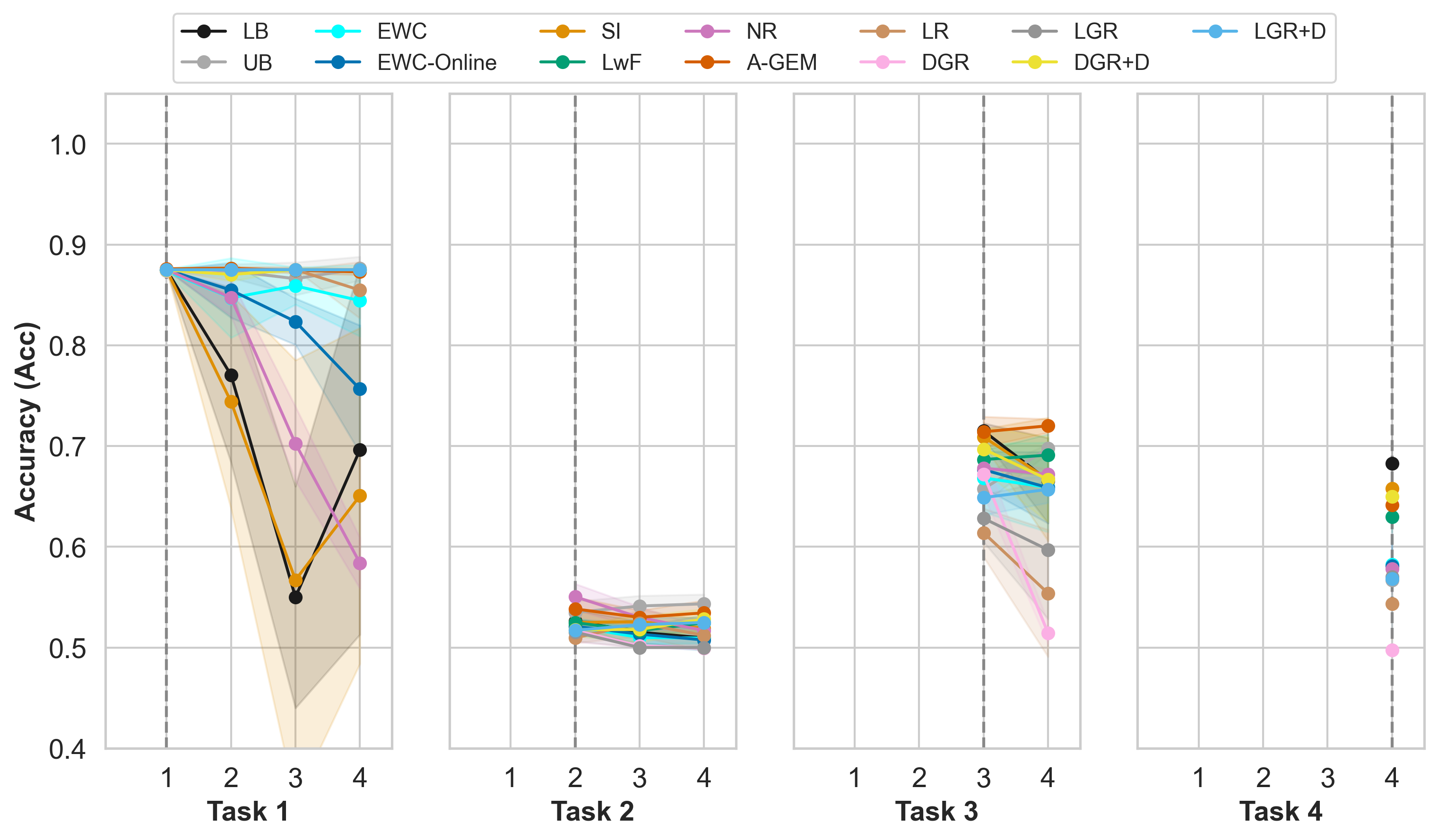}}     \hfill
    \subfloat[\ac{Task-IL} Results following O$3$.\label{fig:icvmefed-task-il-aug-order-2}]{\includegraphics[width=0.5\textwidth]{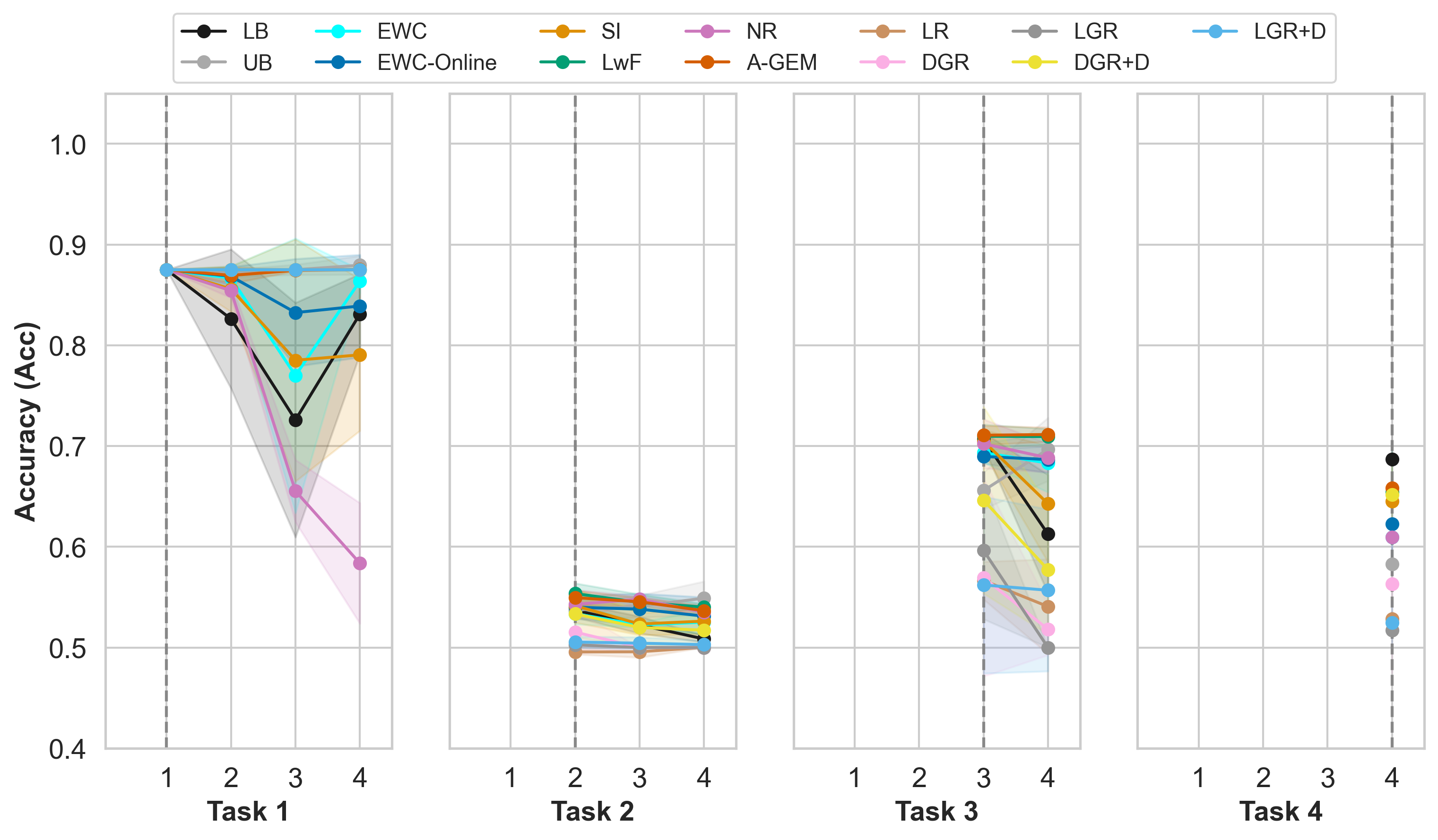}} \\
    \caption[Task-IL Results for \acs{iCV-MEFED}  following Order 2 and Order 3.]{\ac{Task-IL} results for \acs{iCV-MEFED} following (a) Order $2$ and (b) Order $3$, respectively. For each task, test-set accuracy is shown as the learning progresses from when the task is introduced to the end of the overall training procedure.}
    \label{fig:icvMefed-taskil-orders} 
    
\end{figure}

\FloatBarrier

\normalsize
\textit{}
\newpage
\subsubsection{\acf{Class-IL} Results}
\label{app:classilicv}
\begin{table}[h!]
\centering
\setlength\tabcolsep{2.0pt}

\caption[\acs{Class-IL} \acs{Acc} for \acs{iCV-MEFED}  W/ and W/O Data-Augmentation.]{\acs{Class-IL} \acs{Acc} for \acs{iCV-MEFED}  W/ and W/O Data-Augmentation. \textbf{Bold} values denote best (highest) while [\textit{bracketed}] denote second-best values for each column. 
}
\label{tab:class-il-icvMefed-acc}

{
\scriptsize
\begin{tabular}{l|cccccccc}\toprule

\multirow{2}[2]{*}{\makecell[c]{\textbf{Method}}}           & \multicolumn{8}{c}{\textbf{Accuracy W/O Data-Augmentation}} \\ \cmidrule{2-9}

\multicolumn{1}{c|}{ }                          & 
\multicolumn{1}{c|}{\textbf{Class 1}}            & \multicolumn{1}{c|}{\textbf{Class 2}}          &
\multicolumn{1}{c|}{\textbf{Class 3}}            & \multicolumn{1}{c|}{\textbf{Class 4}}           &
\multicolumn{1}{c|}{\textbf{Class 5}}            & \multicolumn{1}{c|}{\textbf{Class 6}}          &
\multicolumn{1}{c|}{\textbf{Class 7}}            & \multicolumn{1}{c}{\textbf{Class 8}}          \\ \midrule

\multicolumn{9}{c}{\textbf{Baseline Approaches}} \\ \midrule
        LB & \cellcolor{gray!25}$\bm{1.00\pm0.00}$ & $0.50\pm0.00$ & $0.33\pm0.00$ & $0.25\pm0.00$ & $0.20\pm0.00$ & $0.17\pm0.00$ & $0.14\pm0.00$ & $0.12\pm0.00$ \\
        UB & \cellcolor{gray!25}$\bm{1.00\pm0.00}$ & $0.51\pm0.02$ & $0.38\pm0.04$ & $0.30\pm0.04$ & $0.25\pm0.04$ & $0.23\pm0.04$ & $0.21\pm0.05$ & [$0.18\pm0.04$] \\ \midrule

\multicolumn{9}{c}{\textbf{Regularisation-Based Approaches}} \\ \midrule

       EWC & \cellcolor{gray!25}$\bm{1.00\pm0.00}$ & $0.50\pm0.00$ & $0.33\pm0.00$ & $0.25\pm0.00$ & $0.20\pm0.00$ & $0.17\pm0.00$ & $0.14\pm0.00$ & $0.12\pm0.00$ \\
EWC-Online & \cellcolor{gray!25}$\bm{1.00\pm0.00}$ & $0.50\pm0.00$ & $0.33\pm0.00$ & $0.25\pm0.00$ & $0.20\pm0.00$ & $0.17\pm0.00$ & $0.14\pm0.00$ & $0.12\pm0.00$ \\
        SI & \cellcolor{gray!25}$\bm{1.00\pm0.00}$ & $0.50\pm0.00$ & $0.33\pm0.00$ & $0.25\pm0.00$ & $0.20\pm0.00$ & $0.17\pm0.00$ & $0.14\pm0.00$ & $0.12\pm0.00$ \\
       LwF & \cellcolor{gray!25}$\bm{1.00\pm0.00}$ & $0.50\pm0.00$ & $0.33\pm0.00$ & $0.25\pm0.00$ & $0.20\pm0.00$ & $0.17\pm0.00$ & $0.14\pm0.00$ & $0.12\pm0.00$ \\ \midrule
\multicolumn{9}{c}{\textbf{Replay-Based Approaches}} \\ \midrule

        NR & \cellcolor{gray!25}$\bm{1.00\pm0.00}$ & \cellcolor{gray!25}$\bm{0.69\pm0.01}$ & [$0.44\pm0.02$] & \cellcolor{gray!25}$\bm{0.34\pm0.02}$ & [$0.27\pm0.01$] & [$0.26\pm0.00$] & [$0.22\pm0.01$] & $0.17\pm0.01$ \\
     A-GEM & \cellcolor{gray!25}$\bm{1.00\pm0.00}$ & $0.50\pm0.00$ & $0.33\pm0.00$ & $0.26\pm0.01$ & $0.20\pm0.00$ & $0.17\pm0.00$ & $0.14\pm0.00$ & $0.12\pm0.00$ \\
        LR & \cellcolor{gray!25}$\bm{1.00\pm0.00}$ & [$0.64\pm0.00$] & \cellcolor{gray!25}$\bm{0.45\pm0.02}$ & \cellcolor{gray!25}$\bm{0.34\pm0.01}$ & \cellcolor{gray!25}$\bm{0.29\pm0.01}$ & \cellcolor{gray!25}$\bm{0.27\pm0.01}$ & \cellcolor{gray!25}$\bm{0.24\pm0.01}$ & \cellcolor{gray!25}$\bm{0.22\pm0.01}$ \\
       DGR & \cellcolor{gray!25}$\bm{1.00\pm0.00}$ & $0.50\pm0.00$ & $0.33\pm0.00$ & $0.25\pm0.00$ & $0.20\pm0.00$ & $0.17\pm0.00$ & $0.14\pm0.00$ & $0.12\pm0.00$ \\
       LGR & \cellcolor{gray!25}$\bm{1.00\pm0.00}$ & $0.57\pm0.02$ & $0.38\pm0.02$ & $0.28\pm0.01$ & $0.23\pm0.01$ & $0.23\pm0.01$ & $0.20\pm0.01$ & $0.17\pm0.01$ \\
     DGR+D & \cellcolor{gray!25}$\bm{1.00\pm0.00}$ & $0.50\pm0.00$ & $0.33\pm0.00$ & $0.25\pm0.00$ & $0.20\pm0.00$ & $0.17\pm0.00$ & $0.14\pm0.00$ & $0.13\pm0.00$ \\
     LGR+D & \cellcolor{gray!25}$\bm{1.00\pm0.00}$ & $0.61\pm0.03$ & $0.41\pm0.04$ & [$0.30\pm0.01$] & $0.23\pm0.01$ & $0.25\pm0.02$ & $0.21\pm0.01$ & $0.17\pm0.02$ \\

\midrule

\multirow{2}[2]{*}{\makecell[c]{\textbf{Method}}}           & \multicolumn{8}{c}{\textbf{Accuracy W/ Data-Augmentation}} \\ \cmidrule{2-9}

\multicolumn{1}{c|}{ }                          & 
\multicolumn{1}{c|}{\textbf{Class 1}}            & \multicolumn{1}{c|}{\textbf{Class 2}}          &
\multicolumn{1}{c|}{\textbf{Class 3}}            & \multicolumn{1}{c|}{\textbf{Class 4}}           &
\multicolumn{1}{c|}{\textbf{Class 5}}            & \multicolumn{1}{c|}{\textbf{Class 6}}          &
\multicolumn{1}{c|}{\textbf{Class 7}}            & \multicolumn{1}{c}{\textbf{Class 8}}          \\ \midrule
\multicolumn{9}{c}{\textbf{Baseline Approaches}} \\ \midrule

        LB & \cellcolor{gray!25}$\bm{1.00\pm0.00}$ & $0.50\pm0.00$ & $0.33\pm0.00$ & $0.25\pm0.00$ & $0.20\pm0.00$ & $0.17\pm0.00$ & $0.14\pm0.00$ & $0.12\pm0.00$ \\
        UB & \cellcolor{gray!25}$\bm{1.00\pm0.00}$ & $0.50\pm0.00$ & $0.34\pm0.01$ & $0.26\pm0.01$ & $0.22\pm0.02$ & [$0.20\pm0.05$] & [$0.18\pm0.05$] & \cellcolor{gray!25}$\bm{0.16\pm0.06}$ \\\midrule

\multicolumn{9}{c}{\textbf{Regularisation-Based Approaches}} \\ \midrule

       EWC & \cellcolor{gray!25}$\bm{1.00\pm0.00}$ & $0.50\pm0.00$ & $0.33\pm0.00$ & $0.25\pm0.00$ & $0.20\pm0.00$ & $0.17\pm0.00$ & $0.14\pm0.00$ & $0.12\pm0.00$ \\
EWC-Online & \cellcolor{gray!25}$\bm{1.00\pm0.00}$ & $0.50\pm0.00$ & $0.33\pm0.00$ & $0.25\pm0.00$ & $0.20\pm0.00$ & $0.17\pm0.00$ & $0.14\pm0.00$ & $0.12\pm0.00$ \\
        SI & \cellcolor{gray!25}$\bm{1.00\pm0.00}$ & $0.50\pm0.00$ & $0.33\pm0.00$ & $0.25\pm0.00$ & $0.20\pm0.00$ & $0.17\pm0.00$ & $0.14\pm0.00$ & $0.12\pm0.00$ \\
       LwF & \cellcolor{gray!25}$\bm{1.00\pm0.00}$ & $0.50\pm0.00$ & $0.33\pm0.00$ & $0.25\pm0.00$ & $0.20\pm0.00$ & $0.17\pm0.00$ & $0.14\pm0.00$ & $0.12\pm0.00$  \\\midrule
\multicolumn{9}{c}{\textbf{Replay-Based Approaches}} \\ \midrule

        NR & \cellcolor{gray!25}$\bm{1.00\pm0.00}$ & \cellcolor{gray!25}$\bm{0.66\pm0.04}$ & \cellcolor{gray!25}$\bm{0.42\pm0.02}$ & \cellcolor{gray!25}$\bm{0.31\pm0.01}$ & [$0.23\pm0.01$] & [$0.20\pm0.01$] & [$0.18\pm0.02$] & [$0.14\pm0.00$] \\
     A-GEM & \cellcolor{gray!25}$\bm{1.00\pm0.00}$ & $0.50\pm0.00$ & $0.33\pm0.00$ & $0.25\pm0.00$ & $0.20\pm0.00$ & $0.17\pm0.00$ & $0.14\pm0.00$ & $0.12\pm0.00$ \\
        LR & \cellcolor{gray!25}$\bm{1.00\pm0.00}$ & [$0.56\pm0.05$] & [$0.40\pm0.05$] & [$0.30\pm0.04$] & \cellcolor{gray!25}$\bm{0.24\pm0.03}$ & \cellcolor{gray!25}$\bm{0.23\pm0.04}$ & \cellcolor{gray!25}$\bm{0.19\pm0.04}$ & \cellcolor{gray!25}$\bm{0.16\pm0.03}$ \\
       DGR & \cellcolor{gray!25}$\bm{1.00\pm0.00}$ & $0.50\pm0.00$ & $0.33\pm0.00$ & $0.25\pm0.00$ & $0.20\pm0.00$ & $0.17\pm0.00$ & $0.14\pm0.00$ & $0.12\pm0.00$ \\
       LGR & \cellcolor{gray!25}$\bm{1.00\pm0.00}$ & $0.52\pm0.03$ & $0.34\pm0.02$ & $0.25\pm0.00$ & $0.21\pm0.01$ & $0.18\pm0.01$ & $0.15\pm0.01$ & $0.13\pm0.01$ \\
     DGR+D & \cellcolor{gray!25}$\bm{1.00\pm0.00}$ & $0.50\pm0.00$ & $0.33\pm0.00$ & $0.25\pm0.00$ & $0.20\pm0.00$ & $0.17\pm0.00$ & $0.14\pm0.00$ & $0.12\pm0.00$ \\
     LGR+D & \cellcolor{gray!25}$\bm{1.00\pm0.00}$ & $0.52\pm0.03$ & $0.35\pm0.03$ & $0.27\pm0.03$ & $0.21\pm0.01$ & $0.19\pm0.03$ & $0.15\pm0.02$ & $0.13\pm0.01$ \\

\bottomrule

\end{tabular}

}
\end{table}
\FloatBarrier

\begin{table}[h!]
\centering
\caption[\acs{Class-IL} \acs{CF} scores for \acs{iCV-MEFED}  W/ and W/O Data-Augmentation.]{\acs{CF} scores for \acs{Class-IL} on \acs{iCV-MEFED}  W/ and W/O Data-Augmentation. \textbf{Bold} values denote best (lowest) while [\textit{bracketed}] denote second-best values for each column.}
\label{tab:class-il-icvMefed-cf}

{
\scriptsize
\setlength\tabcolsep{2.0pt}

\begin{tabular}{l|ccccccc}\toprule

\multirow{2}[2]{*}{\makecell[c]{\textbf{Method}}}           & \multicolumn{7}{c}{\textbf{\acs{CF} W/O Data-Augmentation}} \\ \cmidrule{2-8}

\multicolumn{1}{c|}{ }                          & 
\multicolumn{1}{c|}{\textbf{Class 2}}          &
\multicolumn{1}{c|}{\textbf{Class 3}}            & \multicolumn{1}{c|}{\textbf{Class 4}}           &
\multicolumn{1}{c|}{\textbf{Class 5}}            & \multicolumn{1}{c|}{\textbf{Class 6}}          &
\multicolumn{1}{c|}{\textbf{Class 7}}            & \multicolumn{1}{c}{\textbf{Class 8}}          \\ \midrule
\multicolumn{8}{c}{\textbf{Baseline Approaches}} \\ \midrule

LB & $1.00\pm0.00$ & $1.00\pm0.00$ & $1.00\pm0.00$ & $1.00\pm0.00$ & $1.00\pm0.00$ & $1.00\pm0.00$ & $1.00\pm0.00$ \\
UB & $0.65\pm0.11$ & [$0.50\pm0.12$] & [$0.44\pm0.13$] & [$0.38\pm0.13$] & [$0.33\pm0.12$] & [$0.31\pm0.12$] & [$0.30\pm0.00$] \\\midrule

\multicolumn{8}{c}{\textbf{Regularisation-Based Approaches}} \\ \midrule

       EWC & $1.00\pm0.00$ & $1.00\pm0.00$ & $1.00\pm0.00$ & $1.00\pm0.00$ & $1.00\pm0.00$ & $1.00\pm0.00$ & $1.00\pm0.00$ \\
EWC-Online & $1.00\pm0.00$ & $1.00\pm0.00$ & $1.00\pm0.00$ & $1.00\pm0.00$ & $1.00\pm0.00$ & $1.00\pm0.00$ & $1.00\pm0.00$ \\
        SI & $1.00\pm0.00$ & $1.00\pm0.00$ & $1.00\pm0.00$ & $1.00\pm0.00$ & $1.00\pm0.00$ & $1.00\pm0.00$ & $1.00\pm0.00$ \\
       LwF & $1.00\pm0.00$ & $1.00\pm0.00$ & $1.00\pm0.00$ & $1.00\pm0.00$ & $1.00\pm0.00$ & $1.00\pm0.00$ & $1.00\pm0.00$ \\ \midrule

\multicolumn{8}{c}{\textbf{Replay-Based Approaches}} \\ \midrule

        NR & $0.66\pm0.05$ & $0.62\pm0.04$ & $0.63\pm0.05$ & $0.61\pm0.07$ & $0.62\pm0.08$ & $0.66\pm0.09$ & $0.69\pm0.00$ \\
     A-GEM & $1.00\pm0.01$ & $0.99\pm0.01$ & $0.92\pm0.12$ & $0.93\pm0.09$ & $0.94\pm0.08$ & $0.95\pm0.07$ & $1.00\pm0.00$ \\
        LR & \cellcolor{gray!25}$\bm{0.40\pm0.02}$ & \cellcolor{gray!25}$\bm{0.34\pm0.00}$ & \cellcolor{gray!25}$\bm{0.30\pm0.00}$ & \cellcolor{gray!25}$\bm{0.29\pm0.01}$ & \cellcolor{gray!25}$\bm{0.27\pm0.01}$ & $0.25\pm0.01$ & \cellcolor{gray!25}$\bm{0.27\pm0.00}$ \\
       DGR & $1.00\pm0.00$ & $1.00\pm0.00$ & $1.00\pm0.00$ & $1.00\pm0.00$ & $1.00\pm0.00$ & $1.00\pm0.00$ & $1.00\pm0.00$ \\
       LGR & $0.80\pm0.05$ & $0.82\pm0.04$ & $0.79\pm0.05$ & $0.75\pm0.04$ & $0.74\pm0.04$ & $0.76\pm0.04$ & $0.75\pm0.00$ \\
     DGR+D & $0.83\pm0.24$ & $0.89\pm0.16$ & $0.91\pm0.12$ & $0.93\pm0.10$ & $0.94\pm0.08$ & $0.95\pm0.07$ & $0.95\pm0.00$ \\
     LGR+D & [$0.60\pm0.08$] & $0.66\pm0.06$ & $0.67\pm0.07$ & $0.62\pm0.07$ & $0.63\pm0.05$ & $0.66\pm0.06$ & $0.67\pm0.00$ \\ \midrule

\multirow{2}[2]{*}{\makecell[c]{\textbf{Method}}}          & \multicolumn{7}{c}{\textbf{\acs{CF} W/ Data-Augmentation}} \\ \cmidrule{2-8}

\multicolumn{1}{c|}{ }                          & 
\multicolumn{1}{c|}{\textbf{Class 2}}          &
\multicolumn{1}{c|}{\textbf{Class 3}}            & \multicolumn{1}{c|}{\textbf{Class 4}}           &
\multicolumn{1}{c|}{\textbf{Class 5}}            & \multicolumn{1}{c|}{\textbf{Class 6}}          &
\multicolumn{1}{c|}{\textbf{Class 7}}            & \multicolumn{1}{c}{\textbf{Class 8}}          \\ \midrule

\multicolumn{8}{c}{\textbf{Baseline Approaches}} \\ \midrule

        LB & $1.00\pm0.00$ & $1.00\pm0.00$ & $1.00\pm0.00$ & $1.00\pm0.00$ & $1.00\pm0.00$ & $1.00\pm0.00$ & $1.00\pm0.00$ \\
        UB & $0.89\pm0.16$ & $0.61\pm0.09$ & $0.47\pm0.04$ & $0.46\pm0.10$ & $0.45\pm0.16$ & $0.39\pm0.13$ & $0.32\pm0.08$ \\ \midrule

\multicolumn{8}{c}{\textbf{Regularisation-Based Approaches}} \\ \midrule

       EWC & $1.00\pm0.00$ & $1.00\pm0.00$ & $1.00\pm0.00$ & $1.00\pm0.00$ & $1.00\pm0.00$ & $1.00\pm0.00$ & $1.00\pm0.00$ \\
EWC-Online & $1.00\pm0.00$ & $1.00\pm0.00$ & $1.00\pm0.00$ & $1.00\pm0.00$ & $1.00\pm0.00$ & $1.00\pm0.00$ & $1.00\pm0.00$ \\
        SI & $1.00\pm0.00$ & $1.00\pm0.00$ & $1.00\pm0.00$ & $1.00\pm0.00$ & $1.00\pm0.00$ & $1.00\pm0.00$ & $1.00\pm0.00$ \\
        SI & $1.00\pm0.00$ & $1.00\pm0.00$ & $1.00\pm0.00$ & $1.00\pm0.00$ & $1.00\pm0.00$ & $1.00\pm0.00$ & $1.00\pm0.00$ \\

\multicolumn{8}{c}{\textbf{Replay-Based Approaches}} \\ \midrule

        NR & $0.65\pm0.01$ & $0.66\pm0.05$ & $0.69\pm0.11$ & $0.73\pm0.10$ & $0.74\pm0.10$ & $0.79\pm0.07$ & $0.71\pm0.02$ \\
     A-GEM & $1.00\pm0.00$ & $1.00\pm0.00$ & $1.00\pm0.00$ & $1.00\pm0.00$ & $1.00\pm0.00$ & $1.00\pm0.00$ & $1.00\pm0.00$ \\
        LR & \cellcolor{gray!25}$\bm{0.34\pm0.13}$ & \cellcolor{gray!25}$\bm{0.26\pm0.09}$ & \cellcolor{gray!25}$\bm{0.21\pm0.05}$ & \cellcolor{gray!25}$\bm{0.19\pm0.06}$ & \cellcolor{gray!25}$\bm{0.16\pm0.05}$ & \cellcolor{gray!25}$\bm{0.14\pm0.04}$ & \cellcolor{gray!25}$\bm{0.13\pm0.06}$ \\
       DGR & $1.00\pm0.00$ & $1.00\pm0.01$ & $1.00\pm0.00$ & $1.00\pm0.00$ & $1.00\pm0.00$ & $1.00\pm0.00$ & $1.00\pm0.00$ \\
       LGR & $0.43\pm0.08$ & $0.32\pm0.02$ & $0.25\pm0.01$ & $0.21\pm0.02$ & $0.19\pm0.04$ & $0.17\pm0.05$ & $0.17\pm0.09$ \\
     DGR+D & $1.00\pm0.00$ & $1.00\pm0.00$ & $1.00\pm0.00$ & $1.00\pm0.00$ & $1.00\pm0.00$ & $1.00\pm0.00$ & $1.00\pm0.00$ \\
     LGR+D & [$0.40\pm0.15$] & [$0.30\pm0.15$] & [$0.23\pm0.17$] & [$0.21\pm0.16$] & [$0.17\pm0.17$] & [$0.15\pm0.18$] & [$0.15\pm0.09$] \\

\bottomrule

\end{tabular}

}
\end{table}

\begin{figure}[h!] 
    \centering
    \subfloat[\ac{Class-IL} Results w/o Augmentation.\label{fig:icvMefed-class-il-noaug}]{\includegraphics[width=0.5\textwidth]{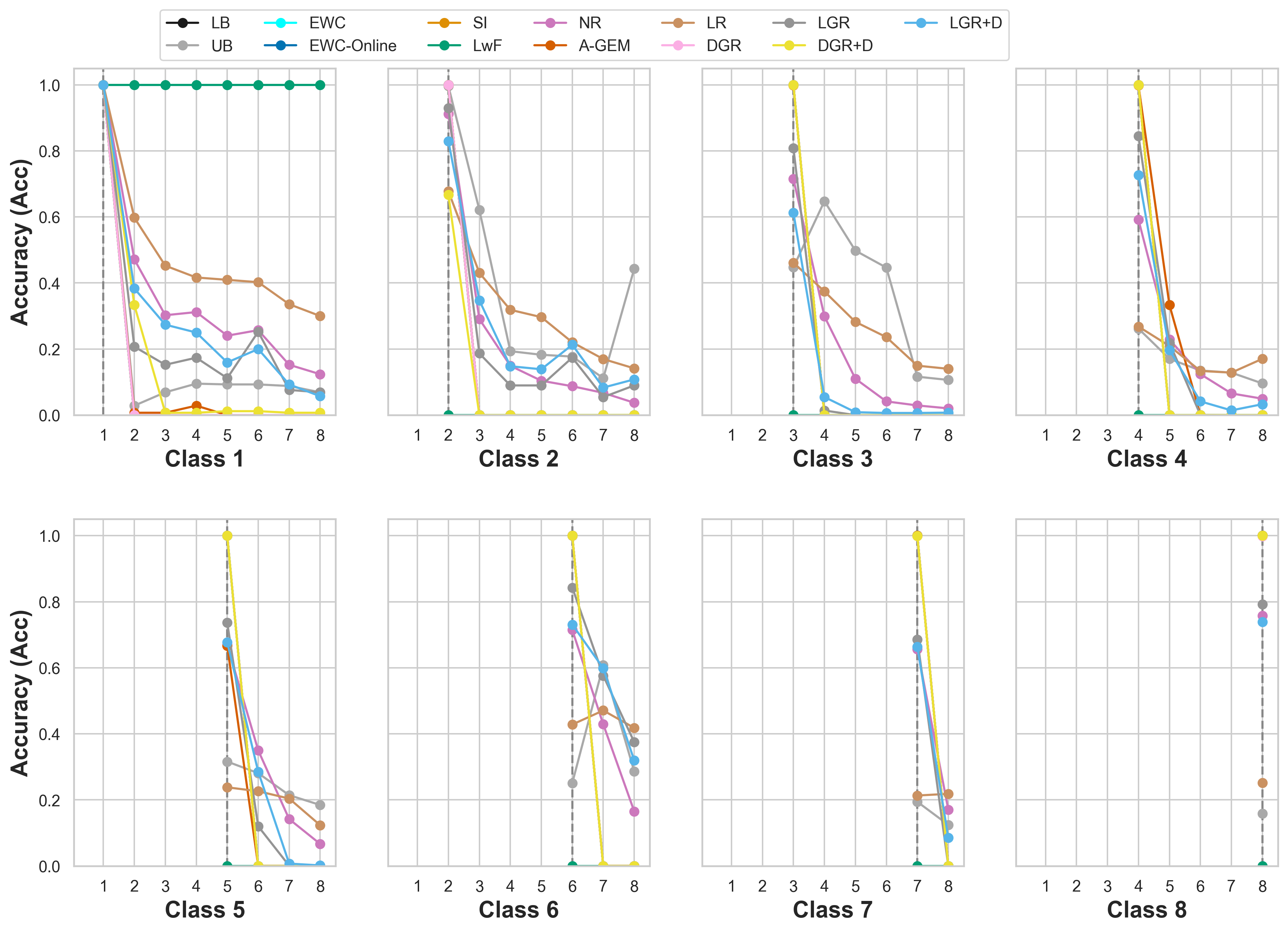}} \hfill
    \subfloat[\ac{Class-IL} Results w/ Augmentation.\label{fig:icvMefed-class-il-aug}]{\includegraphics[width=0.5\textwidth]{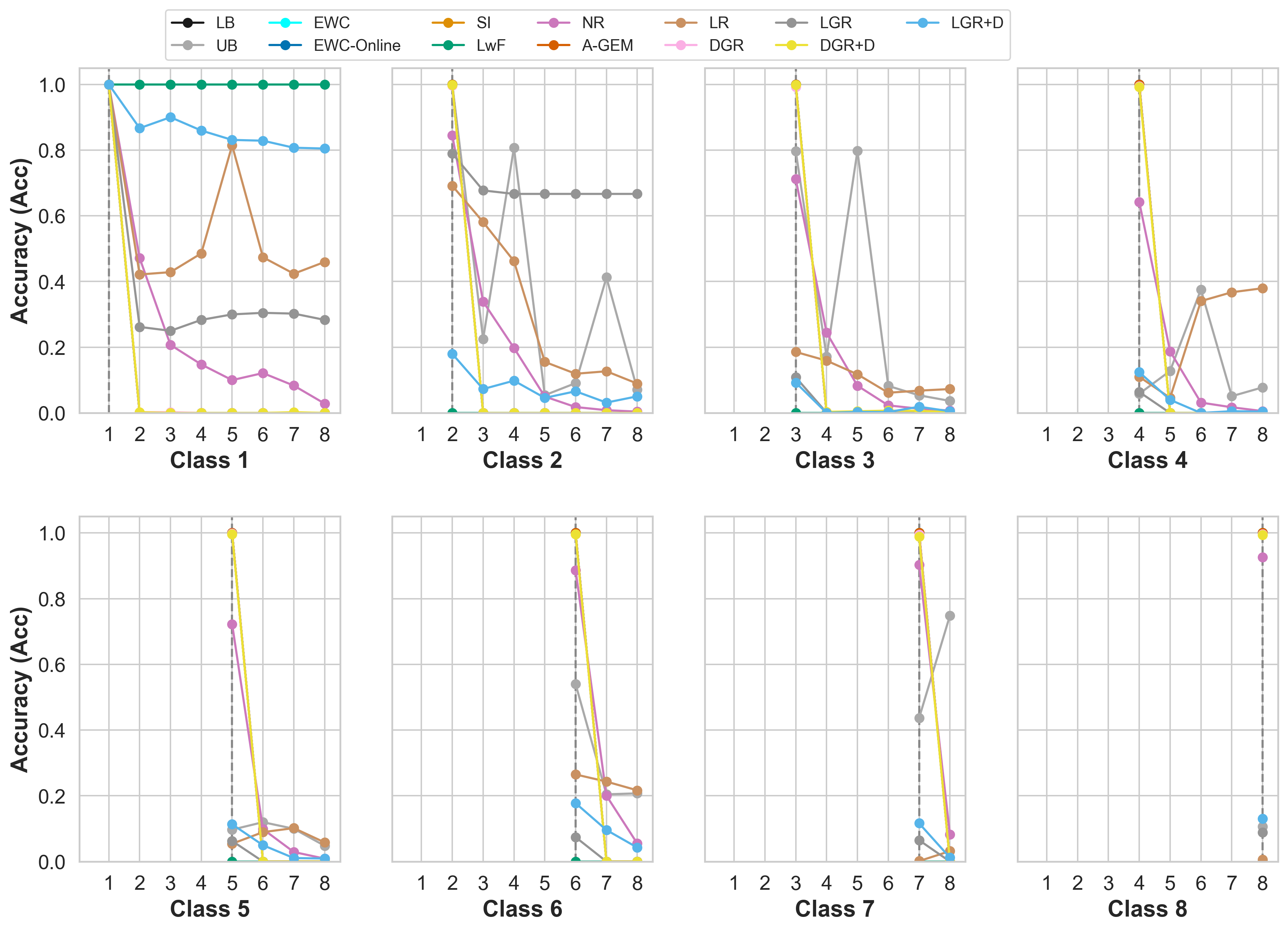}} \\
    \caption[Class-IL Results for iCV-MEFED W/ and W/O Augmentation.]{\ac{Class-IL} results for \acs{iCV-MEFED} (a) without and (b) with augmentation. For each class, test-set accuracy is shown as the learning progresses from when the class is introduced to the end of the overall training procedure.}
    \label{fig:icvMefed-Class} 
    
\end{figure}

\FloatBarrier
\normalsize

\subsubsection{Class-Ordering}
\FloatBarrier
\label{app:classordericv}
\begin{table}[h!]
\centering
\setlength\tabcolsep{2.0pt}

\caption[\acs{Class-IL} \acs{Acc} for \acs{iCV-MEFED} following Orderings O2 and O3.]{\acs{Class-IL} \acs{Acc} for \acs{iCV-MEFED} W/ Data-Augmentation following Orderings O2 and O3. \textbf{Bold} values denote best (highest) while [\textit{bracketed}] denote second-best values for each column.}

\label{tab:class-il-icvMefed-acc-orders}

{
\scriptsize
\begin{tabular}{l|cccccccc}\toprule

\multirow{2}[2]{*}{\makecell[c]{\textbf{Method}}}           & \multicolumn{8}{c}{\textbf{Accuracy following O2}} \\ \cmidrule{2-9}

\multicolumn{1}{c|}{ }                          & 
\multicolumn{1}{c|}{\textbf{Class 1}}            & \multicolumn{1}{c|}{\textbf{Class 2}}          &
\multicolumn{1}{c|}{\textbf{Class 3}}            & \multicolumn{1}{c|}{\textbf{Class 4}}           &
\multicolumn{1}{c|}{\textbf{Class 5}}            & \multicolumn{1}{c|}{\textbf{Class 6}}          &
\multicolumn{1}{c|}{\textbf{Class 7}}            & \multicolumn{1}{c}{\textbf{Class 8}}          \\ \midrule

\multicolumn{9}{c}{\textbf{Baseline Approaches}} \\ \midrule

        LB & \cellcolor{gray!25}$\bm{1.00\pm0.00}$ & $0.50\pm0.00$ & $0.33\pm0.00$ & $0.25\pm0.00$ & $0.20\pm0.00$ & $0.17\pm0.00$ & $0.14\pm0.00$ & $0.12\pm0.00$ \\
        UB & \cellcolor{gray!25}$\bm{1.00\pm0.00}$ & $0.56\pm0.08$ & $0.41\pm0.05$ & $0.29\pm0.06$ & $0.24\pm0.06$ & $0.21\pm0.05$ & [$0.18\pm0.05$] & [$0.15\pm0.05$] \\\midrule

\multicolumn{9}{c}{\textbf{Regularisation-Based Approaches}} \\ \midrule
       EWC & \cellcolor{gray!25}$\bm{1.00\pm0.00}$ & $0.50\pm0.00$ & $0.33\pm0.00$ & $0.25\pm0.00$ & $0.20\pm0.00$ & $0.17\pm0.00$ & $0.14\pm0.00$ & $0.12\pm0.00$ \\
EWC-Online & \cellcolor{gray!25}$\bm{1.00\pm0.00}$ & $0.50\pm0.00$ & $0.33\pm0.00$ & $0.25\pm0.00$ & $0.20\pm0.00$ & $0.17\pm0.00$ & $0.14\pm0.00$ & $0.12\pm0.00$ \\
        SI & \cellcolor{gray!25}$\bm{1.00\pm0.00}$ & $0.50\pm0.00$ & $0.33\pm0.00$ & $0.25\pm0.00$ & $0.20\pm0.00$ & $0.17\pm0.00$ & $0.14\pm0.00$ & $0.12\pm0.00$ \\
       LwF & \cellcolor{gray!25}$\bm{1.00\pm0.00}$ & $0.50\pm0.00$ & $0.33\pm0.00$ & $0.25\pm0.00$ & $0.20\pm0.00$ & $0.17\pm0.00$ & $0.14\pm0.00$ & $0.12\pm0.00$ \\\midrule
\multicolumn{9}{c}{\textbf{Replay-Based Approaches}} \\ \midrule

        NR & \cellcolor{gray!25}$\bm{1.00\pm0.00}$ & \cellcolor{gray!25}$\bm{0.75\pm0.04}$ & \cellcolor{gray!25}$\bm{0.53\pm0.01}$ & \cellcolor{gray!25}$\bm{0.38\pm0.02}$ & \cellcolor{gray!25}$\bm{0.28\pm0.03}$ & [$0.22\pm0.01$] & $0.17\pm0.01$ & $0.14\pm0.01$ \\
     A-GEM & \cellcolor{gray!25}$\bm{1.00\pm0.00}$ & $0.50\pm0.00$ & $0.33\pm0.00$ & $0.25\pm0.00$ & $0.20\pm0.00$ & $0.17\pm0.00$ & $0.14\pm0.00$ & $0.12\pm0.00$ \\
        LR & \cellcolor{gray!25}$\bm{1.00\pm0.00}$ & [$0.65\pm0.10$] & [$0.43\pm0.07$] & [$0.33\pm0.06$] & [$0.27\pm0.05$] & \cellcolor{gray!25}$\bm{0.23\pm0.04}$ & \cellcolor{gray!25}$\bm{0.19\pm0.03}$ & \cellcolor{gray!25}$\bm{0.16\pm0.03}$ \\
       DGR & \cellcolor{gray!25}$\bm{1.00\pm0.00}$ & $0.50\pm0.00$ & $0.33\pm0.00$ & $0.25\pm0.00$ & $0.20\pm0.00$ & $0.17\pm0.00$ & $0.14\pm0.00$ & $0.12\pm0.00$ \\
       LGR & \cellcolor{gray!25}$\bm{1.00\pm0.00}$ & $0.64\pm0.09$ & $0.40\pm0.04$ & $0.27\pm0.03$ & $0.21\pm0.02$ & $0.17\pm0.00$ & $0.15\pm0.01$ & $0.13\pm0.01$ \\
     DGR+D & \cellcolor{gray!25}$\bm{1.00\pm0.00}$ & $0.50\pm0.00$ & $0.33\pm0.00$ & $0.25\pm0.00$ & $0.20\pm0.00$ & $0.17\pm0.00$ & $0.14\pm0.00$ & $0.12\pm0.00$ \\
     LGR+D & \cellcolor{gray!25}$\bm{1.00\pm0.00}$ & $0.60\pm0.07$ & $0.37\pm0.03$ & $0.27\pm0.03$ & $0.22\pm0.02$ & $0.18\pm0.01$ & $0.16\pm0.02$ & $0.13\pm0.02$ \\\midrule

\multirow{2}[2]{*}{\makecell[c]{\textbf{Method}}}           & \multicolumn{8}{c}{\textbf{Accuracy following O3}} \\ \cmidrule{2-9}

\multicolumn{1}{c|}{ }                          & 
\multicolumn{1}{c|}{\textbf{Class 1}}            & \multicolumn{1}{c|}{\textbf{Class 2}}          &
\multicolumn{1}{c|}{\textbf{Class 3}}            & \multicolumn{1}{c|}{\textbf{Class 4}}           &
\multicolumn{1}{c|}{\textbf{Class 5}}            & \multicolumn{1}{c|}{\textbf{Class 6}}          &
\multicolumn{1}{c|}{\textbf{Class 7}}            & \multicolumn{1}{c}{\textbf{Class 8}}          \\ \midrule

\multicolumn{9}{c}{\textbf{Baseline Approaches}} \\ \midrule

        LB & \cellcolor{gray!25}$\bm{1.00\pm0.00}$ & $0.50\pm0.00$ & $0.33\pm0.00$ & $0.25\pm0.00$ & $0.20\pm0.00$ & $0.17\pm0.00$ & $0.14\pm0.00$ & $0.12\pm0.00$ \\
        UB & \cellcolor{gray!25}$\bm{1.00\pm0.00}$ & $0.50\pm0.00$ & $0.35\pm0.02$ & $0.27\pm0.03$ & $0.22\pm0.02$ & [$0.19\pm0.02$] & [$0.17\pm0.04$] & $0.15\pm0.04$ \\\midrule

\multicolumn{9}{c}{\textbf{Regularisation-Based Approaches}} \\ \midrule

       EWC & \cellcolor{gray!25}$\bm{1.00\pm0.00}$ & $0.50\pm0.00$ & $0.33\pm0.00$ & $0.25\pm0.00$ & $0.20\pm0.00$ & $0.17\pm0.00$ & $0.14\pm0.00$ & $0.12\pm0.00$ \\
EWC-Online & \cellcolor{gray!25}$\bm{1.00\pm0.00}$ & $0.50\pm0.00$ & $0.33\pm0.00$ & $0.25\pm0.00$ & $0.20\pm0.00$ & $0.17\pm0.00$ & $0.14\pm0.00$ & $0.12\pm0.00$ \\
        SI & \cellcolor{gray!25}$\bm{1.00\pm0.00}$ & $0.50\pm0.00$ & $0.33\pm0.00$ & $0.25\pm0.00$ & $0.20\pm0.00$ & $0.17\pm0.00$ & $0.14\pm0.00$ & $0.12\pm0.00$ \\
       LwF & \cellcolor{gray!25}$\bm{1.00\pm0.00}$ & $0.50\pm0.00$ & $0.33\pm0.00$ & $0.25\pm0.00$ & $0.20\pm0.00$ & $0.17\pm0.00$ & $0.14\pm0.00$ & $0.12\pm0.00$ \\\midrule
\multicolumn{9}{c}{\textbf{Replay-Based Approaches}} \\ \midrule

        NR & \cellcolor{gray!25}$\bm{1.00\pm0.00}$ & \cellcolor{gray!25}$\bm{0.69\pm0.00}$ & \cellcolor{gray!25}$\bm{0.43\pm0.01}$ & \cellcolor{gray!25}$\bm{0.32\pm0.02}$ & \cellcolor{gray!25}$\bm{0.24\pm0.01}$ & \cellcolor{gray!25}$\bm{0.20\pm0.00}$ & \cellcolor{gray!25}$\bm{0.19\pm0.01}$ & [$0.16\pm0.00$] \\
     A-GEM & \cellcolor{gray!25}$\bm{1.00\pm0.00}$ & $0.50\pm0.00$ & $0.33\pm0.00$ & $0.25\pm0.00$ & $0.20\pm0.00$ & $0.17\pm0.00$ & $0.14\pm0.00$ & $0.12\pm0.00$ \\
        LR & \cellcolor{gray!25}$\bm{1.00\pm0.00}$ & [$0.59\pm0.08$] & [$0.41\pm0.07$] & [$0.29\pm0.03$] & [$0.23\pm0.03$] & \cellcolor{gray!25}$\bm{0.20\pm0.03}$ & \cellcolor{gray!25}$\bm{0.19\pm0.04}$ & \cellcolor{gray!25}$\bm{0.17\pm0.04}$ \\
       DGR & \cellcolor{gray!25}$\bm{1.00\pm0.00}$ & $0.50\pm0.00$ & $0.33\pm0.00$ & $0.25\pm0.00$ & $0.20\pm0.00$ & $0.17\pm0.00$ & $0.14\pm0.00$ & $0.12\pm0.00$ \\
       LGR & \cellcolor{gray!25}$\bm{1.00\pm0.00}$ & $0.56\pm0.02$ & $0.35\pm0.00$ & $0.26\pm0.00$ & $0.21\pm0.01$ & $0.17\pm0.01$ & $0.15\pm0.01$ & $0.13\pm0.00$ \\
     DGR+D & \cellcolor{gray!25}$\bm{1.00\pm0.00}$ & $0.50\pm0.00$ & $0.33\pm0.00$ & $0.25\pm0.00$ & $0.20\pm0.00$ & $0.17\pm0.00$ & $0.14\pm0.00$ & $0.12\pm0.00$ \\
     LGR+D & \cellcolor{gray!25}$\bm{1.00\pm0.00}$ & $0.51\pm0.02$ & $0.33\pm0.00$ & $0.25\pm0.00$ & $0.21\pm0.01$ & $0.17\pm0.00$ & $0.15\pm0.01$ & $0.12\pm0.00$ \\

\bottomrule

\end{tabular}

}
\end{table}

\begin{table}[h!]
\centering
\caption[\acs{Class-IL} \acs{CF} scores for \acs{iCV-MEFED}  following O2 and O3.]{\acs{Class-IL} \acs{CF} scores for \acs{iCV-MEFED} W/ Data-Augmentation following Orderings O2 and O3. \textbf{Bold} values denote best (lowest) while [\textit{bracketed}] denote second-best values for each column.}

\label{tab:class-il-icvMefed-cf-orders}

{
\scriptsize
\setlength\tabcolsep{2.0pt}

\begin{tabular}{l|ccccccc}\toprule

\multirow{2}[2]{*}{\makecell[c]{\textbf{Method}}}           & \multicolumn{7}{c}{\textbf{\acs{CF} following O2}} \\ \cmidrule{2-8}

\multicolumn{1}{c|}{ }                          &  \multicolumn{1}{c|}{\textbf{Class 2}}          &
\multicolumn{1}{c|}{\textbf{Class 3}}            & \multicolumn{1}{c|}{\textbf{Class 4}}           &
\multicolumn{1}{c|}{\textbf{Class 5}}            & \multicolumn{1}{c|}{\textbf{Class 6}}          &
\multicolumn{1}{c|}{\textbf{Class 7}}            & \multicolumn{1}{c}{\textbf{Class 8}}          \\ \midrule
\multicolumn{8}{c}{\textbf{Baseline Approaches}} \\ \midrule

        LB & $1.00\pm0.00$ & $1.00\pm0.00$ & $1.00\pm0.00$ & $1.00\pm0.00$ & $1.00\pm0.00$ & $1.00\pm0.00$ & $1.00\pm0.00$ \\
        UB & $0.62\pm0.08$ & \cellcolor{gray!25}$\bm{0.47\pm0.10}$ & [$0.43\pm0.05$] & [$0.36\pm0.03$] & [$0.30\pm0.02$] & \cellcolor{gray!25}$\bm{0.26\pm0.02}$ & [$0.22\pm0.00$] \\\midrule

\multicolumn{8}{c}{\textbf{Regularisation-Based Approaches}} \\ \midrule

       EWC & $1.00\pm0.00$ & $1.00\pm0.00$ & $1.00\pm0.00$ & $1.00\pm0.00$ & $1.00\pm0.00$ & $1.00\pm0.00$ & $1.00\pm0.00$ \\
EWC-Online & $1.00\pm0.00$ & $1.00\pm0.00$ & $1.00\pm0.00$ & $1.00\pm0.00$ & $1.00\pm0.00$ & $1.00\pm0.00$ & $1.00\pm0.00$ \\
        SI & $1.00\pm0.00$ & $1.00\pm0.00$ & $1.00\pm0.00$ & $1.00\pm0.00$ & $1.00\pm0.00$ & $1.00\pm0.00$ & $1.00\pm0.00$ \\
       LwF & $1.00\pm0.00$ & $1.00\pm0.00$ & $1.00\pm0.00$ & $1.00\pm0.00$ & $1.00\pm0.00$ & $1.00\pm0.00$ & $1.00\pm0.00$  \\\midrule

\multicolumn{8}{c}{\textbf{Replay-Based Approaches}} \\ \midrule

        NR & \cellcolor{gray!25}$\bm{0.52\pm0.08}$ & [$0.50\pm0.10$] & \cellcolor{gray!25}$\bm{0.42\pm0.14}$ & \cellcolor{gray!25}$\bm{0.35\pm0.12}$ & $0.28\pm0.11$ & [$0.28\pm0.10$] & $0.28\pm0.00$ \\
     A-GEM & $1.00\pm0.00$ & $1.00\pm0.00$ & $1.00\pm0.00$ & $1.00\pm0.00$ & $1.00\pm0.00$ & $1.00\pm0.00$ & $1.00\pm0.00$ \\
        LR & [$0.53\pm0.12$] & $0.51\pm0.09$ & $0.44\pm0.06$ & $0.43\pm0.06$ & $0.37\pm0.05$ & \cellcolor{gray!25}$\bm{0.26\pm0.05}$ & \cellcolor{gray!25}$\bm{0.20\pm0.00}$ \\
       DGR & $0.99\pm0.00$ & $0.99\pm0.01$ & $0.99\pm0.01$ & $0.99\pm0.01$ & $0.99\pm0.01$ & $0.98\pm0.02$ & $1.00\pm0.00$ \\
       LGR & $0.55\pm0.16$ & $0.57\pm0.06$ & $0.53\pm0.04$ & $0.54\pm0.14$ & $0.63\pm0.13$ & $0.62\pm0.14$ & $0.64\pm0.00$ \\
     DGR+D & $1.00\pm0.00$ & $1.00\pm0.00$ & $1.00\pm0.00$ & $0.99\pm0.01$ & $0.99\pm0.01$ & $0.99\pm0.01$ & $1.00\pm0.00$ \\
     LGR+D & $0.58\pm0.06$ & $0.58\pm0.05$ & $0.57\pm0.06$ & $0.57\pm0.07$ & $0.66\pm0.05$ & $0.69\pm0.05$ & $0.70\pm0.00$ \\ \midrule

\multirow{2}[2]{*}{\makecell[c]{\textbf{Method}}}           & \multicolumn{7}{c}{\textbf{\acs{CF} following O3}} \\ \cmidrule{2-8}

\multicolumn{1}{c|}{ }                          &  \multicolumn{1}{c|}{\textbf{Class 2}}          &
\multicolumn{1}{c|}{\textbf{Class 3}}            & \multicolumn{1}{c|}{\textbf{Class 4}}           &
\multicolumn{1}{c|}{\textbf{Class 5}}            & \multicolumn{1}{c|}{\textbf{Class 6}}          &
\multicolumn{1}{c|}{\textbf{Class 7}}            & \multicolumn{1}{c}{\textbf{Class 8}}          \\ \midrule
\multicolumn{8}{c}{\textbf{Baseline Approaches}} \\ \midrule

        LB & $1.00\pm0.00$ & $1.00\pm0.00$ & $1.00\pm0.00$ & $1.00\pm0.00$ & $1.00\pm0.00$ & $1.00\pm0.00$ & $1.00\pm0.00$ \\
        UB & $0.54\pm0.05$ & [$0.38\pm0.07$] & [$0.38\pm0.10$] & \cellcolor{gray!25}$\bm{0.31\pm0.08}$ & $0.27\pm0.07$ & $0.23\pm0.06$ & [$0.21\pm0.00$] \\\midrule

\multicolumn{8}{c}{\textbf{Regularisation-Based Approaches}} \\ \midrule

       EWC & $1.00\pm0.00$ & $1.00\pm0.00$ & $1.00\pm0.00$ & $1.00\pm0.00$ & $1.00\pm0.00$ & $1.00\pm0.00$ & $1.00\pm0.00$ \\
EWC-Online & $1.00\pm0.00$ & $1.00\pm0.00$ & $1.00\pm0.00$ & $1.00\pm0.00$ & $1.00\pm0.00$ & $1.00\pm0.00$ & $1.00\pm0.00$ \\
        SI & $1.00\pm0.00$ & $1.00\pm0.00$ & $1.00\pm0.00$ & $1.00\pm0.00$ & $1.00\pm0.00$ & $1.00\pm0.00$ & $1.00\pm0.00$ \\
       LwF & $1.00\pm0.00$ & $1.00\pm0.00$ & $1.00\pm0.00$ & $1.00\pm0.00$ & $1.00\pm0.00$ & $1.00\pm0.00$ & $1.00\pm0.00$ \\\midrule

\multicolumn{8}{c}{\textbf{Replay-Based Approaches}} \\ \midrule
        NR & \cellcolor{gray!25}$\bm{0.51\pm0.07}$ & \cellcolor{gray!25}$\bm{0.36\pm0.11}$ & \cellcolor{gray!25}$\bm{0.40\pm0.14}$ & [$0.33\pm0.13$] & \cellcolor{gray!25}$\bm{0.25\pm0.12}$ & [$0.22\pm0.10$] & \cellcolor{gray!25}$\bm{0.20\pm0.00}$ \\
     A-GEM & $1.00\pm0.00$ & $1.00\pm0.00$ & $1.00\pm0.00$ & $1.00\pm0.00$ & $1.00\pm0.00$ & $1.00\pm0.00$ & $1.00\pm0.00$ \\
        LR & [$0.52\pm0.02$] & $0.42\pm0.05$ & $0.45\pm0.11$ & $0.39\pm0.11$ & [$0.26\pm0.12$] & \cellcolor{gray!25}$\bm{0.19\pm0.11}$ & \cellcolor{gray!25}$\bm{0.20\pm0.00}$ \\
       DGR & $1.00\pm0.00$ & $1.00\pm0.00$ & $1.00\pm0.00$ & $1.00\pm0.00$ & $1.00\pm0.00$ & $1.00\pm0.00$ & $1.00\pm0.00$ \\
       LGR & [$0.52\pm0.23$] & $0.47\pm0.25$ & $0.47\pm0.26$ & $0.47\pm0.27$ & $0.48\pm0.26$ & $0.49\pm0.27$ & $0.48\pm0.00$ \\
     DGR+D & $1.00\pm0.00$ & $1.00\pm0.00$ & $1.00\pm0.00$ & $1.00\pm0.00$ & $1.00\pm0.00$ & $1.00\pm0.00$ & $1.00\pm0.00$ \\
     LGR+D & [$0.52\pm0.23$] & $0.47\pm0.25$ & $0.47\pm0.26$ & $0.47\pm0.27$ & $0.48\pm0.26$ & $0.49\pm0.27$ & $0.48\pm0.00$ \\

\bottomrule

\end{tabular}

}
\end{table}

\begin{figure}[h!]  
    \centering
    \subfloat[\ac{Class-IL} Results following O$2$.\label{fig:icvMefed-Class-il-aug-order-1}]{\includegraphics[width=0.5\textwidth]{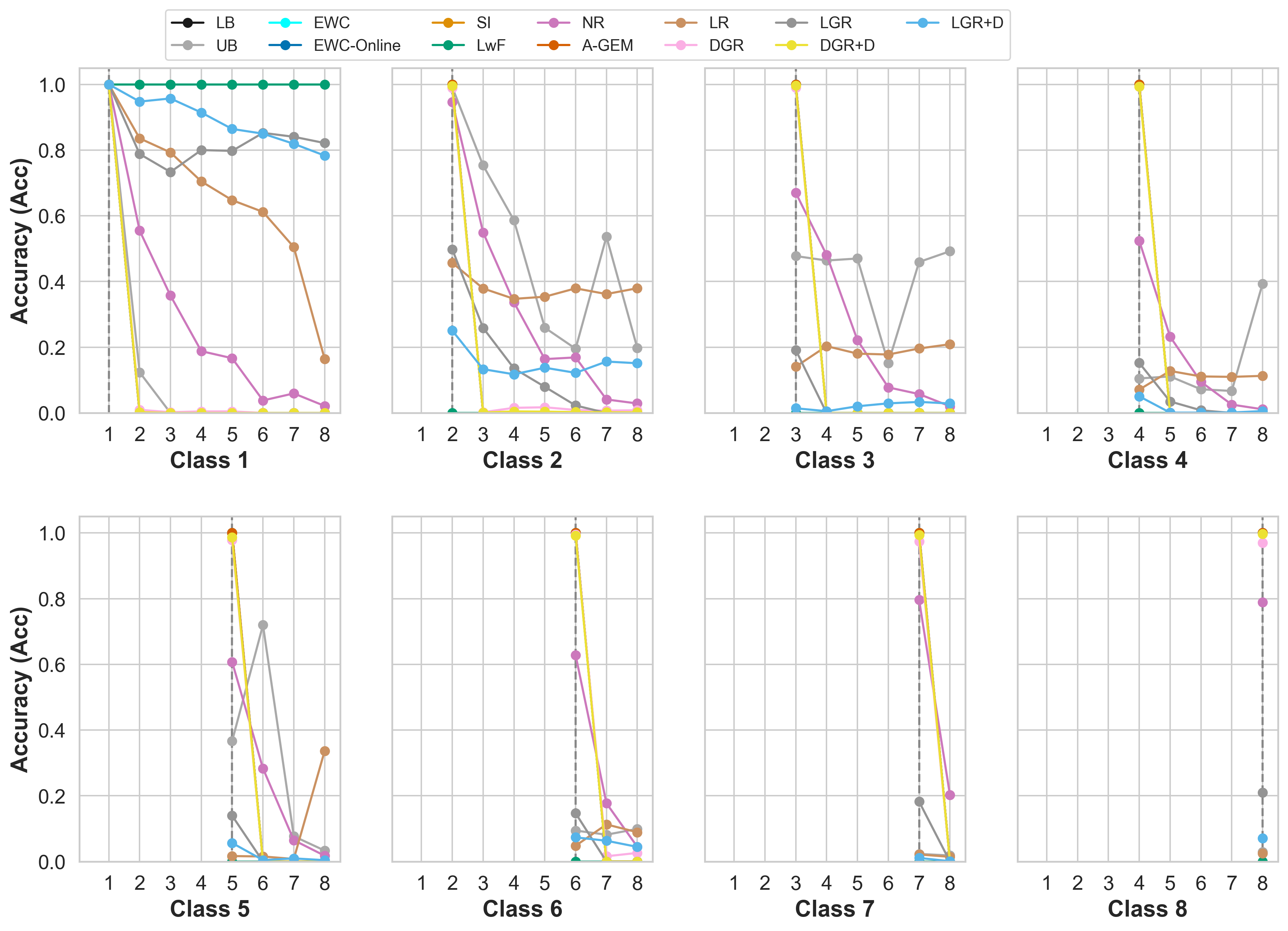}}
    \hfill
    \subfloat[\ac{Class-IL} Results following O$3$.\label{fig:icvMefed-Class-il-aug-order-2}]{\includegraphics[width=0.5\textwidth]{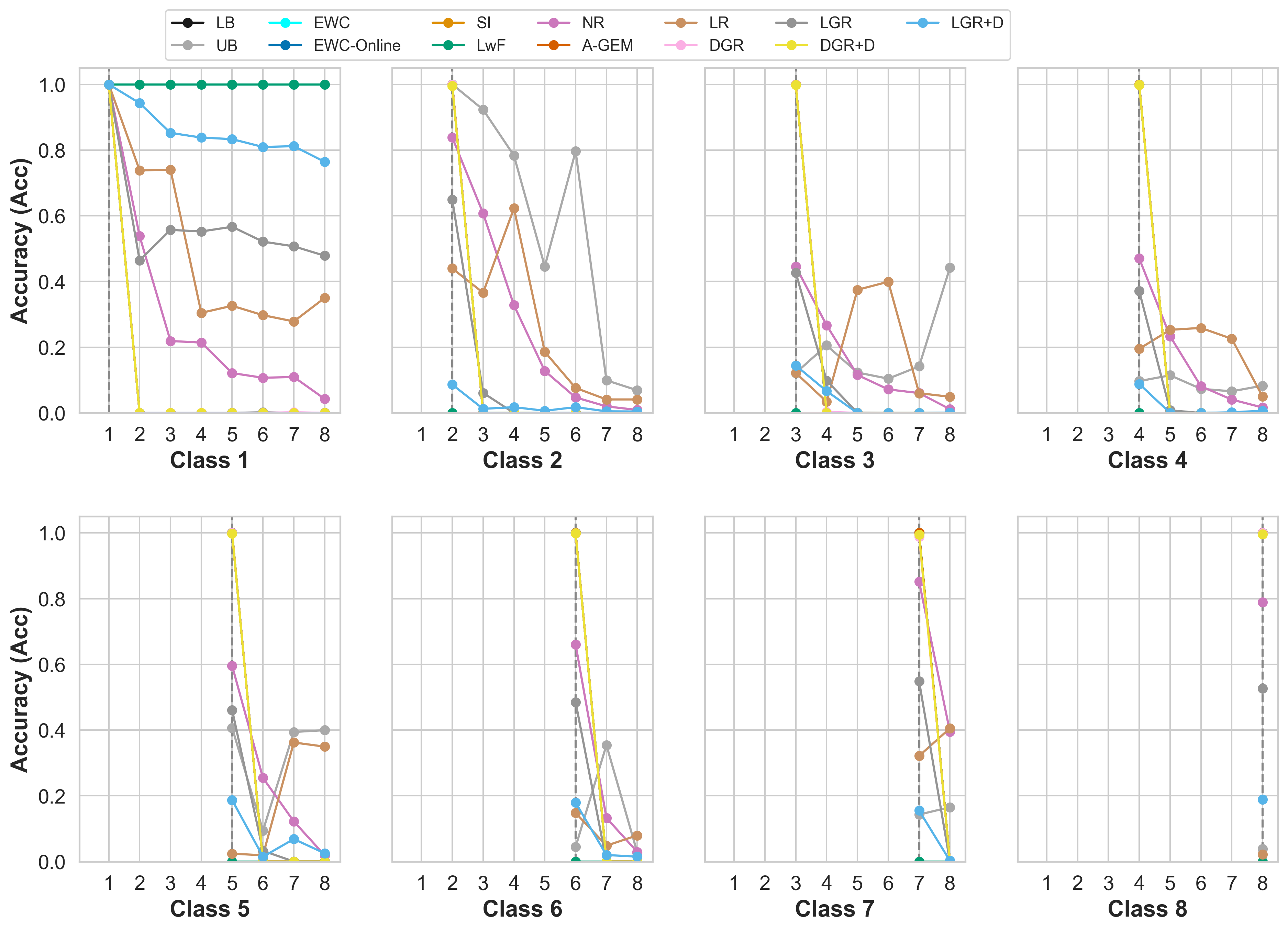}} \\
    \caption[Class-IL Results for \acs{iCV-MEFED} following Order 2 and Order 3.]{\ac{Class-IL} results for \acs{iCV-MEFED} following (a) Order $2$ and (b) Order $3$, respectively. For each class, test-set accuracy is shown as the learning progresses from when the class is introduced to the end of the overall training procedure.}
    \label{fig:icvMefed-classil-orders} 
    
\end{figure}

\FloatBarrier

\normalsize

\subsection{BAUM-1 Dataset}
The BAUM-1 dataset~\citep{ZhalehpourBaum2017} consists of clip-level annotations with $13$ different emotional and mental states for $1184$ video-clips. The video-clips consist of participants spontaneously responding to sequences of images or short videos, designed to elicit specific affective responses in the participants. Using the clip-level label for all the frames in a video-clip, $\approx73K$ sample frames are extracted for $8$ expression categories, namely, \textit{Anger, Surprise, Fearful, Disgust, Happy, Sad, Contempt} and \textit{Neutral} to split the data. Furthermore, $70:30$ training \textit{vs.} test-split is used for training and evaluating the model.

\subsubsection{\acf{Task-IL} Results}
\label{app:taskilbaum}
\begin{table}[h]
\centering
\setlength\tabcolsep{3.5pt}

\caption[\acs{Task-IL} \acs{Acc} for BAUM-1  W/ and W/O Data-Augmentation.]{\acs{Task-IL} \acf{Acc} for BAUM-1  W/ and W/O Data-Augmentation. \textbf{Bold} values denote best (highest) while [\textit{bracketed}] denote second-best values for each column.}
\label{tab:task-il-baum-acc}

{
\scriptsize
\begin{tabular}{l|cccc|cccc}\toprule
\multicolumn{1}{c|}{\textbf{Method}}            & \multicolumn{4}{c|}{\textbf{Accuracy W/O Data-Augmentation}} 
& \multicolumn{4}{c}{\textbf{Accuracy W/ Data-Augmentation}} \\ \midrule

\multicolumn{1}{c|}{ }                          & 
\multicolumn{1}{c|}{\textbf{Task 1}}            & \multicolumn{1}{c|}{\textbf{Task 2}}          &
\multicolumn{1}{c|}{\textbf{Task 3}}            & \multicolumn{1}{c|}{\textbf{Task 4}}           &
\multicolumn{1}{c|}{\textbf{Task 1}}            & \multicolumn{1}{c|}{\textbf{Task 2}}          &
\multicolumn{1}{c|}{\textbf{Task 3}}            & \multicolumn{1}{c}{\textbf{Task 4}}          \\ \midrule

\multicolumn{9}{c}{\textbf{Baseline Approaches}} \\ \midrule

LB & \cellcolor{gray!25}$\bm{0.88\pm0.00}$ & $0.89\pm0.04$ & $0.85\pm0.03$ & $0.73\pm0.03$ & \cellcolor{gray!25}$\bm{0.82\pm0.00}$ & $0.84\pm0.01$ & $0.83\pm0.02$ & $0.75\pm0.00$ \\
UB & \cellcolor{gray!25}$\bm{0.88\pm0.01}$ & \cellcolor{gray!25}$\bm{0.95\pm0.00}$ & [$0.95\pm0.00$] & [$0.95\pm0.01$] & \cellcolor{gray!25}$\bm{0.82\pm0.01}$ & [$0.87\pm0.00$] & \cellcolor{gray!25}$\bm{0.89\pm0.01}$ & \cellcolor{gray!25}$\bm{0.89\pm0.00}$ \\ \midrule

\multicolumn{9}{c}{\textbf{Regularisation-Based Approaches}} \\ \midrule
EWC  & \cellcolor{gray!25}$\bm{0.88\pm0.00}$ & $0.92\pm0.01$ & $0.93\pm0.01$ & $0.91\pm0.03$ & [$0.81\pm0.00$] & $0.83\pm0.01$ & $0.84\pm0.03$ & $0.84\pm0.00$ \\
EWC-Online  & \cellcolor{gray!25}$\bm{0.88\pm0.00}$ & [$0.93\pm0.00$] & $0.92\pm0.01$ & $0.90\pm0.01$ & [$0.81\pm0.00$] & $0.86\pm0.02$ & $0.86\pm0.01$ & $0.85\pm0.00$ \\
SI  & \cellcolor{gray!25}$\bm{0.88\pm0.00}$ & $0.90\pm0.02$ & $0.87\pm0.02$ & $0.86\pm0.03$ & \cellcolor{gray!25}$\bm{0.82\pm0.00}$ & [$0.87\pm0.02$] & $0.83\pm0.02$ & $0.84\pm0.00$ \\
LwF & \cellcolor{gray!25}$\bm{0.88\pm0.00}$ & [$0.93\pm0.01$] & $0.94\pm0.01$ & $0.93\pm0.01$ & \cellcolor{gray!25}$\bm{0.82\pm0.00}$ & [$0.87\pm0.00$] & $0.88\pm0.01$ & $0.86\pm0.01$ \\\midrule
\multicolumn{9}{c}{\textbf{Replay-Based Approaches}} \\ \midrule
NR  & \cellcolor{gray!25}$\bm{0.88\pm0.00}$ & \cellcolor{gray!25}$\bm{0.95\pm0.00}$ & $0.93\pm0.01$ & $0.90\pm0.01$ & \cellcolor{gray!25}$\bm{0.82\pm0.01}$ & \cellcolor{gray!25}$\bm{0.88\pm0.00}$ & $0.82\pm0.01$ & $0.81\pm0.00$ \\
A-GEM  & \cellcolor{gray!25}$\bm{0.88\pm0.00}$ & \cellcolor{gray!25}$\bm{0.95\pm0.00}$ & \cellcolor{gray!25}$\bm{0.96\pm0.00}$ & \cellcolor{gray!25}$\bm{0.96\pm0.00}$ & [$0.81\pm0.00$] & [$0.87\pm0.02$] & [$0.88\pm0.01$] & [$0.88\pm0.01$] \\
LR & [$0.85\pm0.00$] & $0.88\pm0.00$ & $0.86\pm0.00$ & $0.85\pm0.00$ & [$0.81\pm0.00$] & $0.84\pm0.00$ & $0.80\pm0.01$ & $0.78\pm0.00$ \\
DGR  & \cellcolor{gray!25}$\bm{0.88\pm0.01}$ & $0.89\pm0.00$ & $0.90\pm0.00$ & $0.87\pm0.01$ & [$0.81\pm0.00$] & $0.86\pm0.00$ & $0.85\pm0.01$ & $0.84\pm0.00$ \\
LGR  & [$0.85\pm0.00$] & $0.86\pm0.00$ & $0.86\pm0.00$ & $0.85\pm0.00$ & \cellcolor{gray!25}$\bm{0.82\pm0.00}$ & $0.83\pm0.00$ & $0.84\pm0.00$ & $0.84\pm0.00$ \\
DGR+D  & \cellcolor{gray!25}$\bm{0.88\pm0.00}$ & $0.90\pm0.00$ & $0.91\pm0.01$ & $0.90\pm0.01$ & [$0.81\pm0.00$] & [$0.87\pm0.01$] & $0.85\pm0.00$ & $0.84\pm0.00$ \\
LGR+D  & $0.84\pm0.01$ & $0.86\pm0.01$ & $0.87\pm0.01$ & $0.86\pm0.01$ & [$0.81\pm0.00$] & $0.83\pm0.00$ & $0.84\pm0.00$ & $0.84\pm0.00$ \\

\bottomrule

\end{tabular}

}
\end{table}

\begin{table}[h]
\centering
\caption[\acs{Task-IL} \acs{CF} scores for BAUM-1  W/ and W/O Data-Augmentation.]{\acs{CF} scores for \acs{Task-IL} on BAUM-1  W/ and W/O Data-Augmentation. \textbf{Bold} values denote best (lowest) while [\textit{bracketed}] denote second-best values for each column. }
\label{tab:task-il-baum-cf}

{
\scriptsize

\begin{tabular}{l|ccc|ccc}\toprule
\multicolumn{1}{c|}{\textbf{Method}}            & \multicolumn{3}{c|}{\textbf{\acs{CF} W/O Data-Augmentation}} 
& \multicolumn{3}{c}{\textbf{\acs{CF}  W/ Data-Augmentation}} \\ \midrule

\multicolumn{1}{c|}{ }                          & \multicolumn{1}{c|}{\textbf{Task 2}}          &
\multicolumn{1}{c|}{\textbf{Task 3}}            & \multicolumn{1}{c|}{\textbf{Task 4}}           & 
\multicolumn{1}{c|}{\textbf{Task 2}}          &
\multicolumn{1}{c|}{\textbf{Task 3}}            & \multicolumn{1}{c}{\textbf{Task 4}}          \\ \cmidrule{2-7}
\multicolumn{7}{c}{\textbf{Baseline Approaches}} \\ \midrule

LB&  $0.30\pm0.10$ &  $0.24\pm0.05$ &  $0.09\pm0.07$ &  $0.13\pm0.04$ &  $0.06\pm0.00$ &  $0.04\pm0.02$ \\
UB& \cellcolor{gray!25}$\bm{-0.06\pm0.03}$ & \cellcolor{gray!25}$\bm{-0.05\pm0.01}$ & \cellcolor{gray!25}$\bm{-0.04\pm0.01}$ & \cellcolor{gray!25}$\bm{-0.06\pm0.02}$ & \cellcolor{gray!25}$\bm{-0.06\pm0.00}$ & [$-0.01\pm0.01$] \\\midrule

\multicolumn{7}{c}{\textbf{Regularisation-Based Approaches}} \\ \midrule
EWC &  $0.04\pm0.01$ &  $0.07\pm0.04$ &  $0.01\pm0.02$ &  $0.10\pm0.10$ &  $0.04\pm0.00$ &  $0.06\pm0.01$ \\
EWC-Online &  $0.05\pm0.03$ &  $0.08\pm0.02$ & $0.00\pm0.01$ &  $0.05\pm0.03$ &  $0.04\pm0.01$ &  $0.01\pm0.02$ \\
SI &  $0.24\pm0.06$ &  $0.20\pm0.06$ &  $0.07\pm0.03$ &  $0.16\pm0.06$ &  $0.08\pm0.03$ &  $0.01\pm0.02$ \\
LwF&  $0.02\pm0.01$ &  $0.03\pm0.01$ &  $0.01\pm0.01$ &  $0.02\pm0.00$ &  $0.02\pm0.00$ &  $0.00\pm0.00$ \\\midrule

\multicolumn{7}{c}{\textbf{Replay-Based Approaches}} \\ \midrule

NR &  $0.05\pm0.02$ &  $0.08\pm0.01$ & [$-0.03\pm0.01$] &  $0.18\pm0.01$ &  $0.11\pm0.01$ & [$-0.01\pm0.01$] \\
A-GEM & [$-0.02\pm0.00$] & [$-0.01\pm0.01$] & $-0.02\pm0.01$ & [$-0.01\pm0.04$] & [$-0.01\pm0.02$] & \cellcolor{gray!25}$\bm{-0.02\pm0.01}$ \\
LR&  $0.10\pm0.01$ &  $0.07\pm0.00$ & $0.00\pm0.00$ &  $0.14\pm0.03$ &  $0.12\pm0.01$ & [$-0.01\pm0.01$] \\
DGR &  $0.12\pm0.01$ &  $0.13\pm0.02$ &  $0.07\pm0.01$ &  $0.07\pm0.01$ &  $0.05\pm0.02$ &  $0.00\pm0.00$ \\
LGR &  $0.10\pm0.01$ &  $0.08\pm0.00$ &  $0.04\pm0.01$ &  $0.01\pm0.01$ &  $0.01\pm0.01$ &  $0.00\pm0.00$ \\
DGR+D &  $0.12\pm0.03$ &  $0.11\pm0.02$ &  $0.06\pm0.02$ &  $0.09\pm0.01$ &  $0.07\pm0.01$ &  $0.00\pm0.00$ \\
LGR+D &  $0.04\pm0.01$ &  $0.03\pm0.01$ &  $0.01\pm0.00$ &  $0.02\pm0.01$ &  $0.01\pm0.01$ &  $0.00\pm0.00$ \\

\bottomrule

\end{tabular}

}
\end{table}

\begin{figure}  
    \centering
    \subfloat[\ac{Task-IL} Results w/o Augmentation.\label{fig:Baum-task-il-noaug}]{\includegraphics[width=0.5\textwidth]{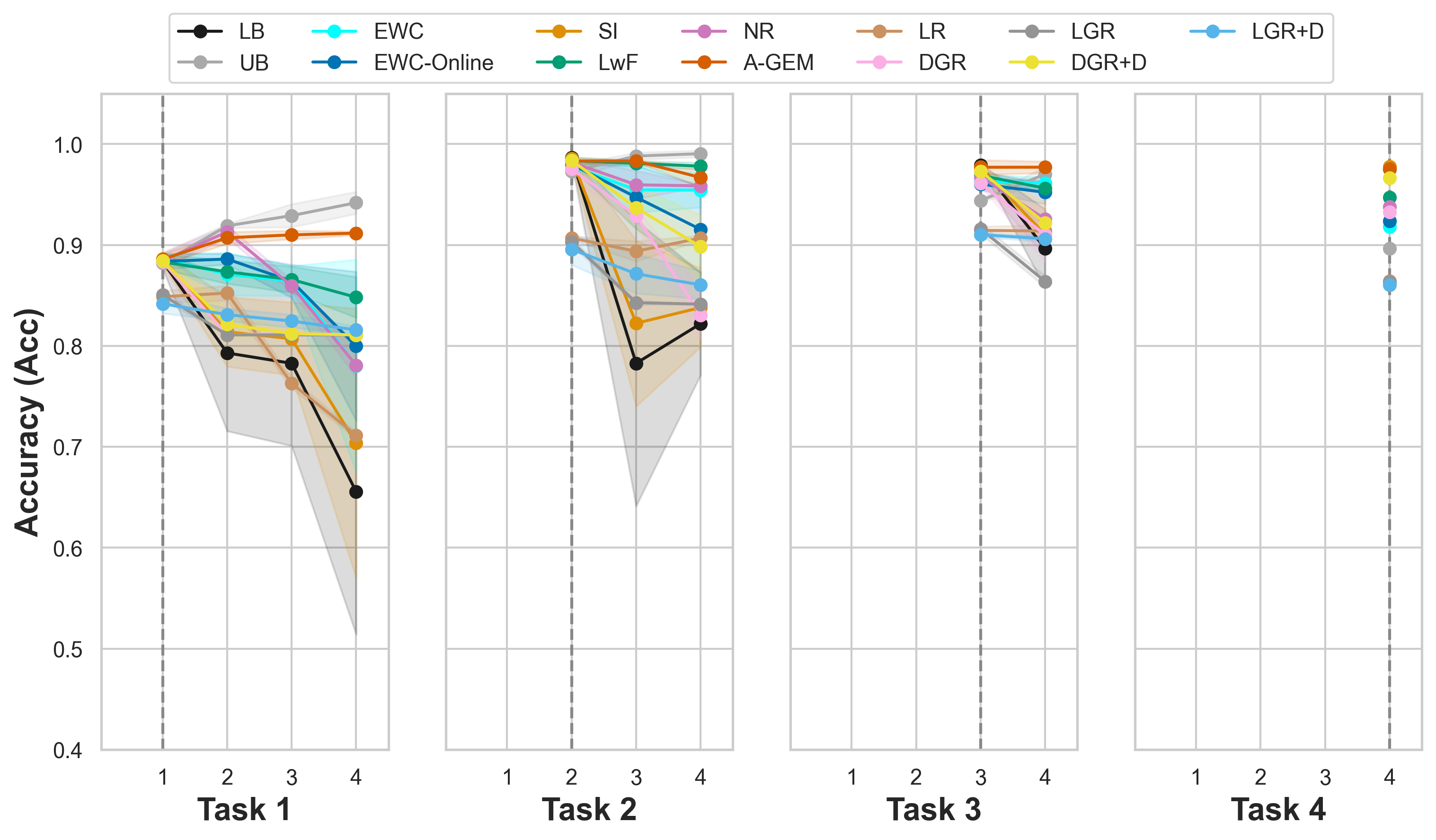}} \hfill
    \subfloat[\ac{Task-IL} Results w/ Augmentation.\label{fig:Baum-taskil-aug}]{\includegraphics[width=0.5\textwidth]{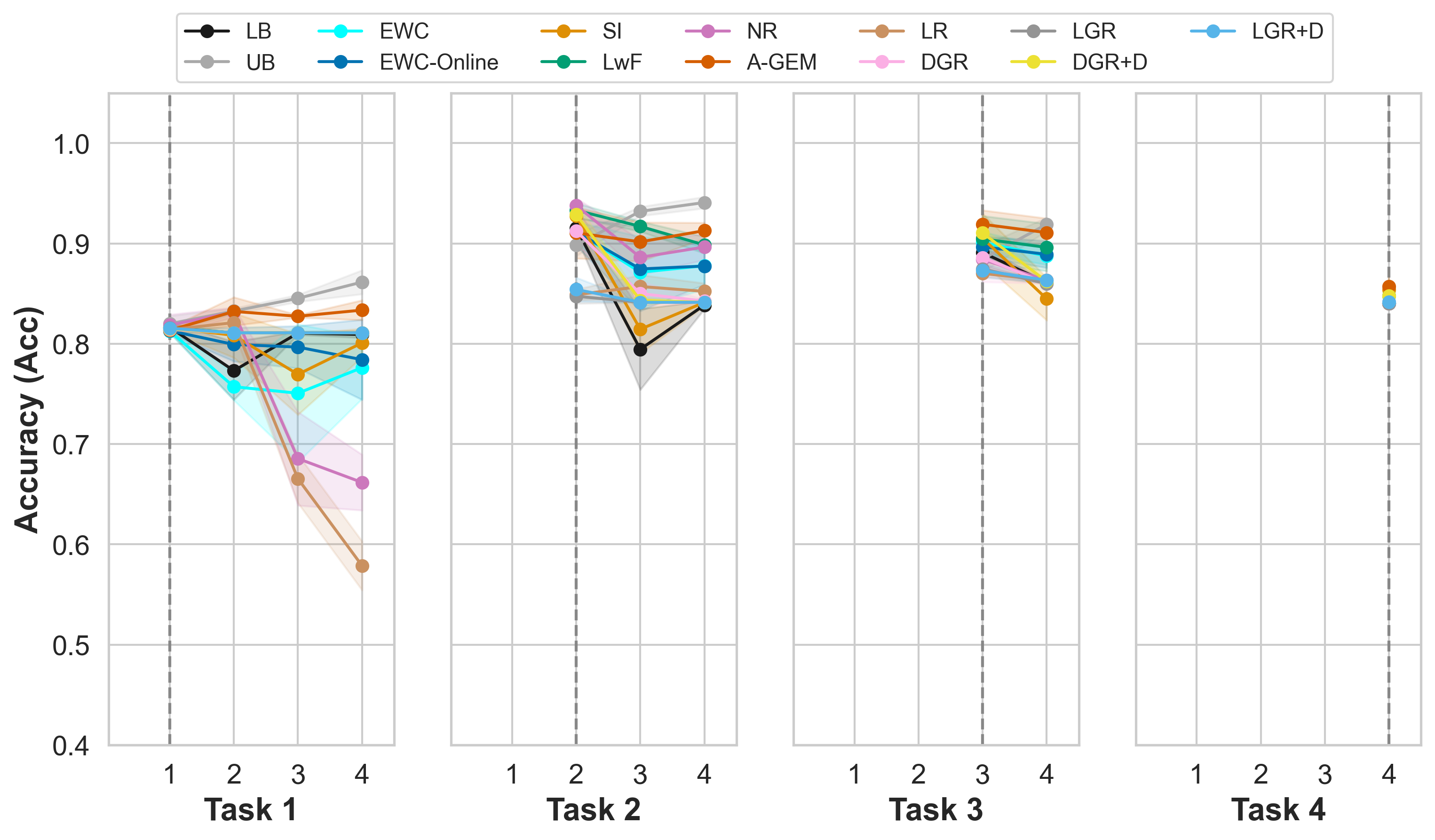}} \\
    \caption[Task-IL Results for BAUM-1  W/ and W/O Augmentation.]{\ac{Task-IL} results for BAUM-1 (a) without and (b) with augmentation. For each task, test-set accuracy is shown as the learning progresses from when the task is introduced to the end of the overall training procedure.}
    \label{fig:Baum-taskil} 
    
\end{figure}
\FloatBarrier

\subsubsection{Task-Ordering}
\label{app:taskorderbaum}
\begin{table}[h!]
\centering
\setlength\tabcolsep{2.0pt}

\caption[\acs{Task-IL} \acs{Acc} for BAUM-1 following Orderings O2 and O3.]{\acs{Task-IL} \acs{Acc} for BAUM-1 W/ Data-Augmentation following Orderings O2 and O3. \textbf{Bold} values denote best (highest) while [\textit{bracketed}] denote second-best values for each column.}

\label{tab:task-il-baum-orders-acc}

{
\scriptsize
\begin{tabular}{l|cccc|cccc}\toprule
\multicolumn{1}{c|}{\textbf{Method}}            & \multicolumn{4}{c|}{\textbf{Accuracy following O2}} 
& \multicolumn{4}{c}{\textbf{Accuracy following O3}} \\  \cmidrule{2-9}

\multicolumn{1}{c|}{ }                          & 
\multicolumn{1}{c|}{\textbf{Task 1}}            & \multicolumn{1}{c|}{\textbf{Task 2}}          &
\multicolumn{1}{c|}{\textbf{Task 3}}            & \multicolumn{1}{c|}{\textbf{Task 4}}           &
\multicolumn{1}{c|}{\textbf{Task 1}}            & \multicolumn{1}{c|}{\textbf{Task 2}}          &
\multicolumn{1}{c|}{\textbf{Task 3}}            & \multicolumn{1}{c}{\textbf{Task 4}}          \\ \midrule

\multicolumn{9}{c}{\textbf{Baseline Approaches}} \\ \midrule
LB & [$0.81\pm0.00$] & $0.84\pm0.01$ & $0.80\pm0.04$ & $0.77\pm0.01$ & [$0.81\pm0.01$] & $0.81\pm0.03$ & $0.77\pm0.07$ & $0.83\pm0.02$ \\
UB & \cellcolor{gray!25}$\bm{0.82\pm0.00}$ & [$0.86\pm0.02$] & \cellcolor{gray!25}$\bm{0.87\pm0.02}$ & \cellcolor{gray!25}$\bm{0.87\pm0.02}$ & \cellcolor{gray!25}$\bm{0.82\pm0.01}$ & $0.85\pm0.02$ & \cellcolor{gray!25}$\bm{0.88\pm0.01}$ & \cellcolor{gray!25}$\bm{0.89\pm0.01}$ \\
 \midrule

\multicolumn{9}{c}{\textbf{Regularisation-Based Approaches}} \\ \midrule

EWC  & [$0.81\pm0.00$] & [$0.86\pm0.02$ ]& $0.85\pm0.01$ & $0.85\pm0.01$ & \cellcolor{gray!25}$\bm{0.82\pm0.01}$ & $0.85\pm0.03$ & $0.85\pm0.01$ & $0.83\pm0.02$ \\
EWC-Online  & [$0.81\pm0.00$] & \cellcolor{gray!25}$\bm{0.87\pm0.02}$ & $0.85\pm0.02$ & [$0.86\pm0.01$] & \cellcolor{gray!25}$\bm{0.82\pm0.01}$ & $0.85\pm0.01$ & $0.85\pm0.01$ & $0.83\pm0.01$ \\
SI  & [$0.81\pm0.00$] & $0.85\pm0.02$ & $0.76\pm0.05$ & $0.77\pm0.03$ & \cellcolor{gray!25}$\bm{0.82\pm0.01}$ & $0.84\pm0.03$ & $0.73\pm0.07$ & $0.83\pm0.01$ \\
LwF  & [$0.81\pm0.00$] & \cellcolor{gray!25}$\bm{0.87\pm0.00}$ & \cellcolor{gray!25}$\bm{0.87\pm0.01}$ & [$0.86\pm0.01$] & \cellcolor{gray!25}$\bm{0.82\pm0.01}$ & \cellcolor{gray!25}$\bm{0.87\pm0.01}$ & [$0.87\pm0.01$] & $0.86\pm0.01$ \\ \midrule
\multicolumn{9}{c}{\textbf{Replay-Based Approaches}} \\ \midrule

NR & [$0.81\pm0.00$] & [$0.86\pm0.02$] & $0.76\pm0.10$ & $0.77\pm0.06$ & \cellcolor{gray!25}$\bm{0.82\pm0.01}$ & \cellcolor{gray!25}$\bm{0.87\pm0.01}$ & $0.81\pm0.02$ & $0.80\pm0.01$ \\
A-GEM  & [$0.81\pm0.00$] & $0.85\pm0.02$ & [$0.86\pm0.02$] & $0.85\pm0.02$ & \cellcolor{gray!25}$\bm{0.82\pm0.01}$ & \cellcolor{gray!25}$\bm{0.87\pm0.01}$ & \cellcolor{gray!25}$\bm{0.88\pm0.02}$ & [$0.88\pm0.01$] \\
LR  & \cellcolor{gray!25}$\bm{0.82\pm0.00}$ & $0.84\pm0.01$ & $0.80\pm0.01$ & $0.80\pm0.01$ & [$0.81\pm0.00$] & $0.84\pm0.00$ & $0.79\pm0.00$ & $0.79\pm0.00$ \\
DGR  & [$0.81\pm0.00$] & $0.83\pm0.00$ & $0.84\pm0.00$ & $0.84\pm0.00$ & [$0.81\pm0.00$] & $0.85\pm0.00$ & $0.85\pm0.00$ & $0.84\pm0.00$ \\
LGR  & \cellcolor{gray!25}$\bm{0.82\pm0.01}$ & $0.83\pm0.00$ & $0.84\pm0.00$ & $0.84\pm0.00$ & [$0.81\pm0.00$] & $0.83\pm0.00$ & $0.84\pm0.00$ & $0.84\pm0.00$ \\
DGR+D  & [$0.81\pm0.00$] & [$0.86\pm0.01$] & $0.85\pm0.00$ & $0.85\pm0.00$ & [$0.81\pm0.00$] & [$0.86\pm0.00$] & $0.85\pm0.00$ & $0.84\pm0.00$ \\
LGR+D  & \cellcolor{gray!25}$\bm{0.82\pm0.00}$ & $0.83\pm0.00$ & $0.84\pm0.00$ & $0.84\pm0.00$ & [$0.81\pm0.00$] & $0.83\pm0.00$ & $0.84\pm0.00$ & $0.84\pm0.00$ \\

\bottomrule

\end{tabular}

}
\end{table}

\begin{table}[h!]
\centering
\caption[\acs{Task-IL} \acs{CF} scores for BAUM-1 following O2 and O3.]{\acs{Task-IL} \acs{CF} scores for BAUM-1 W/ Data-Augmentation following Orderings O2 and O3. \textbf{Bold} values denote best (lowest) while [\textit{bracketed}] denote second-best values for each column.}

\label{tab:task-il-baum-orders-cf}

{
\scriptsize
\setlength\tabcolsep{2.0pt}

\begin{tabular}{l|ccc|ccc}\toprule
\multicolumn{1}{c|}{\textbf{Method}}            & \multicolumn{3}{c|}{\textbf{CF following O2}} 
& \multicolumn{3}{c}{\textbf{CF following O3}} \\ \cmidrule{2-7}
\multicolumn{1}{c|}{ }                          & 
\multicolumn{1}{c|}{\textbf{Task 2}}          &
\multicolumn{1}{c|}{\textbf{Task 3}}            & \multicolumn{1}{c|}{\textbf{Task 4}}           &
\multicolumn{1}{c|}{\textbf{Task 2}}          &
\multicolumn{1}{c|}{\textbf{Task 3}}            & \multicolumn{1}{c}{\textbf{Task 4}}          \\ \midrule

\multicolumn{7}{c}{\textbf{Baseline Approaches}} \\ \midrule
LB               &  $0.15\pm0.17$ &  $0.03\pm0.04$ & $-0.00\pm0.00$ &  $0.31\pm0.20$ &  $0.08\pm0.05$ &  $0.10\pm0.07$ \\
UB               & \cellcolor{gray!25}$\bm{-0.03\pm0.03}$ & \cellcolor{gray!25}$\bm{-0.03\pm0.02}$ & \cellcolor{gray!25}$\bm{-0.01\pm0.01}$ & \cellcolor{gray!25}$\bm{-0.09\pm0.01}$ & \cellcolor{gray!25}$\bm{-0.08\pm0.01}$ & \cellcolor{gray!25}$\bm{-0.02\pm0.02}$ \\\midrule

\multicolumn{7}{c}{\textbf{Regularisation-Based Approaches}} \\ \midrule

EWC  &  $0.03\pm0.04$ &  $0.02\pm0.02$ & $0.00\pm0.00$ &  $0.07\pm0.03$ &  $0.06\pm0.06$ &  $0.04\pm0.05$ \\
EWC-Online  &  $0.06\pm0.07$ &  [$0.01\pm0.01$] & $0.00\pm0.00$ &  $0.06\pm0.04$ &  $0.06\pm0.07$ &  $0.03\pm0.03$ \\
SI  &  $0.32\pm0.14$ &  $0.19\pm0.05$ & $0.00\pm0.00$ &  $0.42\pm0.21$ &  $0.07\pm0.03$ &  $0.05\pm0.05$ \\
LwF  &  $0.03\pm0.03$ &  $0.02\pm0.02$ & $0.00\pm0.00$ &  $0.02\pm0.01$ &  $0.03\pm0.01$ &  $0.00\pm0.00$ \\\midrule

\multicolumn{7}{c}{\textbf{Replay-Based Approaches}} \\ \midrule

NR &  $0.31\pm0.22$ &  $0.18\pm0.09$ & $0.00\pm0.01$ &  $0.17\pm0.03$ &  $0.12\pm0.01$ & $0.00\pm0.01$ \\
A-GEM  & [$-0.01\pm0.02$] &  [$0.01\pm0.01$] & \cellcolor{gray!25}$\bm{-0.01\pm0.01}$ & [$-0.01\pm0.02$] & [$-0.01\pm0.02$] & [$-0.01\pm0.00$] \\
LR  &  $0.13\pm0.02$ &  $0.10\pm0.01$ & $0.00\pm0.00$ &  $0.16\pm0.01$ &  $0.10\pm0.01$ & [$-0.01\pm0.00$] \\
DGR  &  $0.01\pm0.02$ &  [$0.01\pm0.01$] &  $0.00\pm0.00$ &  $0.05\pm0.01$ &  $0.04\pm0.01$ &  $0.00\pm0.00$ \\
LGR  &  $0.01\pm0.01$ &  [$0.01\pm0.01$] &  $0.01\pm0.01$ &  $0.01\pm0.00$ &  $0.01\pm0.00$ &  $0.00\pm0.00$ \\
DGR+D  &  $0.07\pm0.01$ &  $0.05\pm0.01$ &  $0.00\pm0.00$ &  $0.06\pm0.02$ &  $0.04\pm0.01$ &  $0.00\pm0.00$ \\
LGR+D  &  $0.01\pm0.01$ &  [$0.01\pm0.01$] &  $0.01\pm0.00$ &  $0.01\pm0.00$ &  $0.01\pm0.00$ &  $0.00\pm0.00$ \\

\bottomrule

\end{tabular}

}
\end{table}
\begin{figure}[h!]
    \centering
    \subfloat[\ac{Task-IL} Results following O$2$.\label{fig:Baum-task-il-aug-order-1}]{\includegraphics[width=0.5\textwidth]{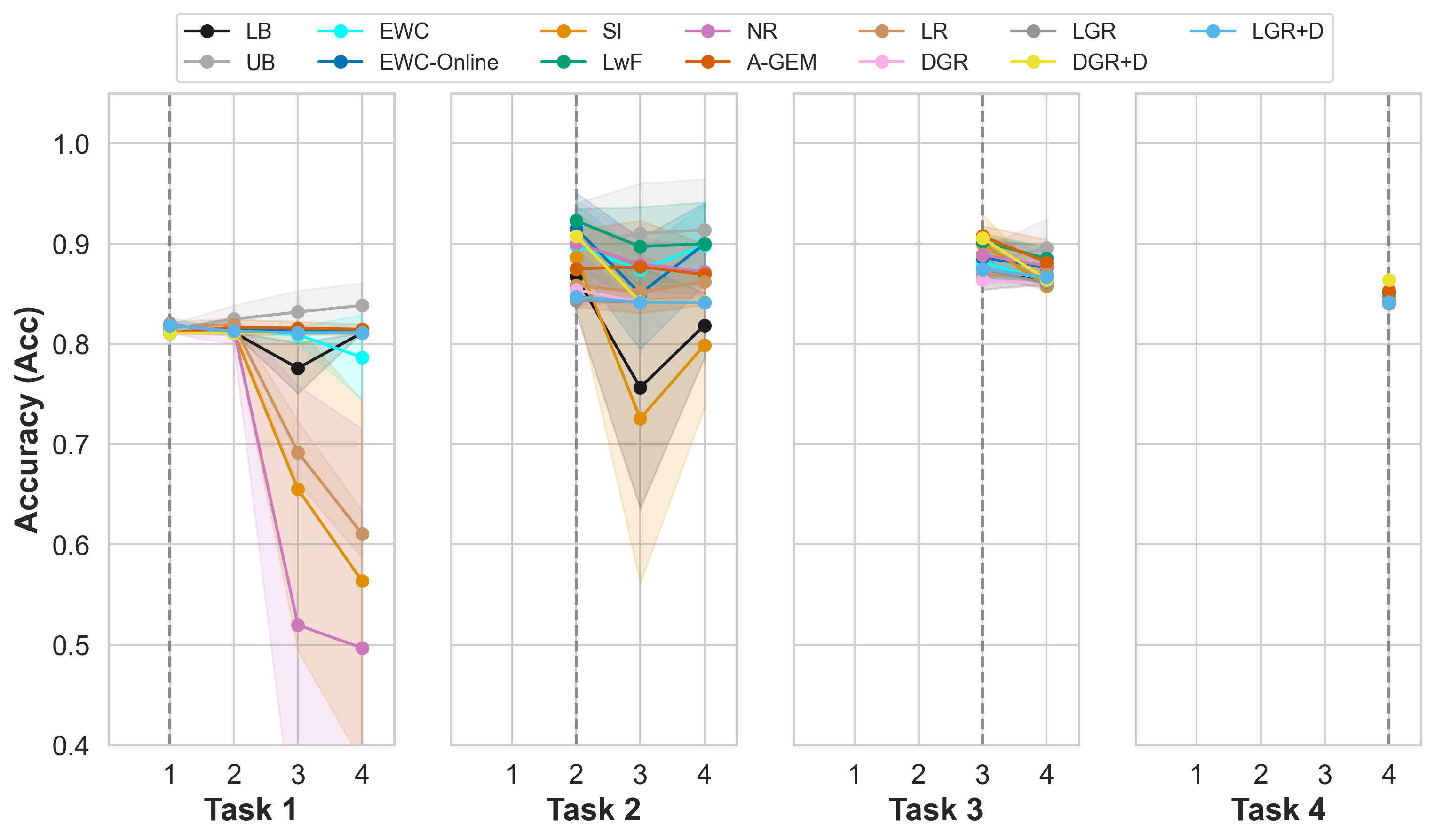}} \hfill
    \subfloat[\ac{Task-IL} Results following O$3$.\label{fig:Baum-task-il-aug-order-2}]{\includegraphics[width=0.5\textwidth]{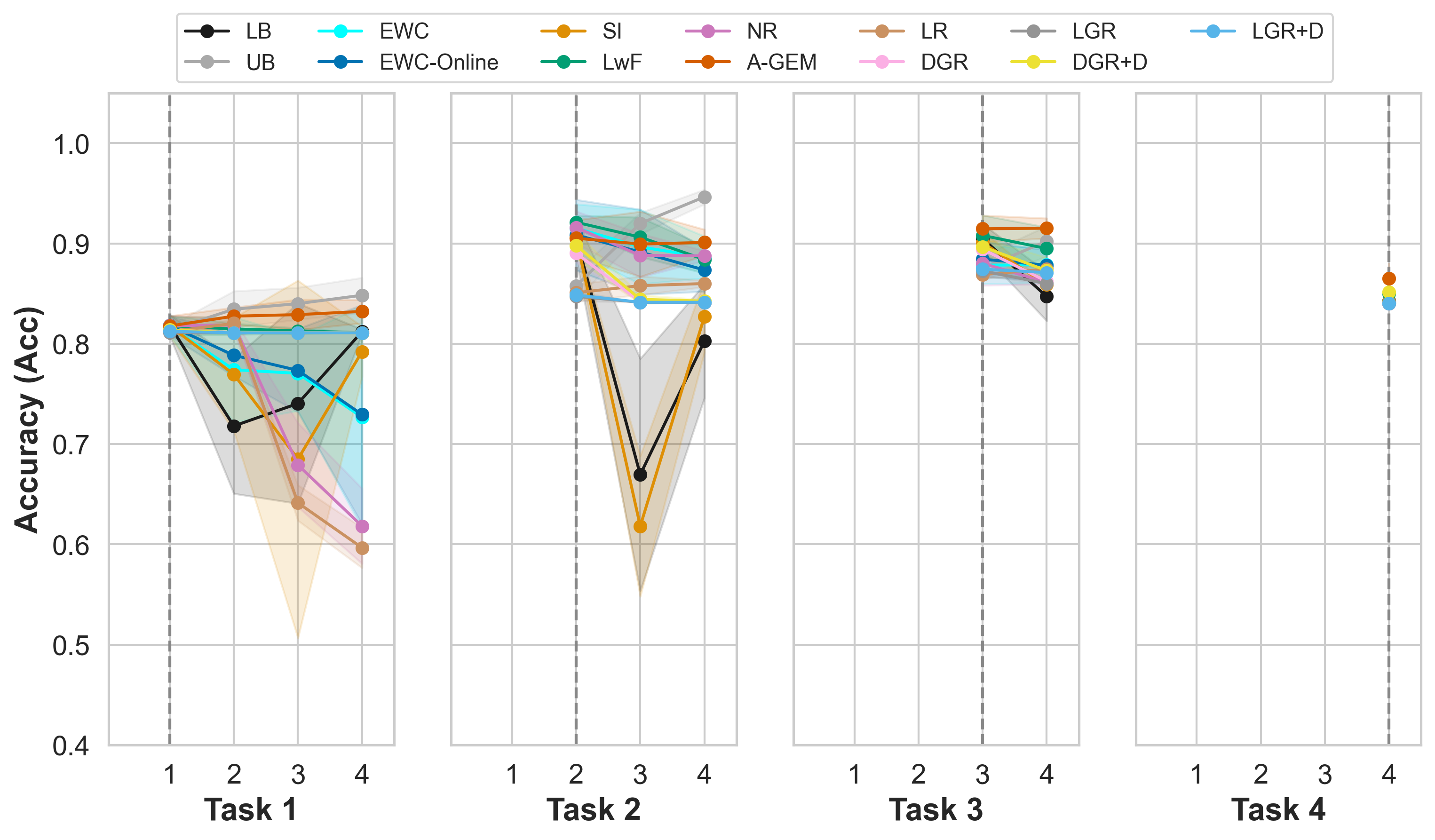}} \\
    \caption[Task-IL Results for BAUM-1 following Order 2 and Order 3.]{\ac{Task-IL} results for  BAUM-1 following (a) Order $2$ and (b) Order $3$, respectively. For each task, test-set accuracy is shown as the learning progresses from when the task is introduced to the end of the overall training procedure.}
    \label{fig:Baum-taskil-orders} 
\end{figure}
\FloatBarrier

\normalsize
\textit{}
\newpage
\subsubsection{\acf{Class-IL} Results}
\label{app:classilbaum}
\begin{table}[h!]
\centering
\setlength\tabcolsep{2.0pt}

\caption[\acs{Class-IL} \acs{Acc} for BAUM-1  W/ and W/O Data-Augmentation.]{\acs{Class-IL} \acs{Acc} for BAUM-1 W/ and W/O Data-Augmentation. \textbf{Bold} values denote best (highest) while [\textit{bracketed}] denote second-best values for each column. 
}
\label{tab:class-il-baum-acc}

{
\scriptsize
\begin{tabular}{l|cccccccc}\toprule

\multirow{2}[2]{*}{\makecell[c]{\textbf{Method}}}           & \multicolumn{8}{c}{\textbf{Accuracy W/O Data-Augmentation}} \\ \cmidrule{2-9}

\multicolumn{1}{c|}{ }                          & 
\multicolumn{1}{c|}{\textbf{Class 1}}            & \multicolumn{1}{c|}{\textbf{Class 2}}          &
\multicolumn{1}{c|}{\textbf{Class 3}}            & \multicolumn{1}{c|}{\textbf{Class 4}}           &
\multicolumn{1}{c|}{\textbf{Class 5}}            & \multicolumn{1}{c|}{\textbf{Class 6}}          &
\multicolumn{1}{c|}{\textbf{Class 7}}            & \multicolumn{1}{c}{\textbf{Class 8}}          \\ \midrule

\multicolumn{9}{c}{\textbf{Baseline Approaches}} \\ \midrule
LB & \cellcolor{gray!25}$\bm{1.00\pm0.00}$ & $0.50\pm0.00$ & $0.33\pm0.00$ & $0.25\pm0.00$ & $0.20\pm0.00$ & $0.17\pm0.00$ & $0.14\pm0.00$ & $0.12\pm0.00$ \\
UB & \cellcolor{gray!25}$\bm{1.00\pm0.00}$ & [$0.81\pm0.01]$ & [$0.83\pm0.03$] & \cellcolor{gray!25}$\bm{0.86\pm0.02}$ & \cellcolor{gray!25}$\bm{0.83\pm0.03}$ & \cellcolor{gray!25}$\bm{0.81\pm0.01}$ & \cellcolor{gray!25}$\bm{0.78\pm0.02}$ & \cellcolor{gray!25}$\bm{0.73\pm0.01}$ \\ \midrule

\multicolumn{9}{c}{\textbf{Regularisation-Based Approaches}} \\ \midrule

EWC  & \cellcolor{gray!25}$\bm{1.00\pm0.00}$ & $0.50\pm0.00$ & $0.33\pm0.00$ & $0.25\pm0.00$ & $0.20\pm0.00$ & $0.17\pm0.00$ & $0.14\pm0.00$ & $0.12\pm0.00$ \\
EWC-Online  & \cellcolor{gray!25}$\bm{1.00\pm0.00}$ & $0.50\pm0.00$ & $0.33\pm0.00$ & $0.25\pm0.00$ & $0.20\pm0.00$ & $0.17\pm0.00$ & $0.14\pm0.00$ & $0.12\pm0.00$ \\
SI  & \cellcolor{gray!25}$\bm{1.00\pm0.00}$ & $0.50\pm0.00$ & $0.33\pm0.00$ & $0.25\pm0.00$ & $0.20\pm0.00$ & $0.17\pm0.00$ & $0.14\pm0.00$ & $0.12\pm0.00$ \\
LwF  & \cellcolor{gray!25}$\bm{1.00\pm0.00}$ & $0.50\pm0.00$ & $0.33\pm0.00$ & $0.25\pm0.00$ & $0.20\pm0.00$ & $0.17\pm0.00$ & $0.14\pm0.00$ & $0.12\pm0.00$ \\\midrule
\multicolumn{9}{c}{\textbf{Replay-Based Approaches}} \\ \midrule

NR  & \cellcolor{gray!25}$\bm{1.00\pm0.00}$ & \cellcolor{gray!25}$\bm{0.91\pm0.00}$ & \cellcolor{gray!25}$\bm{0.91\pm0.01}$ & [$0.83\pm0.01$] & [$0.79\pm0.01$] & [$0.70\pm0.01$] & [$0.65\pm0.01$] & [$0.64\pm0.01$] \\
A-GEM  & \cellcolor{gray!25}$\bm{1.00\pm0.00}$ & $0.52\pm0.02$ & $0.33\pm0.00$ & $0.25\pm0.00$ & $0.20\pm0.00$ & $0.17\pm0.00$ & $0.14\pm0.00$ & $0.12\pm0.00$ \\
LR  & \cellcolor{gray!25}$\bm{1.00\pm0.00}$ & $0.75\pm0.02$ & $0.72\pm0.04$ & $0.64\pm0.05$ & $0.57\pm0.05$ & $0.56\pm0.05$ & $0.50\pm0.05$ & $0.46\pm0.05$ \\
DGR  & \cellcolor{gray!25}$\bm{1.00\pm0.00}$ & $0.50\pm0.00$ & $0.33\pm0.00$ & $0.25\pm0.00$ & $0.20\pm0.00$ & $0.17\pm0.00$ & $0.14\pm0.00$ & $0.12\pm0.00$ \\
LGR  & \cellcolor{gray!25}$\bm{1.00\pm0.00}$ & $0.60\pm0.04$ & $0.63\pm0.02$ & $0.47\pm0.03$ & $0.33\pm0.05$ & $0.33\pm0.04$ & $0.29\pm0.02$ & $0.24\pm0.03$ \\
DGR+D  & \cellcolor{gray!25}$\bm{1.00\pm0.00}$ & $0.50\pm0.00$ & $0.34\pm0.01$ & $0.25\pm0.00$ & $0.20\pm0.00$ & $0.17\pm0.00$ & $0.14\pm0.00$ & $0.12\pm0.00$ \\
LGR+D  & \cellcolor{gray!25}$\bm{1.00\pm0.00}$ & $0.69\pm0.03$ & $0.65\pm0.04$ & $0.52\pm0.03$ & $0.40\pm0.07$ & $0.38\pm0.06$ & $0.32\pm0.05$ & $0.26\pm0.04$ \\

\midrule

\multirow{2}[2]{*}{\makecell[c]{\textbf{Method}}}           & \multicolumn{8}{c}{\textbf{Accuracy W/ Data-Augmentation}} \\ \cmidrule{2-9}

\multicolumn{1}{c|}{ }                          & 
\multicolumn{1}{c|}{\textbf{Class 1}}            & \multicolumn{1}{c|}{\textbf{Class 2}}          &
\multicolumn{1}{c|}{\textbf{Class 3}}            & \multicolumn{1}{c|}{\textbf{Class 4}}           &
\multicolumn{1}{c|}{\textbf{Class 5}}            & \multicolumn{1}{c|}{\textbf{Class 6}}          &
\multicolumn{1}{c|}{\textbf{Class 7}}            & \multicolumn{1}{c}{\textbf{Class 8}}          \\ \midrule

\multicolumn{9}{c}{\textbf{Baseline Approaches}} \\ \midrule

LB & \cellcolor{gray!25}$\bm{1.00\pm0.00}$ & $0.50\pm0.00$ & $0.33\pm0.00$ & $0.25\pm0.00$ & $0.20\pm0.00$ & $0.17\pm0.00$ & $0.14\pm0.00$ & $0.12\pm0.00$ \\
UB & \cellcolor{gray!25}$\bm{1.00\pm0.00}$ & $0.51\pm0.01$ & $0.35\pm0.03$ & $0.33\pm0.11$ & $0.30\pm0.14$ & $0.26\pm0.13$ & $0.24\pm0.14$ & $0.22\pm0.14$ \\ \midrule

\multicolumn{9}{c}{\textbf{Regularisation-Based Approaches}} \\ \midrule

EWC  & \cellcolor{gray!25}$\bm{1.00\pm0.00}$ & $0.50\pm0.00$ & $0.33\pm0.00$ & $0.25\pm0.00$ & $0.20\pm0.00$ & $0.17\pm0.00$ & $0.14\pm0.00$ & $0.12\pm0.00$ \\
EWC-Online  & \cellcolor{gray!25}$\bm{1.00\pm0.00}$ & $0.50\pm0.00$ & $0.33\pm0.00$ & $0.25\pm0.00$ & $0.20\pm0.00$ & $0.17\pm0.00$ & $0.14\pm0.00$ & $0.12\pm0.00$ \\
SI  & \cellcolor{gray!25}$\bm{1.00\pm0.00}$ & $0.50\pm0.00$ & $0.33\pm0.00$ & $0.25\pm0.00$ & $0.20\pm0.00$ & $0.17\pm0.00$ & $0.14\pm0.00$ & $0.12\pm0.00$ \\
LwF  & \cellcolor{gray!25}$\bm{1.00\pm0.00}$ & $0.50\pm0.00$ & $0.33\pm0.00$ & $0.25\pm0.00$ & $0.20\pm0.00$ & $0.17\pm0.00$ & $0.14\pm0.00$ & $0.12\pm0.00$ \\ \midrule
\multicolumn{9}{c}{\textbf{Replay-Based Approaches}} \\ \midrule

NR  & \cellcolor{gray!25}$\bm{1.00\pm0.00}$ & \cellcolor{gray!25}$\bm{0.77\pm0.01}$ & \cellcolor{gray!25}$\bm{0.78\pm0.00}$ & \cellcolor{gray!25}$\bm{0.64\pm0.02}$ & \cellcolor{gray!25}$\bm{0.56\pm0.02}$ & \cellcolor{gray!25}$\bm{0.48\pm0.03}$ & \cellcolor{gray!25}$\bm{0.40\pm0.03}$ & \cellcolor{gray!25}$\bm{0.38\pm0.02}$ \\
A-GEM  & \cellcolor{gray!25}$\bm{1.00\pm0.00}$ & $0.50\pm0.00$ & $0.41\pm0.11$ & $0.25\pm0.00$ & $0.20\pm0.00$ & $0.17\pm0.00$ & $0.14\pm0.00$ & $0.12\pm0.00$ \\
LR  & \cellcolor{gray!25}$\bm{1.00\pm0.00}$ & [$0.60\pm0.07$] & [$0.53\pm0.10$] & [$0.44\pm0.11$] & [$0.35\pm0.09$] & [$0.34\pm0.09$] & [$0.30\pm0.08$] & [$0.27\pm0.08$] \\
DGR  & \cellcolor{gray!25}$\bm{1.00\pm0.00}$ & $0.50\pm0.00$ & $0.34\pm0.00$ & $0.25\pm0.00$ & $0.21\pm0.00$ & $0.17\pm0.00$ & $0.14\pm0.00$ & $0.13\pm0.00$ \\
LGR  & \cellcolor{gray!25}$\bm{1.00\pm0.00}$ & $0.57\pm0.03$ & $0.42\pm0.08$ & $0.27\pm0.02$ & $0.20\pm0.02$ & $0.19\pm0.01$ & $0.19\pm0.04$ & $0.13\pm0.03$ \\
DGR+D  & \cellcolor{gray!25}$\bm{1.00\pm0.00}$ & $0.50\pm0.00$ & $0.35\pm0.00$ & $0.25\pm0.00$ & $0.20\pm0.00$ & $0.17\pm0.00$ & $0.14\pm0.00$ & $0.13\pm0.00$ \\
LGR+D  & \cellcolor{gray!25}$\bm{1.00\pm0.00}$ & $0.57\pm0.05$ & $0.46\pm0.09$ & $0.32\pm0.05$ & $0.21\pm0.01$ & $0.22\pm0.04$ & $0.20\pm0.04$ & $0.14\pm0.01$ \\

\bottomrule

\end{tabular}

}
\end{table}

\begin{table}[h!]
\centering
\caption[\acs{Class-IL} \acs{CF} scores for BAUM-1 W/ and W/O Data-Augmentation.]{\acs{CF} scores for \acs{Class-IL} for BAUM-1  W/ and W/O Data-Augmentation. \textbf{Bold} values denote best (lowest) while [\textit{bracketed}] denote second-best values for each column.}
\label{tab:class-il-baum-cf}

{
\scriptsize
\setlength\tabcolsep{1.2pt}

\begin{tabular}{l|ccccccc}\toprule

\multirow{2}[2]{*}{\makecell[c]{\textbf{Method}}}           & \multicolumn{7}{c}{\textbf{\acs{CF} W/O Data-Augmentation}} \\ \cmidrule{2-8}

\multicolumn{1}{c|}{ }                          & 
\multicolumn{1}{c|}{\textbf{Class 2}}          &
\multicolumn{1}{c|}{\textbf{Class 3}}            & \multicolumn{1}{c|}{\textbf{Class 4}}           &
\multicolumn{1}{c|}{\textbf{Class 5}}            & \multicolumn{1}{c|}{\textbf{Class 6}}          &
\multicolumn{1}{c|}{\textbf{Class 7}}            & \multicolumn{1}{c}{\textbf{Class 8}}          \\ \midrule
\multicolumn{8}{c}{\textbf{Baseline Approaches}} \\ \midrule

LB &  $1.00\pm0.00$ &  $1.00\pm0.00$ &  $1.00\pm0.00$ &  $1.00\pm0.00$ &  $1.00\pm0.00$ &  $1.00\pm0.00$ & $1.00\pm0.00$ \\
UB & \cellcolor{gray!25}$\bm{-0.05\pm0.03}$ & \cellcolor{gray!25}$\bm{-0.07\pm0.02}$ & \cellcolor{gray!25}$\bm{-0.11\pm0.00}$ & \cellcolor{gray!25}$\bm{-0.05\pm0.02}$ & \cellcolor{gray!25}$\bm{-0.01\pm0.05}$ & \cellcolor{gray!25}$\bm{-0.06\pm0.03}$ & \cellcolor{gray!25}$\bm{0.01\pm0.00}$ \\\midrule

\multicolumn{8}{c}{\textbf{Regularisation-Based Approaches}} \\ \midrule

EWC  &  $1.00\pm0.00$ &  $1.00\pm0.00$ &  $1.00\pm0.00$ &  $1.00\pm0.00$ &  $1.00\pm0.00$ &  $1.00\pm0.00$ & $1.00\pm0.00$ \\
EWC-Online  &  $1.00\pm0.00$ &  $1.00\pm0.00$ &  $1.00\pm0.00$ &  $1.00\pm0.00$ &  $1.00\pm0.00$ &  $1.00\pm0.00$ & $1.00\pm0.00$ \\
SI  &  $1.00\pm0.00$ &  $1.00\pm0.00$ &  $1.00\pm0.00$ &  $1.00\pm0.00$ &  $1.00\pm0.00$ &  $1.00\pm0.00$ & $1.00\pm0.00$ \\
LwF  &  $1.00\pm0.00$ &  $1.00\pm0.00$ &  $1.00\pm0.00$ &  $1.00\pm0.00$ &  $1.00\pm0.00$ &  $1.00\pm0.00$ & $1.00\pm0.00$ \\ \midrule

\multicolumn{8}{c}{\textbf{Replay-Based Approaches}} \\ \midrule

NR  &  $0.10\pm0.01$ &  [$0.19\pm0.00$] &  $0.24\pm0.01$ &  $0.33\pm0.02$ &  $0.38\pm0.01$ &  $0.38\pm0.01$ & $0.00\pm0.00$ \\
A-GEM  &  $1.00\pm0.00$ &  $1.00\pm0.00$ &  $1.00\pm0.00$ &  $1.00\pm0.00$ &  $1.00\pm0.00$ &  $0.95\pm0.07$ & $1.00\pm0.00$ \\
LR  &  [$0.22\pm0.04$] &  [$0.19\pm0.03$] &  [$0.18\pm0.02$] &  [$0.17\pm0.03$] &  [$0.18\pm0.02$] &  [$0.16\pm0.01$] & [$0.15\pm0.00$] \\
DGR  &  $1.00\pm0.00$ &  $1.00\pm0.00$ &  $1.00\pm0.00$ &  $1.00\pm0.00$ &  $1.00\pm0.00$ &  $1.00\pm0.00$ & $1.00\pm0.00$ \\
LGR  &  $0.50\pm0.04$ &  $0.62\pm0.04$ &  $0.73\pm0.05$ &  $0.70\pm0.04$ &  $0.72\pm0.03$ &  $0.76\pm0.04$ & $0.70\pm0.00$ \\
DGR+D  &  $0.99\pm0.01$ &  $1.00\pm0.00$ &  $1.00\pm0.00$ &  $1.00\pm0.00$ &  $1.00\pm0.00$ &  $1.00\pm0.00$ & $1.00\pm0.00$ \\
LGR+D  &  $0.41\pm0.04$ &  $0.46\pm0.01$ &  $0.56\pm0.08$ &  $0.54\pm0.07$ &  $0.60\pm0.05$ &  $0.66\pm0.05$ & $0.65\pm0.00$ \\ \midrule

\multirow{2}[2]{*}{\makecell[c]{\textbf{Method}}}          & \multicolumn{7}{c}{\textbf{\acs{CF} W/ Data-Augmentation}} \\ \cmidrule{2-8}

\multicolumn{1}{c|}{ }                          & 
\multicolumn{1}{c|}{\textbf{Class 2}}          &
\multicolumn{1}{c|}{\textbf{Class 3}}            & \multicolumn{1}{c|}{\textbf{Class 4}}           &
\multicolumn{1}{c|}{\textbf{Class 5}}            & \multicolumn{1}{c|}{\textbf{Class 6}}          &
\multicolumn{1}{c|}{\textbf{Class 7}}            & \multicolumn{1}{c}{\textbf{Class 8}}          \\ \midrule

\multicolumn{8}{c}{\textbf{Baseline Approaches}} \\ \midrule
LB &  $1.00\pm0.00$ &  $1.00\pm0.00$ &  $1.00\pm0.00$ &  $1.00\pm0.00$ &  $1.00\pm0.00$ &  $1.00\pm0.00$ & $1.00\pm0.00$ \\
UB & \cellcolor{gray!25}$\bm{-0.02\pm0.03}$ & \cellcolor{gray!25}$\bm{-0.04\pm0.05}$ & \cellcolor{gray!25}$\bm{-0.07\pm0.10}$ & \cellcolor{gray!25}$\bm{-0.02\pm0.03}$ & \cellcolor{gray!25}$\bm{-0.01\pm0.01}$ & \cellcolor{gray!25}$\bm{-0.02\pm0.03}$ & \cellcolor{gray!25}$\bm{0.01\pm0.00}$ \\ \midrule

\multicolumn{8}{c}{\textbf{Regularisation-Based Approaches}} \\ \midrule

EWC  &  $1.00\pm0.00$ &  $1.00\pm0.00$ &  $1.00\pm0.00$ &  $1.00\pm0.00$ &  $1.00\pm0.00$ &  $1.00\pm0.00$ & $1.00\pm0.00$ \\
EWC-Online  &  $1.00\pm0.00$ &  $1.00\pm0.00$ &  $1.00\pm0.00$ &  $1.00\pm0.00$ &  $1.00\pm0.00$ &  $1.00\pm0.00$ & $1.00\pm0.00$ \\
SI  &  $1.00\pm0.00$ &  $1.00\pm0.00$ &  $1.00\pm0.00$ &  $1.00\pm0.00$ &  $1.00\pm0.00$ &  $1.00\pm0.00$ & $1.00\pm0.00$ \\
LwF  &  $1.00\pm0.00$ &  $1.00\pm0.00$ &  $1.00\pm0.00$ &  $1.00\pm0.00$ &  $1.00\pm0.00$ &  $1.00\pm0.00$ & $1.00\pm0.00$ \\ \midrule

\multicolumn{8}{c}{\textbf{Replay-Based Approaches}} \\ \midrule

NR  &  [$0.21\pm0.02$] &  $0.35\pm0.05$ &  $0.45\pm0.05$ &  $0.52\pm0.05$ &  $0.59\pm0.06$ &  $0.60\pm0.05$ & $0.48\pm0.00$ \\
A-GEM  &  $0.88\pm0.17$ &  $1.00\pm0.01$ &  $1.00\pm0.00$ &  $1.00\pm0.00$ &  $1.00\pm0.00$ &  $1.00\pm0.00$ & $1.00\pm0.00$ \\
LR  &  $0.41\pm0.15$ &  [$0.30\pm0.08$] &  [$0.26\pm0.04$] &  [$0.25\pm0.05$] &  [$0.22\pm0.02$] &  [$0.22\pm0.01$] & [$0.22\pm0.00$] \\
DGR  &  $0.99\pm0.00$ &  $1.00\pm0.00$ &  $0.99\pm0.00$ &  $1.00\pm0.00$ &  $1.00\pm0.00$ &  $1.00\pm0.00$ & $1.00\pm0.00$ \\
LGR  &  $0.62\pm0.14$ &  $0.68\pm0.28$ &  $0.69\pm0.31$ &  $0.69\pm0.32$ &  $0.65\pm0.27$ &  $0.68\pm0.31$ & $0.65\pm0.00$ \\
DGR+D  &  $0.97\pm0.00$ &  $1.00\pm0.00$ &  $1.00\pm0.00$ &  $1.00\pm0.00$ &  $1.00\pm0.00$ &  $1.00\pm0.00$ & $1.00\pm0.00$ \\
LGR+D  &  $0.25\pm0.20$ &  $0.35\pm0.28$ &  $0.46\pm0.34$ &  $0.42\pm0.31$ &  $0.43\pm0.32$ &  $0.50\pm0.36$ & $0.50\pm0.00$ \\

\bottomrule

\end{tabular}

}
\end{table}

\begin{figure}[h!]
    \centering
    \subfloat[\ac{Class-IL} Results w/o Augmentation.\label{fig:baum-class-il-noaug}]{\includegraphics[width=0.5\textwidth]{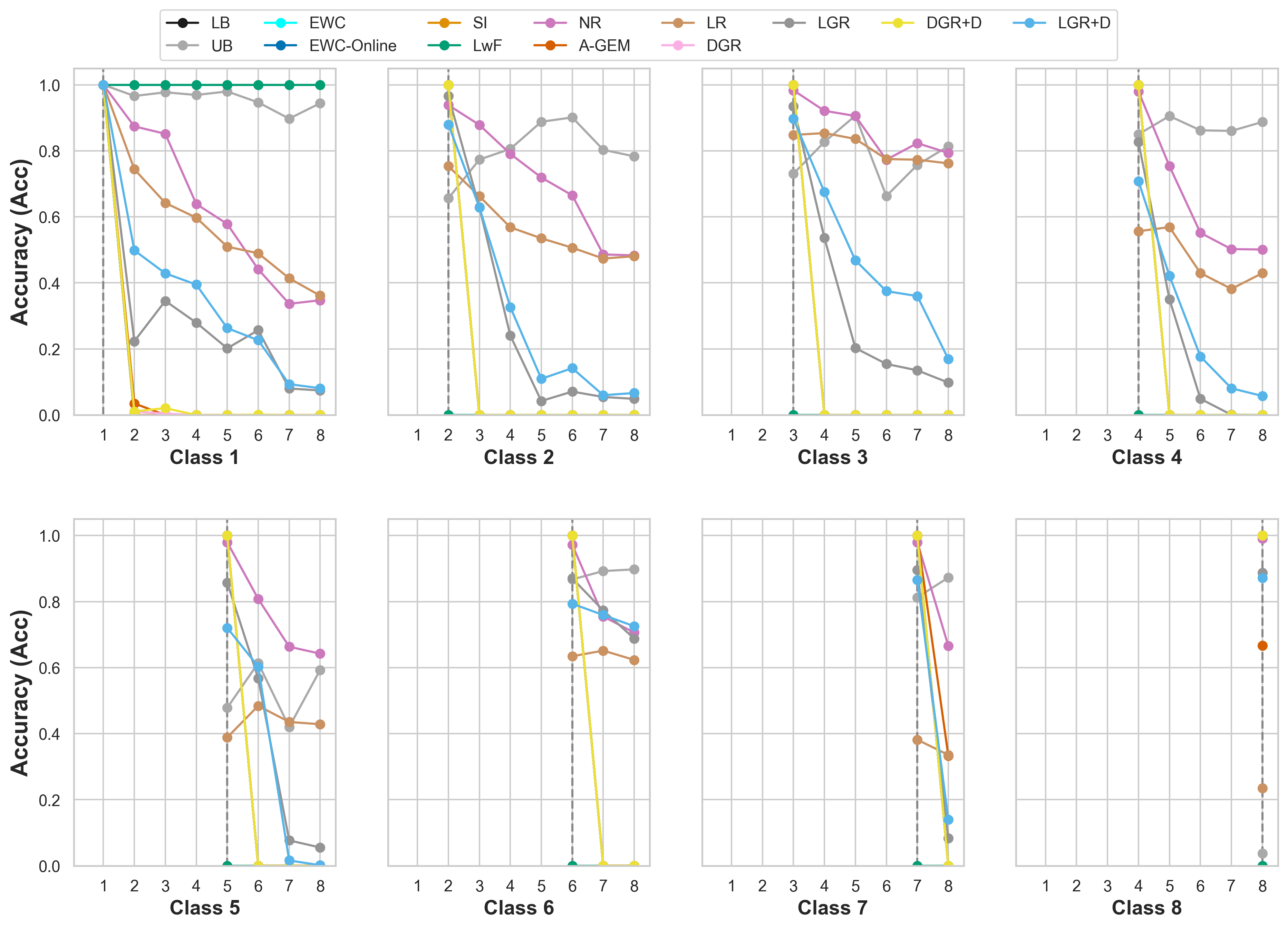}} \hfill
    \subfloat[\ac{Class-IL} Results w/ Augmentation.\label{fig:baum-class-il-aug}]{\includegraphics[width=0.5\textwidth]{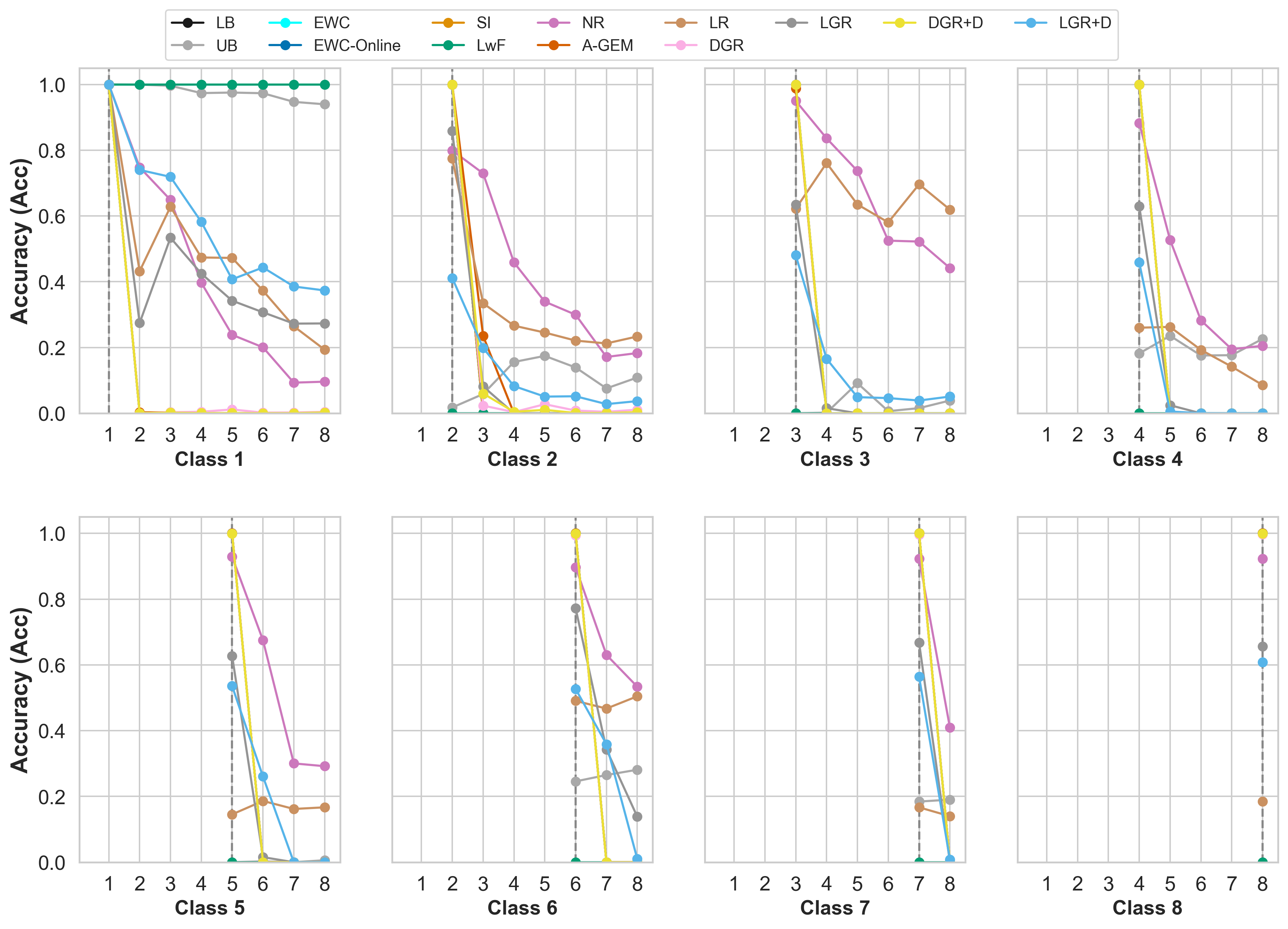}} \\
    \caption[Class-IL Results for BAUM-1 W/ and W/O Augmentation.]{\ac{Class-IL} results for BAUM-1 (a) without and (b) with augmentation. For each class, test-set accuracy is shown as the learning progresses from when the class is introduced to the end of the overall training procedure.}
    \label{fig:baum-Class} 
    
\end{figure}

\FloatBarrier
\normalsize

\subsubsection{Class-Ordering}
\label{app:classorderbaum}
\begin{table}[h!]
\centering
\setlength\tabcolsep{2.0pt}

\caption[\acs{Class-IL} \acs{Acc} for BAUM-1 following Orderings O2 and O3.]{\acs{Class-IL} \acs{Acc} for BAUM-1 W/ Data-Augmentation following Orderings O2 and O3. \textbf{Bold} values denote best (highest) while [\textit{bracketed}] denote second-best values for each column.}

\label{tab:class-il-baum-acc-orders}

{
\scriptsize
\begin{tabular}{l|cccccccc}\toprule

\multirow{2}[2]{*}{\makecell[c]{\textbf{Method}}}           & \multicolumn{8}{c}{\textbf{Accuracy following O2}} \\ \cmidrule{2-9}

\multicolumn{1}{c|}{ }                          & 
\multicolumn{1}{c|}{\textbf{Class 1}}            & \multicolumn{1}{c|}{\textbf{Class 2}}          &
\multicolumn{1}{c|}{\textbf{Class 3}}            & \multicolumn{1}{c|}{\textbf{Class 4}}           &
\multicolumn{1}{c|}{\textbf{Class 5}}            & \multicolumn{1}{c|}{\textbf{Class 6}}          &
\multicolumn{1}{c|}{\textbf{Class 7}}            & \multicolumn{1}{c}{\textbf{Class 8}}          \\ \midrule

LB & \cellcolor{gray!25}$\bm{1.00\pm0.00}$ & $0.50\pm0.00$ & $0.33\pm0.00$ & $0.25\pm0.00$ & $0.20\pm0.00$ & $0.17\pm0.00$ & $0.14\pm0.00$ & $0.12\pm0.00$ \\
UB & \cellcolor{gray!25}$\bm{1.00\pm0.00}$ & $0.50\pm0.00$ & $0.39\pm0.05$ & $0.36\pm0.08$ & $0.36\pm0.11$ & $0.33\pm0.11$ & $0.32\pm0.13$ & $0.30\pm0.13$ \\
 \midrule

\makecell[c]{}                   & \multicolumn{8}{c}{\textbf{Regularisation-Based Approaches}} \\ \midrule
EWC  & \cellcolor{gray!25}$\bm{1.00\pm0.00}$ & $0.50\pm0.00$ & $0.33\pm0.00$ & $0.25\pm0.00$ & $0.20\pm0.00$ & $0.17\pm0.00$ & $0.14\pm0.00$ & $0.12\pm0.00$ \\
EWC-Online  & \cellcolor{gray!25}$\bm{1.00\pm0.00}$ & $0.50\pm0.00$ & $0.33\pm0.00$ & $0.25\pm0.00$ & $0.20\pm0.00$ & $0.17\pm0.00$ & $0.14\pm0.00$ & $0.12\pm0.00$ \\
SI  & \cellcolor{gray!25}$\bm{1.00\pm0.00}$ & $0.50\pm0.00$ & $0.33\pm0.00$ & $0.25\pm0.00$ & $0.20\pm0.00$ & $0.17\pm0.00$ & $0.14\pm0.00$ & $0.12\pm0.00$ \\
LwF  & \cellcolor{gray!25}$\bm{1.00\pm0.00}$ & $0.50\pm0.00$ & $0.33\pm0.00$ & $0.25\pm0.00$ & $0.20\pm0.00$ & $0.17\pm0.00$ & $0.14\pm0.00$ & $0.12\pm0.00$ \\\midrule
\makecell[c]{}                           & \multicolumn{8}{c}{\textbf{Replay-Based Approaches}} \\ \midrule

NR  & \cellcolor{gray!25}$\bm{1.00\pm0.00}$ & \cellcolor{gray!25}$\bm{0.77\pm0.02}$ & \cellcolor{gray!25}$\bm{0.75\pm0.03}$ & \cellcolor{gray!25}$\bm{0.63\pm0.04}$ & \cellcolor{gray!25}$\bm{0.54\pm0.04}$ & \cellcolor{gray!25}$\bm{0.43\pm0.04}$ & \cellcolor{gray!25}$\bm{0.40\pm0.04}$ & \cellcolor{gray!25}$\bm{0.35\pm0.05}$ \\
A-GEM  & \cellcolor{gray!25}$\bm{1.00\pm0.00}$ & $0.51\pm0.01$ & $0.33\pm0.00$ & $0.25\pm0.00$ & $0.20\pm0.00$ & $0.17\pm0.00$ & $0.14\pm0.00$ & $0.12\pm0.00$ \\
LR  & \cellcolor{gray!25}$\bm{1.00\pm0.00}$ & [$0.67\pm0.01$] & [$0.62\pm0.02$] & [$0.52\pm0.03$] & [$0.43\pm0.03$] & [$0.42\pm0.04$] & [$0.37\pm0.03$] & [$0.33\pm0.03$] \\
DGR  & \cellcolor{gray!25}$\bm{1.00\pm0.00}$ & $0.50\pm0.00$ & $0.35\pm0.01$ & $0.25\pm0.00$ & $0.21\pm0.00$ & $0.17\pm0.00$ & $0.14\pm0.00$ & $0.13\pm0.00$ \\
LGR  & \cellcolor{gray!25}$\bm{1.00\pm0.00}$ & $0.56\pm0.03$ & $0.56\pm0.01$ & $0.32\pm0.01$ & $0.24\pm0.03$ & $0.22\pm0.05$ & $0.22\pm0.02$ & $0.17\pm0.04$ \\
DGR+D  & \cellcolor{gray!25}$\bm{1.00\pm0.00}$ & $0.50\pm0.00$ & $0.35\pm0.01$ & $0.25\pm0.00$ & $0.20\pm0.00$ & $0.17\pm0.00$ & $0.14\pm0.00$ & $0.12\pm0.00$ \\
LGR+D  & \cellcolor{gray!25}$\bm{1.00\pm0.00}$ & $0.61\pm0.02$ & $0.56\pm0.02$ & $0.38\pm0.05$ & $0.28\pm0.06$ & $0.29\pm0.05$ & $0.24\pm0.03$ & $0.19\pm0.05$ \\ \midrule

\multirow{2}[2]{*}{\makecell[c]{\textbf{Method}}}           & \multicolumn{8}{c}{\textbf{Accuracy following O3}} \\ \cmidrule{2-9}

\multicolumn{1}{c|}{ }                          & 
\multicolumn{1}{c|}{\textbf{Class 1}}            & \multicolumn{1}{c|}{\textbf{Class 2}}          &
\multicolumn{1}{c|}{\textbf{Class 3}}            & \multicolumn{1}{c|}{\textbf{Class 4}}           &
\multicolumn{1}{c|}{\textbf{Class 5}}            & \multicolumn{1}{c|}{\textbf{Class 6}}          &
\multicolumn{1}{c|}{\textbf{Class 7}}            & \multicolumn{1}{c}{\textbf{Class 8}}          \\ \midrule

\multicolumn{9}{c}{\textbf{Baseline Approaches}} \\ \midrule
LB & \cellcolor{gray!25}$\bm{1.00\pm0.00}$ & $0.50\pm0.00$ & $0.33\pm0.00$ & $0.25\pm0.00$ & $0.20\pm0.00$ & $0.17\pm0.00$ & $0.14\pm0.00$ & $0.12\pm0.00$ \\
UB & \cellcolor{gray!25}$\bm{1.00\pm0.00}$ & $0.50\pm0.00$ & $0.33\pm0.00$ & $0.25\pm0.00$ & $0.20\pm0.00$ & $0.17\pm0.00$ & $0.17\pm0.04$ & $0.15\pm0.04$ \\

 \midrule

\multicolumn{9}{c}{\textbf{Regularisation-Based Approaches}} \\ \midrule

EWC  & \cellcolor{gray!25}$\bm{1.00\pm0.00}$ & $0.50\pm0.00$ & $0.33\pm0.00$ & $0.25\pm0.00$ & $0.20\pm0.00$ & $0.17\pm0.00$ & $0.14\pm0.00$ & $0.12\pm0.00$ \\
EWC-Online  & \cellcolor{gray!25}$\bm{1.00\pm0.00}$ & $0.50\pm0.00$ & $0.33\pm0.00$ & $0.25\pm0.00$ & $0.20\pm0.00$ & $0.17\pm0.00$ & $0.14\pm0.00$ & $0.12\pm0.00$ \\
SI  & \cellcolor{gray!25}$\bm{1.00\pm0.00}$ & $0.50\pm0.00$ & $0.33\pm0.00$ & $0.25\pm0.00$ & $0.20\pm0.00$ & $0.17\pm0.00$ & $0.14\pm0.00$ & $0.12\pm0.00$ \\
LwF  & \cellcolor{gray!25}$\bm{1.00\pm0.00}$ & $0.50\pm0.00$ & $0.33\pm0.00$ & $0.25\pm0.00$ & $0.20\pm0.00$ & $0.17\pm0.00$ & $0.14\pm0.00$ & $0.12\pm0.00$ \\ \midrule
\multicolumn{9}{c}{\textbf{Replay-Based Approaches}} \\ \midrule

NR  & \cellcolor{gray!25}$\bm{1.00\pm0.00}$ & \cellcolor{gray!25}$\bm{0.91\pm0.01}$ & \cellcolor{gray!25}$\bm{0.68\pm0.02}$ & \cellcolor{gray!25}$\bm{0.60\pm0.04}$ & \cellcolor{gray!25}$\bm{0.54\pm0.04}$ & \cellcolor{gray!25}$\bm{0.47\pm0.01}$ & \cellcolor{gray!25}$\bm{0.43\pm0.01}$ & \cellcolor{gray!25}$\bm{0.39\pm0.01}$ \\
A-GEM  & \cellcolor{gray!25}$\bm{1.00\pm0.00}$ & $0.63\pm0.03$ & $0.33\pm0.00$ & $0.25\pm0.00$ & $0.20\pm0.00$ & $0.17\pm0.00$ & $0.14\pm0.00$ & $0.12\pm0.00$ \\
LR  & \cellcolor{gray!25}$\bm{1.00\pm0.00}$ & [$0.80\pm0.01$] & [$0.58\pm0.01$] & [$0.51\pm0.00$] & [$0.43\pm0.01$] & [$0.38\pm0.00$] & [$0.38\pm0.01$] & [$0.33\pm0.00$] \\
DGR  & \cellcolor{gray!25}$\bm{1.00\pm0.00}$ & $0.54\pm0.02$ & $0.33\pm0.00$ & $0.25\pm0.00$ & $0.20\pm0.00$ & $0.17\pm0.00$ & $0.14\pm0.00$ & $0.12\pm0.00$ \\
LGR  & \cellcolor{gray!25}$\bm{1.00\pm0.00}$ & $0.75\pm0.01$ & $0.51\pm0.01$ & $0.28\pm0.03$ & $0.21\pm0.03$ & $0.21\pm0.03$ & $0.19\pm0.04$ & $0.20\pm0.02$ \\
DGR+D  & \cellcolor{gray!25}$\bm{1.00\pm0.00}$ & $0.59\pm0.00$ & $0.33\pm0.00$ & $0.25\pm0.00$ & $0.20\pm0.00$ & $0.17\pm0.00$ & $0.14\pm0.00$ & $0.13\pm0.00$ \\
LGR+D  & \cellcolor{gray!25}$\bm{1.00\pm0.00}$ & $0.75\pm0.02$ & $0.51\pm0.01$ & $0.30\pm0.01$ & $0.21\pm0.00$ & $0.21\pm0.02$ & $0.23\pm0.01$ & $0.22\pm0.00$ \\

\bottomrule

\end{tabular}

}
\end{table}

\begin{table}[h!]
\centering
\caption[\acs{Class-IL} \acs{CF} scores for BAUM-1 following O2 and O3.]{\acs{Class-IL} \acs{CF} scores for BAUM-1 W/ Data-Augmentation following Orderings O2 and O3. \textbf{Bold} values denote best (lowest) while [\textit{bracketed}] denote second-best values for each column.}

\label{tab:class-il-baum-cf-orders}

{
\scriptsize
\setlength\tabcolsep{2.8pt}

\begin{tabular}{l|crrrrrrr}\toprule

\multirow{2}[2]{*}{\makecell[c]{\textbf{Method}}}           & \multicolumn{7}{c}{\textbf{\acs{CF} following O2}} \\ \cmidrule{2-8}

\multicolumn{1}{c|}{ }                          & \multicolumn{1}{c|}{\textbf{Class 2}}          &
\multicolumn{1}{c|}{\textbf{Class 3}}            & \multicolumn{1}{c|}{\textbf{Class 4}}           &
\multicolumn{1}{c|}{\textbf{Class 5}}            & \multicolumn{1}{c|}{\textbf{Class 6}}          &
\multicolumn{1}{c|}{\textbf{Class 7}}            & \multicolumn{1}{c}{\textbf{Class 8}}          \\ \midrule
\multicolumn{8}{c}{\textbf{Baseline Approaches}} \\ \midrule
LB  &  $1.00\pm0.00$ &  $1.00\pm0.00$ &  $1.00\pm0.00$ &  $1.00\pm0.00$ &  $1.00\pm0.00$ &  $1.00\pm0.00$ & $1.00\pm0.00$ \\
UB  & \cellcolor{gray!25}$\bm{-0.08\pm0.07}$ & \cellcolor{gray!25}$\bm{-0.03\pm0.03}$ & \cellcolor{gray!25}$\bm{-0.11\pm0.08}$ & \cellcolor{gray!25}$\bm{-0.03\pm0.02}$ & \cellcolor{gray!25}$\bm{-0.01\pm0.01}$ & \cellcolor{gray!25}$\bm{-0.03\pm0.02}$ & \cellcolor{gray!25}$\bm{0.01\pm0.00}$ \\ \midrule

\multicolumn{8}{c}{\textbf{Regularisation-Based Approaches}} \\ \midrule

EWC   &  $1.00\pm0.00$ &  $1.00\pm0.00$ &  $1.00\pm0.00$ &  $1.00\pm0.00$ &  $1.00\pm0.00$ &  $1.00\pm0.00$ & $1.00\pm0.00$ \\
EWC-Online   &  $1.00\pm0.00$ &  $1.00\pm0.00$ &  $1.00\pm0.00$ &  $1.00\pm0.00$ &  $1.00\pm0.00$ &  $1.00\pm0.00$ & $0.00\pm0.00$ \\
SI   &  $1.00\pm0.00$ &  $1.00\pm0.00$ &  $1.00\pm0.00$ &  $1.00\pm0.00$ &  $1.00\pm0.00$ &  $1.00\pm0.00$ & $1.00\pm0.00$ \\
LwF  &  $1.00\pm0.00$ &  $1.00\pm0.00$ &  $1.00\pm0.00$ &  $1.00\pm0.00$ &  $1.00\pm0.00$ &  $1.00\pm0.00$ & $1.00\pm0.00$ \\ \midrule

\multicolumn{8}{c}{\textbf{Replay-Based Approaches}} \\ \midrule

NR   &  $0.27\pm0.05$ &  $0.39\pm0.09$ &  $0.48\pm0.08$ &  $0.58\pm0.10$ &  $0.60\pm0.11$ &  $0.64\pm0.11$ & $0.60\pm0.00$ \\
A-GEM   &  $1.00\pm0.00$ &  $1.00\pm0.00$ &  $0.92\pm0.12$ &  $0.93\pm0.09$ &  $0.94\pm0.08$ &  $0.95\pm0.07$ & $0.94\pm0.00$ \\
LR   &  [$0.26\pm0.03$] &  [$0.22\pm0.01$] &  [$0.18\pm0.01$] &  [$0.19\pm0.01$] &  [$0.19\pm0.01$] &  [$0.20\pm0.01$] & [$0.20\pm0.00$] \\
DGR   &  $0.98\pm0.01$ &  $1.00\pm0.00$ &  $0.99\pm0.01$ &  $1.00\pm0.00$ &  $1.00\pm0.00$ &  $1.00\pm0.00$ & $1.00\pm0.00$ \\
LGR   &  $0.53\pm0.11$ &  $0.77\pm0.09$ &  $0.81\pm0.04$ &  $0.78\pm0.00$ &  $0.77\pm0.04$ &  $0.82\pm0.00$ & $0.79\pm0.00$ \\
DGR+D   &  $0.97\pm0.02$ &  $1.00\pm0.00$ &  $0.99\pm0.00$ &  $1.00\pm0.00$ &  $1.00\pm0.00$ &  $1.00\pm0.00$ & $1.00\pm0.00$ \\
LGR+D   &  $0.34\pm0.13$ &  $0.48\pm0.14$ &  $0.54\pm0.10$ &  $0.51\pm0.10$ &  $0.56\pm0.10$ &  $0.61\pm0.10$ & $0.60\pm0.00$ \\ \midrule

\multirow{2}[2]{*}{\makecell[c]{\textbf{Method}}}           & \multicolumn{7}{c}{\textbf{\acs{CF} following O3}} \\ \cmidrule{2-8}

\multicolumn{1}{c|}{ }                          & \multicolumn{1}{c|}{\textbf{Class 2}}          &
\multicolumn{1}{c|}{\textbf{Class 3}}            & \multicolumn{1}{c|}{\textbf{Class 4}}           &
\multicolumn{1}{c|}{\textbf{Class 5}}            & \multicolumn{1}{c|}{\textbf{Class 6}}          &
\multicolumn{1}{c|}{\textbf{Class 7}}            & \multicolumn{1}{c}{\textbf{Class 8}}          \\ \midrule

\multicolumn{8}{c}{\textbf{Baseline Approaches}} \\ \midrule

LB  &  $1.00\pm0.00$ &  $1.00\pm0.00$ &  $1.00\pm0.00$ &  $1.00\pm0.00$ &  $1.00\pm0.00$ &  $1.00\pm0.00$ & $1.00\pm0.00$ \\
UB  &  $1.00\pm0.00$ &  $1.00\pm0.00$ &  $1.00\pm0.00$ &  $1.00\pm0.00$ &  $1.00\pm0.00$ &  $0.50\pm0.00$ & $0.55\pm0.00$ \\ \midrule

\multicolumn{8}{c}{\textbf{Regularisation-Based Approaches}} \\ \midrule

EWC   &  $1.00\pm0.00$ &  $1.00\pm0.00$ &  $1.00\pm0.00$ &  $1.00\pm0.00$ &  $1.00\pm0.00$ &  $1.00\pm0.00$ & $1.00\pm0.00$ \\
EWC-Online   &  $1.00\pm0.00$ &  $1.00\pm0.00$ &  $1.00\pm0.00$ &  $1.00\pm0.00$ &  $1.00\pm0.00$ &  $1.00\pm0.00$ & $1.00\pm0.00$ \\
SI   &  $1.00\pm0.00$ &  $1.00\pm0.00$ &  $1.00\pm0.00$ &  $1.00\pm0.00$ &  $1.00\pm0.00$ &  $1.00\pm0.00$ & $1.00\pm0.00$ \\
LwF   &  $1.00\pm0.00$ &  $1.00\pm0.00$ &  $1.00\pm0.00$ &  $1.00\pm0.00$ &  $1.00\pm0.00$ &  $1.00\pm0.00$ & $1.00\pm0.00$ \\ \midrule

\multicolumn{8}{c}{\textbf{Replay-Based Approaches}} \\ \midrule

NR   &  \cellcolor{gray!25}$\bm{0.28\pm0.02}$ &  [$0.27\pm0.04$] &  [$0.29\pm0.01$] &  [$0.36\pm0.03$] &  [$0.40\pm0.02$] &  [$0.46\pm0.02$] & [$0.44\pm0.00$] \\
A-GEM   &  $1.00\pm0.00$ &  $1.00\pm0.00$ &  $1.00\pm0.00$ &  $1.00\pm0.00$ &  $0.94\pm0.08$ &  $0.95\pm0.07$ & $0.93\pm0.00$ \\
LR   &  $0.29\pm0.01$ &  \cellcolor{gray!25}$\bm{0.26\pm0.03}$ &  \cellcolor{gray!25}$\bm{0.22\pm0.02}$ &  \cellcolor{gray!25}$\bm{0.20\pm0.01}$ &  \cellcolor{gray!25}$\bm{0.20\pm0.01}$ &  \cellcolor{gray!25}$\bm{0.21\pm0.01}$ & \cellcolor{gray!25}$\bm{0.20\pm0.00}$ \\
DGR   &  $1.00\pm0.00$ &  $1.00\pm0.00$ &  $1.00\pm0.00$ &  $1.00\pm0.00$ &  $1.00\pm0.00$ &  $1.00\pm0.00$ & $1.00\pm0.00$ \\
LGR   &  $0.63\pm0.03$ &  $0.86\pm0.04$ &  $0.86\pm0.06$ &  $0.82\pm0.07$ &  $0.81\pm0.08$ &  $0.78\pm0.05$ & $0.75\pm0.00$ \\
DGR+D   &  $1.00\pm0.00$ &  $1.00\pm0.00$ &  $1.00\pm0.00$ &  $1.00\pm0.00$ &  $1.00\pm0.00$ &  $1.00\pm0.00$ & $1.00\pm0.00$ \\
LGR+D   &  $0.47\pm0.01$ &  $0.68\pm0.02$ &  $0.74\pm0.05$ &  $0.72\pm0.02$ &  $0.69\pm0.04$ &  $0.69\pm0.04$ & $0.66\pm0.00$ \\

\bottomrule

\end{tabular}

}
\end{table}

\begin{figure}[h!]  
    \centering
    \subfloat[\ac{Class-IL} Results following O$2$.\label{fig:baum-Class-il-aug-order-1}]{\includegraphics[width=0.5\textwidth]{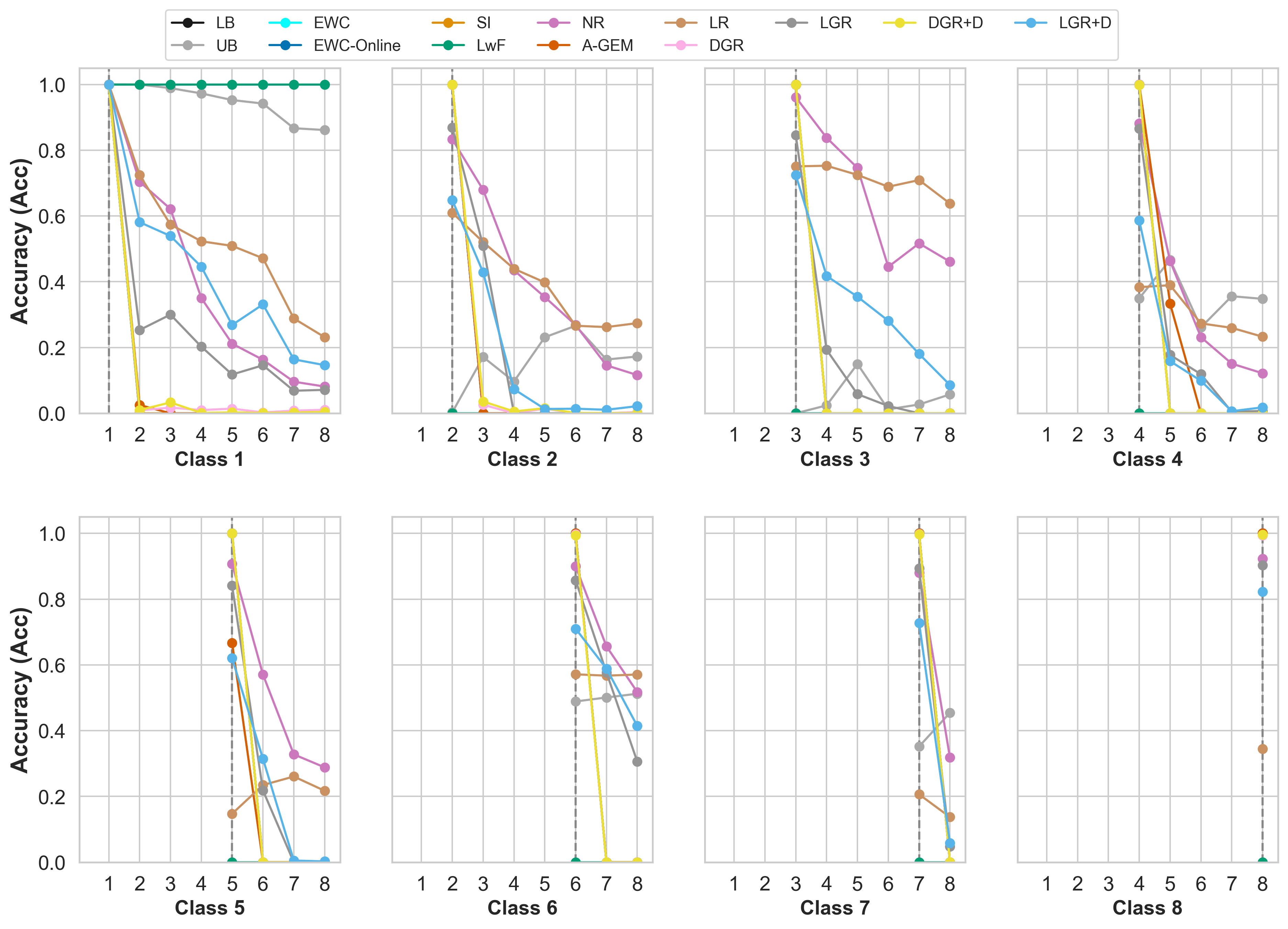}}
    \hfill
    \subfloat[\ac{Class-IL} Results following O$3$.\label{fig:baum-Class-il-aug-order-2}]{\includegraphics[width=0.5\textwidth]{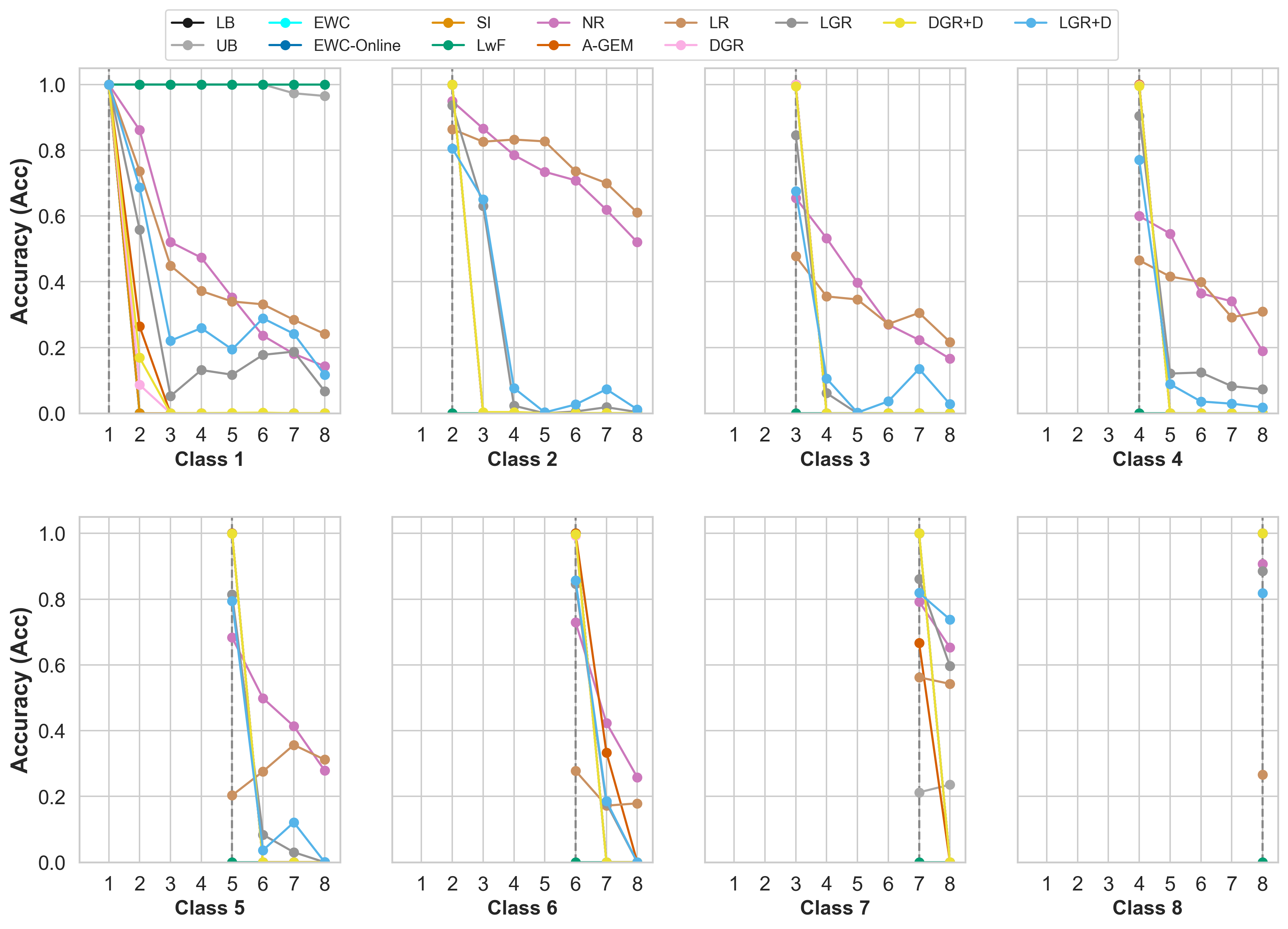}} \\
    \caption[Class-IL Results for BAUM-1 following Order 2 and Order 3.]{\ac{Class-IL} results for BAUM-1 following (a) Order $2$ and (b) Order $3$, respectively. For each class, test-set accuracy is shown as the learning progresses from when the class is introduced to the end of the overall training procedure.}
    \label{fig:baum-classil-orders} 
    
\end{figure}

\FloatBarrier

\normalsize

\end{document}